\pdfoutput=1
\RequirePackage{fix-cm}
\documentclass[twocolumn]{svjour3}          
\smartqed  
\usepackage{graphicx}
\usepackage{epsfig}
\usepackage{amsmath}
\usepackage{amssymb}
\usepackage{fancyhdr}
\usepackage[nolist]{acronym}
\usepackage{subcaption}
\usepackage[sort&compress,round]{natbib}
\usepackage{multirow}   
\usepackage[table]{xcolor}
\usepackage{pifont}
\usepackage{algorithm}
\usepackage{algpseudocode}
\newcommand{\cmark}{\ding{51}}%
\newcommand{\vect}[1]{\boldsymbol{#1}}
\newcommand*\rot{\rotatebox[origin=c]{90}}
\begin{acronym}
\acro{AMC}{Airborne Mobile Camera}
\acro{IMU}{Inertial Measurement unit}
\acro{UAV}{Unmanned Aerial Vehicle}
\acro{MAV}{Micro Aerial Vehicle}
\acro{UAS}{Unmanned Aerial System}
\acro{CCTV}{Closed Circuit Television} 
\acro{PSNR}{Peak Signal to Noise Ratio}
\acro{SSIM}{Structural Similarity Index}
\acro{BSIA}{British Security Industry Association}
\acro{CEN}{European Committee for Standardization}
\acro{AC}{Axis Communication}
\acro{ROI}{Region of Interest}
\acro{VSN}{Visual Sensor Network}
\acro{IR}{Infra Red}
\acro{EO}{Electro Optics}
\acro{UTC}{Coordinated Universal Time}
\acro{GPS}{Global Positioning System}
\acro{IMU}{Inertial Measurement Unit}
\acro{RFID}{Radio Frequency Identification}
\acro{OCR}{Optical Character Recognition}
\acro{PRM}{Proposed Research Module}
\acro{FOV}{Field of View}
\acro{AHGMM}{Adaptive Hopping Gaussian Mixture Model}
\acro{PSF}{Point Spread Function}
\acro{PRNG}{pseudorandom number generator}
\acro{LFW}{Labelled Faces in the Wild}
\acro{3DMM}{3D Morphable Model} 
\acro{AGB}{Adaptive Gaussian Blur}
\acro{SVGB}{Space Variant Gaussian Blur}
\acro{FGB}{Fixed Gaussian Blur}
\acro{CCD}{Charge Coupled Device}
\acro{SLM}{Spatial Light Modulator}
\acro{ROC}{Receiver Operating Curve}
\end{acronym}

\begin{document}

\title{Concealing the identity of faces in oblique images with adaptive hopping Gaussian mixtures
}


\author{Omair Sarwar \and Bernhard Rinner \and Andrea Cavallaro 
}


\institute{O. Sarwar \at
              Institute of Networked and Embedded Systems, Alpen-Adria-Universit\"{a}t Klagenfurt, Austria, and Centre for Intelligent Sensing, Queen Mary University
of London, UK\\
              \email{omair.sarwar@aau.at}           
           \and
           B. Rinner \at
Institute of Networked and Embedded Systems, Alpen-Adria-Universit\"{a}t Klagenfurt, Austria\\
\email{bernhard.rinner@aau.at}
		\and
		A. Cavallaro \at
Centre for Intelligent Sensing, Queen Mary University
of London, UK\\
\email{a.cavallaro@qmul.ac.uk}
}

\date{Received: date / Accepted: date}

\maketitle

\begin{abstract}
Cameras mounted on \acp{MAV} are increasingly used for recreational photography. However, aerial photographs of public places often contain faces of bystanders thus leading to a perceived or actual violation of  privacy. To address this issue, we propose to pseudo-randomly modify the appearance of face regions in the images using a privacy filter that prevents a human or a face recogniser from inferring the identities of people. The filter, which is applied only when the resolution is high enough for a face to be recognisable, adaptively distorts the face appearance as a function of its resolution. Moreover, the proposed filter locally changes its parameters to discourage attacks that use parameter estimation. The filter exploits both global adaptiveness  to reduce distortion and local hopping of the parameters to make their estimation difficult for an attacker. In order to evaluate the efficiency of the proposed approach, we use a state-of-the-art face recognition algorithm and synthetically generated face data with 3D geometric image transformations that mimic faces captured from an \ac{MAV} at different heights and pitch angles. Experimental results show that the proposed filter protects privacy while reducing distortion and exhibits resilience against attacks.
\end{abstract}

\section{Introduction}
\label{sec:introduction}
\acp{MAV} are becoming common platforms for a number of civilian applications such as search and rescue \citep{Waharte2010}, disaster management \citep{Quaritsch2010} and news reporting \citep{Babiceanu2015}. Moreover, individuals use \acp{MAV} equipped with high resolution cameras for  recreational photography and videography in public places during sports activities and social gatherings \citep{Hexoplus2016_url, AirDog2016_url}. Such use in public places raises privacy concerns as bystanders who happen to be within the field of view of the camera are captured as well. The identity of bystanders could be protected by locating and removing (or sufficiently distorting) key image regions, such as faces, using algorithms called privacy filters. However, in order to maintain the aesthetic value of an image, only a minimal distortion of the image content should be allowed. 

A privacy filter for recreational aerial photography should satisfy the following properties: (a) introduce only a minimal distortion; (b) be robust against attacks; and (c) be computationally efficient. {\em Minimal distortion} is necessary to maintain quality of a protected image close to the unprotected one so that the attention of a viewer is not diverted. Therefore blanking out a face \citep{Schiff2007} is not a desirable option. {\em Robustness} is important to avoid privacy violations by various attacks, e.g.~brute-force, na\"{i}ve, parrot and reconstruction attacks \citep{Kundur1996, Boult2005, Newton2005, Dufaux2008, Adam2014, Korshunov2014, Dong2016}. A brute-force attack tries to decipher the protected probe images by an exhaustive search \citep{Boult2005, Dufaux2008}. Other attacks use gallery images in addition to the protected probe images \citep{Newton2005, Adam2014, Korshunov2014, Dong2016}. In a na\"{i}ve attack, the protected probe images are compared against the unprotected gallery images \citep{Newton2005, Adam2014, Korshunov2014}. In a parrot attack, the attacker has knowledge about the privacy filter and can transform the gallery images into the distorted domain \citep{Newton2005}. In a reconstruction attack, the attacker has some knowledge of how to (partially) reconstruct the probe image from the protected to the unprotected domain \citep{Kundur1996}. Examples of reconstruction methods include inverse filtering and super-resolution techniques \citep{Kundur1996, Dong2016}. Finally, {\em computational efficiency} is desirable when the filter operates using the limited computational and battery power of a MAV.

\begin{figure*}[t]
\centering
\includegraphics[width=\textwidth]{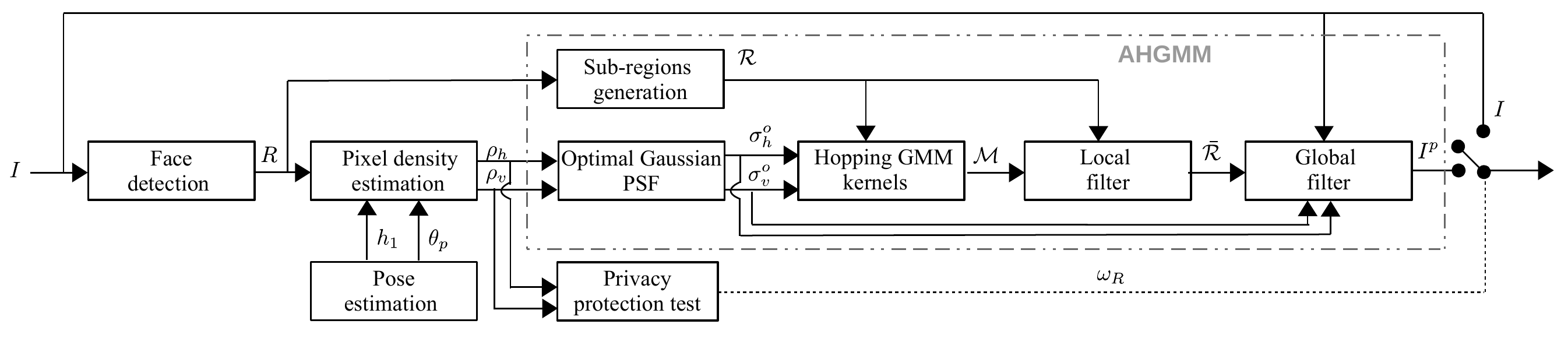}\\
\caption{Block diagram of the proposed Adaptive Hopping Gaussian Mixture Model filter. KEY -- $\rho_h$, $\rho_v$: number of pixels (px) per unit distance (cm) (pixel densities) of a sensitive region $R$; $h_1$, $\theta_P$: altitude and tilt angle of the camera used to calculate the pixel densities; $\omega_R$: control signal generated from the pixel densities to decide when to protect $R$; $\mathcal{R}$: sub-regions  of $R$; $\sigma_o^o$, $\sigma_v^o$: standard deviations for the hopping Gaussian mixture model $\mathcal{M}$ that filters $\mathcal{R}$ to generate the protected sub-regions $\bar{\mathcal{R}}$;  $I^p$: protected image.}
\label{fig:flow_chart}
\end{figure*}

Privacy filters for aerial photography need to face challenges caused by the ego-motion of the camera, changing illumination conditions, and variable face orientation and resolution. Recent frameworks that support facial privacy-preservation in airborne cameras are Generic Data Encryption  \citep{Kim2014}, Unmanned Aircraft Systems-Visual Privacy Guard \citep{Babiceanu2015} and Adaptive Gaussian Blur \citep{sarwar2016}. Generic Data Encryption sends an encrypted face region to a privacy server that Gaussian blurs or mosaics the face and then forwards it to an end-user. Unmanned Aircraft Systems-Visual Privacy Guard \citep{Babiceanu2015} and Adaptive Gaussian Blur \citep{sarwar2016} are aimed instead at on-board implementation with an objective to reduce latency and discourage brute-force attacks on the  server \citep{Kim2014}. Adaptive Gaussian Blur adaptively configures the Gaussian kernel depending upon the face resolution in order to minimise distortion, while Unmanned Aircraft Systems-Visual Privacy Guard blurs faces with a fixed filter. These methods are prone to parrot attacks \citep{Newton2005} on the Gaussian blur.  

In this paper, we present a novel privacy protection filter to be used on-board an \ac{MAV}. The proposed filter distorts a face region with secret parameters to be robust to na\"{i}ve, parrot and reconstruction attacks. The distortion is minimal and adaptive to the resolution of the captured face: we select the smallest Gaussian kernel that reduces the face resolution below a certain threshold. The selected threshold protects the face against the na\"{i}ve attack as well as maintains its resolution at a specified level. To prevent other attacks, we then insert supplementary Gaussian kernels in the selected Gaussian kernel and hop their parameters locally using a \ac{PRNG} so their estimation is difficult from the filtered face image. The block diagram of the proposed filter is shown in Figure~\ref{fig:flow_chart}.

In contrast to airborne photography, an updated work based on the proposed filter is presented in \cite{sarwar2018}, specifically for the airborne videography. The main contributions of this paper are: (1) basic idea of the Gaussian hopping kernels and their details, (2) a large-scale synthetic face image data set emulating faces captured from an \ac{MAV}, and (3) extensive experiments to validate the proposed Gaussian hopping kernels, including the reconstruction attacks.

The paper is organised as follows. Sec.~\ref{sec:literature_review} covers the state-of-the-art in visual privacy protection filters.  Sec.~\ref{sec:problem_formulation} defines the problem. Sec.~\ref{sec:proposed_approach} describes the proposed algorithm, and discusses its computational complexity and security level. Sec.~\ref{sec:data_set} presents our face data set generation and Sec.~\ref{sec:experimental_results} discuss the experimental results. Finally, Sec.~\ref{sec:conclusion} concludes the paper.

\section{Background}
\label{sec:literature_review}
\begin{figure}[t]
\centering
\includegraphics[width=0.5\textwidth]{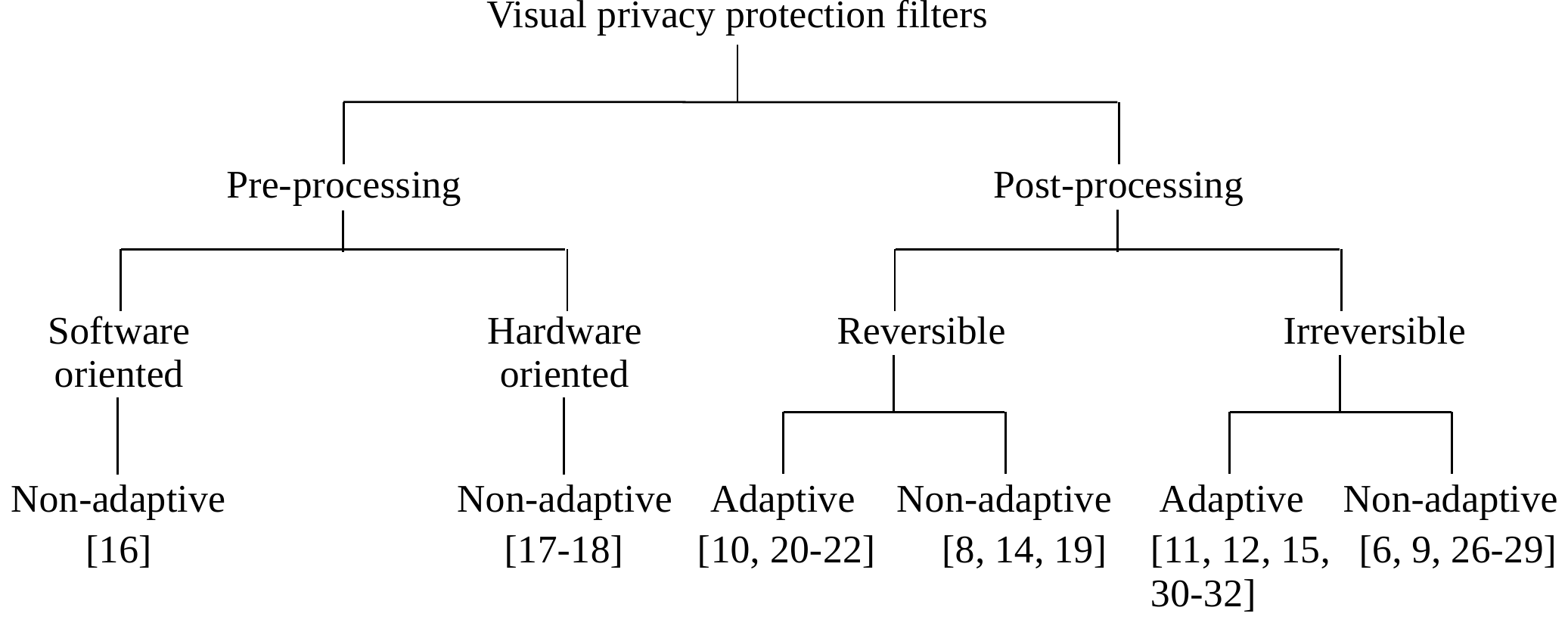}\\
\caption{A taxonomy of visual privacy protection filters.}
\label{SoA_taxonomy}
\end{figure}

Visual privacy protection filters can be applied as pre-processing or post-processing (Fig. \ref{SoA_taxonomy}). 
{\em Pre-processing privacy filters} are irreversible and operate during image acquisition to prevent a camera from capturing sensitive regions. These filters disable the software or hardware of the camera or notify about photography prohibition \citep{SafeHaven2003}. Hardware based filters prevent the camera from taking images for example by bursting back an intense light for flash photography \citep{EagleEye1997, Zhu2017} or by detecting human faces using an infrared sensor and then obfuscating using a spatial light modulator sensor placed in front of the \ac{CCD} sensor \citep{Zhang2014}. 

{\em Post-processing privacy filters}  protect sensitive regions after image acquisition and can be reversible or irreversible.  Reversible filters  conceal sensitive regions using a private key, which can later be used to recover the original sensitive region. Irreversible filters deform the features of a sensitive region permanently. Both reversible and irreversible filters can be non-adaptive or adaptive. 

{\em Reversible non-adaptive filters} are based on generic encryption \citep{Boult2005, Chattopadhyay2007, Rahman2010, Winkler2011, Zhang2018}.
{\em Reversible adaptive filters} include scrambling \citep{Dufaux2006, Dufaux2008, Baaziz2007, Sohn2011, Ruchaud2017}, warping \citep{Korshunov2013} and morphing \citep{Korshunov2013b}.  While reversible adaptive filters are robust against a parrot attack, their protected faces can be compromised by spatial-domain \citep{Jiang2016, Jiang2016b} or frequency-domain attacks \citep{Hatem2015}. 

{\em Irreversible non-adaptive filters}  blank out \citep{Schiff2007, Koelle2018}  or replace a face with a de-identified representation \citep{Newton2005}. For example, to maintain k-anonymity, the algorithm "k-Same" \citep{Newton2005} replaces k faces with their average face. Variants of this algorithm use additional specialised detectors to then preserve attributes such as facial expressions, pose, gender, race, age \citep{Gross2006, Du2014, Lin2012, Letournel2015, Meden2018}.   
Irreversible non-adaptive filters are robust to parrot attacks. {\em Irreversible adaptive filters} lower the  resolution of a sensitive region so that humans or algorithms cannot recognise the identity. Examples include pixelation \citep{Chinomi2008}, Gaussian blur \citep{Wickramasuriya2004} and cartooning \citep{Adam2014}. The kernel size of the privacy filters can be manually selected \citep{Korshunov2014, Adam2014} or the centre kernel size is manually selected and then the Space Variant Gaussian
Blur (SVBG) filter \citep{Saini2012} automatically decreases the kernel size from the centre to the boundary of the detected face. AGB \citep{sarwar2016} exploits the different horizontal and vertical resolutions that are typical in aerial photography, and automatically adapts an anisotropic kernel based on the resolution of the detected face. However, irreversible adaptive filters are vulnerable to parrot attacks. 

\begin{table*}[t]
\centering
\caption{Post-processing privacy filters. KEY -- DCT-S: Discrete Cosine Transform Scrambling \citep{Dufaux2008}; PICO: Privacy through Invertible Cryptographic Obscuration \citep{Boult2005}; GARP:  Gender, Age and Race Preservation \citep{Du2014}; UAS-VPG: Unmanned Aircraft Systems-Visual Privacy Guard \citep{Babiceanu2015}; Cartooning \citep{Adam2014}; SVGB: Space Variant Gaussian Blur \citep{Saini2012}; ODBVP: Optimal Distortion-Based Visual Privacy \citep{Korshunov2014}; AGB: Adaptive Gaussian Blur \citep{sarwar2016}. Adaptive control modulates the strength of a privacy filter.}
\label{tab:state_of_the_Art_critical}
\resizebox{0.9\textwidth}{!}{%
\begin{tabular}{|l|l|l|c|c|c|c|c|c|c|c|c|c|c|}
\cline{4-12}
\multicolumn{3}{c|}{}& \rotatebox{90}{DCT-S} & \rotatebox{90}{PICO} & \rotatebox{90}{GARP}& \rotatebox{90}{UAS-VPG} & \rotatebox{90}{Cartooning } & \rotatebox{90}{SVGB} & \rotatebox{90}{ODBVP} & \rotatebox{90}{AGB} & \rotatebox{90}{Proposed} \\
\hline
\multirow{4}{*}{Distortion}&\multirow{2}{*}{adaptive control}  &image based   &\cmark&&&  & \cmark  & \cmark & \cmark &  &  \\
\cline{3-12}
& & navigation sensors &&&&  &  &  &  & \cmark & \cmark \\
\cline{2-12}
&\multirow{2}{*}{2D kernel}  & isotropic&&&&   &  \cmark &  \cmark &  \cmark &  &  \\
\cline{3-12}
&& anisotropic&&&&  &  &  &  & \cmark & \cmark \\
\hline
\multirow{4}{*}{Robustness} & \multicolumn{2}{l|}{to brute-force attack}& &&  \cmark & \cmark &  \cmark & \cmark & \cmark & \cmark & \cmark \\
\cline{2-12}
 & \multicolumn{2}{l|}{to na\"{i}ve attack}& \cmark& \cmark & \cmark &  \cmark&  \cmark&  \cmark&  \cmark&  \cmark& \cmark \\
\cline{2-12}
 & \multicolumn{2}{l|}{to inverse filter attack}& &  & \cmark &  &  \cmark &  &  &  & \cmark \\
\cline{2-12}
 & \multicolumn{2}{l|}{to super-resolution attack}& \cmark&  \cmark & \cmark &  &  &  &  &  & \cmark\\ 
\cline{2-12}
 & \multirow{2}{*}{to parrot attack}&with detectors&& & \cmark &  &  &  &  &  & \\
 \cline{3-12}
 && without detectors&\cmark&\cmark &  &  &  &  &  &  & \cmark \\ 
\hline
\multicolumn{3}{|l|}{Computational simplicity}  &  &&  & \cmark  &  & \cmark & \cmark & \cmark &  \\
\hline
\end{tabular}}

\end{table*}

As a summary, Table \ref{tab:state_of_the_Art_critical} compares representative filters for the following categories: reversible \& adaptive \citep{Dufaux2008}, reversible \&  non-adaptive \citep{Boult2005}, and irreversible \& non-adaptive  filters \citep{Du2014}. The rest \citep{Babiceanu2015, Adam2014, Saini2012, Korshunov2014,sarwar2016} and proposed are irreversible \& adaptive filters.

\section{Problem Definition}
\label{sec:problem_formulation} 

Let the set $\mathcal{D} = \{\mathcal{R}_k\}_{k=1}^K$ contain face data of $K$ subjects, where $k$ represents the identity (labels). Let each subject $k$ appear in at most $Z$ images, i.e. $\mathcal{R}_k = \{R_i| i \leq Z\}$. Let $\mathcal{R_G}, \mathcal{R_P} \subset \mathcal{D}$ be the gallery and probe  sets, respectively.  Usually $|\mathcal{R_G}| > |\mathcal{R_P}|$, where $|.|$ is the cardinality of a set, and $\mathcal{R_G} \cap \mathcal{R_P}=\emptyset$. 

Let a privacy filter $F_{\Omega_j}: \mathcal{R_P} \rightarrow \mathcal{\bar{R}_P}$ distort image features in order to reduce the  probability $P$ for an attacker to correctly predict  labels. This operation produces a protected probe set $\mathcal{\bar{R}_P}$, whose distortion depends on $\Omega_j$, where $j\in\{h,v\}$ indicates the horizontal and vertical direction in an image. 
Let the {distortion} generated by $F_{\Omega_j}$ be measured by the \ac{PSNR}:
\begin{equation}
PSNR = 20 \log_{10} \frac{R_{max}}{\sqrt{MSE}},
\label{eq:psnr}
\end{equation}
where $R_{max}$ is the dynamic range of the pixel values.  The mean square error, MSE, between the pixel intensities of an unprotected, $R \in \mathcal{R_P}$, and protected, $\bar{R} \in \mathcal{\bar{R}_P}$, face is
\begin{equation}
MSE = \frac{1}{|\mathcal{R_P}|WH} \sum_{r=1}^{|\mathcal{R_P}|} \sum_{w=1}^{W} \sum_{h=1}^{H} ||R(w,h) - \bar{R}(w,h)||^2_r,
\label{eq:psnr}
\end{equation}
where $W$ and $H$ are width and height of $R$, respectively. 

We express the {privacy level} of a face region as the accuracy $\eta$ of a face recogniser  \citep{Adam2014, Korshunov2014}.
The value of $\eta$ is the commutative rank-n in face identification or the Equal Error rate (EER) in face verification. We consider in this paper face verification, thus
\begin{equation}
\eta = \frac{TP+TN}{|\mathcal{R_P}|},
\label{eq:accuracy}
\end{equation}
where $TP$ and $TN$ are true positives and true negatives, respectively. Our target is to force a face recogniser of an attacker to have the accuracy of random classifier, which for face verification is $\epsilon = 0.5$. 

We therefore aim to design $F_{\Omega_j}$ that irreversibly but minimally distorts the appearance of $R$ so that the identity is not recognisable with a probability higher than a random guess. If $E(w,h)=F_{\Omega_j}(R(w,h))-R(w,h)$, the ideal distortion parameter, $ \Omega_j^o $, should be derived as:
%
\small
\begin{equation}
\Omega_j^o=\underset{\Omega_j}{\arg\min}\left(\frac{1}{WH}\sum_{w=1}^{W} \sum_{h=1}^{H}E(w,h) + \\(P(F_{\Omega_j}(R)|\mathcal{B})-0.5) \right),
\end{equation}
\normalsize
where $\mathcal{B} \in \{\mathcal{R_G}, \mathcal{\bar{R}_G}, \mathcal{\hat{R}_G}\}$. The  first term aims to introduce  a minimal distortion, whereas the second term leads the classification results to be equivalent to that of a random classifier, irrespective of whether the filtered or reconstructed face is compared against the unprotected, filtered or reconstructed gallery data sets. The second term objective is dependent upon the recognition capability of a face recogniser and is heuristically addressed for a given face recogniser \citep{Chriskos2016, Adam2017, Pittaluga2017}.

The content of $R$ should be protected against na\"{i}ve-T, parrot-T and reconstruction attacks. Let an attacker have access to $\mathcal{B} \in \{\mathcal{R_G}, \mathcal{\bar{R}_G}, \mathcal{\hat{R}_G}\}$, where $\mathcal{\bar{R}_G}$ is the filtered gallery data set and $\mathcal{\hat{R}_G}$ is the filtered and reconstructed gallery data set. An attacker can modify $\mathcal{\bar{R}_P}$, $\mathcal{R_G}$, or both, to correctly predict $\tilde{K}$ of $\mathcal{\bar{R}_P}$. 
In a  {na\"{i}ve attack} (here referred to as na\"{i}ve-T attack), a privacy filter is applied on $\mathcal{R_P}$ to generate a protected probe data set $\mathcal{\bar{R}_P}$, while the unaltered $\mathcal{R_G}$ is used for training \citep{Newton2005}. A {parrot attack} (here referred to as parrot-T attack), learns the privacy filter type and its parameters $\Omega_j$ (e.g.~Gaussian blur of certain standard deviation used to generate $\mathcal{\bar{R}_P}$). Then, the learned filter is applied on $\mathcal{R_G}$ to generate a privacy protected gallery data set $\mathcal{\bar{R}_G}$. Finally, $\mathcal{\bar{R}_G}$ and $\mathcal{\bar{R}_P}$ are used for training and testing, respectively \citep{Newton2005}. In a reconstruction attack, the discriminating features of $\mathcal{\bar{R}_P}$ are first restored (e.g.~using an inverse filter or a super-resolution algorithm) to generate a reconstructed probe data set $\mathcal{\hat{R}_P}$ and then compared against $\mathcal{R_G}$ or a reconstructed gallery data set $\mathcal{\hat{R}_G}$. An inverse filter first estimates the parameters of a privacy filter using $\mathcal{\bar{R}_P}$ and then performs an inverse operation to reconstruct the original faces \citep{Kundur1996}. Similarly, a super-resolution algorithm first learns embeddings between the high-resolution and their corresponding low-resolution faces and then reconstructs the high-resolution faces for $\mathcal{\bar{R}_P}$ \citep{Dong2016}. 
\section{Proposed Approach}
\label{sec:proposed_approach}
\begin{figure}[t]
\centering
\includegraphics[width=\columnwidth]{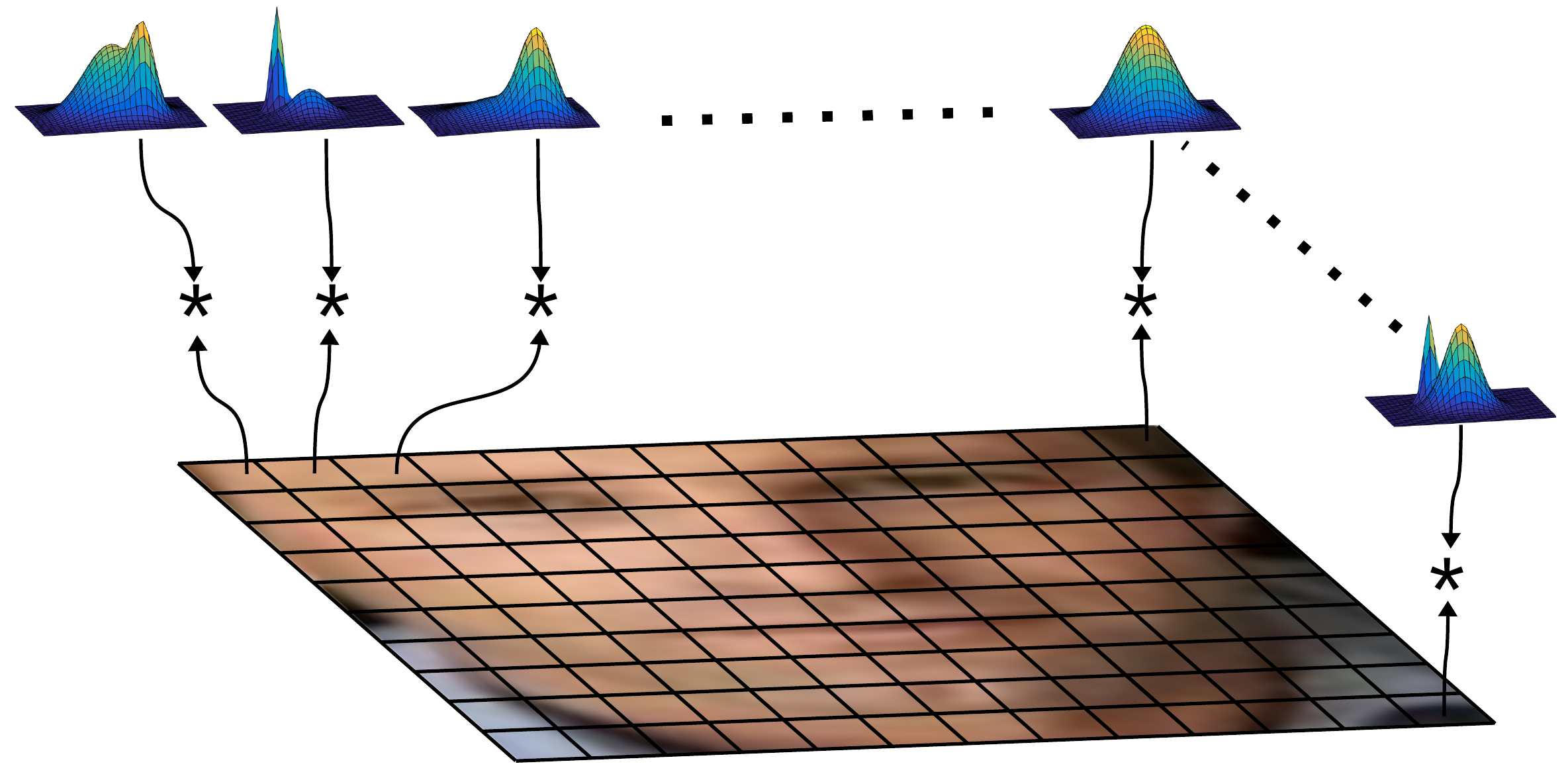}\\
\label{fig:optimal_kernel}
\caption{Visualisation of local filtering in AHGMM. The face region $R$ is divided into $N$ sub-regions and each sub-region $R_{n}$ is convolved ($*$) with a hopping Gaussian mixture model kernel $M_n$, which is made by an optimal Gaussian function and one (or more) supplementary Gaussian function added inside the optimal Gaussian function. While convolving with each sub-region of the face, the optimal and the supplementary Gaussian functions change their parameters, i.e.~mean and standard deviation, which consequently changes the shape of the Gaussian mixture model based kernel.}
\label{fig:AHGMM}
\end{figure}

In order to minimally distort $R$ as well as to achieve robustness against brute-force, na\"{i}ve-T, parrot-T and reconstruction attacks, we propose the \ac{AHGMM} algorithm. The AHGMM consists of a globally estimated optimal Gaussian \ac{PSF} and supplementary Gaussian \acp{PSF} added inside the optimal Gaussian \ac{PSF}. For a single supplementary Gaussian \ac{PSF} inside an optimal Gaussian \ac{PSF}, the AHGMM is illustrated in Fig. \ref{fig:AHGMM}, while the pseudo-code is given in Algorithm \ref{alg:ahgmm}. A list of important notations is presented in Appendix \ref{ap:notations}. 

Figure \ref{fig:flow_chart} shows the processing diagram of our proposed framework and the different blocks of it are explained in more details in the following subsections.

\subsection{Pixel Density Estimation}
\label{subsec:pixel_density_estimation}

\begin{figure}[t]
\begin{center}
\includegraphics[width=0.95\linewidth]{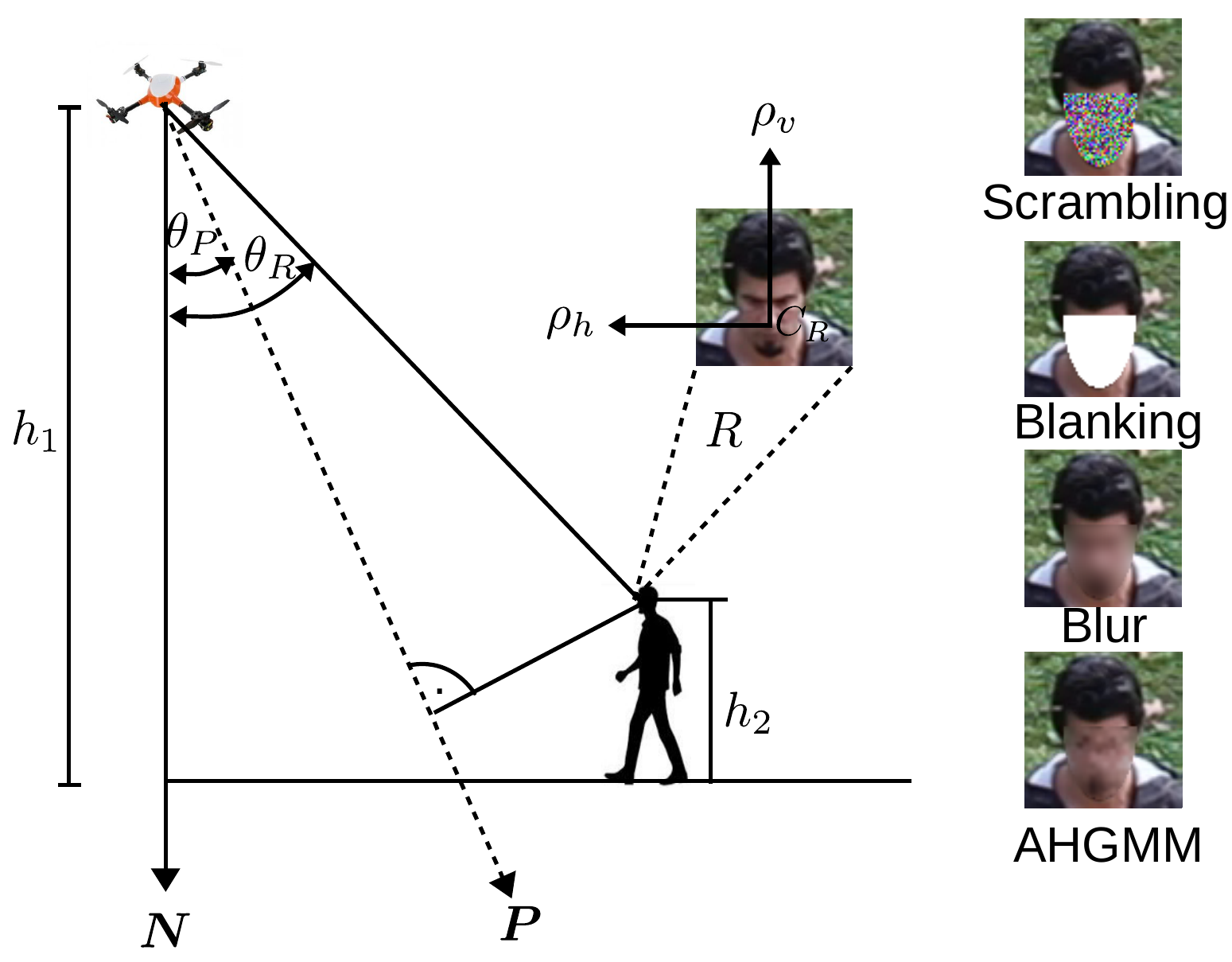}
\end{center}
\caption{Capturing an image with an airborne camera at height $h_1$. The principal axis $\vect{P}$ of the camera is tilted by $\theta_P$ from the nadir direction $\vect{N}$. The face region $R$, at height $h_2$ above the ground, is viewed at an angle $\theta_R$. The variables $\rho_h$ and $\rho_v$ represent the horizontal and vertical pixel density of $R$ at its centre $C_R$ in the captured image. Four sample images show a scrambled, blanked, Gaussian blurred and AHGMM filtered image, which is captured at $\theta_P=\theta_R=50^\circ$.}
\label{fig:face_detection_and_recognition_in_AVC}
\end{figure}

Let an \ac{MAV} capture an image $I$ while flying at an altitude of $h_1$ meters. Let the principal axis $\vect{P}$ of its on-board camera be tilted by $\theta_P$ from the nadir direction $\vect{N}$ (see~Figure~\ref{fig:face_detection_and_recognition_in_AVC}). We assume that height $h_1$ and tilt angle $\theta_P$ of the camera can be estimated. 

A value of $\theta_P \neq 0$ generates an oblique image. Let $h_2$ be the height of the face above ground\footnote{While each image $I$ could contain $L$ faces, for simplicity we consider in this paper only the case $L=1$.}. We represent the face region in the image as $R \in \mathcal{R}_k \subset \mathcal{D}$, which is viewed at an angle $\theta_R$. 

Let $\rho_{j}$ represent the pixel density (px/cm) around the centre $C_R$ of $R$. If $p_h$ and $p_v$ represent the physical dimensions of a pixel in the horizontal and vertical direction, respectively and $f$ is the focal length of the camera, the horizontal density $\rho_h$ for a pixel around $C_R$ \citep{sarwar2016} is

\begin{equation}
\rho_h = \dfrac{f cos(\theta_R)}{p_h (h_1-h_2)},
\label{eq:Dh}
\end{equation} 
and the vertical density $\rho_v$, by exploiting the small angle approximation for a single pixel of the image sensor \citep{sarwar2016}, is
\begin{equation}
\rho_v \approx \dfrac{f cos(\theta_R) sin(\theta_R)}{p_v (h_1-h_2)}.
\label{eq:Dv}
\end{equation} 
Let $\omega_R\in\{0,1\}$ define whether $R$ is naturally protected ($\omega_R=0$) because of a low horizontal and vertical density, or not ($\omega_R=1$) \citep{sarwar2016}: 
\begin{equation}
\omega_R = \begin{cases} 1 \quad \quad if \quad \rho_h > \rho_h^o \quad \text{and} \quad \rho_v > \rho_v^o \\ 
0 \quad \quad \text{otherwise} \end{cases}
\label{eq:V}
\end{equation}
where $\rho_h^o$ and $\rho_v^o$ are pixel densities at which a state-of-the-art machine algorithm starts recognising human faces, and simply called thresholds. If $\omega_R=0$, then the original frame $I$ can be transmitted without any modifications. Otherwise, $R$ should be protected by a privacy filter to reduce its pixel densities below $\rho_h^o$ and $\rho_v^o$. When $R$ is not inherently protected, we assume that the corresponding bounding box is given.
%
\begin{algorithm}[t]
\small
\caption{AHGMM}
\label{alg:ahgmm}
\hspace*{\algorithmicindent} \textbf{Input:}  $I$ \emph{unprotected image}\\
\hspace*{\algorithmicindent} \hspace{0.87cm}  $R$ \emph{detected face region}\\   
\hspace*{\algorithmicindent} \hspace{0.87cm}  $\rho_j$ \emph{pixel density}, where $j\in\{h,v\}$\\
\hspace*{\algorithmicindent} \textbf{Output:} $I^p$ \emph{protected image}
\begin{algorithmic}[1]
\Procedure{FilterAHGMM($I, R, \rho_h, \rho_v$)}{}
\For{\texttt{$j=h:v$}}
    \State \texttt{$\mu_{j}^o \gets 0$}
    \State \texttt{$\sigma_{j}^o \gets \frac{3 \rho_j}{\pi \rho_j^o}$}
\EndFor
\State \texttt{$\mathcal{R} \gets$ $N$ sub-regions of $R$}

\For{\texttt{$n=1:N$}}
	\For{\texttt{$m=0:M$}}
		\For{\texttt{$j=h:v$}}

			\If {$m=0$}
		    		\State $\mu_{jm} \gets \pm  \alpha_{jm}  \sigma_{j}^o$
		    		\State $\sigma_{jm} \gets (1 \pm \beta_{jm})  \sigma_{j}^o$
			\Else
	        		\State $\mu_{jm} \gets \pm  \alpha_{jm}  \sigma_{j}^o \gamma_{jm}$
	        		\State $\sigma_{jm} \gets (1 \pm \beta_{jm})  \sigma_{j}^o \gamma_{jm}$
			\EndIf
    	    	\EndFor
		\State \texttt{$X_n \gets (\mu_{jm}, \sigma_{jm})_n$}
    	    \State \texttt{$G_{nm} \gets$ compute Gaussian functions}
    	    \State \texttt{$\phi_{nm} \gets$ generate weights}
    	\EndFor
    	\State \texttt{${M}_{n} \gets$ Gaussian mixture model }
    	\State $\bar{R}_{n} \gets  R_{n} * M_n$
\EndFor
\State \texttt{$\bar{R} \gets$ apply global filter on $\bar{\mathcal{R}}$}  
\State \texttt{$I^p \gets$ replace $R$ with $\bar{R}$ in $I$}  
\State \Return $I^p$   
\EndProcedure
  \end{algorithmic}
\end{algorithm}

\subsection{Optimal Gaussian PSF}
\label{subsec:optimal_gaussian_psf}

A 2D \ac{PSF} $g(h,v)$, or impulse response, is the output of a filter when the input is a point source. In the discrete domain \citep{Oppenheim1996}, it is given as  
$
g(h,v) = \delta(h,v) * g (h,v)
$,
where $*$ is the convolution operation and 
\begin{equation}
\delta(h,v) = \begin{cases}
    1 & \text{if } h = v = 0,\\
    0              & \text{otherwise}.
\end{cases}
\label{eq:delta}
\end{equation}
%

%
In the case of Gaussian blur, $g(h,v)$ is an approximated Gaussian function of mean $\mu_j=0$ and standard deviation $\sigma_j>0$ \citep{Saini2012, Korshunov2014, sarwar2016}, and thus called a Gaussian \ac{PSF} of parameter $\Omega_j = (\mu_j, \sigma_j)$. More specifically, the parameter $\Omega_{j} \in \{\Omega_{jl}| l \in \mathbb{N}, \Omega_{jl+1} > \Omega_{jl}\}$ controls the distortion strength of $F_{\Omega_j}$ and provides pixel density $\rho_j \in \{\rho_{jl}|l \in \mathbb{N}, \rho_{jl+1} < \rho_{jl}\}$ in $\bar{R}$, respectively. 

As a higher $\Omega_{j}$ results into lower $\rho_j$, we first find the minimum value called optimal parameter $\Omega_{j}^o$ of $\Omega_{j}$ that makes $\rho_j < \rho^o_j$. As a result, $\Omega_{j}^o$ provides the minimum distortion in $\bar{R}$ while making it robust against the na\"{i}ve-T attack (i.e.~$P (\bar{R}| \mathcal{R_G}) \rightarrow \epsilon$). Increasing $\Omega_{j}$ beyond $\Omega_j^o$ increases the distortion without improving the privacy level as the recogniser performance is already at the level of the random classifier. For a face captured from an \ac{MAV} with pixel densities $\rho_j$, we calculate $\Omega_j^o = (\mu_{j}^o ,\sigma_{j}^o)$ of an optimal Gaussian PSF (lines 2-5 in Algorithm \ref{alg:ahgmm}), where $\mu_{j}^o=0$ like in traditional Gaussian blur \citep{Saini2012, Korshunov2014, sarwar2016} and $\sigma_{j}^o$ \citep{sarwar2016} is estimated as follows:

A Gaussian \ac{PSF} of standard deviation $\sigma_{j}^o$ in the spatial domain is another Gaussian \ac{PSF} of standard deviation $\acute{\sigma_{j}^o}$ in the frequency domain and both the Gaussian \acp{PSF} are related as
\begin{equation}
\acute{\sigma_{j}^o} = \dfrac{\rho_j}{2\pi\sigma_{j}^o},
\label{eq:sigma_j}
\end{equation} 
where $\acute{\sigma_{j}^o}$ is measured in cycles/cm, $\sigma_{j}^o$ in px and $\rho_{j}$ in px/cm. Let $f_s$ represents the Nyquist frequency of $\rho_j$. Let $f_s^o < f_s$ is the highest spatial frequency component that we want to completely remove using a low pass filter, i.e.~Gaussian blur. In other words, $f_s^o$ is the Nyquist frequency of $\rho_j^o$, i.e.~pixel density after filtering. Both $\rho_j^o$ and $f_s^o$ are related as   
\begin{equation}
\rho_j^o = 2 f_s^o.
\label{eq:bar_rho_j}
\end{equation}
As we are interested in removing frequency components beyond $f_s^o$, we can select $f_s^o =3\acute{\sigma_{j}^o}$ because the amplitude response of a Gaussian \ac{PSF} at three times of its standard deviation is very close to zero and multiplication (convolution in space domain) with such a Gaussian {PSF} will suppress frequencies larger than $f_s^o$. Substituting $f_s^o =3\acute{\sigma_{j}^o}$ in Eq.~\ref{eq:bar_rho_j}, in the resulting relation Eq.~\ref{eq:sigma_j} and finally rearranging gives the optimal standard deviation of Gaussian PSF as 
\begin{equation}
\sigma_{j}^o = \dfrac{3\rho_j}{\pi\rho_j^o}.
\label{eq:sigma_i}
\end{equation}
\subsection{Hopping GMM Kernels}
\label{subsec:ahgmm}

Filtering $R$ with the optimal Gaussian \ac{PSF} defined by $\Omega_j^o$ would only protect $R$ from a na\"{i}ve-T attack but not from a parrot-T attack and a reconstruction attack. To ensure that the probability of correctly predicting the label of $\bar{R}$ is not increased in case of the parrot-T attack (i.e.~$P(\bar{R}|\mathcal{\bar{R}_G}) \rightarrow \epsilon$) as well as the reconstruction attack (i.e.~$P(\hat{R}|\mathcal{R_G}) \rightarrow \epsilon$ or $P (\hat{R}|\mathcal{\hat{R}_G) \rightarrow \epsilon}$), we secretly modify $\Omega_{j}^o$ to $\bar{\Omega}_j^o$ while generating $\bar{R}$ so that an adversary is unable to accurately reconstruct face region $\hat{R}$, or even generate $\mathcal{\hat{R}_G}$ and $\mathcal{\bar{R}_G}$. For this purpose, we generate a set $\mathcal{R}$ which consists of $N$ sub-regions in such a way that each sub-region covers a small area of $R$:
\begin{equation}
\mathcal{R}=\Big \{ R_{n}| n \in [1,N]\Big\}.
\label{eq:R_hat}
\end{equation}
The size of $R_{n}$ (in pixels) affects the total number of sub-regions $N$ per face region $R$, which could influence its privacy level. Smaller values of $N$ (larger sub-regions) result in a reduced distortion.

After finding $\Omega_j^o = (\mu_j^o$, $\sigma_{j}^o)$ and generating $\mathcal{R}$, we make a hopping mixture of Gaussian for each sub-region, i.e.~we pseudo-randomly change $\Omega_j^o$ to $\bar{\Omega}_j^o$ for each $R_{n}$. Moreover, we select supplementary Gaussian \acp{PSF} inside this optimal Gaussian \ac{PSF} and vary their parameters based on pseudo-random weights (lines 9-17 in Algorithm \ref{alg:ahgmm}). 

Let set $\mathcal{X}$ contains the parameters of the modified optimal and supplementary Gaussian \acp{PSF} for each sub-region, and is represented as
\begin{equation}
\mathcal{X} = \Big\{(\mu_{jm}, \sigma_{jm})_n|n\in[1,N], j\in\{h,v\}, m\in[0,M]\Big\},
\label{eq:X}
\end{equation}
where $M$ is the number of the supplementary Gaussian \acp{PSF}. The element $m=0$ represents the modified optimal Gaussian \ac{PSF} given by 
\begin{equation}
\mu_{j0} = \pm  \alpha_{j0}  \sigma_{j}^o,
\label{eq:mu_j0}
\end{equation}
\begin{equation}
\sigma_{j0} = (1 \pm \beta_{j0})  \sigma_{j}^o,
\label{eq:sigma_j0}
\end{equation}
while the remaining elements (i.e.~$m\in(0,M]$) belong to the supplementary Gaussian \acp{PSF}. These elements are calculated as
\begin{equation}
\mu_{jm} = \pm  \alpha_{jm}  \sigma_{j}^o \gamma_{jm},
\label{eq:mu_jm}
\end{equation}
\begin{equation}
\sigma_{jm} = (1 \pm \beta_{jm})  \sigma_{j}^o \gamma_{jm},
\label{eq:sigma_jm}
\end{equation}
where, $\alpha_{jm} \in [0, 1]$ and $\beta_{jm} \in [0, 1]$ are normalised pseudo-randomly generated numbers and control the local distortion in filtering. The variable $\gamma_{jm} \in (0, 1]$ controls the relative size of the supplementary Gaussian \ac{PSF} w.r.t. the optimal Gaussian \ac{PSF}.

\begin{figure}
\centering
\begin{subfigure}{.29\linewidth}
  \centering
  \includegraphics[width=\textwidth]{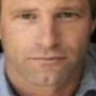}\\
   \caption{}
   \label{fig:Dh_raw}
\end{subfigure}
\begin{subfigure}{.29\linewidth}
  \centering
  \includegraphics[width=\textwidth]{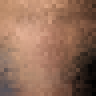}
   \caption{}
   \label{fig:Dv_raw}
\end{subfigure}
\begin{subfigure}{.29\linewidth}
  \centering
  \includegraphics[width=\textwidth]{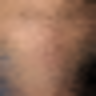}\\
   \caption{}
   \label{fig:V_binary}
\end{subfigure}
\caption{Minimising blocking artefacts of spatially hopping Gaussian functions in AHGMM filter by a convolution with a global kernel. (a) Original image of $96 \times 96$ pixels from the LFW data set, (b) image after local filtering in AHGMM showing blocking artefacts and (c) image after the local filtering followed by the global filtering in AHGMM.}
\label{fig:minimize_blocking_artefacts}
\end{figure} 
After generating the parameters of the Gaussian \acp{PSF}, a set $\mathcal{G}$ representing 2D anisotropic-discretised Gaussian \acp{PSF} corresponding to $\mathcal{X}$ is created as  
\begin{equation}
\mathcal{G} = \Big\{G_{nm}|n\in[1,N], m\in[0,M]\Big\},
\label{eq:G}
\end{equation}
where each $G_{nm}$ is calculated (line 19 in Algorithm \ref{alg:ahgmm}) as \citep{Popkin2010}
\begin{equation}
G_{nm} \approx A_{nm} e^ { - \Big( {\frac{(h- \mu_{hnm})^2}{2\sigma_{hnm}^2 }} + {\frac{(v- \mu_{vnm})^2}{2\sigma_{vnm}^2 }}\Big)},
\label{eq:G_nm}
\end{equation}
%
%
\begin{figure}[!t]
\centering
\resizebox{\columnwidth}{!}{%
\begin{tabular}{cccccccc}
\includegraphics[width=0.125\textwidth]{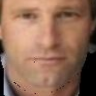} 
&
\includegraphics[width=0.125\textwidth]{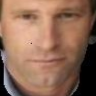} 
&
\includegraphics[width=0.125\textwidth]{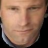} 
&
\includegraphics[width=0.125\textwidth]{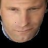} 
&
\includegraphics[width=0.125\textwidth]{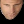} 
&
\includegraphics[width=0.125\textwidth]{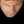}  
&
\includegraphics[width=0.125\textwidth]{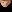} 
&
\includegraphics[width=0.125\textwidth]{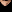} 
\\
(6.21, 4.63) & (6.21, 4.56) & (3.11, 2.17) & (3.11, 2.00) & (1.55, 0.89)&(1.55, 0.74)  & (0.78, 0.29)& (0.78, 0.20)  
\\
 &  & & (a) & & &&
\\
\includegraphics[width=0.125\textwidth]{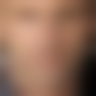} 
&
\includegraphics[width=0.125\textwidth]{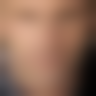} 
&
\includegraphics[width=0.125\textwidth]{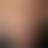} 
&
\includegraphics[width=0.125\textwidth]{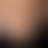} 
&
\includegraphics[width=0.125\textwidth]{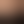} 
&
\includegraphics[width=0.125\textwidth]{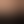}  
&
\includegraphics[width=0.125\textwidth]{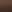} 
&
\includegraphics[width=0.125\textwidth]{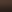} 
\\
\includegraphics[width=0.125\textwidth]{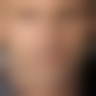} 
&
\includegraphics[width=0.125\textwidth]{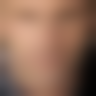} 
&
\includegraphics[width=0.125\textwidth]{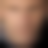} 
&
\includegraphics[width=0.125\textwidth]{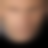} 
&
\includegraphics[width=0.125\textwidth]{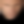} 
&
\includegraphics[width=0.125\textwidth]{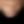}  
&
\includegraphics[width=0.125\textwidth]{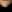} 
&
\includegraphics[width=0.125\textwidth]{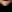} 
\\  
\includegraphics[width=0.125\textwidth]{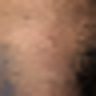} 
&
\includegraphics[width=0.125\textwidth]{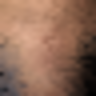} 
&
\includegraphics[width=0.125\textwidth]{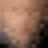} 
&
\includegraphics[width=0.125\textwidth]{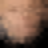} 
&
\includegraphics[width=0.125\textwidth]{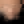} 
&
\includegraphics[width=0.125\textwidth]{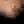}  
&
\includegraphics[width=0.125\textwidth]{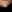} 
&
\includegraphics[width=0.125\textwidth]{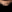} 
\\  
 \multicolumn{8}{c}{(b) $\rho_h^0 = \rho_v^0 = 0.7$ px/cm} 
 \\
\includegraphics[width=0.125\textwidth]{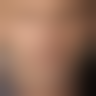} 
&
\includegraphics[width=0.125\textwidth]{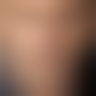} 
&
\includegraphics[width=0.125\textwidth]{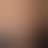} 
&
\includegraphics[width=0.125\textwidth]{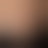} 
&
\includegraphics[width=0.125\textwidth]{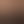} 
&
\includegraphics[width=0.125\textwidth]{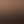}  
&
\includegraphics[width=0.125\textwidth]{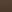} 
&
\includegraphics[width=0.125\textwidth]{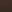} 
\\
\includegraphics[width=0.125\textwidth]{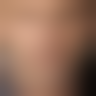} 
&
\includegraphics[width=0.125\textwidth]{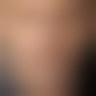} 
&
\includegraphics[width=0.125\textwidth]{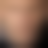} 
&
\includegraphics[width=0.125\textwidth]{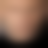} 
&
\includegraphics[width=0.125\textwidth]{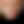} 
&
\includegraphics[width=0.125\textwidth]{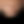}  
&
\includegraphics[width=0.125\textwidth]{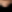} 
&
\includegraphics[width=0.125\textwidth]{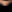} 
\\  
\includegraphics[width=0.125\textwidth]{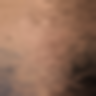} 
&
\includegraphics[width=0.125\textwidth]{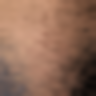} 
&
\includegraphics[width=0.125\textwidth]{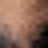} 
&
\includegraphics[width=0.125\textwidth]{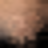} 
&
\includegraphics[width=0.125\textwidth]{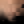} 
&
\includegraphics[width=0.125\textwidth]{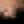}  
&
\includegraphics[width=0.125\textwidth]{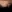} 
&
\includegraphics[width=0.125\textwidth]{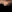} 
\\  
 \multicolumn{8}{c}{(c) $\rho_h^0 = \rho_v^0 = 0.5$ px/cm}
\\
\includegraphics[width=0.125\textwidth]{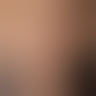} 
&
\includegraphics[width=0.125\textwidth]{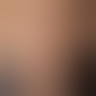} 
&
\includegraphics[width=0.125\textwidth]{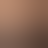} 
&
\includegraphics[width=0.125\textwidth]{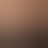} 
&
\includegraphics[width=0.125\textwidth]{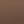} 
&
\includegraphics[width=0.125\textwidth]{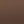}  
&
\includegraphics[width=0.125\textwidth]{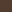} 
&
\includegraphics[width=0.125\textwidth]{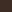} 
\\
\includegraphics[width=0.125\textwidth]{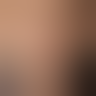} 
&
\includegraphics[width=0.125\textwidth]{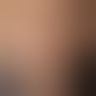} 
&
\includegraphics[width=0.125\textwidth]{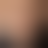} 
&
\includegraphics[width=0.125\textwidth]{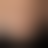} 
&
\includegraphics[width=0.125\textwidth]{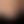} 
&
\includegraphics[width=0.125\textwidth]{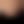}  
&
\includegraphics[width=0.125\textwidth]{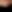} 
&
\includegraphics[width=0.125\textwidth]{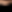} 
\\  
\includegraphics[width=0.125\textwidth]{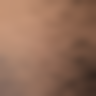} 
&
\includegraphics[width=0.125\textwidth]{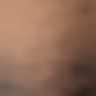} 
&
\includegraphics[width=0.125\textwidth]{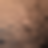} 
&
\includegraphics[width=0.125\textwidth]{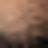} 
&
\includegraphics[width=0.125\textwidth]{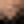} 
&
\includegraphics[width=0.125\textwidth]{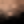}  
&
\includegraphics[width=0.125\textwidth]{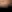} 
&
\includegraphics[width=0.125\textwidth]{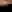} 
\\  
 \multicolumn{8}{c}{(d) $\rho_h^0 = \rho_v^0 = 0.3$ px/cm}
\\
\end{tabular}
}
\caption{Visual comparison between fixed Gaussian blur (FGB), AGB \citep{sarwar2016} and AHGMM on the multi-resolution synthetically generated face data set. (a) Original images with pixel densities decreasing from left to right due different height and pitch angle. (.,.) indicates the horizontal and vertical pixel density in px/cm, respectively. (b-d) For various thresholds ($\rho_h^o$, $\rho_v^o$), results of FGB (first row), AGB (second row) and AHGMM filter (third row). For each threshold, FGB is selected w.r.t. the highest pixel density image in the data set. FGB does not adapt its parameters and therefore results into almost blanking out the image with smaller pixel density. In contrast, both AGB and AHGMM maintain high smoothness by varying their parameters depending upon the pixel densities of an image. Comparatively, AGB produces smoother images, while AHGMM filter creates blocking artefacts due to spatial switching of its parameters.}
\label{fig:visual_comparison}
\end{figure}
%
%
\begin{figure*}[t]
\resizebox{\textwidth}{!}{%
\begin{subfigure}{0.5\textwidth}
\resizebox{\textwidth}{!}{
\begin{tabular}{cccccccc}
\includegraphics[width=0.125\textwidth]{raw_samples_Aaron_Eckhart_0001_96_00.png} 
&
\includegraphics[width=0.125\textwidth]{raw_samples_Aaron_Eckhart_0001_96_10.png} 
&
\includegraphics[width=0.125\textwidth]{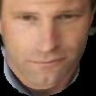} 
&
\includegraphics[width=0.125\textwidth]{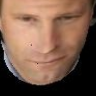} 
&
\includegraphics[width=0.125\textwidth]{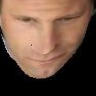} 
&
\includegraphics[width=0.125\textwidth]{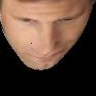}  
&
\includegraphics[width=0.125\textwidth]{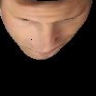} 
&
\includegraphics[width=0.125\textwidth]{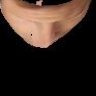} 
\\  
\includegraphics[width=0.125\textwidth]{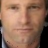} 
&
\includegraphics[width=0.125\textwidth]{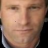} 
&
\includegraphics[width=0.125\textwidth]{raw_samples_Aaron_Eckhart_0001_48_20.png}  
&
\includegraphics[width=0.125\textwidth]{raw_samples_Aaron_Eckhart_0001_48_30.png} 
&
\includegraphics[width=0.125\textwidth]{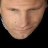} 
&
\includegraphics[width=0.125\textwidth]{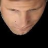} 
&
\includegraphics[width=0.125\textwidth]{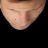}  
&
\includegraphics[width=0.125\textwidth]{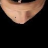}
\\ 
\includegraphics[width=0.125\textwidth]{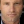} 
&
\includegraphics[width=0.125\textwidth]{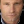} 
&
\includegraphics[width=0.125\textwidth]{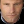} 
&
\includegraphics[width=0.125\textwidth]{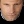}  
&
\includegraphics[width=0.125\textwidth]{raw_samples_Aaron_Eckhart_0001_24_40.png} 
&
\includegraphics[width=0.125\textwidth]{raw_samples_Aaron_Eckhart_0001_24_50.png} 
&
\includegraphics[width=0.125\textwidth]{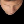} 
&
\includegraphics[width=0.125\textwidth]{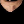}  
\\  
\includegraphics[width=0.125\textwidth]{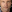} 
&
\includegraphics[width=0.125\textwidth]{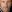} 
&
\includegraphics[width=0.125\textwidth]{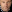}  
&
\includegraphics[width=0.125\textwidth]{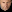} 
&
\includegraphics[width=0.125\textwidth]{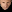} 
&
\includegraphics[width=0.125\textwidth]{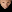} 
&
\includegraphics[width=0.125\textwidth]{raw_samples_Aaron_Eckhart_0001_12_60.png}  
&
\includegraphics[width=0.125\textwidth]{raw_samples_Aaron_Eckhart_0001_12_70.png} 
\\
\includegraphics[width=0.125\textwidth]{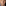} 
&
\includegraphics[width=0.125\textwidth]{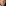} 
&
\includegraphics[width=0.125\textwidth]{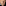}  
&
\includegraphics[width=0.125\textwidth]{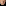} 
&
\includegraphics[width=0.125\textwidth]{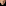} 
&
\includegraphics[width=0.125\textwidth]{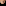} 
&
\includegraphics[width=0.125\textwidth]{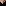}  
&
\includegraphics[width=0.125\textwidth]{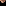} 
\\
\end{tabular}
}
\caption{}
\end{subfigure}
\begin{subfigure}{0.5\textwidth}
\resizebox{\textwidth}{!}{
\begin{tabular}{cccccccc}
\includegraphics[width=0.125\textwidth]{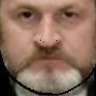} 
&
\includegraphics[width=0.125\textwidth]{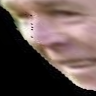} 
&
\includegraphics[width=0.125\textwidth]{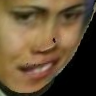} 
&
\includegraphics[width=0.125\textwidth]{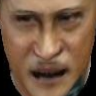} 
&
\includegraphics[width=0.125\textwidth]{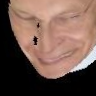} 
&
\includegraphics[width=0.125\textwidth]{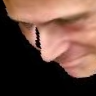}  
&
\includegraphics[width=0.125\textwidth]{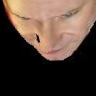} 
&
\includegraphics[width=0.125\textwidth]{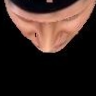} 
\\  
\includegraphics[width=0.125\textwidth]{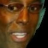} 
&
\includegraphics[width=0.125\textwidth]{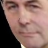} 
&
\includegraphics[width=0.125\textwidth]{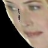}  
&
\includegraphics[width=0.125\textwidth]{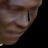} 
&
\includegraphics[width=0.125\textwidth]{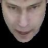} 
&
\includegraphics[width=0.125\textwidth]{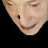} 
&
\includegraphics[width=0.125\textwidth]{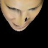}  
&
\includegraphics[width=0.125\textwidth]{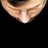}
\\
\includegraphics[width=0.125\textwidth]{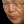} 
&
\includegraphics[width=0.125\textwidth]{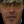} 
&
\includegraphics[width=0.125\textwidth]{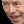} 
&
\includegraphics[width=0.125\textwidth]{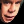}  
&
\includegraphics[width=0.125\textwidth]{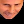} 
&
\includegraphics[width=0.125\textwidth]{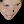} 
&
\includegraphics[width=0.125\textwidth]{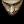} 
&
\includegraphics[width=0.125\textwidth]{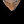}  
\\  
\includegraphics[width=0.125\textwidth]{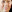} 
&
\includegraphics[width=0.125\textwidth]{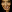} 
&
\includegraphics[width=0.125\textwidth]{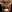}  
&
\includegraphics[width=0.125\textwidth]{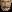} 
&
\includegraphics[width=0.125\textwidth]{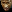} 
&
\includegraphics[width=0.125\textwidth]{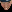} 
&
\includegraphics[width=0.125\textwidth]{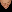}  
&
\includegraphics[width=0.125\textwidth]{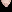} 
\\
\includegraphics[width=0.125\textwidth]{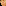} 
&
\includegraphics[width=0.125\textwidth]{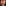} 
&
\includegraphics[width=0.125\textwidth]{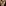}  
&
\includegraphics[width=0.125\textwidth]{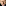} 
&
\includegraphics[width=0.125\textwidth]{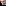} 
&
\includegraphics[width=0.125\textwidth]{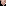} 
&
\includegraphics[width=0.125\textwidth]{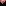}  
&
\includegraphics[width=0.125\textwidth]{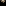} 
\\
\end{tabular}
}
\caption{}
\end{subfigure}
}
\caption{Sample images belonging to (a) a single subject and (b) multiple subjects from our synthetically generated airborne data set based on the LFW data set \citep{LFWTech2007}. In each row, the pitch angle varies from $0^\circ$ to $70^\circ$ in $10^\circ$ steps from left to right, while the image resolution remains constant, i.e.~first row: $96 \times 96$ pixels, second row: $48 \times 48$ pixels, third row: $24 \times 24$ pixels, fourth row: $12 \times 12$ pixels and fifth row: $6 \times 6$ pixels.}
\label{fig:raw_images}
\end{figure*}
%
where
\begin{equation}
A_{nm} = {1}\bigg / {\sum_{(h,v) \in {d}}^{} e^ { - \Big( {\frac{(h- \mu_{hnm})^2}{2(\sigma_{hnm})^2 }} + {\frac{(v- \mu_{vnm})^2}{2(\sigma_{vnm})^2 }}\Big)}},
\label{eq:A_nm}
\end{equation}
and
\begin{equation}
\scalebox{0.95}[1]{${d} = \Big\{(h,v) \in \mathbb{Z}^2: \Big\lceil \frac{-\psi_h}{2} \Big\rceil \leq h \leq \Big\lceil \frac{\psi_h}{2} \Big\rceil, \Big\lceil \frac{-\psi_v}{2} \Big\rceil \leq v \leq \Big\lceil \frac{\psi_v}{2} \Big\rceil  \Big\}$},
\end{equation}
with $\psi_{j} = 2 \Big\lceil 3 \sigma_{j} \Big\rceil + 1$. In order to develop a mixture model from the $M$ discretised Gaussian \acp{PSF} of each sub-region, a set of weights $\mathcal{\phi}$ is required. We again utilise a \ac{PRNG} to generate $\mathcal{\phi}$ such that
\begin{equation}
\mathcal{\phi}=\Big\{\phi_{nm}|n\in[1,N], m\in[0,M], \sum_{m=0}^{M}\phi_{nm} = 1\Big\}.
\label{eq:phi}
\end{equation} 
Finally, a set of mixture models is generated for each sub-region (line 22 in Algorithm \ref{alg:ahgmm}) as 
\begin{equation}
{\mathcal{M}} = \Big\{M_{n}|n\in[1,N]\Big\},
\label{eq:M}
\end{equation}
where each element is calculated as
\begin{equation}
M_n =  \sum_{m=0}^{M}\phi_{nm} G_{nm}.
\label{eq:M_n}
\end{equation} 
\subsection{Local and Global Filtering}
\label{subsec:local_and_global_filtering}
We have now $N$ discretised Gaussian mixture models in $\mathcal{M}$ for $N$ sub-regions of $R$. We locally convolve each sub-region $R_{n}$ (Eq. \ref{eq:R_hat}) with their respective $M_n$ to make a protected sub-region $\bar{R}_{n}$:
\begin{equation}
\bar{\mathcal{R}}=\Big\{\bar{R}_{n}| n \in [1,N]\Big\},
\label{eq:R_pn}
\end{equation}
where $\bar{R}_{n} =  R_{n} * M_n$. Changing the convolutional kernel for each sub-region generates blocking artefacts (see Fig. \ref{fig:minimize_blocking_artefacts}). To smooth these artefacts, we apply a global convolution filter (line 25 in Algorithm \ref{alg:ahgmm}) with a Gaussian kernel of zero mean and standard deviation 
\begin{equation}
{\bar{\sigma}_j} = \frac{\sigma_{j}^o}{Q_j},
\label{eq:sigma_global}
\end{equation}
where $Q_j$ represents the sub-region size in pixels. As a result, a smoothed protected face $\bar{R}$ is developed which is replaced in the captured image $I$ to generate a privacy protected image $I^p$. Fig. \ref{fig:visual_comparison} shows few sample images filtered by AHGMM at different thresholds. 

\subsection{Computational Complexity}
\label{subsec:computational_complexity}

The generation of a convolutional kernel is more complex in AHGMM than in the adaptive Gaussian blur filter \citep{sarwar2016}. In fact, the latter only needs to compute a single Gaussian function, while \ac{AHGMM} requires the computation of $N \cdot M$ Gaussian functions. Moreover, the adaptive Gaussian blur exploits the separability property of 2D convolutional kernels, i.e. $\psi = \psi_h * \psi_v$, to reduce the number of multiplications and additions from $W\cdot H\cdot|\psi_h|\cdot|\psi_v|$ to $W \cdot H \cdot (|\psi_h|$ + $|\psi_v|)$ ($W$ and $H$ represent the width and height of $R$ in pixels, respectively). Instead, {AHGMM} dynamically reconfigures the convolutional kernel after processing each sub-region and therefore requires exactly $W\cdot H \cdot |\psi_h| \cdot |\psi_v|$ multiplications and additions.
\section{Dataset Generation}
\label{sec:data_set}

To the best of our knowledge, there is no large publicly available face dataset collected from an \ac{MAV}. We therefore generate face images as if they were captured from an \ac{MAV} via geometric transformation and down-sampling of the LFW dataset \citep{LFWTech2007}. The LFW dataset was collected in an unconstrained environment with extreme illumination conditions and extreme poses. 
%
We use the standard verification benchmark test of the LFW dataset (12000 images of 4281 subjects), divided into 10-folds for cross-validation. Each fold contains 600 images of the same subject and 600 images of different subjects. We use the deep funnelled version of the LFW dataset.

\begin{figure}
\centering
\begin{subfigure}{.19\linewidth}
\includegraphics[width=\textwidth]{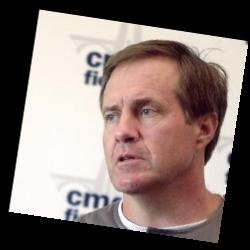}
\caption{}
\end{subfigure}
\begin{subfigure}{.19\linewidth}
\includegraphics[width=\textwidth]{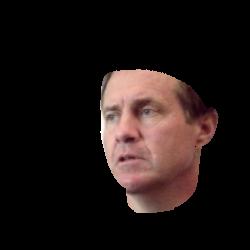}
\caption{}
\end{subfigure}
\begin{subfigure}{.19\linewidth}
\includegraphics[width=\textwidth]{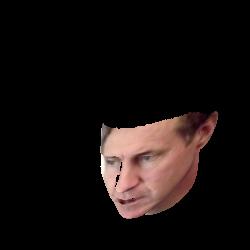}
\caption{}
\end{subfigure}
\begin{subfigure}{.19\linewidth}
\includegraphics[width=\textwidth]{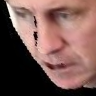}
\caption{}
\end{subfigure}
\begin{subfigure}{.19\linewidth}
\includegraphics[width=\textwidth]{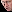}
\caption{}
\end{subfigure}
\caption{Sample images at different stages during the data set generation process. (a) Original image $250\times250$ pixels, (b) image after fitting a 3D morphable model at $0^\circ$ pitch angle, (c) image with synthetic pitch effect produced by applying a 3D geometric transformation, (d) aligned image of $96\times96$ pixels produced by applying an affine transformation computed by detecting eyes and nose location and (e) down-sampled image emulating an image captured at a different height.}
\label{fig:data_set_generation_steps}
\end{figure}

Figure~\ref{fig:data_set_generation_steps} shows sample images of the stages of the dataset generation pipeline. We fit a \ac{3DMM} \citep{Anil2016} on an input image to detect 68 facial landmarks \citep{Zhu2012} and then iteratively fit a \ac{3DMM} to generate a 3D image representation\footnote{Among the 12000 images, the landmark detector \citep{Zhu2012} was unable to detect 68 facial landmarks on 74 images. Therefore, we were unable to fit a 3DMM and used the original 74 images in order to comply with the standard verification test script of the LFW data set.}. As there may be only a few degrees pitch of the subject captured in the images (e.g.~a person looking slightly downward or upward),  we rotate the 3D image at $0^\circ$ pitch by applying a geometric transformation computed from the estimated pose of the fitted 3DMM. This disturbs the image alignment of the original data set, so a realignment is required, which we perform after generating the pitch effect.
The synthetic pitch angles start from $0^\circ$ to $70^\circ$ with a step size of $10^\circ$  and project it back to generate a corresponding 2D image. In order to align this image so that the eyes and nose appear at the same place among the images belonging to the same pitch angle, we apply an affine transformation computed by detecting eyes and nose tip using Dlib library \citep{dlib2009} such that the transformed face has a resolution of $96 \times 96$ pixels. As the detection accuracy of the eyes and nose decrease with increasing pitch angle, we generate a ground truth (location of eyes and nose tip) of the $0^\circ$ pitch angle images and uses it for the higher pitch angle images. 

Finally, to introduce different height effects for the $8$ synthetically generated images, we down-sample them with a factor of $2$, $4$, $8$ and $16$ generating images of $48 \times 48$, $24 \times 24$, $12 \times 12$, $6 \times 6$ pixels, respectively. Thus, we increase the size of the original standard verification test of the LFW data set by $40$ times, i.e.~from $12000$ images to $480,000$ images. Fig. \ref{fig:raw_images} shows the 40 sample images belonging to the same and different subjects.

We manually determined the values of $\rho_h$ and $\rho_v$ by 
\begin{equation}
\rho_h = S_c / S_h,
\label{eq:exp_rho_h}
\end{equation}
\begin{equation}
\rho_v = S_c cos (\gamma) / S_v,
\label{eq:exp_rho_v}
\end{equation} 
where $S_c$ is the cropped face size in pixels, $\gamma=90^\circ - \theta_R$ is the pitch angle of the image and $S_h$ and $S_v$ are the average human face dimensions, i.e.~the bitragion breadth of 15.45 cm and menton-crinion length of 20.75 cm, respectively \citep{DOD_HFE_TAG2000}.

\section{Experimental Results}
\label{sec:experimental_results}

\subsection{Experimental Set up}
\label{subsec:experimental_setup}

We compare \ac{AHGMM} against \ac{SVGB} \citep{Saini2012}, \ac{AGB} \citep{sarwar2016} and \ac{FGB}, which uses a constant Gaussian kernel defined with respect to the highest resolution face. Thus, we estimate the kernel for \ac{FGB} as in \citep{sarwar2016} for the face with $96\times96$ pixels at $0^\circ$ pitch angle. For the SVGB filter, we divide the face into four concentric circles and reduce the kernel size by $5\%$ while radially moving out between two consecutive regions as in \citep{Saini2012}. Although the kernel for the innermost region was manually selected in the original work, we choose the anisotropic kernel as estimated by the AGB \citep{sarwar2016} and convert it into an isotropic kernel for a fair comparison. We use a block size of $4\times4$ and $m=1$ for the AHGMM.

To compare privacy filters, we measure the face verfication accuracy using OpenFace \citep{amos2016}, an open source implementation of Google's face recognition algorithm FaceNet \citep{schroff2016}. OpenFace uses a deep Convolutional Neural Network (CNN) as a feature extractor, which is trained by a large face data set (500k images). This feature extractor is applied on the training and test images for their representations (embeddings) which are used for classification \citep{schroff2016}. 

To measure distortion as in \citep{Adam2014, Nawaz2015}, we apply the \ac{PSNR}, the power ratio of the original image with respect to the filtered image.

We perform experiments with 480,000 images (consisting of 5 different resolutions and 8 different pitch angles) to determine the validity of the proposed \ac{AHGMM} to protect the identity information of an individual. For this purpose, we analyse the effect of a na\"{i}ve-T attack, a parrot-T attack, an inverse filter attack and a super-resolution attack. Moreover, we quantify the corresponding fidelity degradation caused by the AHGMM. 

As AGB and SVGB do not use any secret key, we evaluate them only using their accurate parameters in the parrot-T, inverse filter and super-resolution attacks. In contrast, any of these attacks on AHGMM can be further divided into three sub-attacks: optimal kernel, pseudo AHGMM and accurate AHGMM. In the optimal kernel sub-attack, we assume that an attacker is able to estimate the parameters of the optimal kernel and applies the optimal kernel to the entire face. In the pseudo AHGMM sub-attack, we assume that the attacker knows the optimal kernel and randomly modifies the filter parameter for the $N$ sub-regions. In the accurate AHGMM sub-attack, we assume that the attacker has access to the secret key and can decipher all filter parameters for the $N$ sub-regions. As this prior-knowledge can be exploited for both probe and gallery images, we therefore evaluate AHGMM under 13 different scenarios stated in Table \ref{tab:attack_types}.

\begin{table}[t]
\centering
\caption{Attacks used to evaluate the privacy level of the proposed AHGMM algorithm. Both the gallery faces and the probe faces can be protected or unprotected (na\"{i}ve-BL). Moreover, the protected faces could be either unchanged or reconstructed (e.g.~through an inverse-filter (IF) or super-resolution (SR)). Finally, any AHGMM attack could be further divided into three sub-attacks corresponding to the prior-knowledge of an attacker: optimal, pseudo and accurate.}
\label{tab:attack_types}
\resizebox{1.0\columnwidth}{!}{%
\begin{tabular}{cccc|c|c|c|c|}
\cline{5-8}
                                                    &                                                &                                                     &                  & \multicolumn{4}{c|}{Gallery images}                                                                                                                                                                                                                                                  \\ \cline{5-8} 
                                                    &                                                &                                                     &                  & unprotected & \multicolumn{3}{c|}{protected}                                                                                                                                                      \\ \cline{6-8} 
                                                    &                                                &                                                     &                  &                                                                                                 & unchanged & \multicolumn{2}{c|}{reconstructed}                                                            \\ \cline{7-8} 
                                                    &                                                &                                                     &                  &                                                                                                 &                                                                                    & IF & SR                                                             \\ \hline
\multicolumn{1}{|l|}{\multirow{13}{*}{\rot{Probe images}}} & \multicolumn{1}{l}{\rot{unprotected}}                       &                                &                  &                                                                                                na\"{i}ve-BL & N/A                                                                                   &     N/A           &  N/A                                                                            \\ \cline{2-8} 
\multicolumn{1}{|l|}{}                              & \multicolumn{1}{l|}{\multirow{9}{*}{\rot{protected}}} & \multicolumn{1}{l}{\rot{unchanged}}                          &                  & na\"{i}ve-T                                                                                           & \begin{tabular}[c]{@{}l@{}}parrot-T\\ - optimal\\ - pseudo\\ - accurate\end{tabular} &       ---         &                                                                             --- \\ \cline{3-8} 
\multicolumn{1}{|l|}{}                              & \multicolumn{1}{l|}{} & \multicolumn{1}{l|}{\multirow{6}{*}{\rot{reconstructed}}} & \rot{IF}   & \begin{tabular}[c]{@{}l@{}}na\"{i}ve-IF\\ -optimal\\ -pseudo\\ -accurate\end{tabular}          &                                                                                   --- &       \begin{tabular}[c]{@{}l@{}}parrot-IF\\ -accurate\end{tabular}         &           --- \\ \cline{4-8} 
\multicolumn{1}{|l|}{}                              & \multicolumn{1}{l|}{}                          & \multicolumn{1}{l|}{}                               & \rot{SR} & \begin{tabular}[c]{@{}l@{}}na\"{i}ve-SR\\ -optimal\\ -pseudo\\ -accurate\end{tabular} &                                                                                   --- &          ---      & \begin{tabular}[c]{@{}l@{}}parrot-SR\\  -accurate\end{tabular} \\ \hline
\end{tabular}
}
\end{table}

We assume that an attacker is able to determine the pitch angle of a protected face using the background information of an image captured from an \ac{MAV} and can apply a geometric transformation to transform the gallery images at that pitch angle. Therefore, in all the following attacks, both the gallery and the probe images are at the same pitch angle which can be protected or unprotected depending upon the attack type. Moreover, we use the same resolution for both the gallery images and the probe images. 

\subsection{Na\"{i}ve-T Attack}
\label{subsec:naive_attack}

\begin{figure}[t]
\centering
\includegraphics[width=\columnwidth]{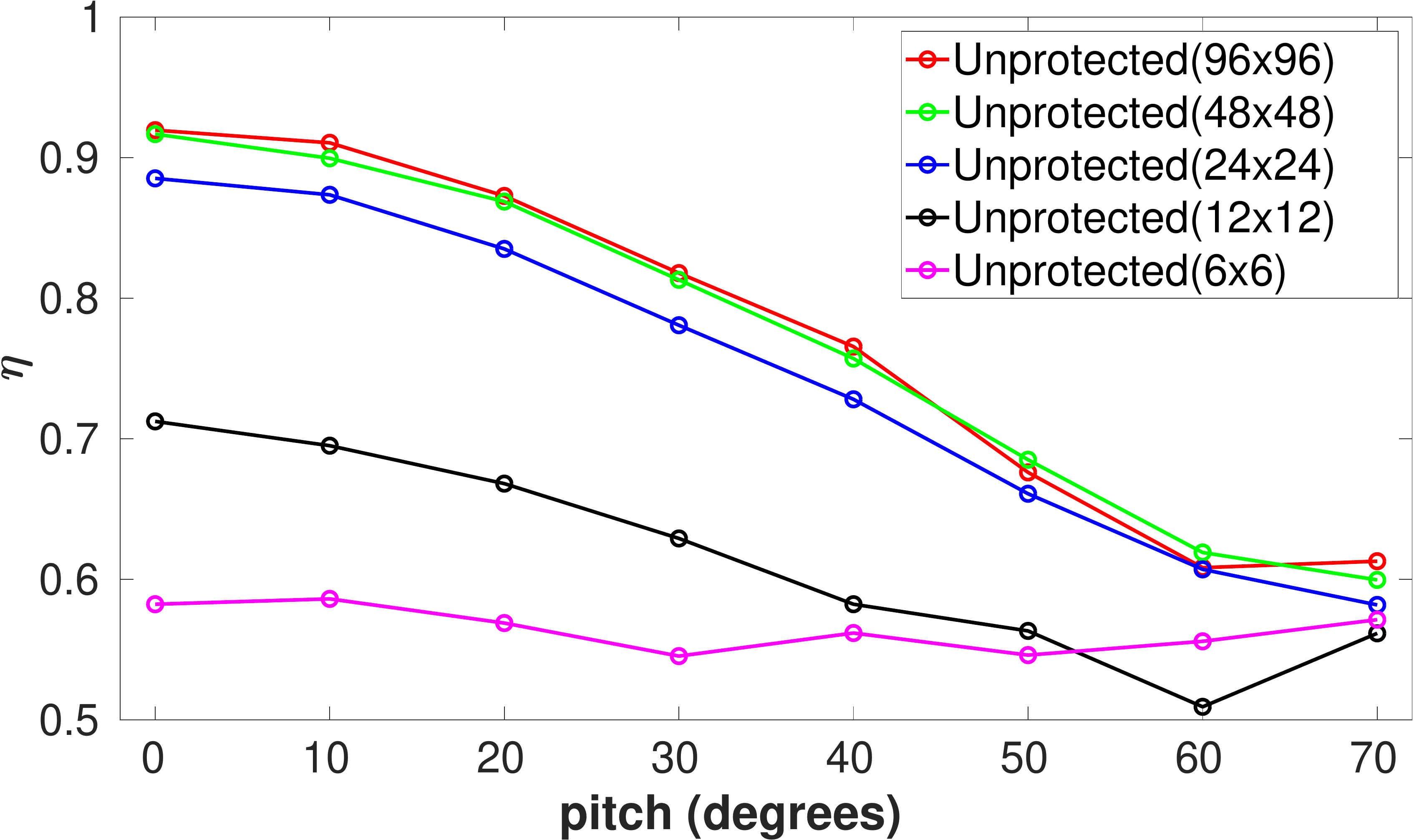}\\
\caption{Face verification accuracy $\eta$ of a na\"{i}ve-BL attack on our synthetically generated face data set. In general, $\eta$ increases with increasing the face size except at high pitch angles of $60$ and $70$ degrees where it slightly fluctuates randomly. For $6\times6$ pixels faces, $\eta$ is the lowest and rather independent of the pitch angle.}
\label{fig:privacy_raw}
\end{figure}

\begin{figure*}[!htb]
\centering
\resizebox{0.97\textwidth}{!}{%
\begin{tabular}{cccc}
Na\"{i}ve-T & Parrot-T & Parrot-T & Parrot-T\\
& (optimal kernel) & (pseudo AHGMM) & (accurate AHGMM)\\
\includegraphics[width=0.25\linewidth]{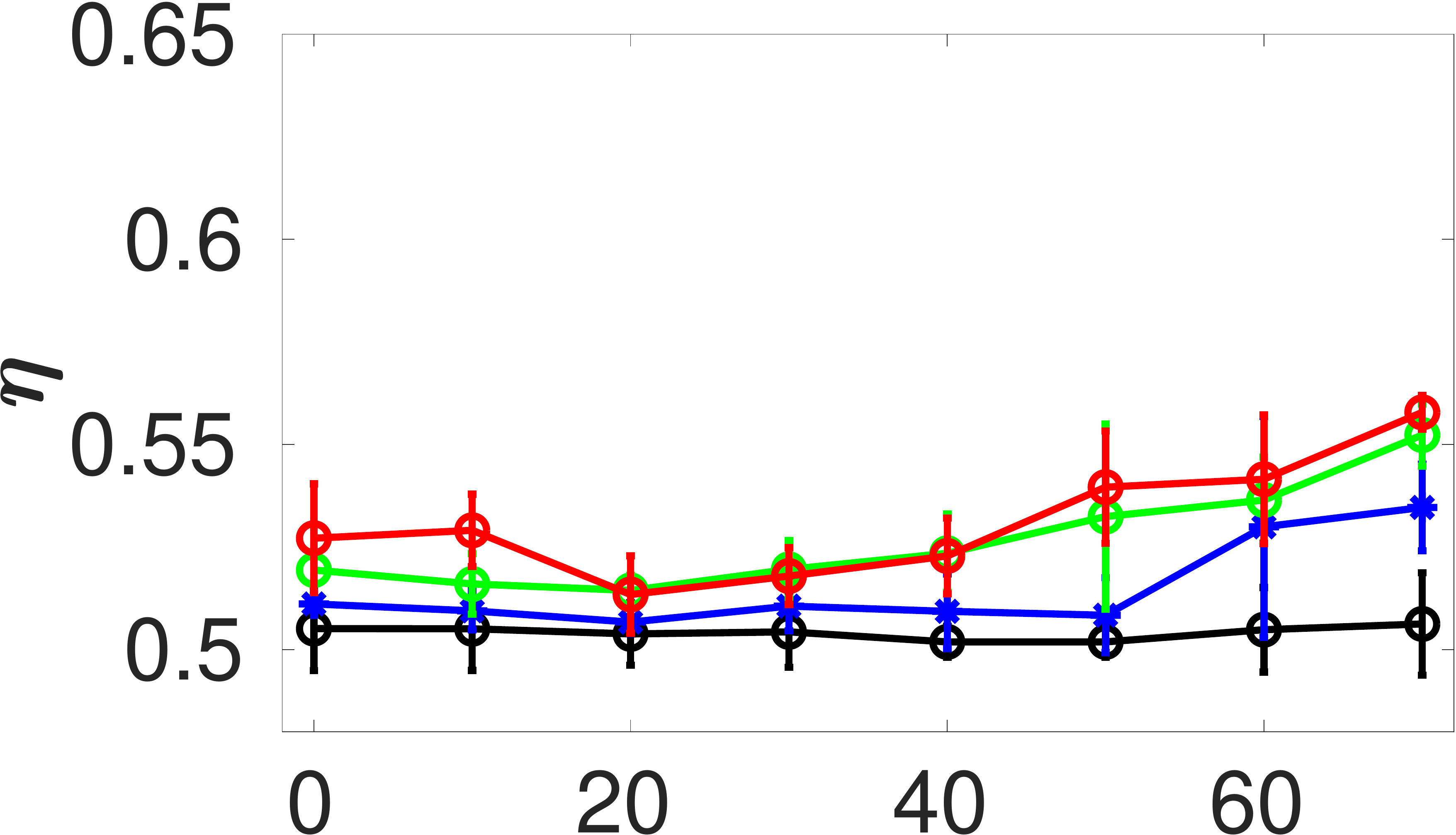}
&
\includegraphics[width=0.25\linewidth]{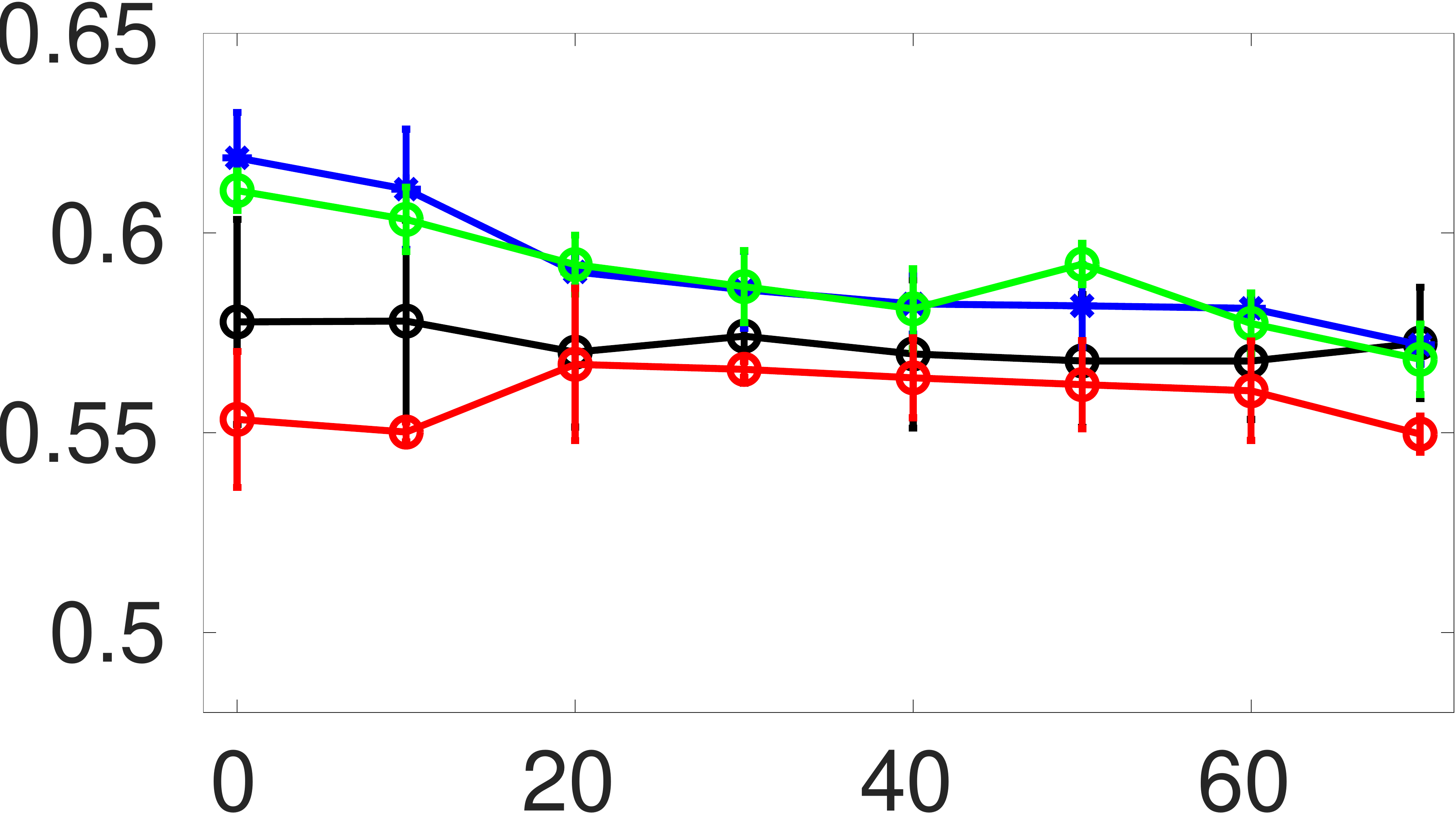}
&
\includegraphics[width=0.25\linewidth]{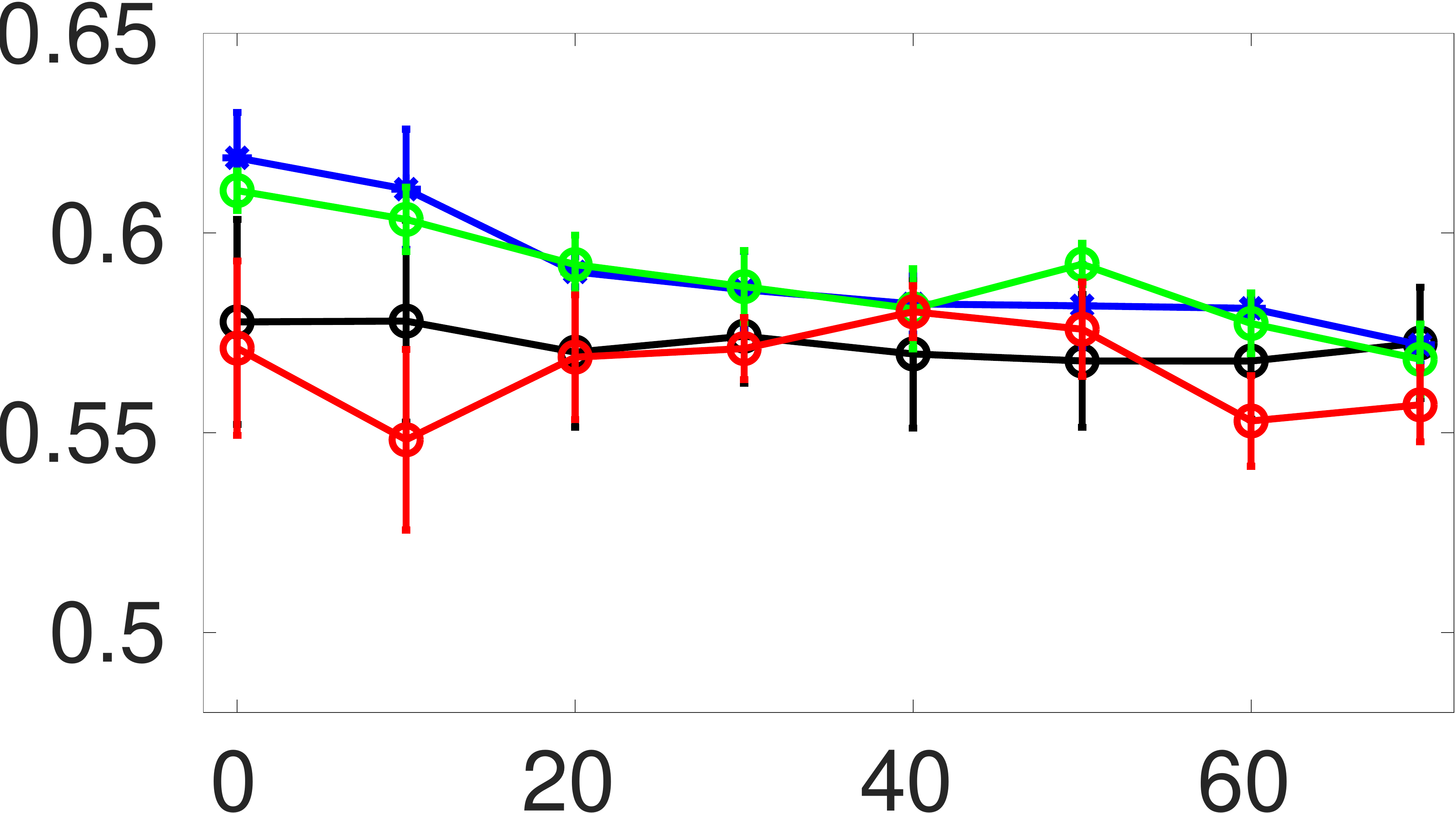}
&
\includegraphics[width=0.25\linewidth]{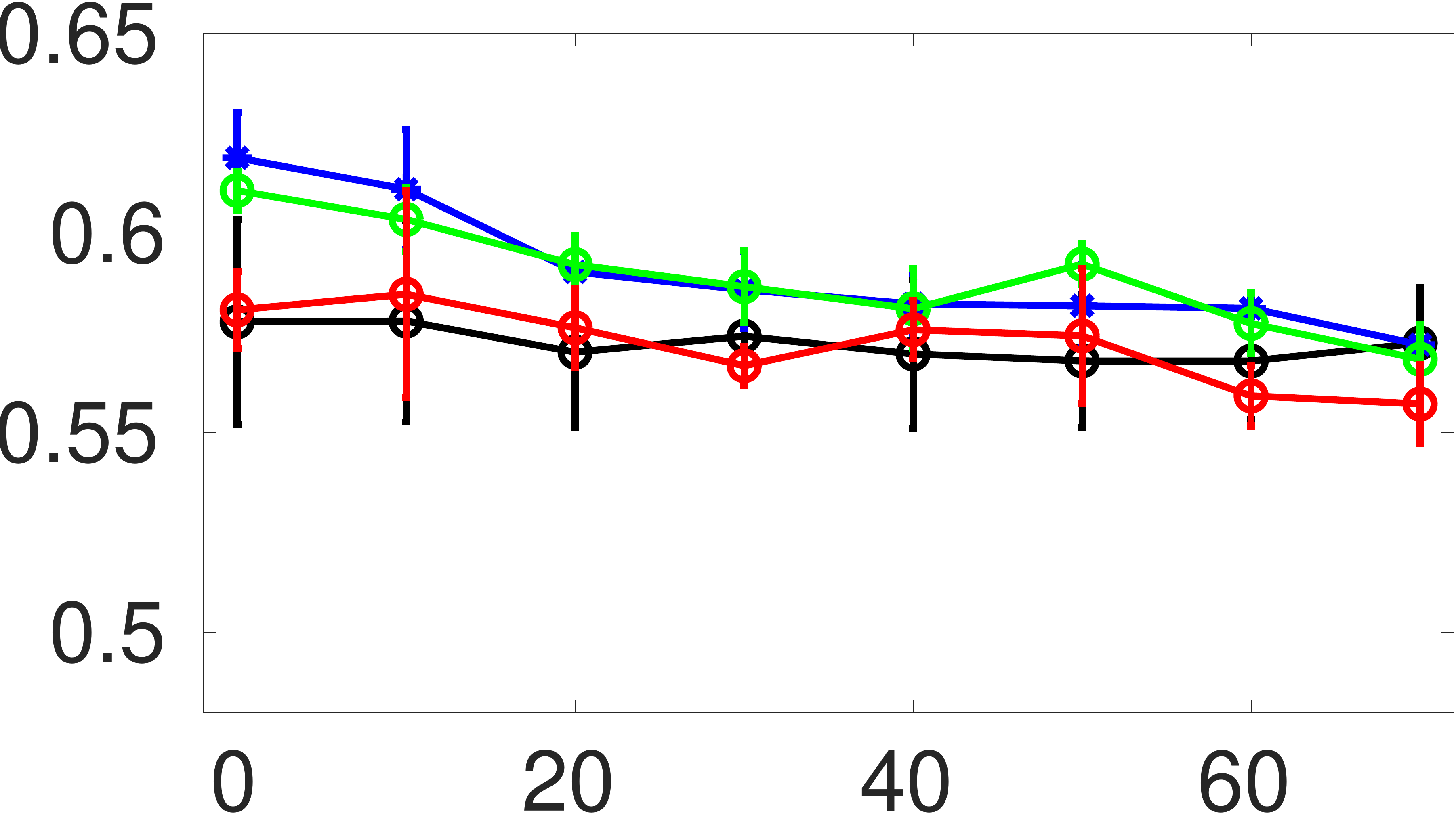}
\\
\includegraphics[width=0.25\linewidth]{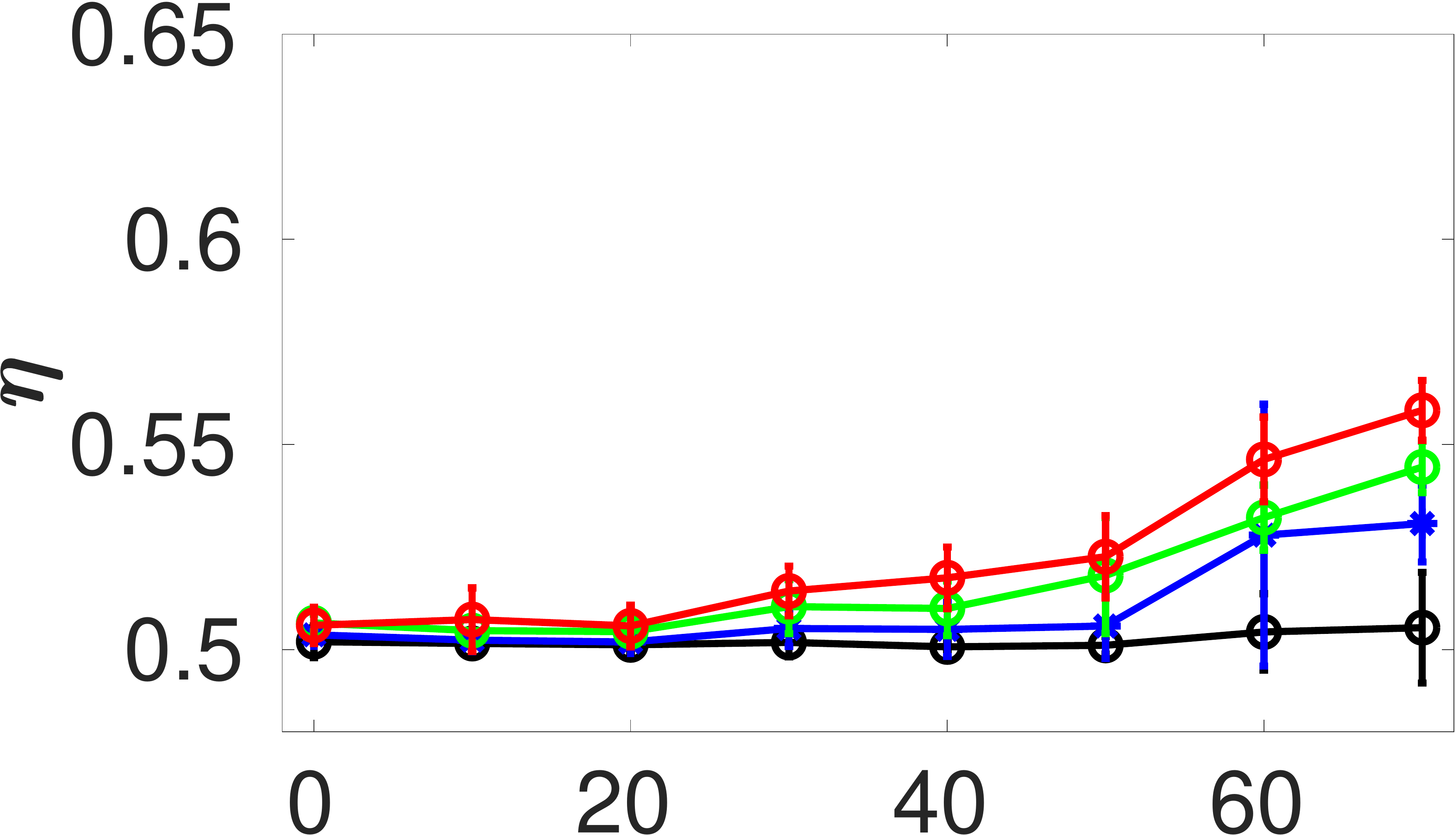}
&
\includegraphics[width=0.25\linewidth]{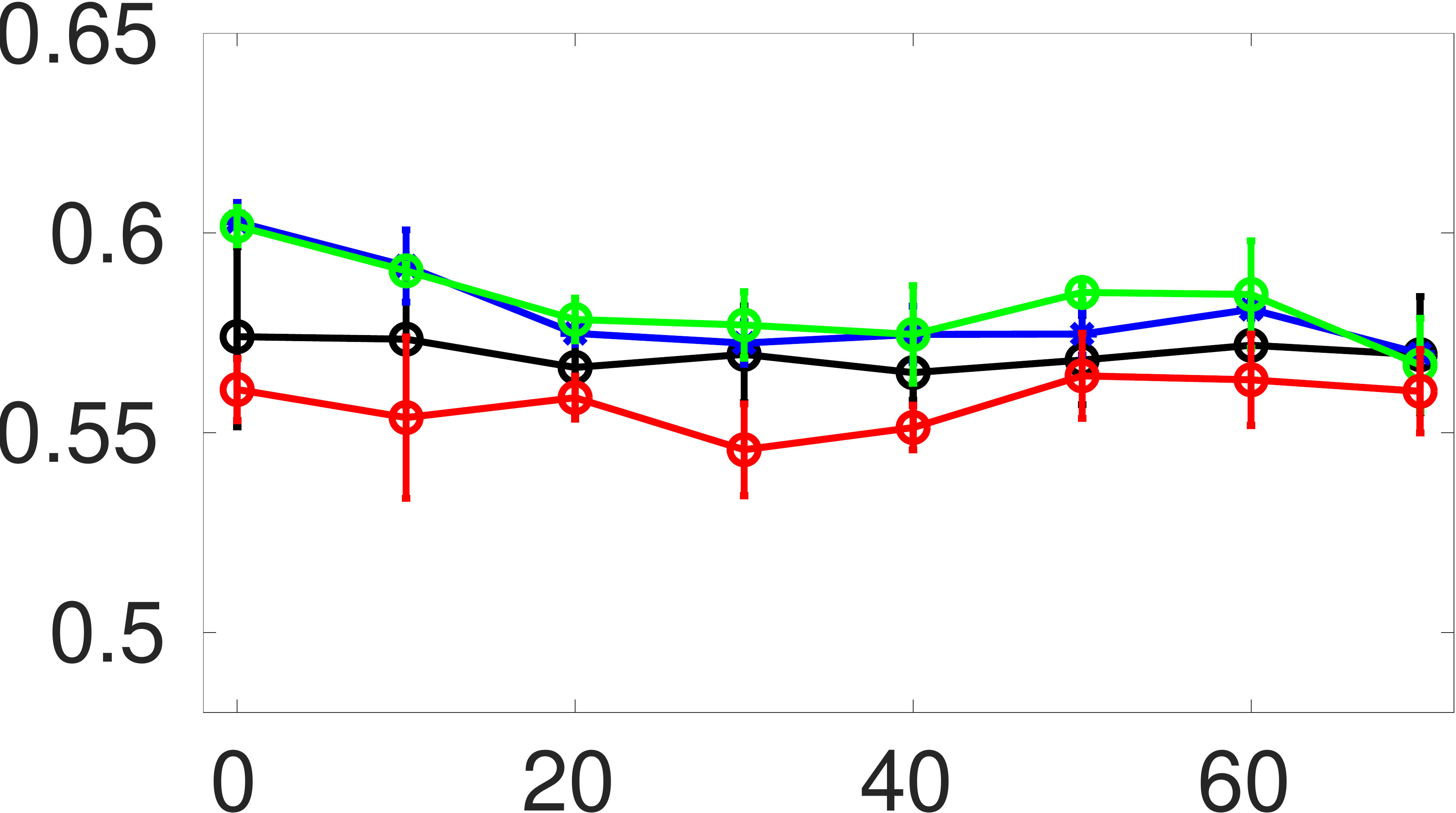}
&
\includegraphics[width=0.25\linewidth]{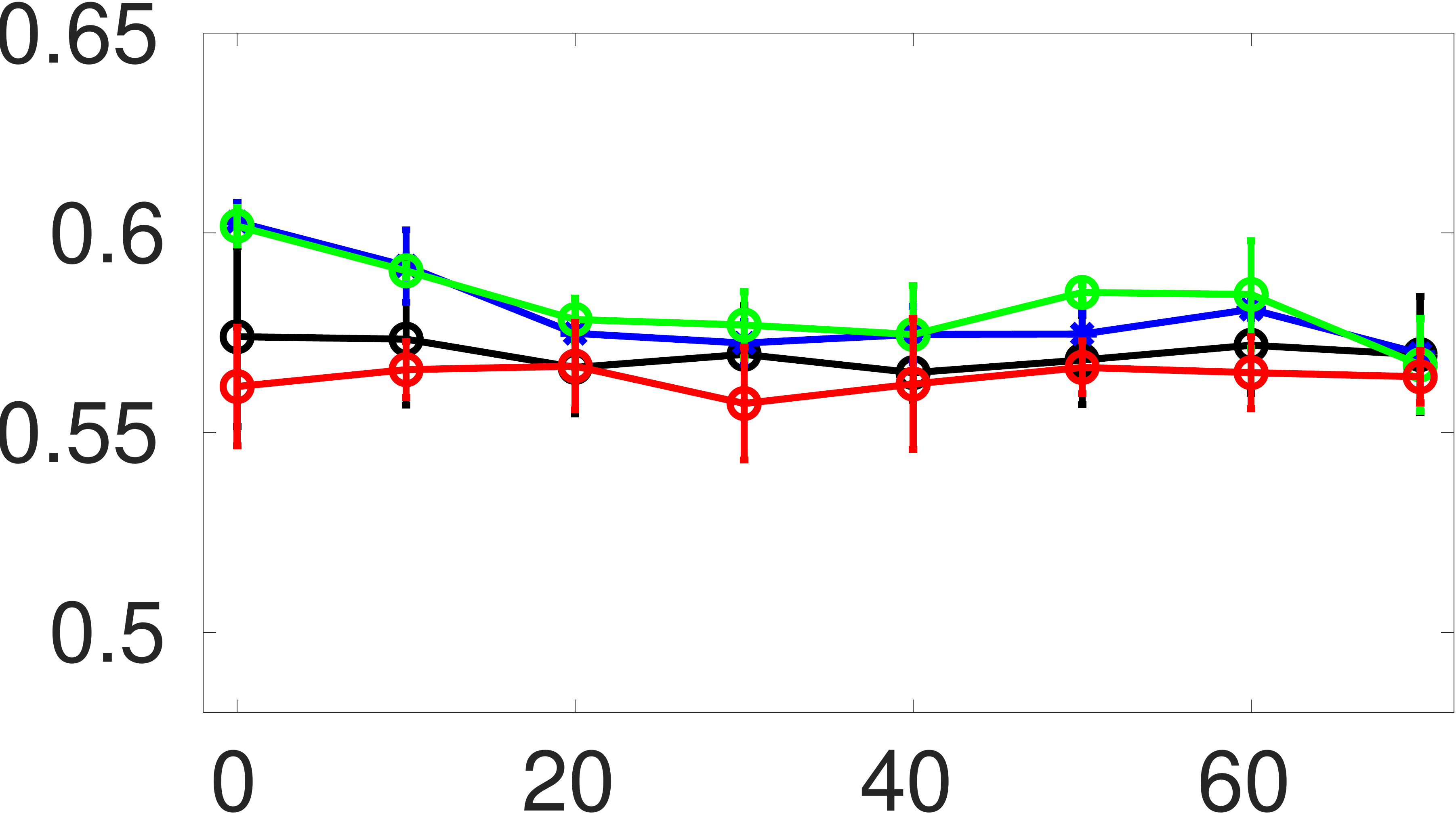}
&
\includegraphics[width=0.25\linewidth]{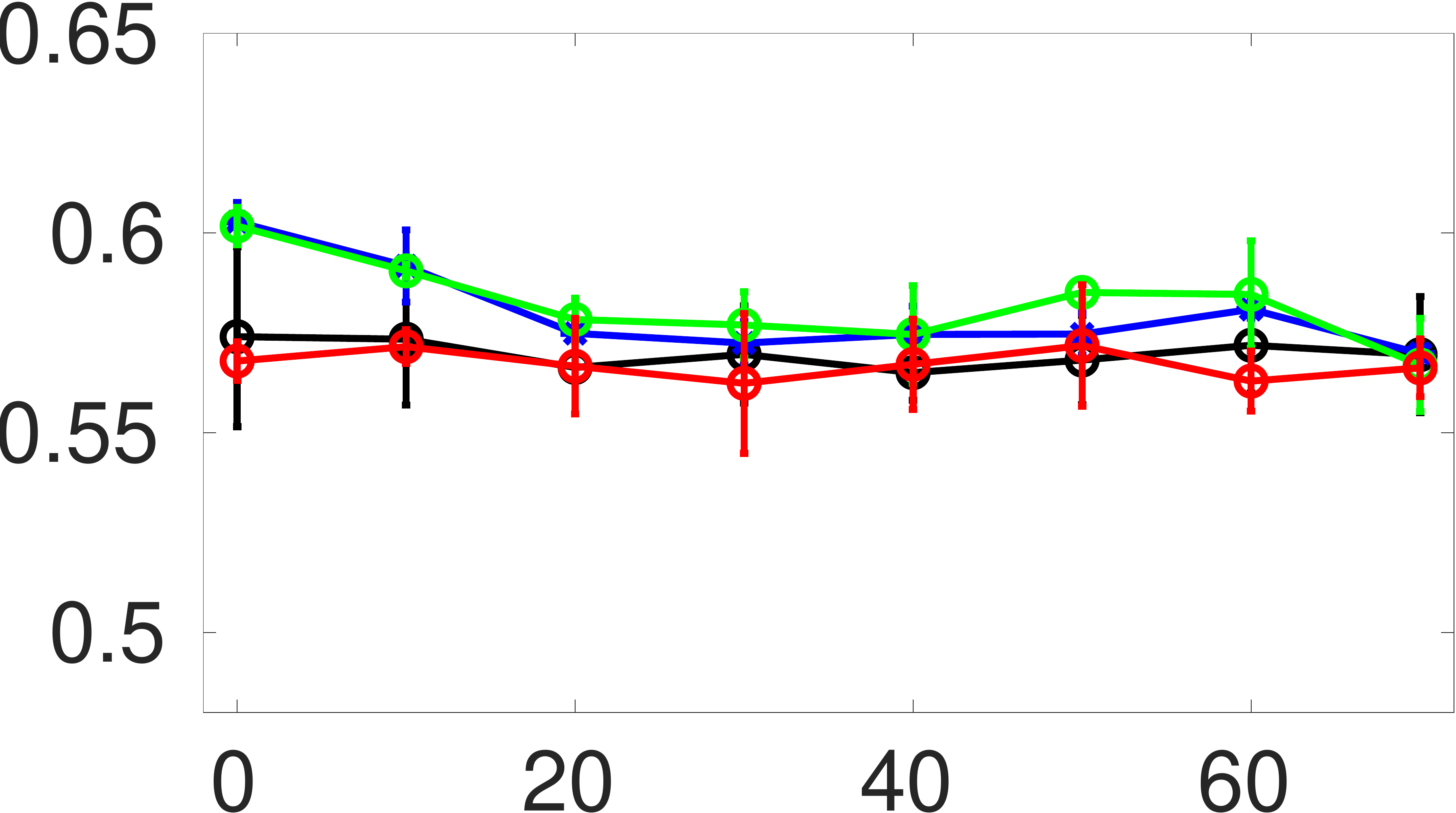}
\\
\includegraphics[width=0.25\linewidth]{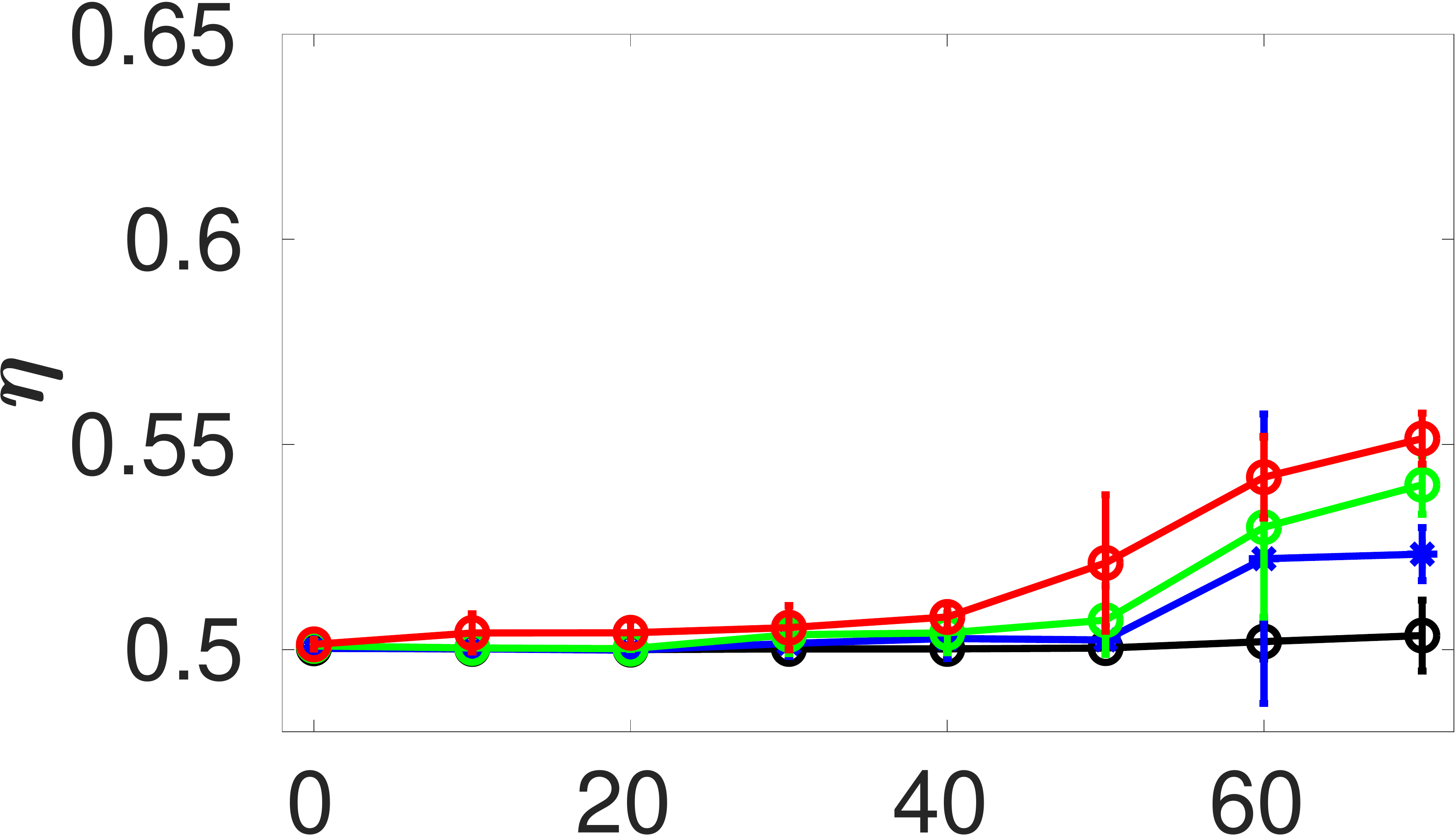}
&
\includegraphics[width=0.25\linewidth]{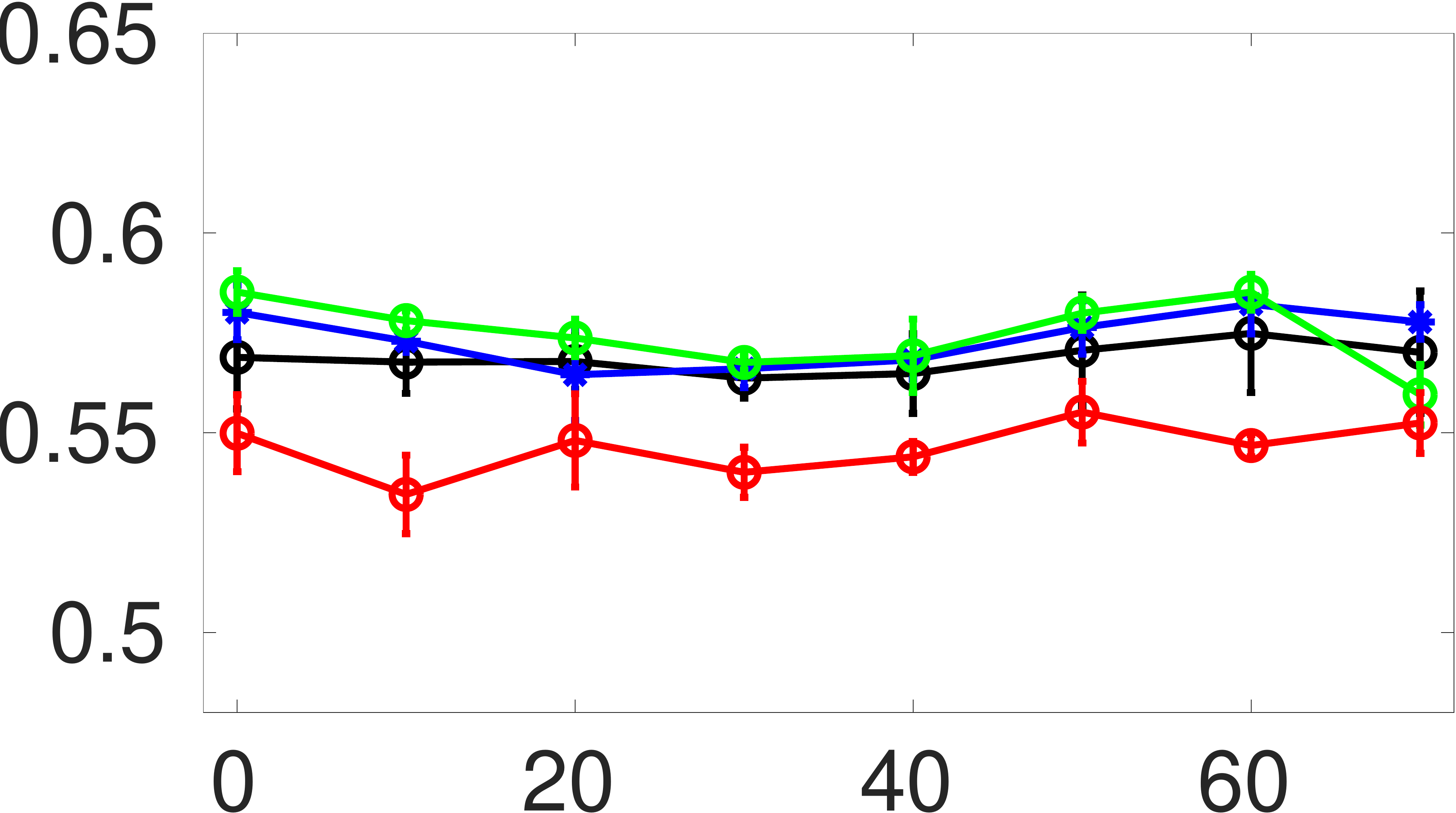}
&
\includegraphics[width=0.25\linewidth]{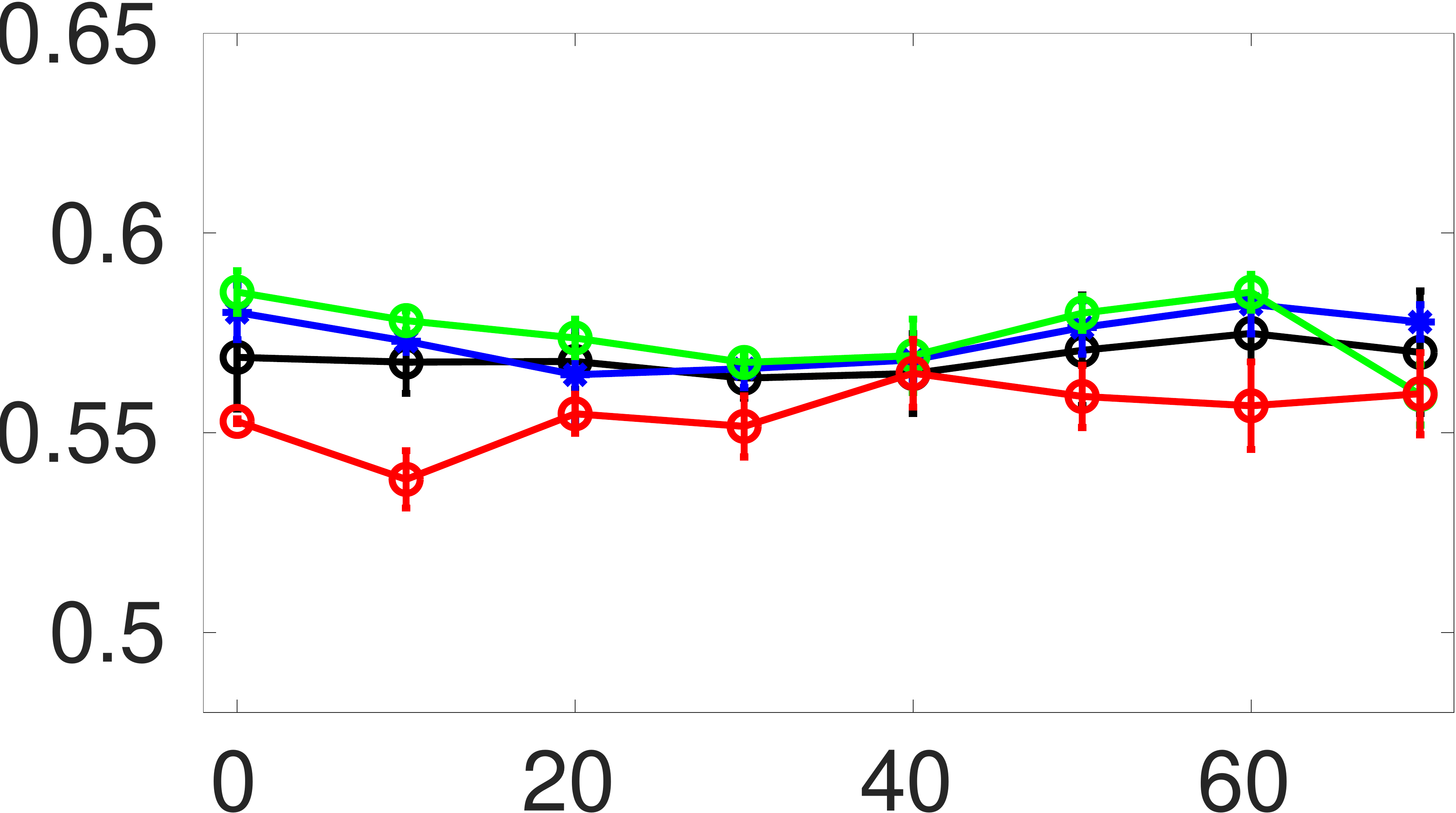}
&
\includegraphics[width=0.25\linewidth]{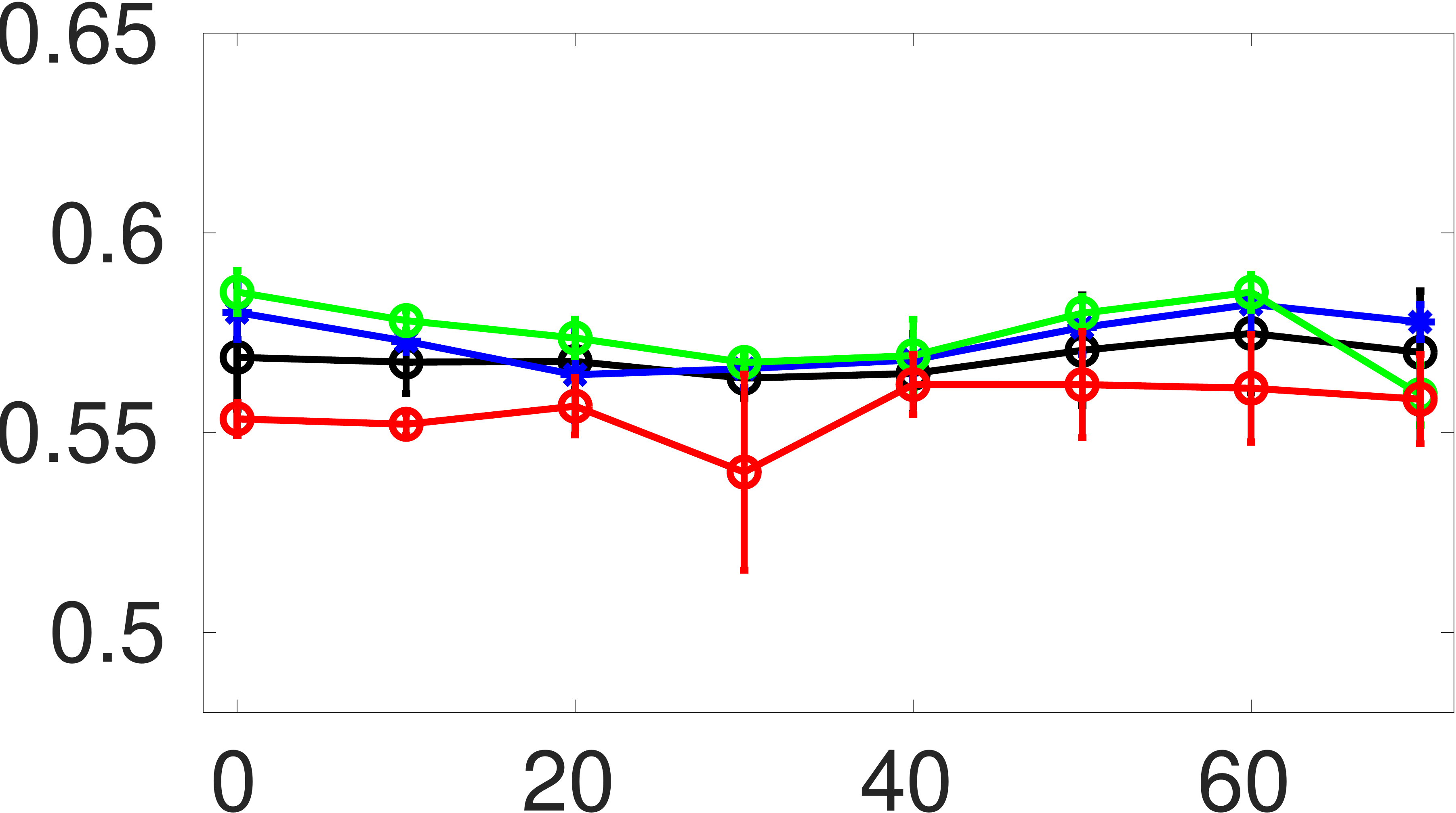}
\\
\includegraphics[width=0.25\linewidth]{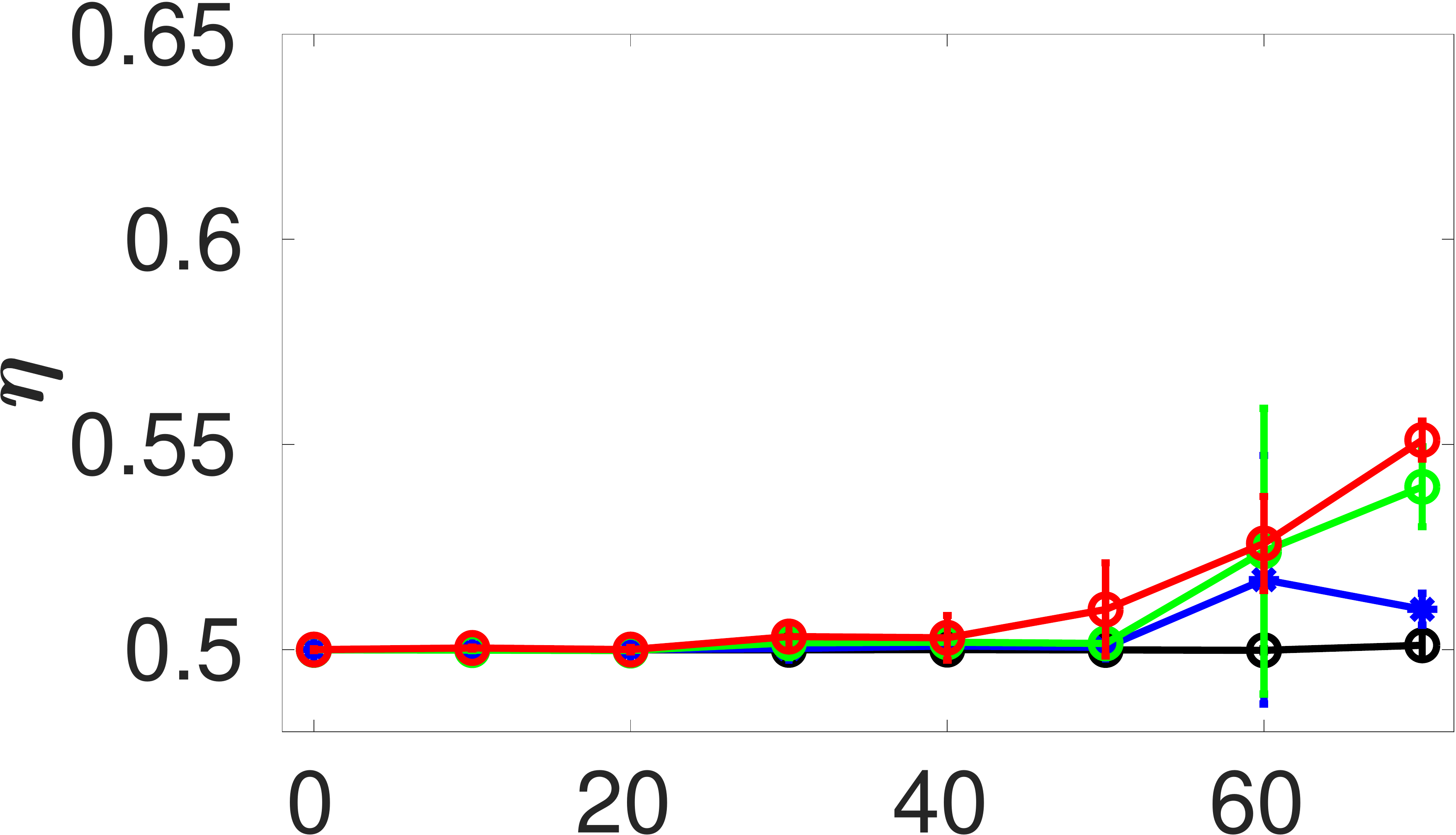}
&
 \includegraphics[width=0.25\linewidth]{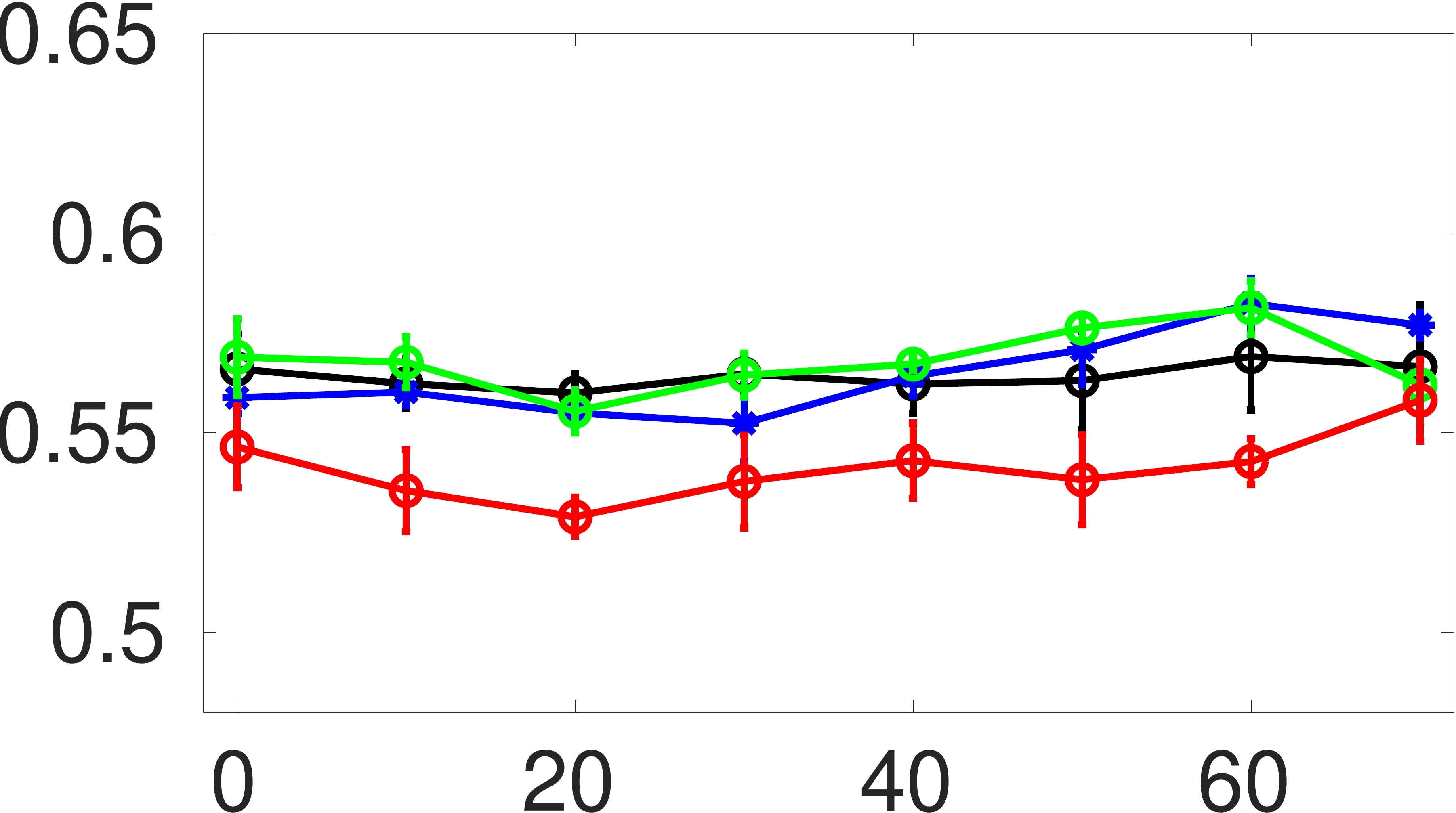}
&
\includegraphics[width=0.25\linewidth]{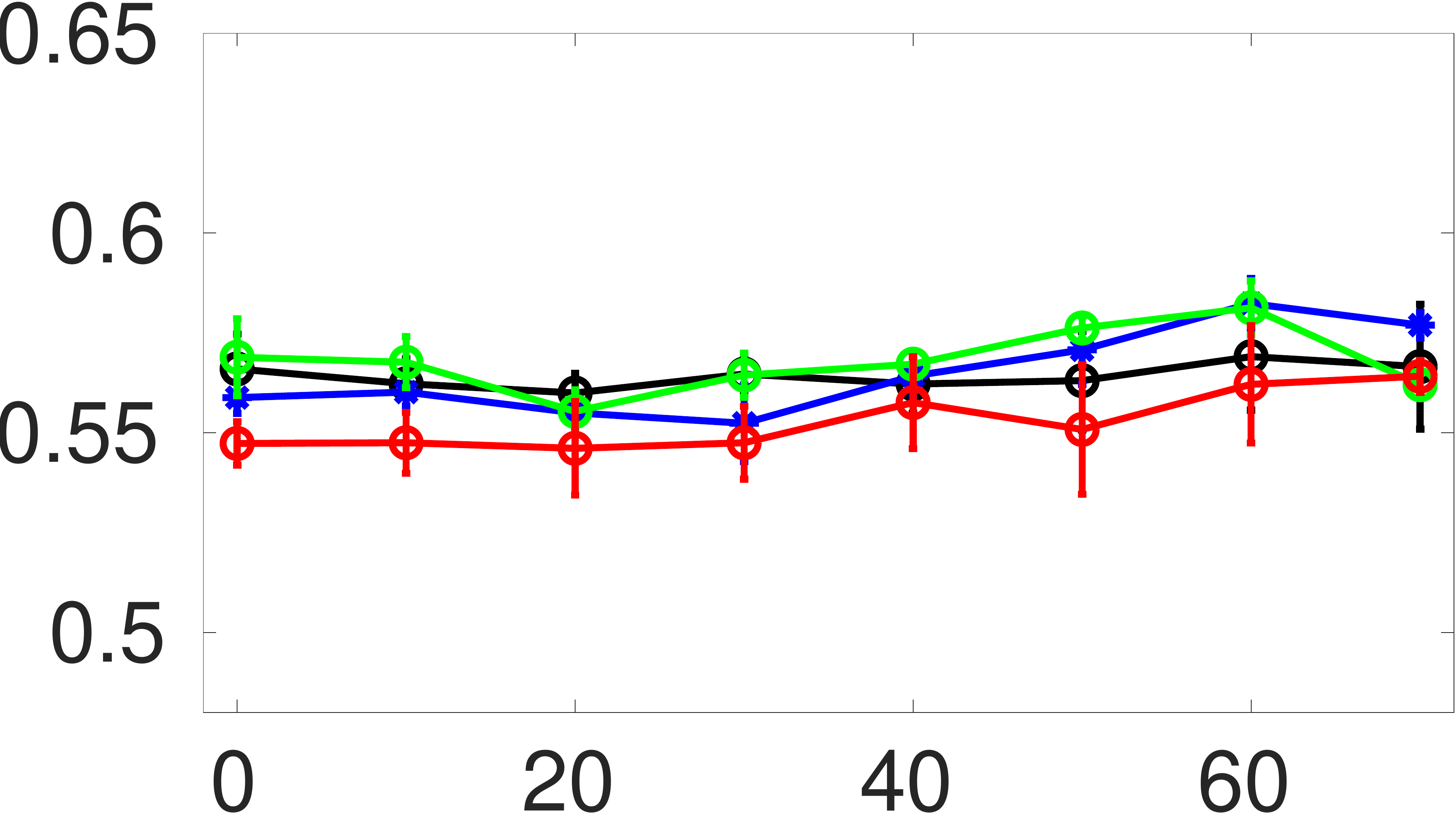}
&
\includegraphics[width=0.25\linewidth]{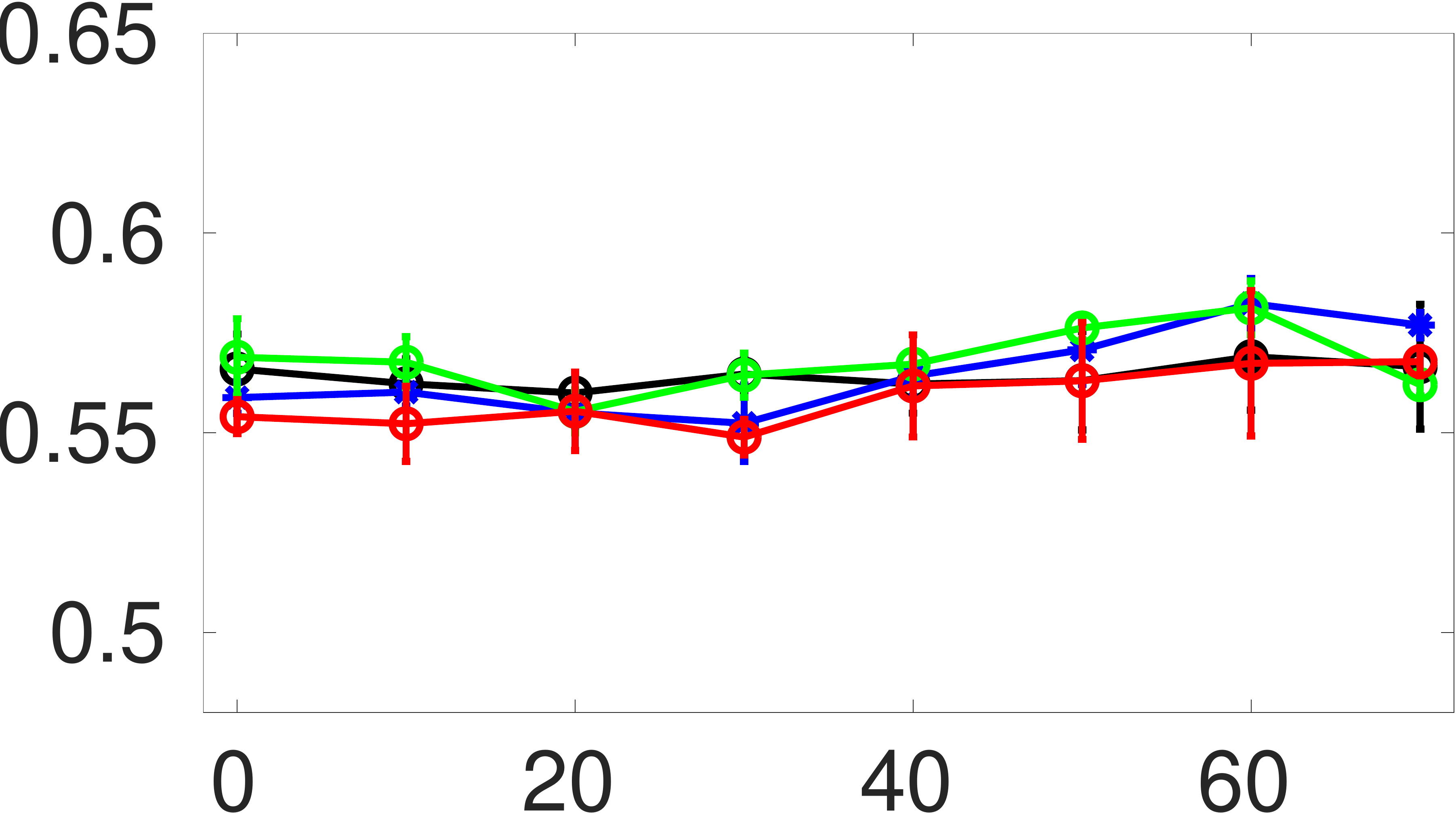}
\\
\includegraphics[width=0.25\linewidth]{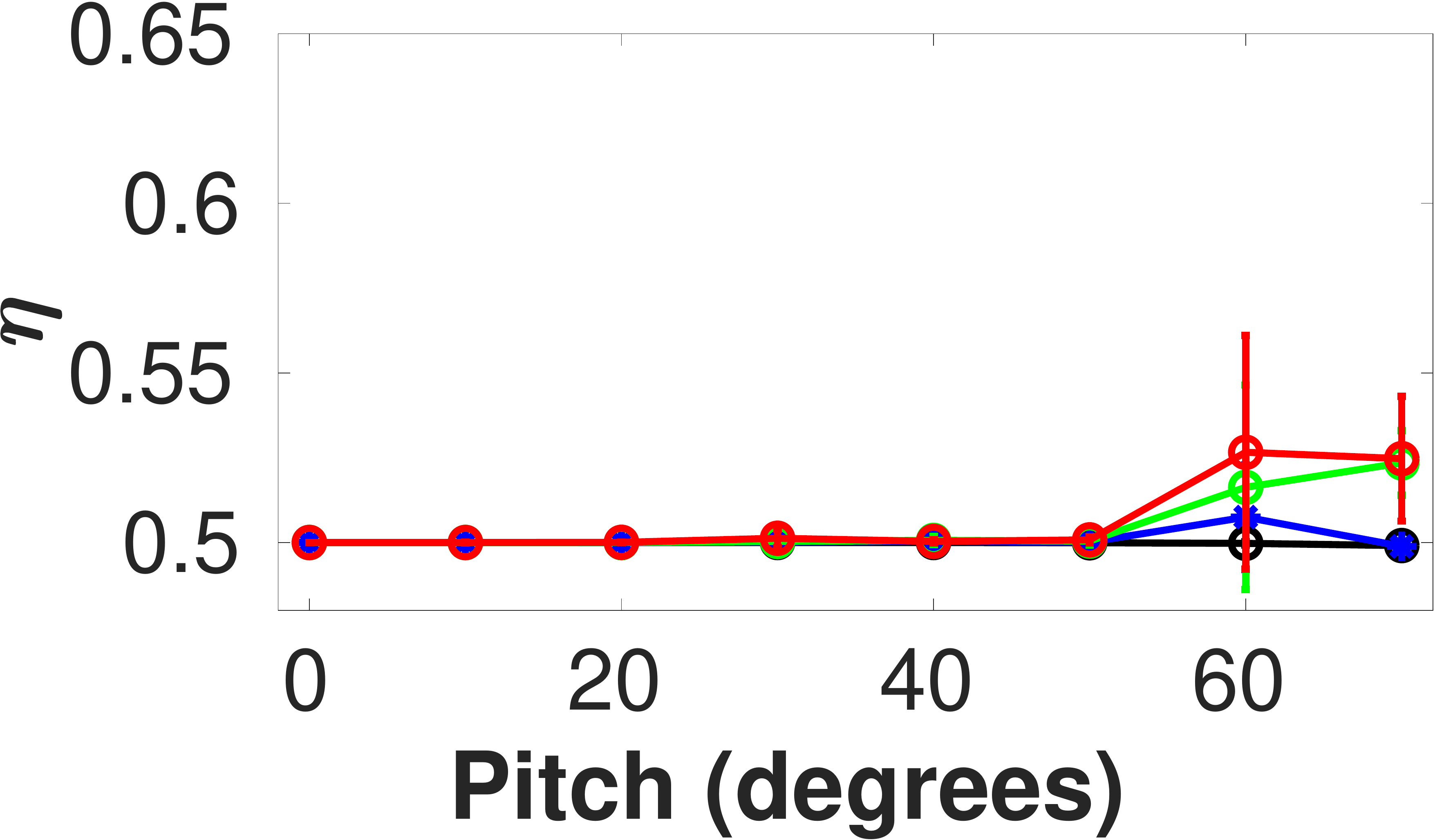}
&
\includegraphics[width=0.25\linewidth]{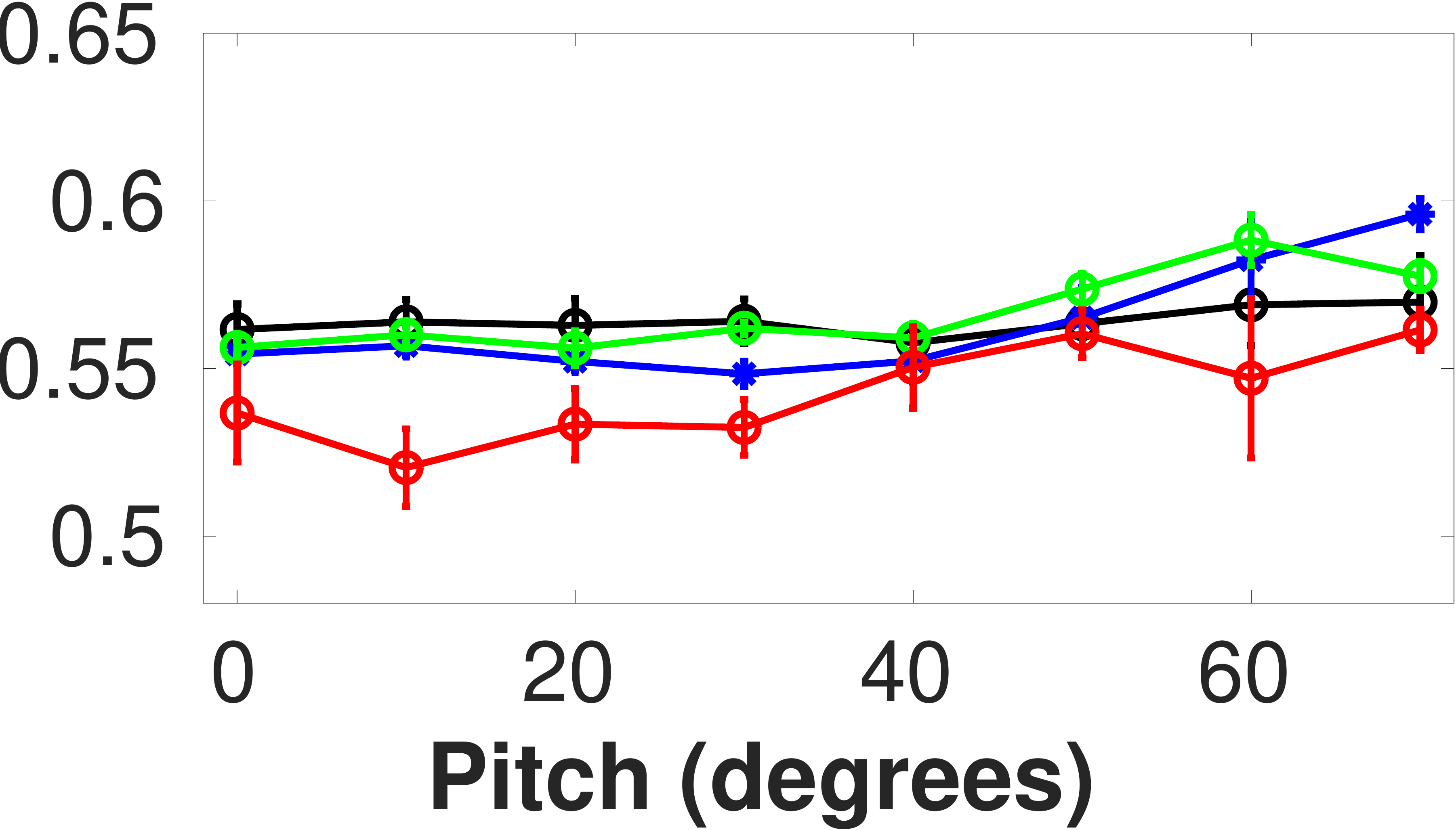}
&
\includegraphics[width=0.25\linewidth]{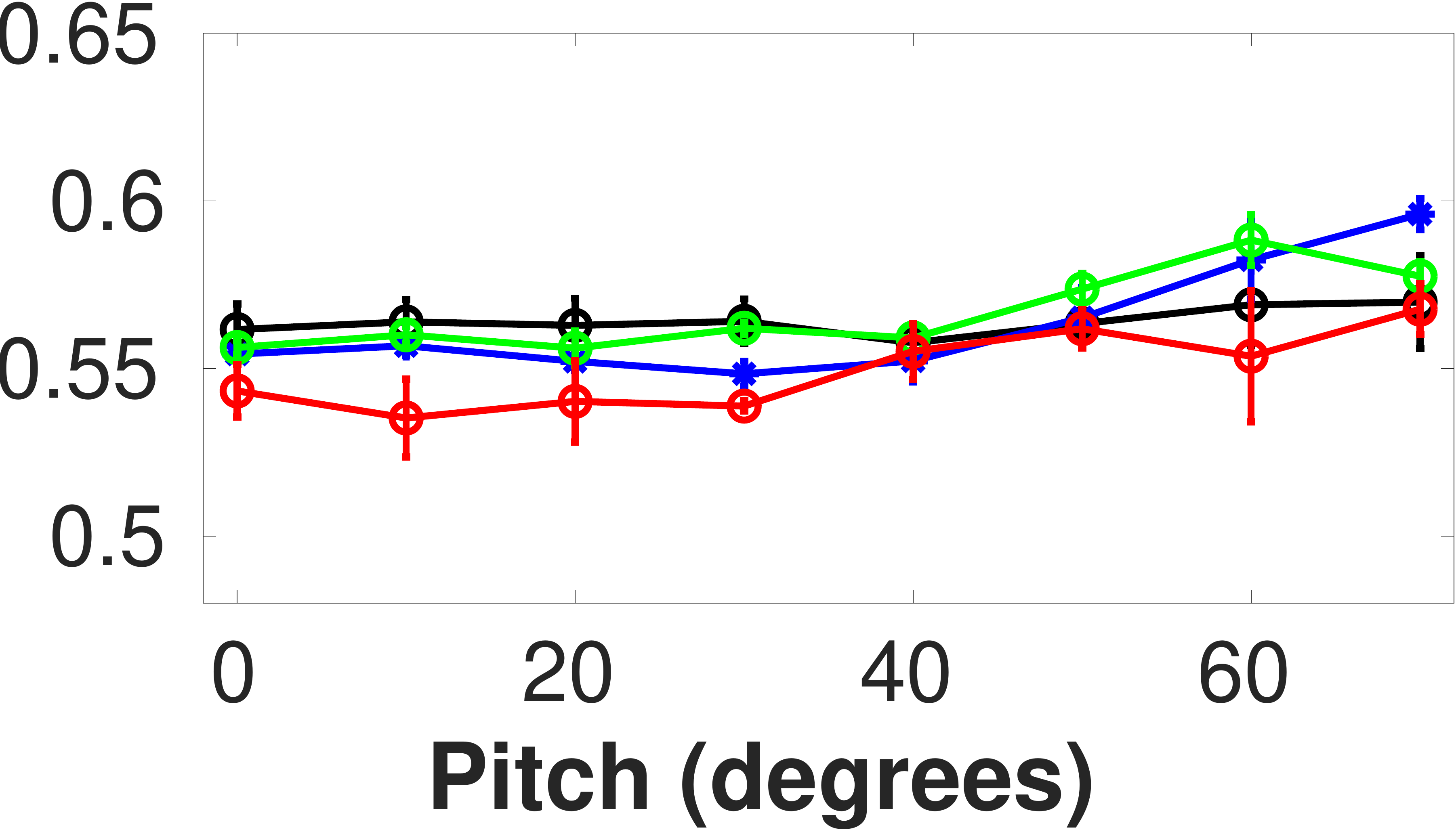}
&
\includegraphics[width=0.25\linewidth]{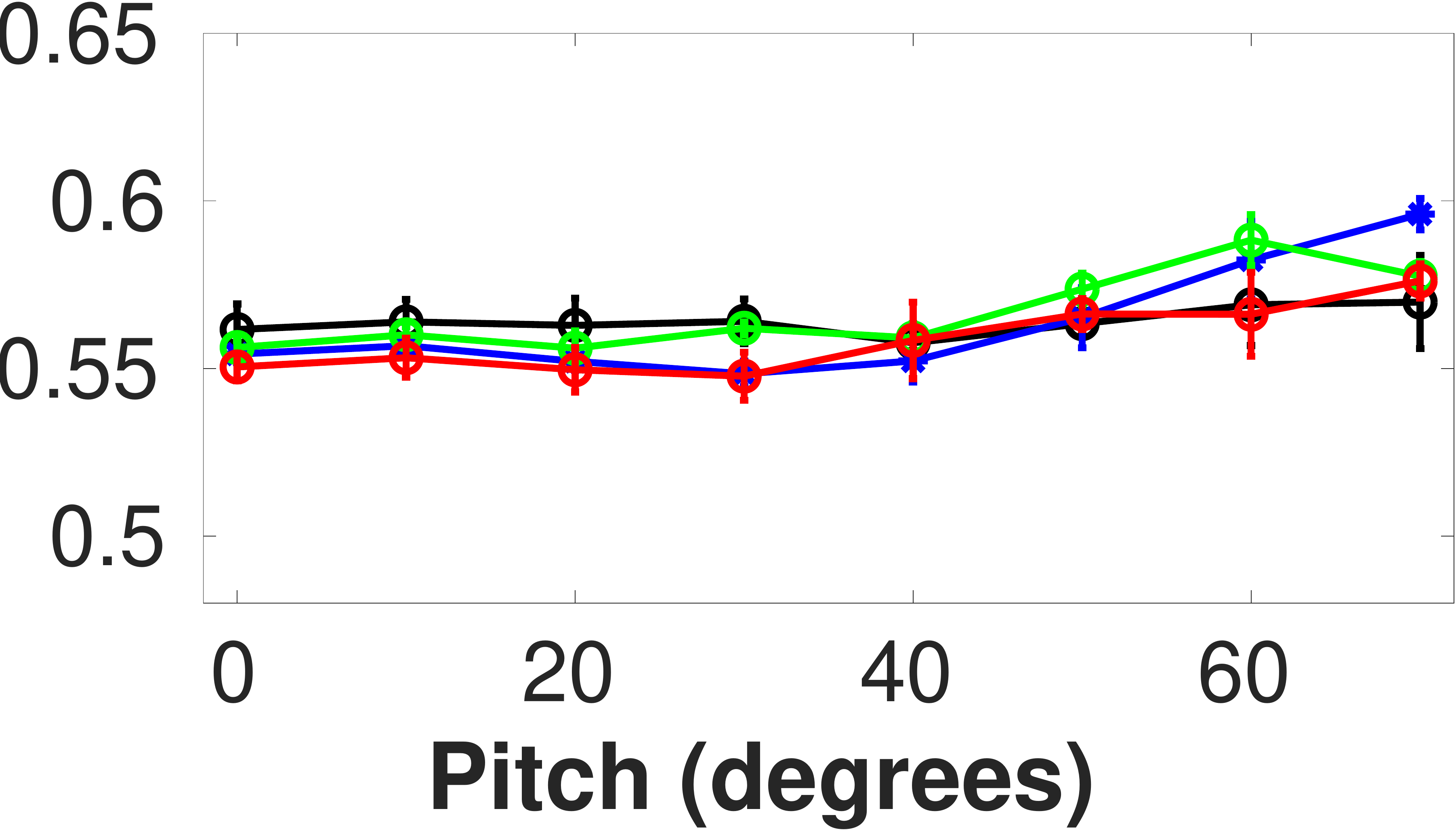}
\\
\end{tabular}
}
\caption{Face verification accuracy $\eta$ achieved by na\"{i}ve and parrot attacks on images protected by four different privacy protection filters at different thresholds $\rho_j^o$: first row: $\rho_j^o=0.7$ px/cm, second row: $\rho_j^o=0.6$ px/cm, third row: $\rho_j^o=0.5$ px/cm, fourth row: $\rho_j^o=0.4$ px/cm, fifth row: $\rho_j^o=0.3$ px/cm. The filled marker shows the mean and the vertical bar indicates the standard deviation of $\eta$ for the multi-resolution images ($96\times96$, $48\times48$, $24\times24$, $12\times12$). Legend: \textcolor{red}{\Huge ---} AHGMM, \textcolor{green}{\Huge{---}} \ac{AGB} \citep{sarwar2016}, \textcolor{blue}{\Huge{---}} \ac{SVGB} \citep{Saini2012}, \textcolor{black}{\Huge{---}} \ac{FGB}. Under the na\"{i}ve-T attack, AHGMM posses the highest $\eta$ which converges towards $\eta=0.5$ as the $\rho_j^o$ is decreased and finally at $\rho_j^o \leq 0.5$ px/cm, the difference between $\eta$ of AHGMM, AGB, SVGB and FGB becomes negligible, except unexpectedly at pitch angles $60^{\circ}$ and $70^{\circ}$ degrees. The parrot-T attack on AHGMM is divided into three sub-attacks: optimal kernel parrot-T attack, pseudo AHGMM parrot-T attack and accurate AHGMM parrot-T attack. In contrast to {na\"{i}ve-T} attack, AHGMM provides the lowest $\eta$ under any type of the three parrot-T attacks and this fact becomes negligible at $\rho_j^o = 0.3$ px/cm under accurate AHGMM parrot-T attack.}
\label{fig:naive_attack}
\end{figure*}

First of all, we perform a na\"{i}ve-BL attack which shows the baseline face verfication accuracy when both the probe data set and the gallery data set are unprotected. The results of the na\"{i}ve-BL attack are given in Fig.~\ref{fig:privacy_raw}. After that we perform a na\"{i}ve-T attack in which the gallery images are unprotected, while the probe images are protected using FGB, SVGB \citep{Saini2012}, AGB \citep{sarwar2016} and AHGMM. The results of this attack are given in Fig. \ref{fig:naive_attack} at different thresholds $\rho_j^o$.

The na\"{i}ve-BL attack shows that the accuracy $\eta$ of our synthetically generated data set decreases with the decrease of the face resolution and with the increase in the face pitch angle. However, this trend vanishes at high pitch angles, i.e.~$60^\circ$ and $70^\circ$, where it shows slight randomness. Finally, for the low resolution faces ($6\times6$ pixels), the accuracy does not show any effect of the pitch angle and slightly oscillates. Therefore, we consider $6\times6$ pixels inherently privacy protected and remove these images from the analysis of the privacy filters.

\begin{figure*}
\centering
\resizebox{0.97\textwidth}{!}{%
\begin{tabular}{cccc}
\includegraphics[width=\linewidth]{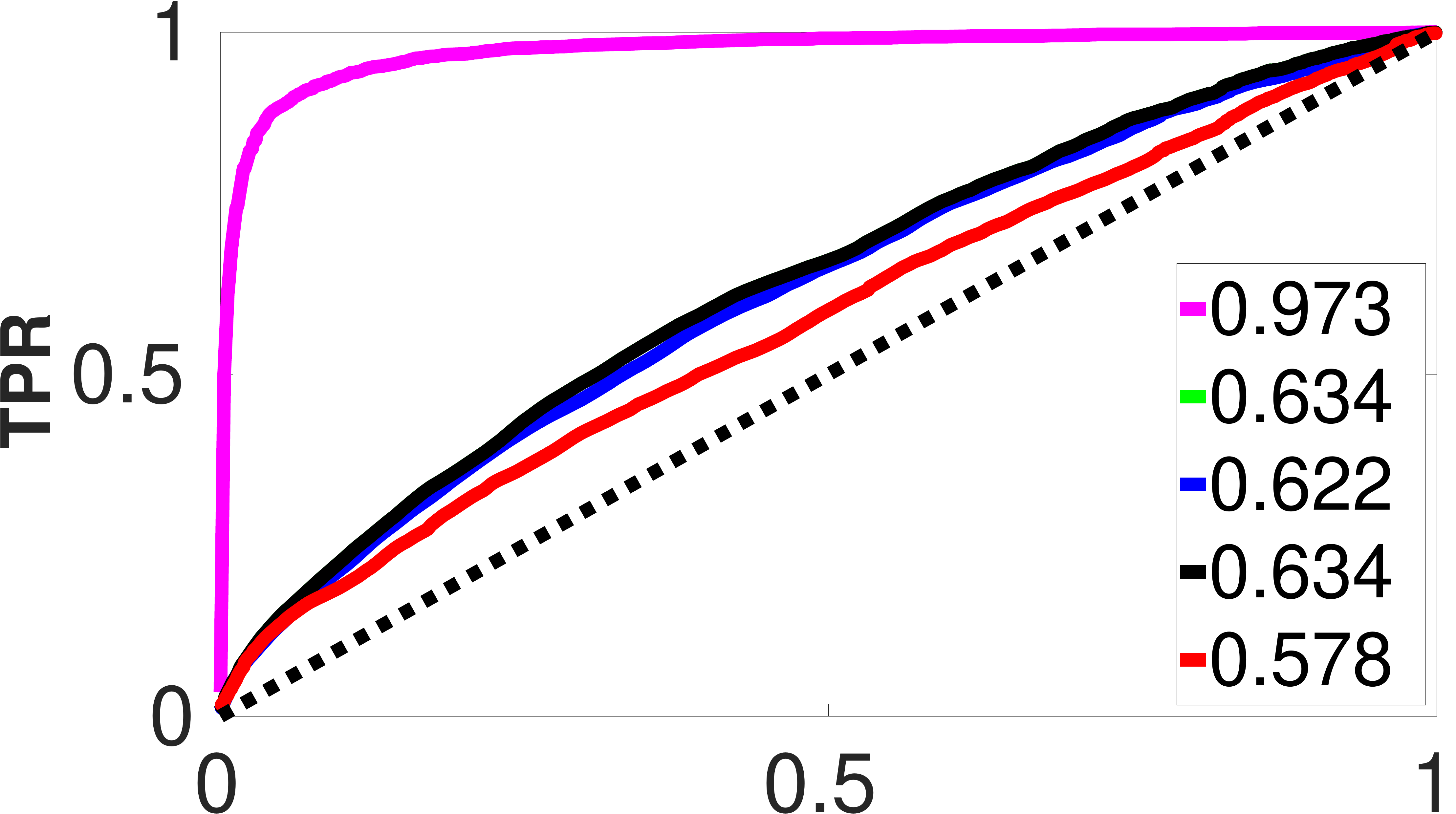}
&
 \includegraphics[width=\linewidth]{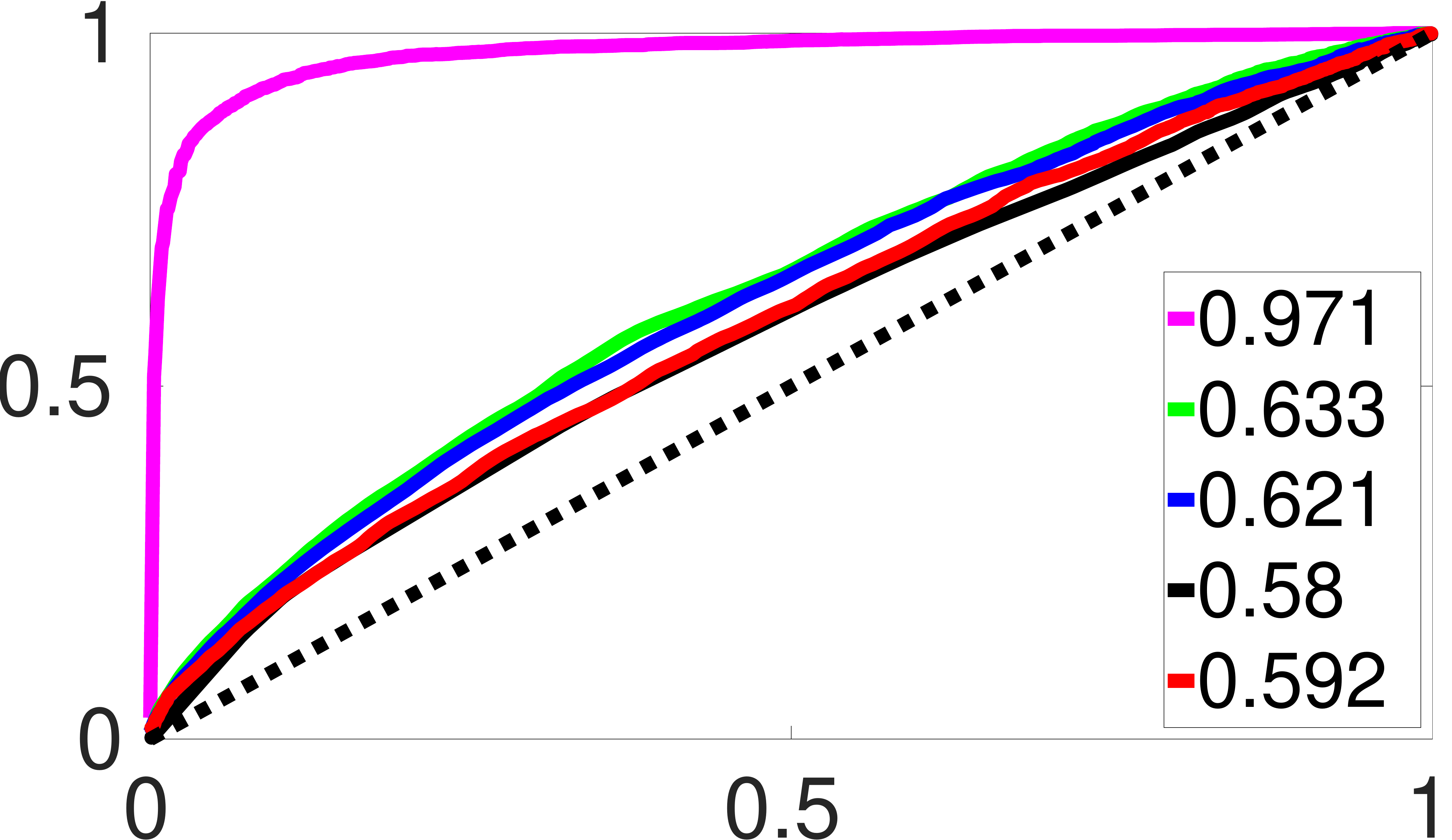}
&
\includegraphics[width=\linewidth]{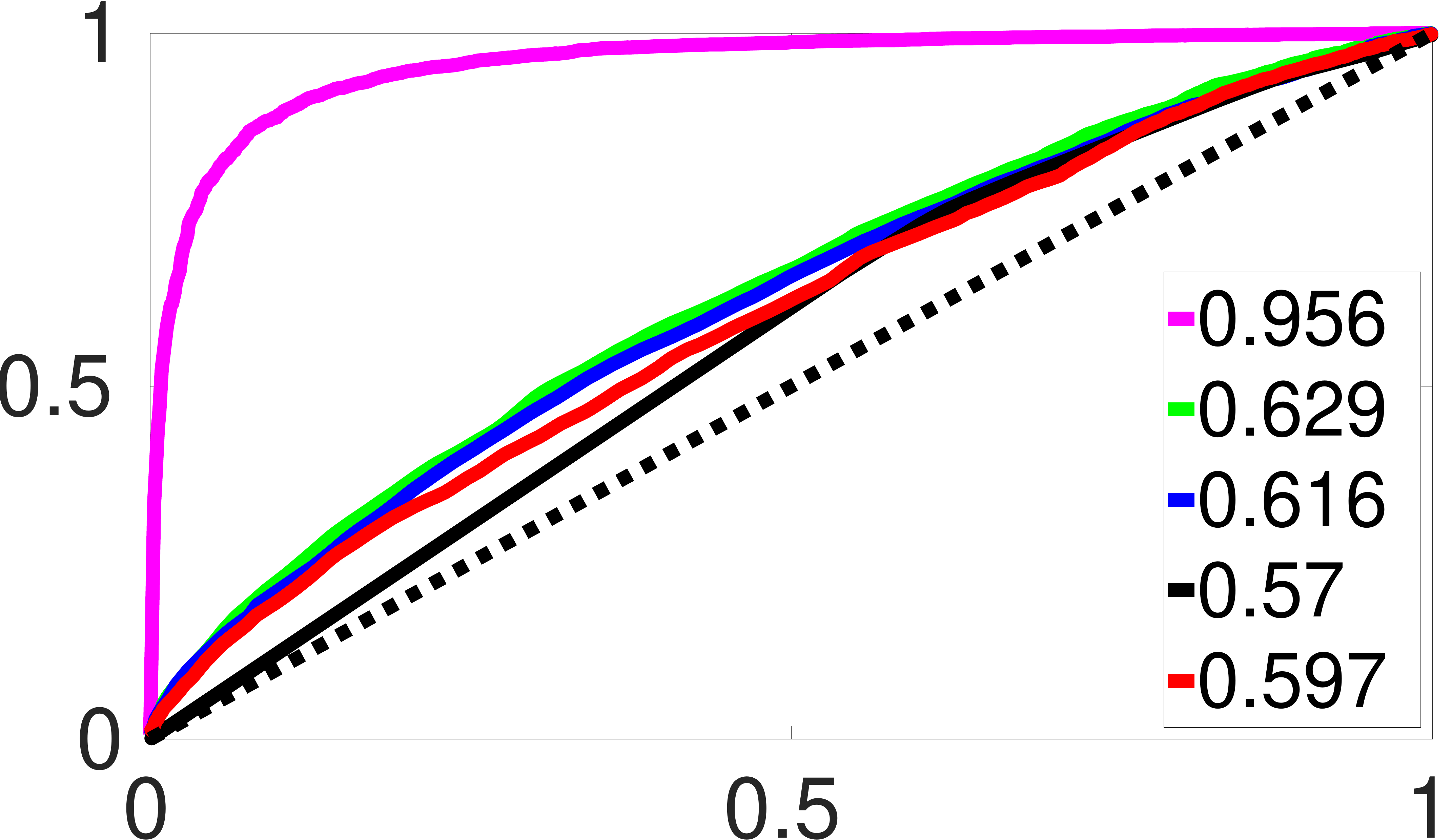}
&
\includegraphics[width=\linewidth]{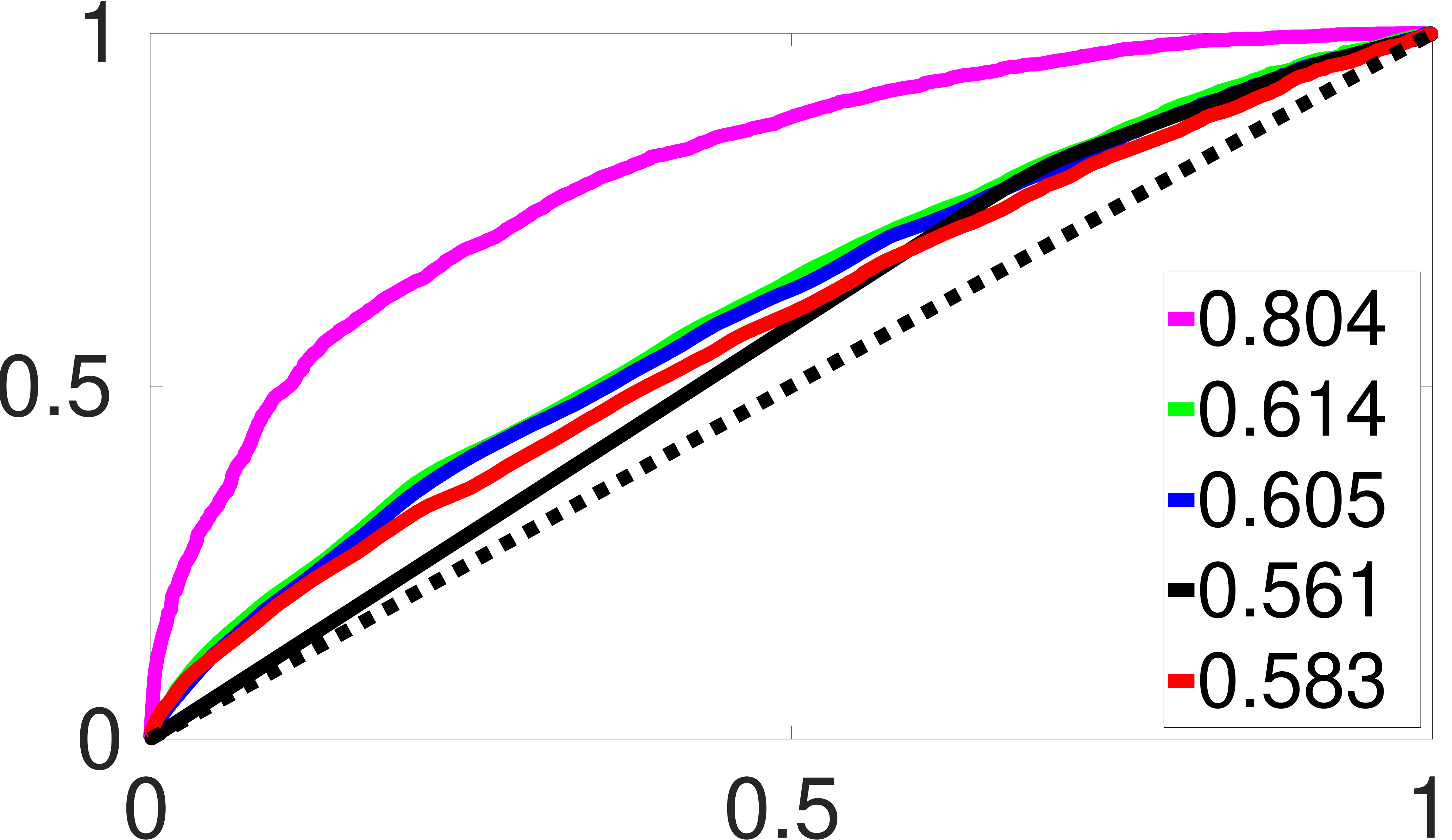}
\\
\includegraphics[width=\linewidth]{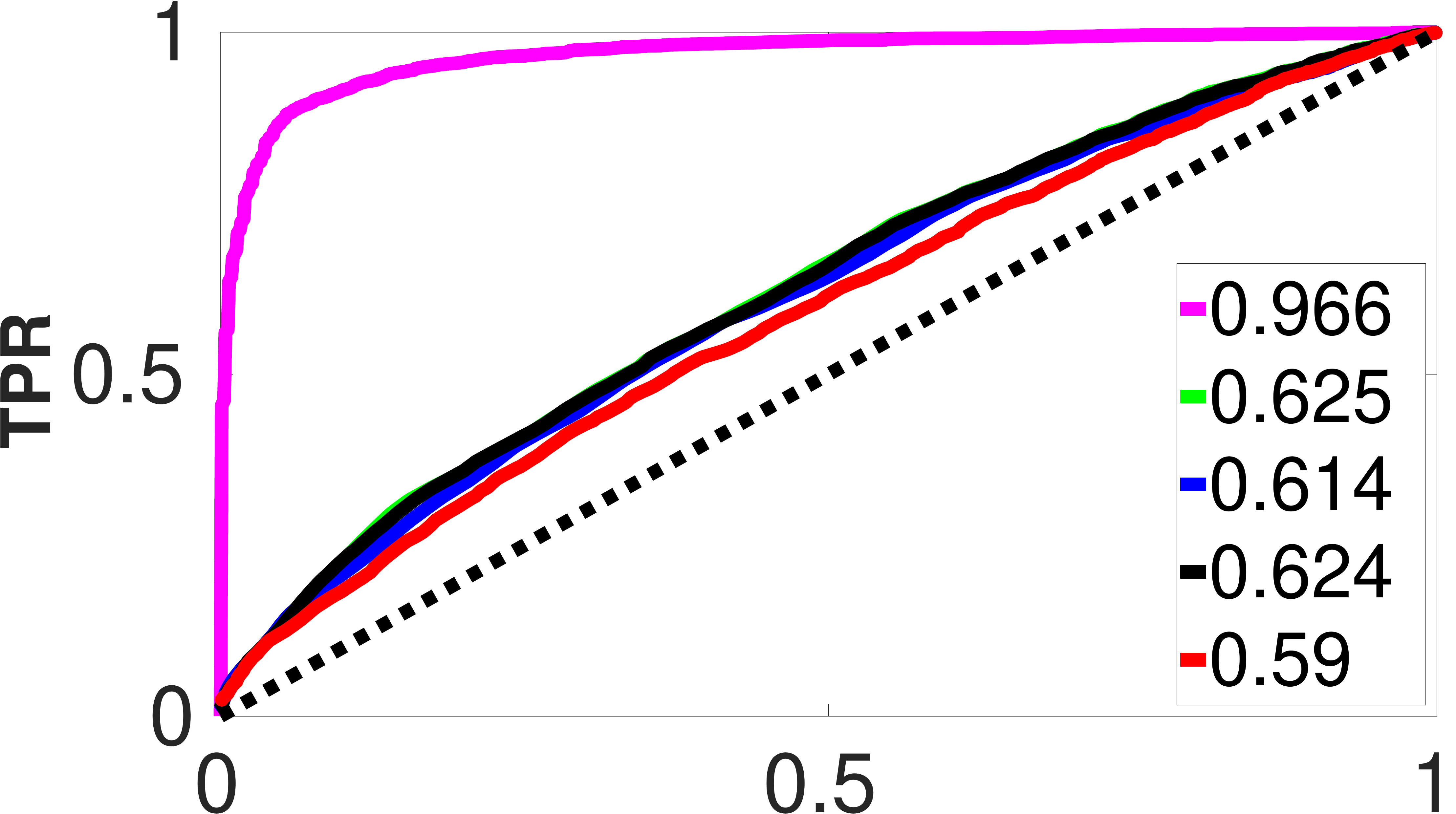}
&
 \includegraphics[width=\linewidth]{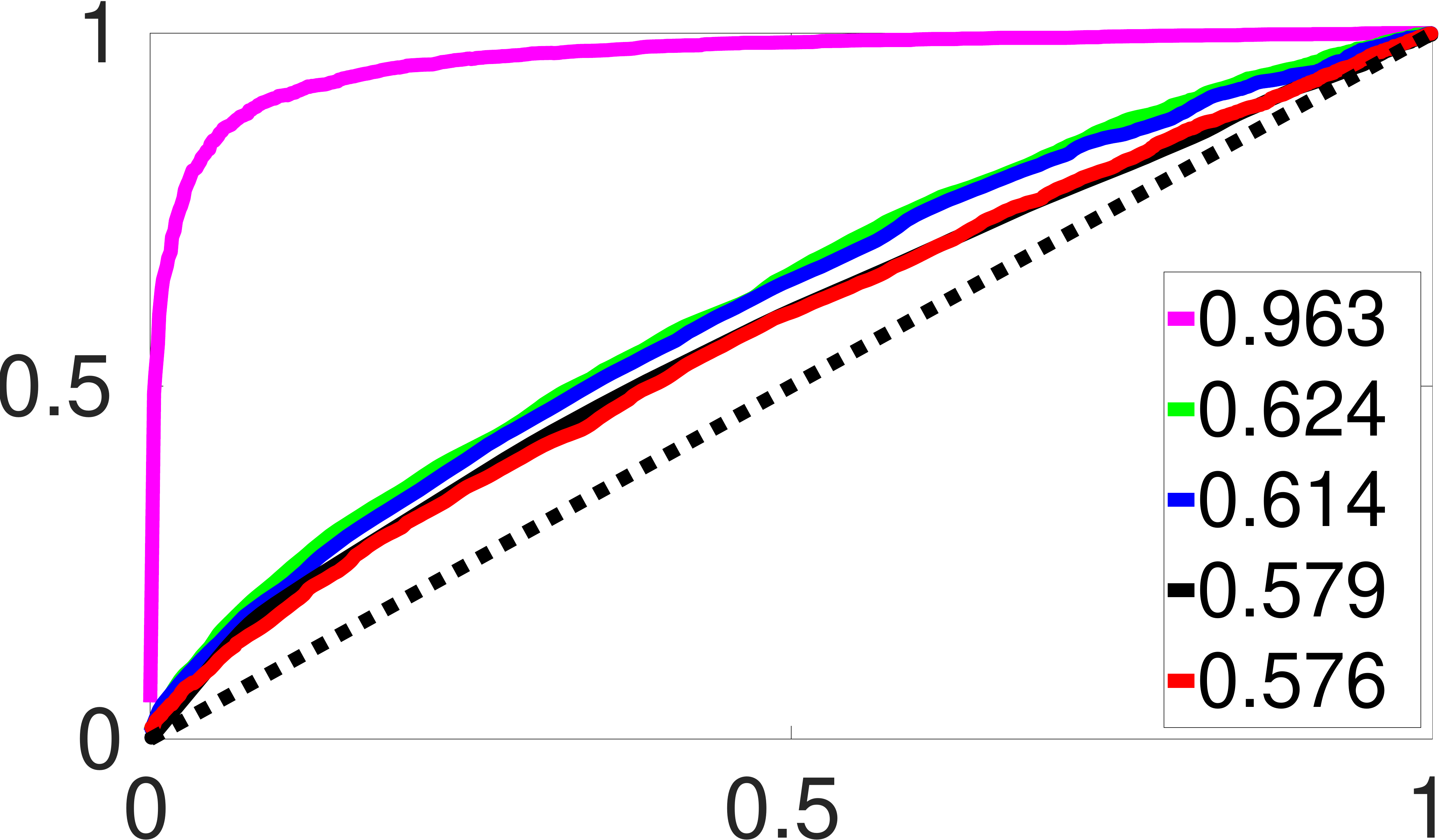}
&
\includegraphics[width=\linewidth]{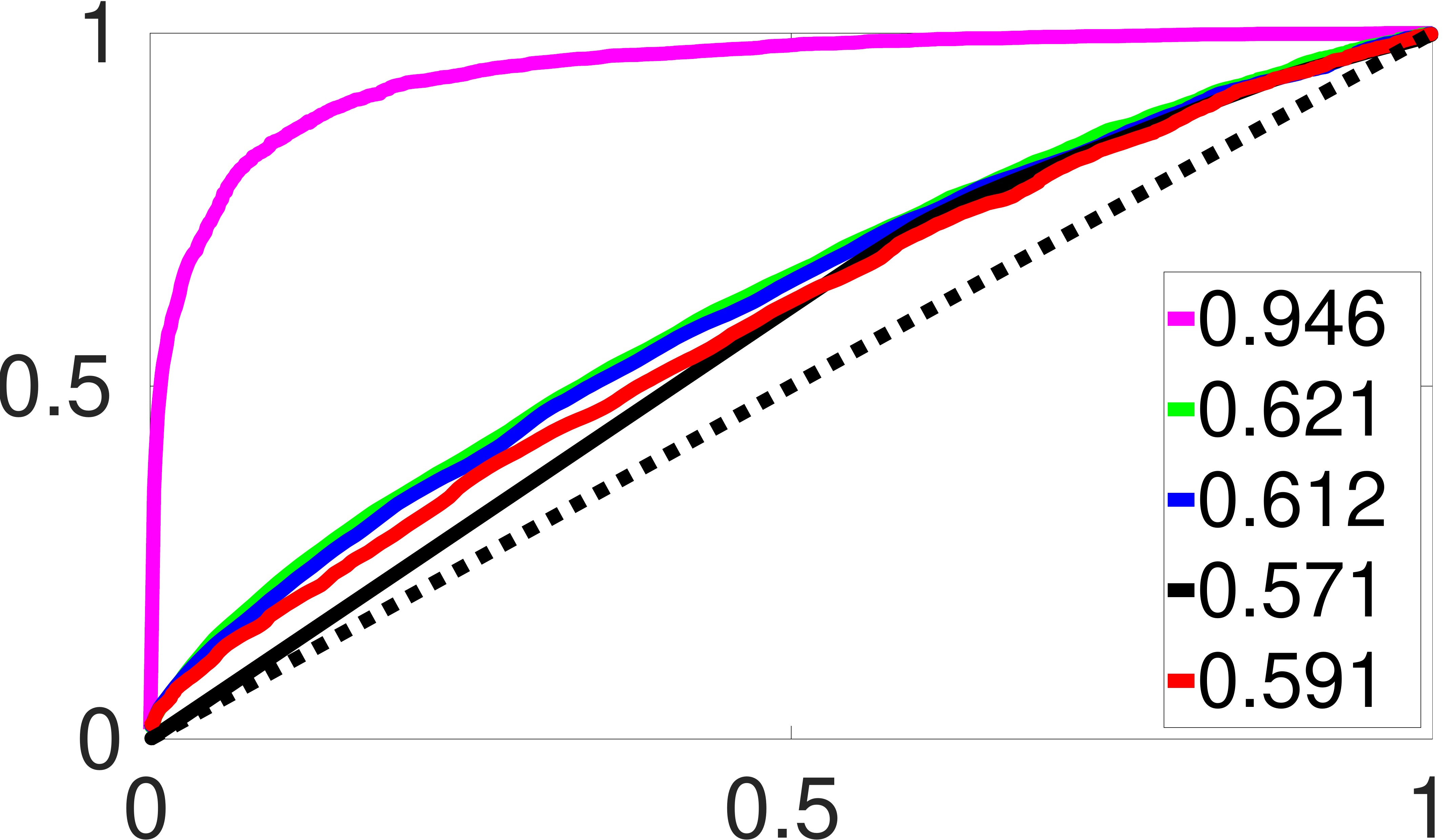}
&
\includegraphics[width=\linewidth]{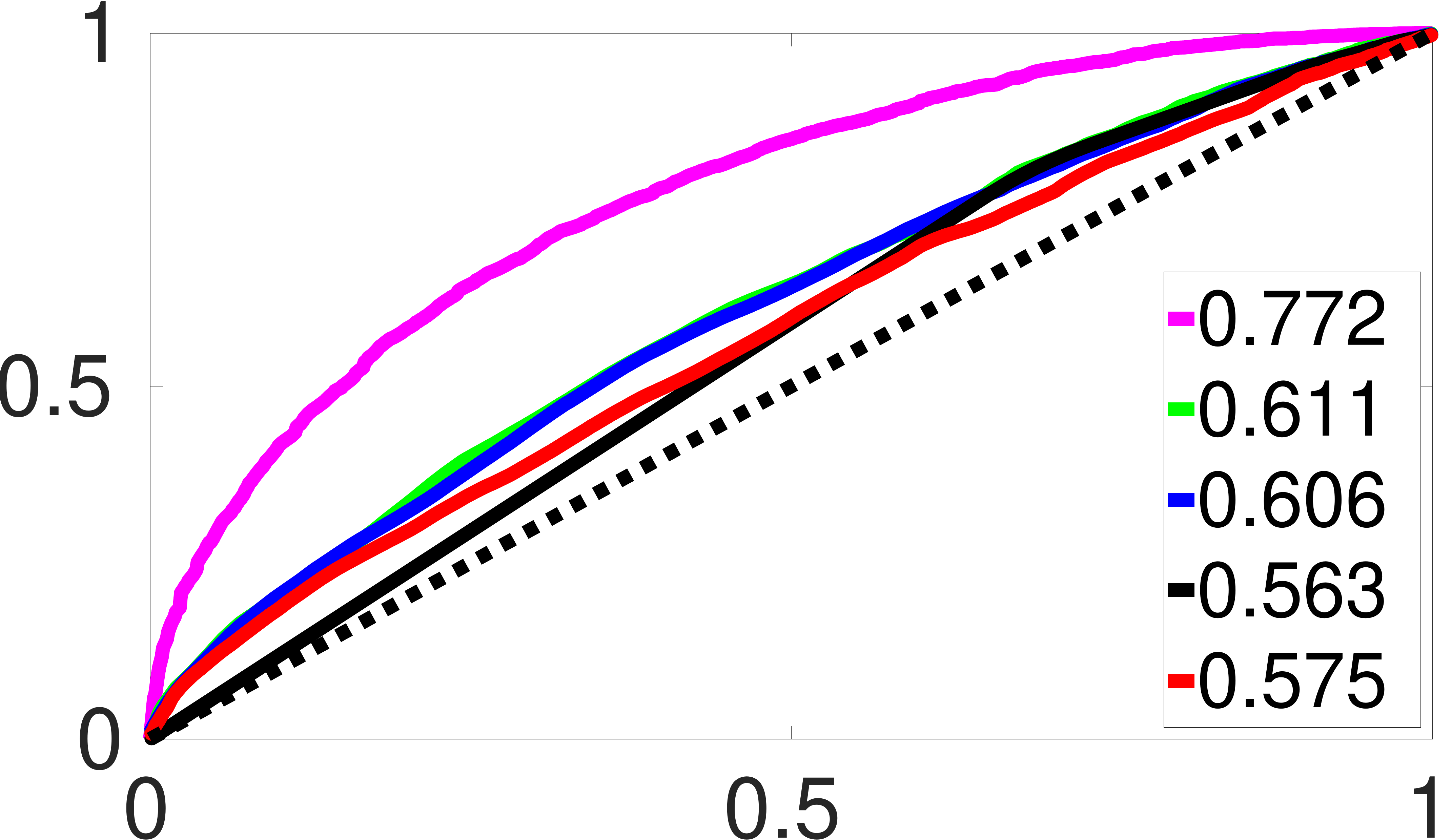}
\\
\includegraphics[width=\linewidth]{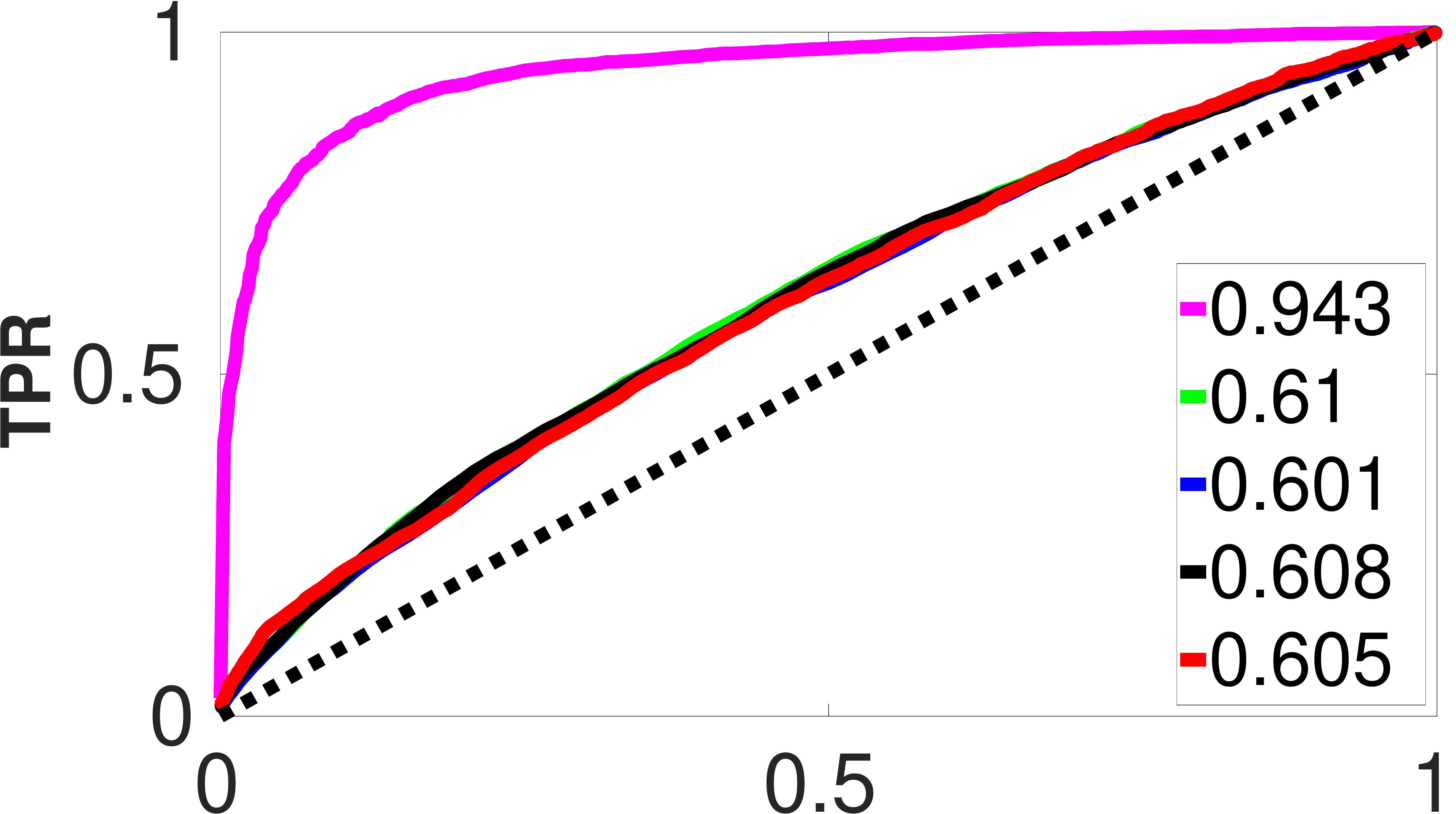}
&
 \includegraphics[width=\linewidth]{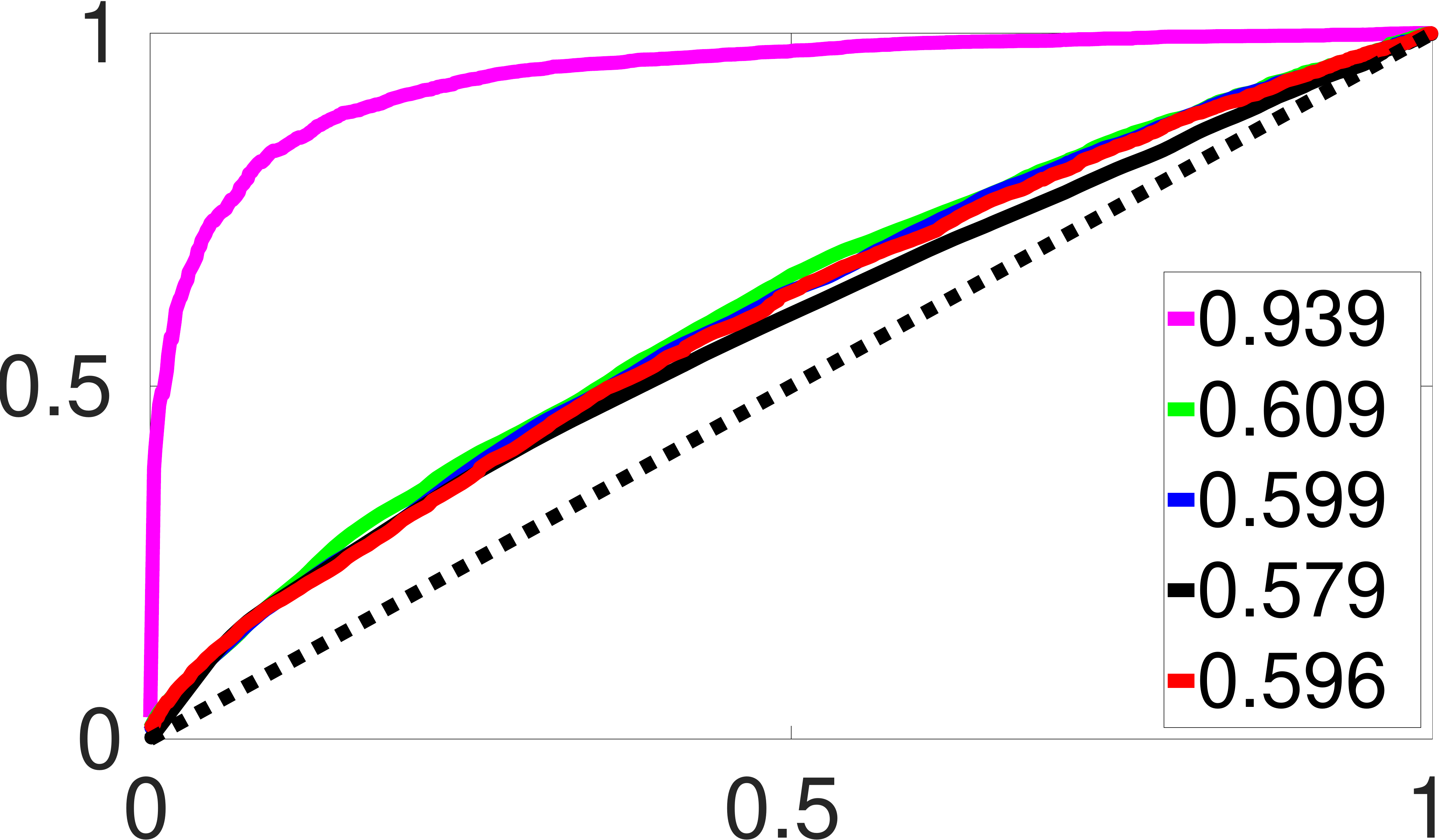}
&
\includegraphics[width=\linewidth]{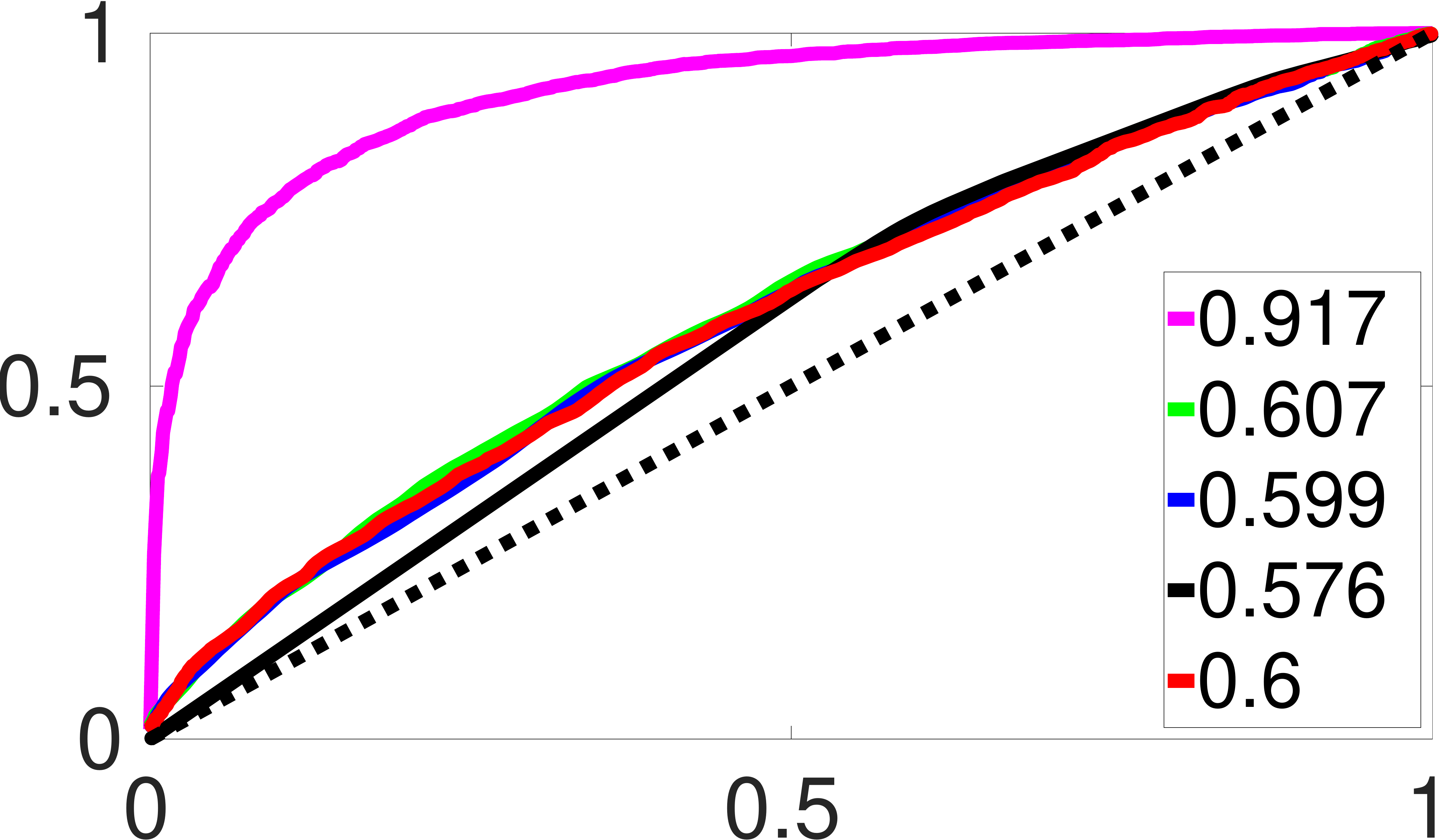}
&
\includegraphics[width=\linewidth]{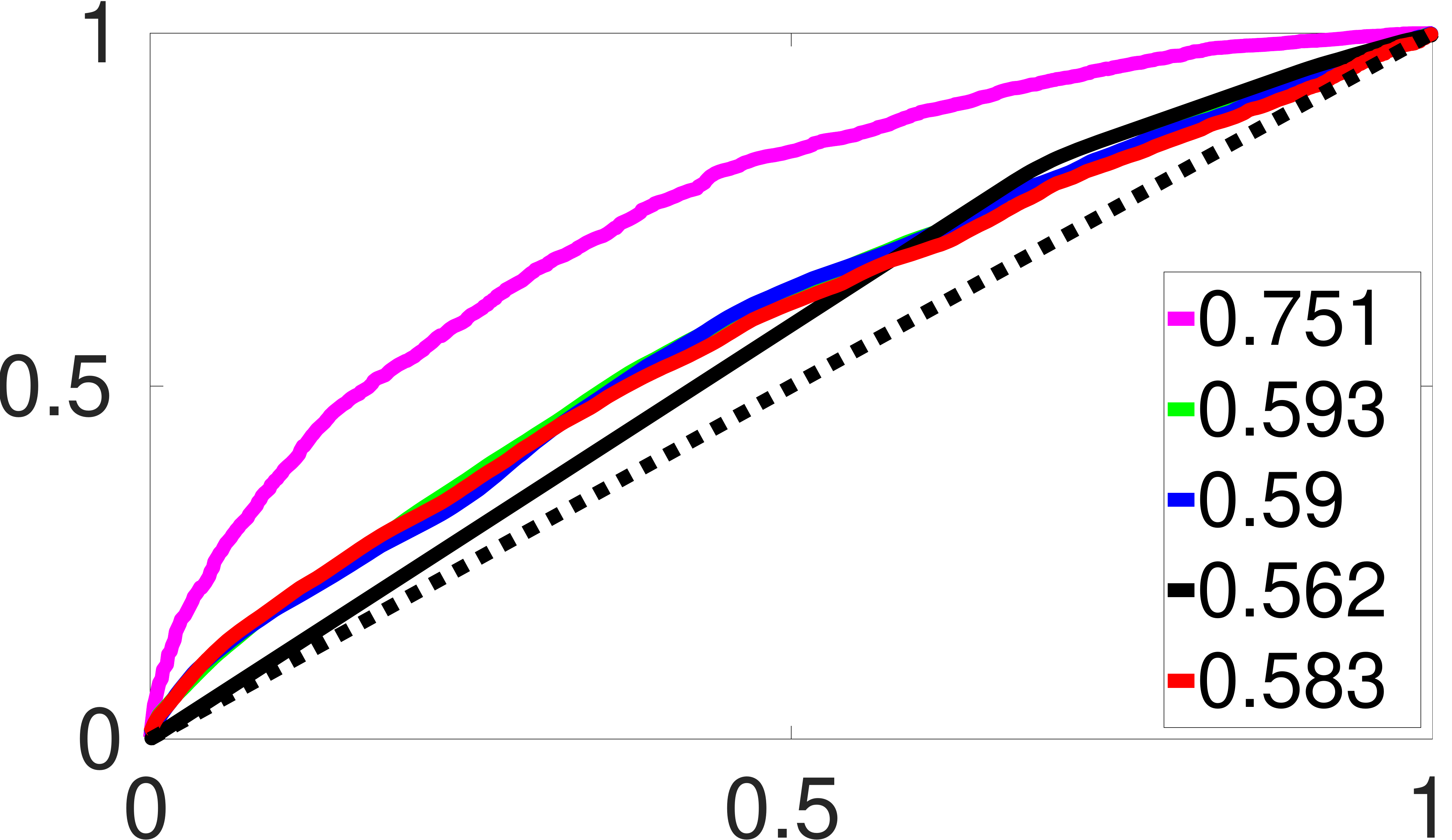}
\\
\includegraphics[width=\linewidth]{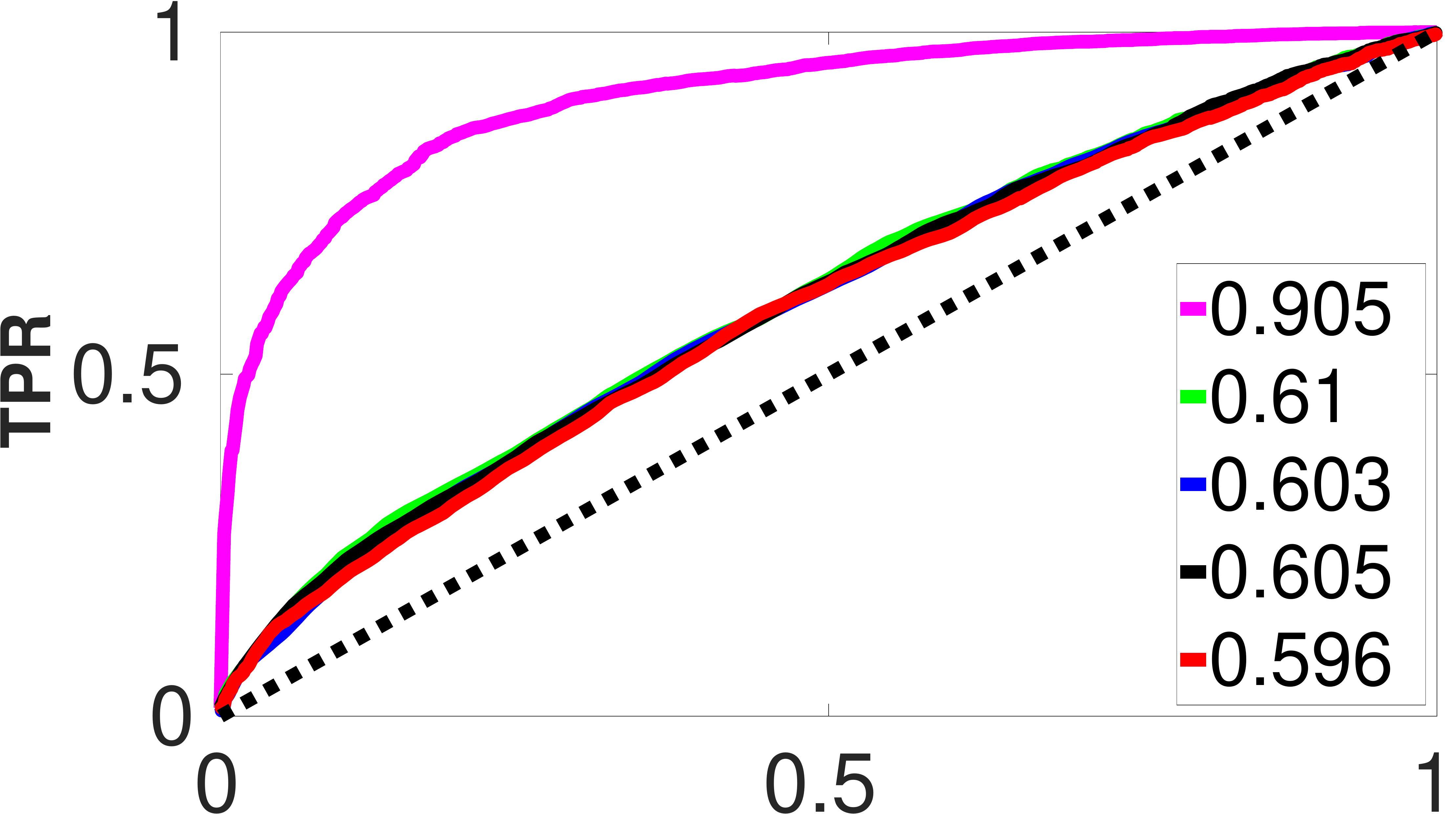}
&
\includegraphics[width=\linewidth]{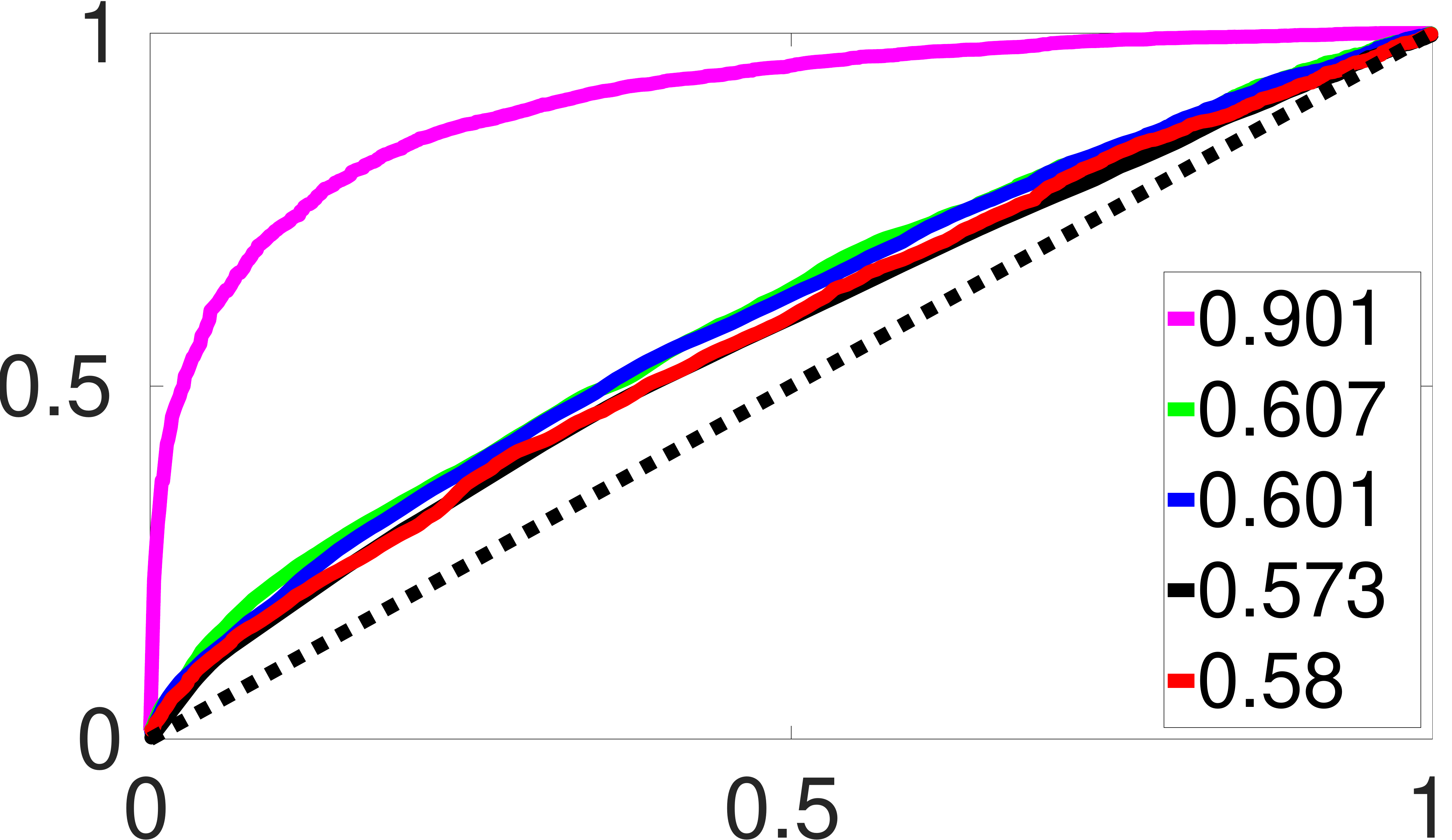}
&
\includegraphics[width=\linewidth]{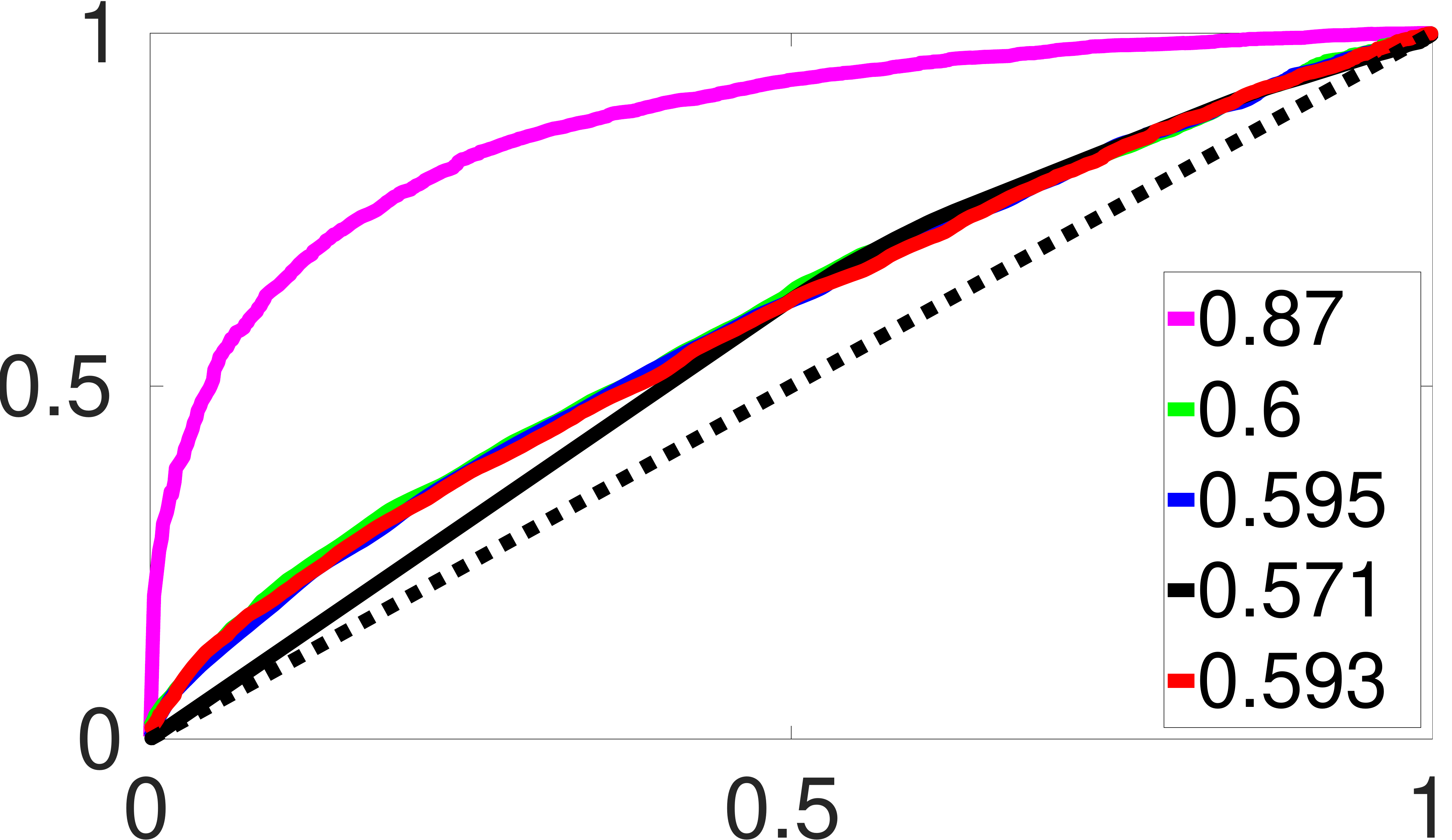}
&
\includegraphics[width=\linewidth]{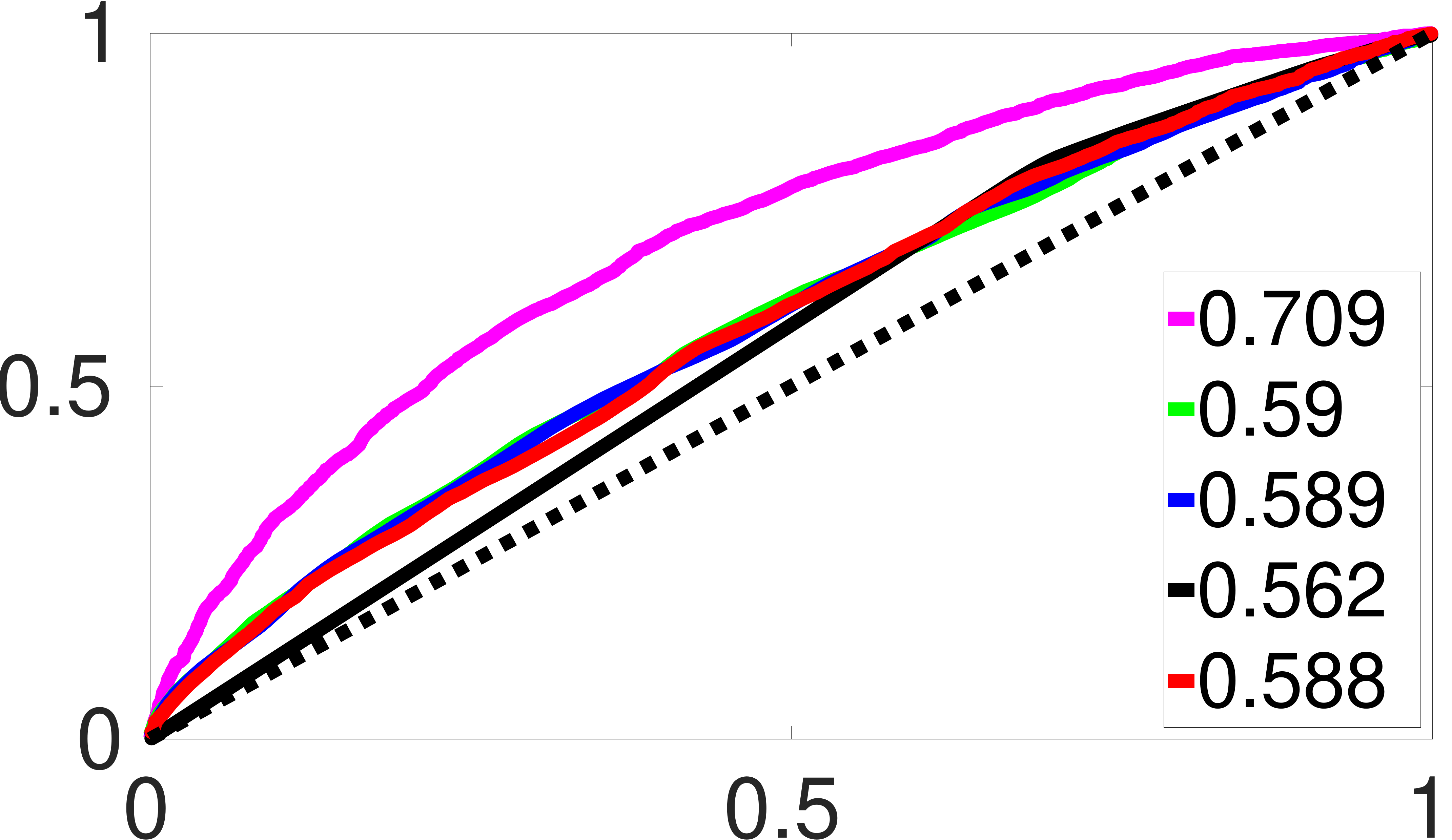}
\\
\includegraphics[width=\linewidth]{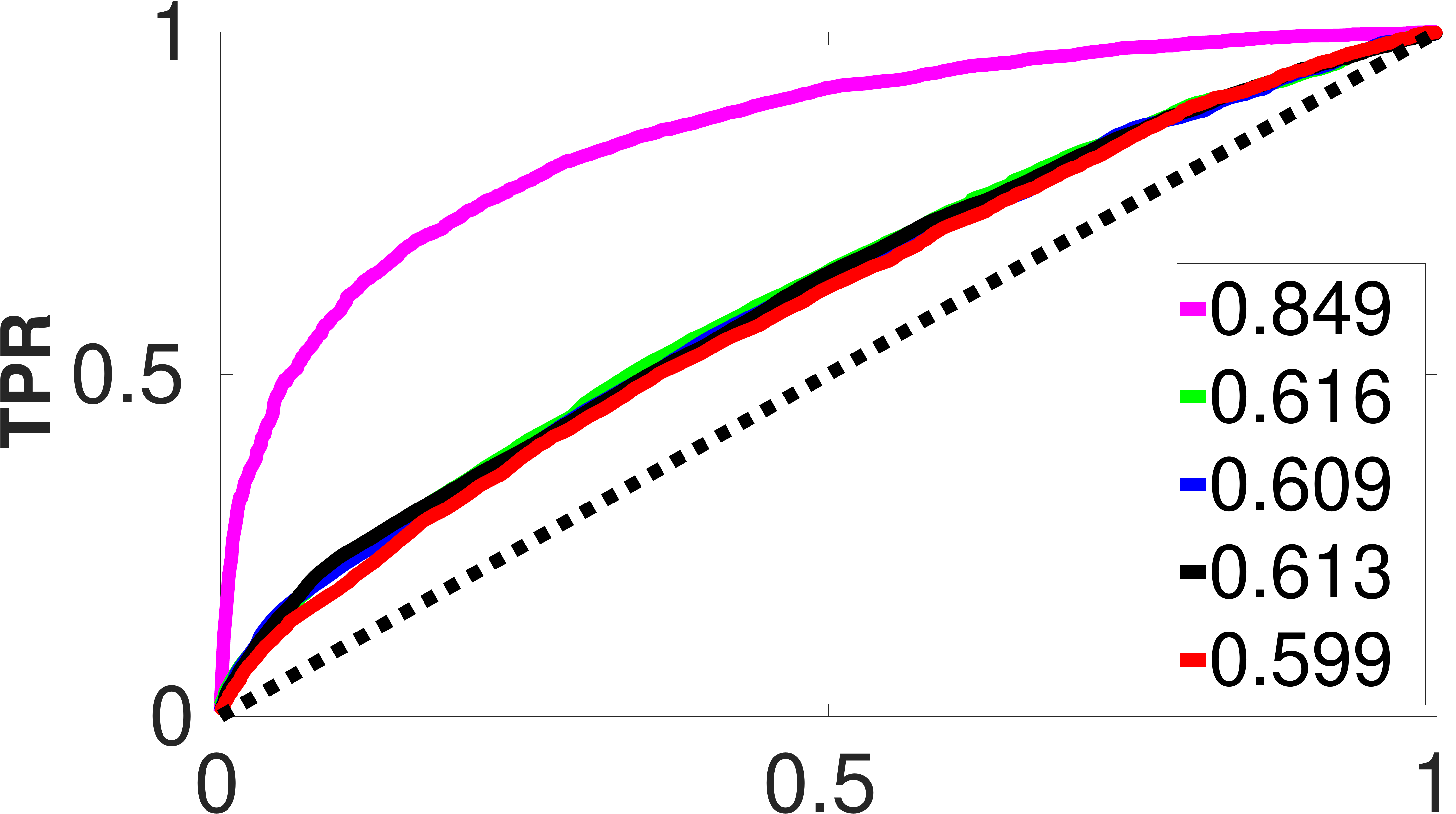}
&
 \includegraphics[width=\linewidth]{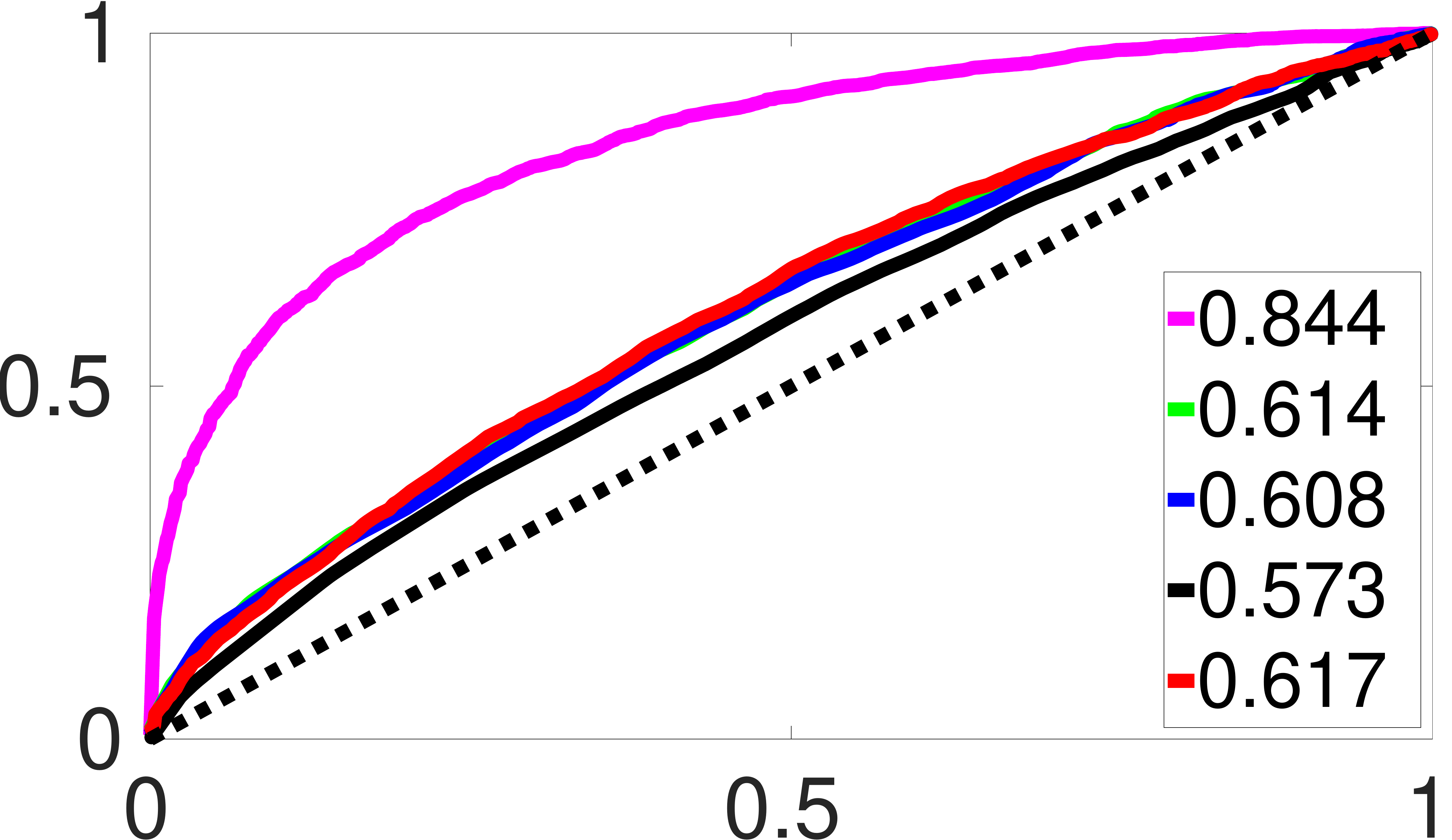}
&
\includegraphics[width=\linewidth]{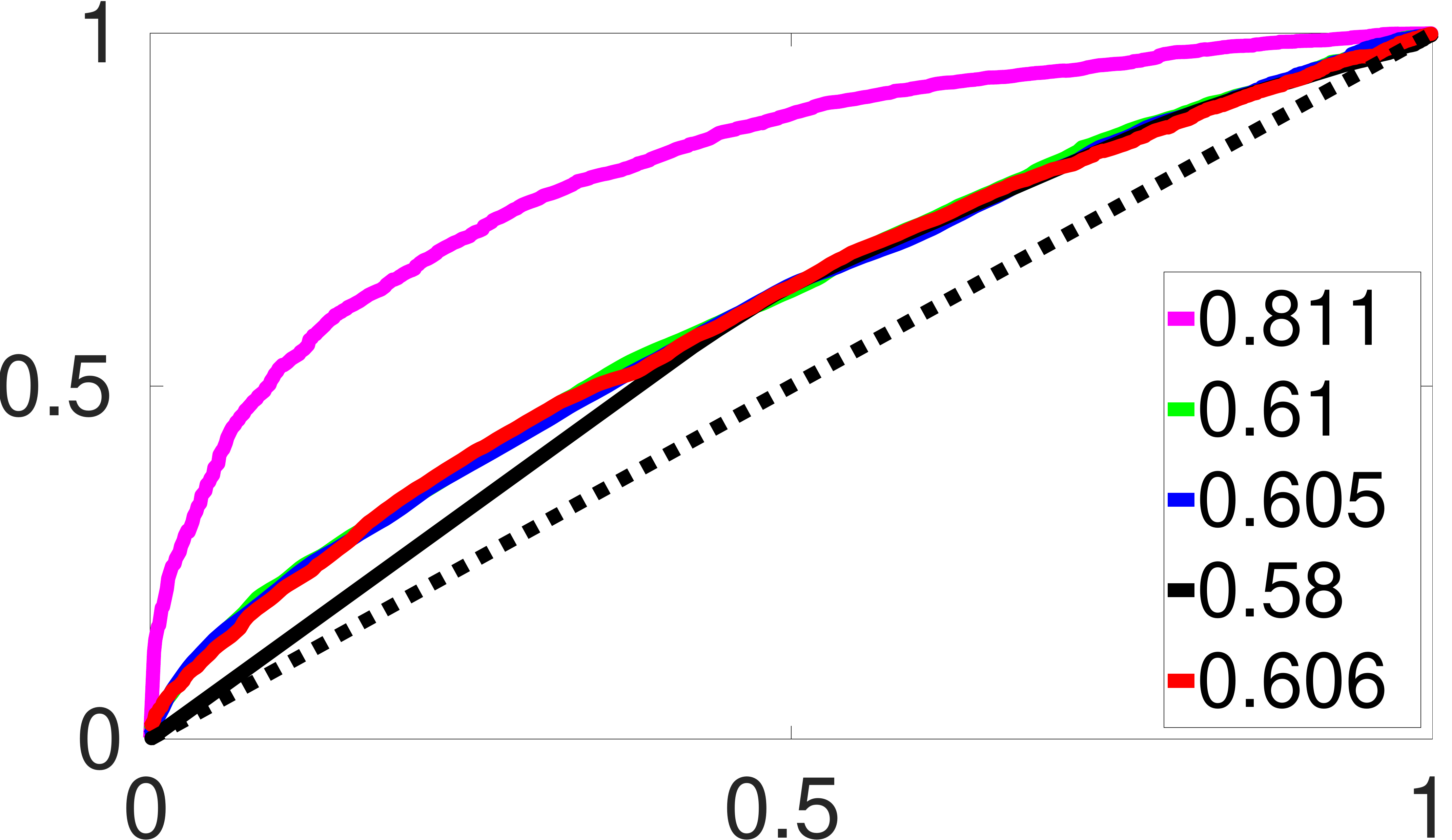}
&
\includegraphics[width=\linewidth]{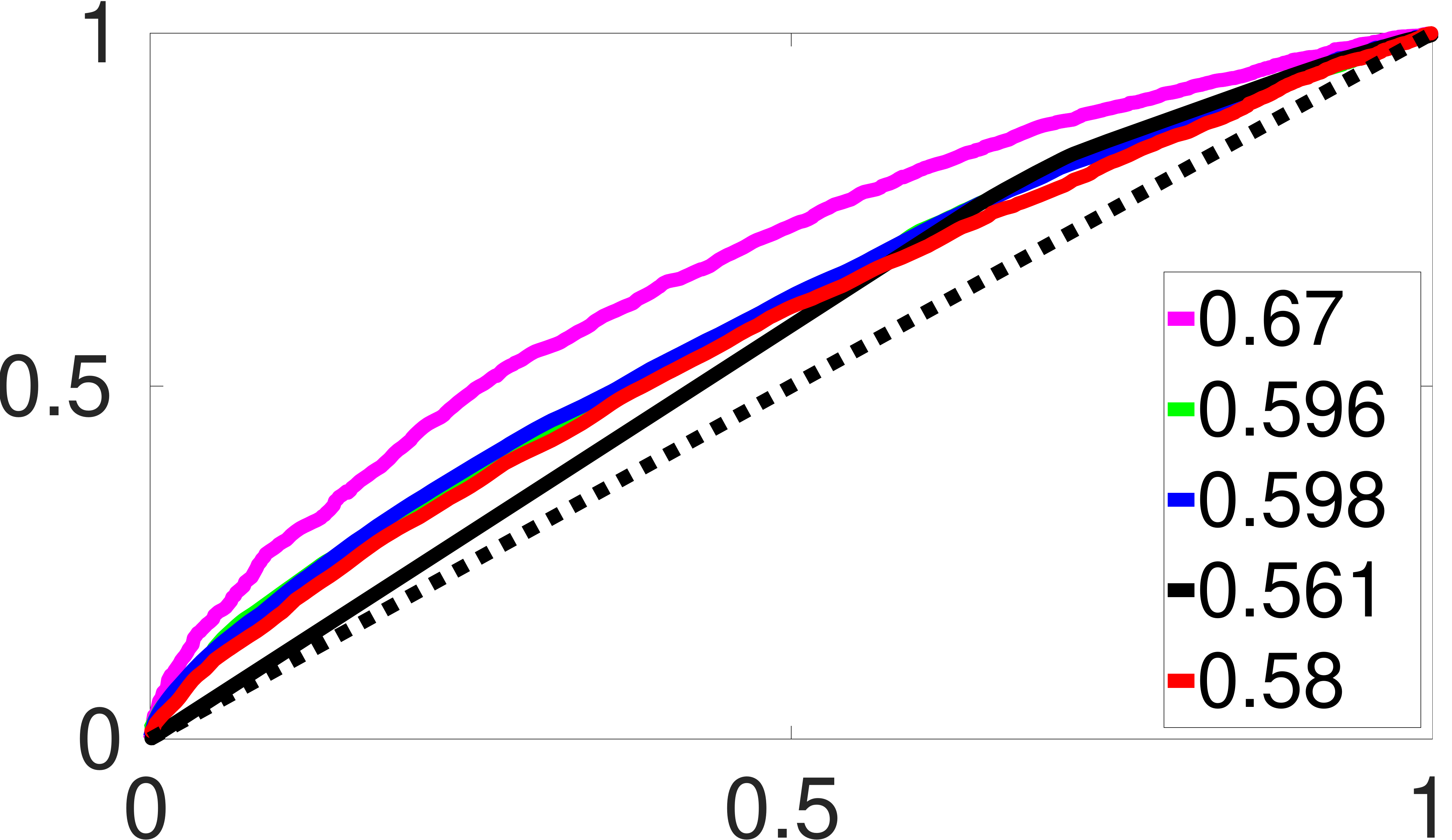}
\\
\includegraphics[width=\linewidth]{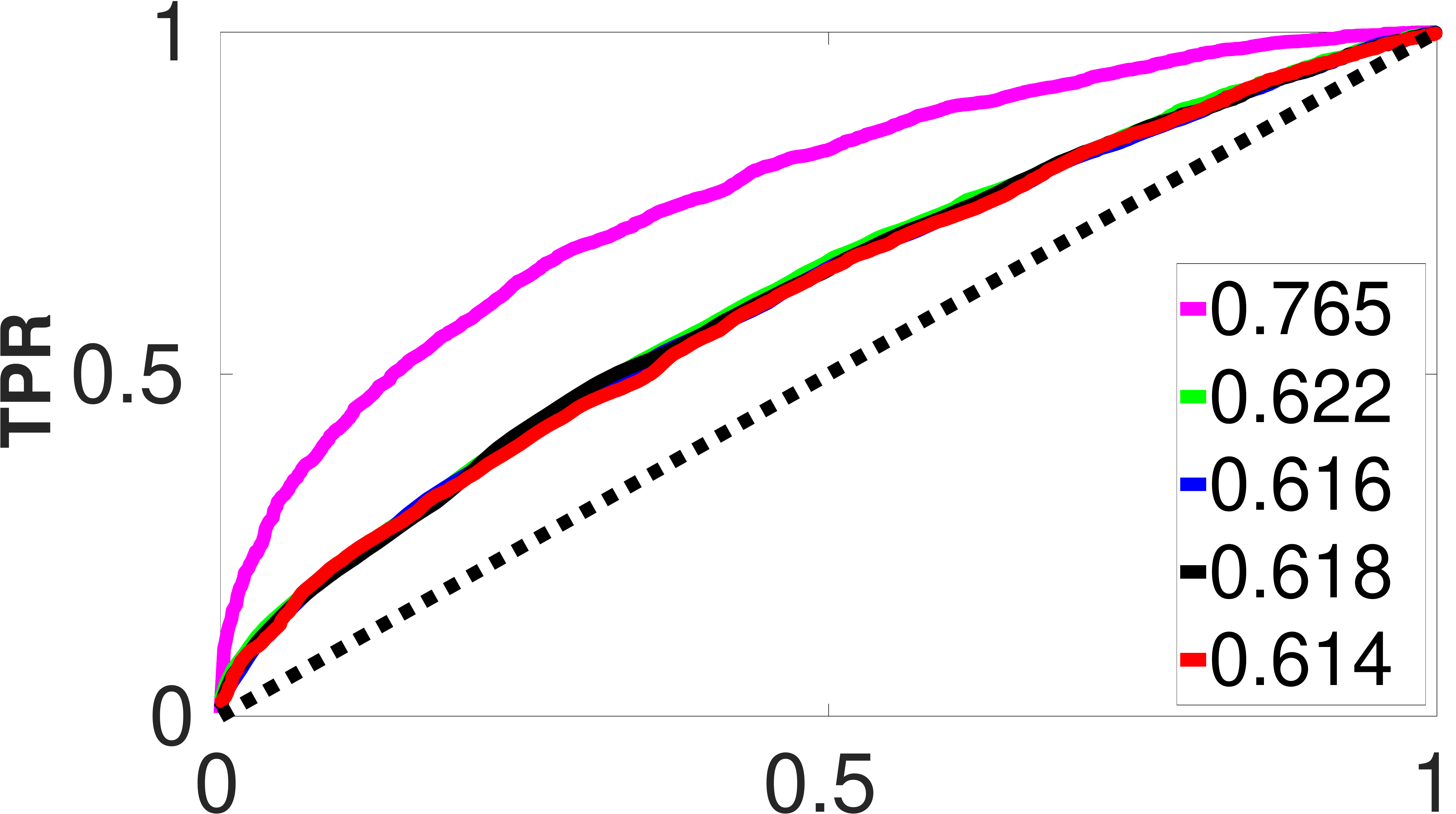}
&
 \includegraphics[width=\linewidth]{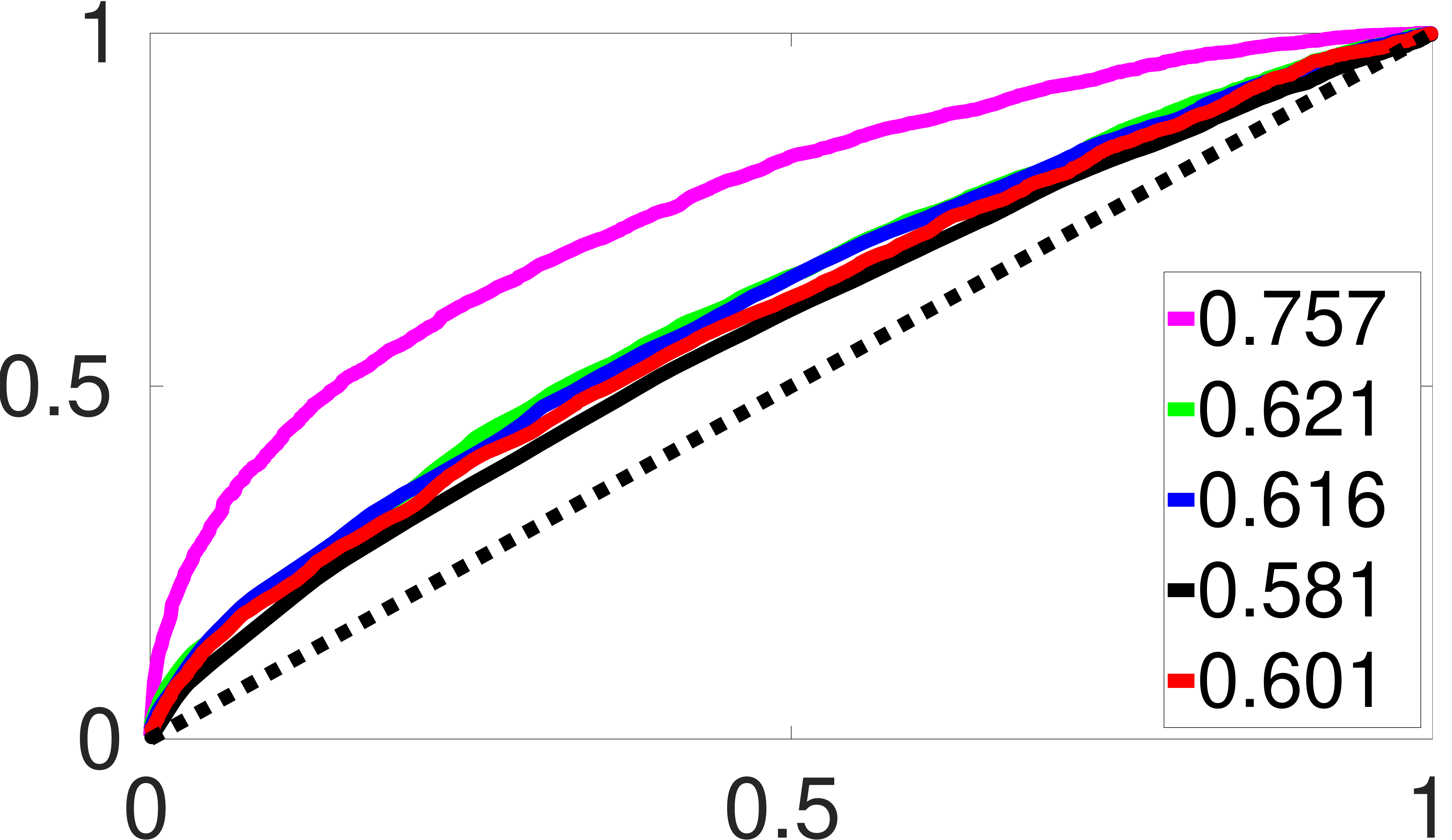}
&
\includegraphics[width=\linewidth]{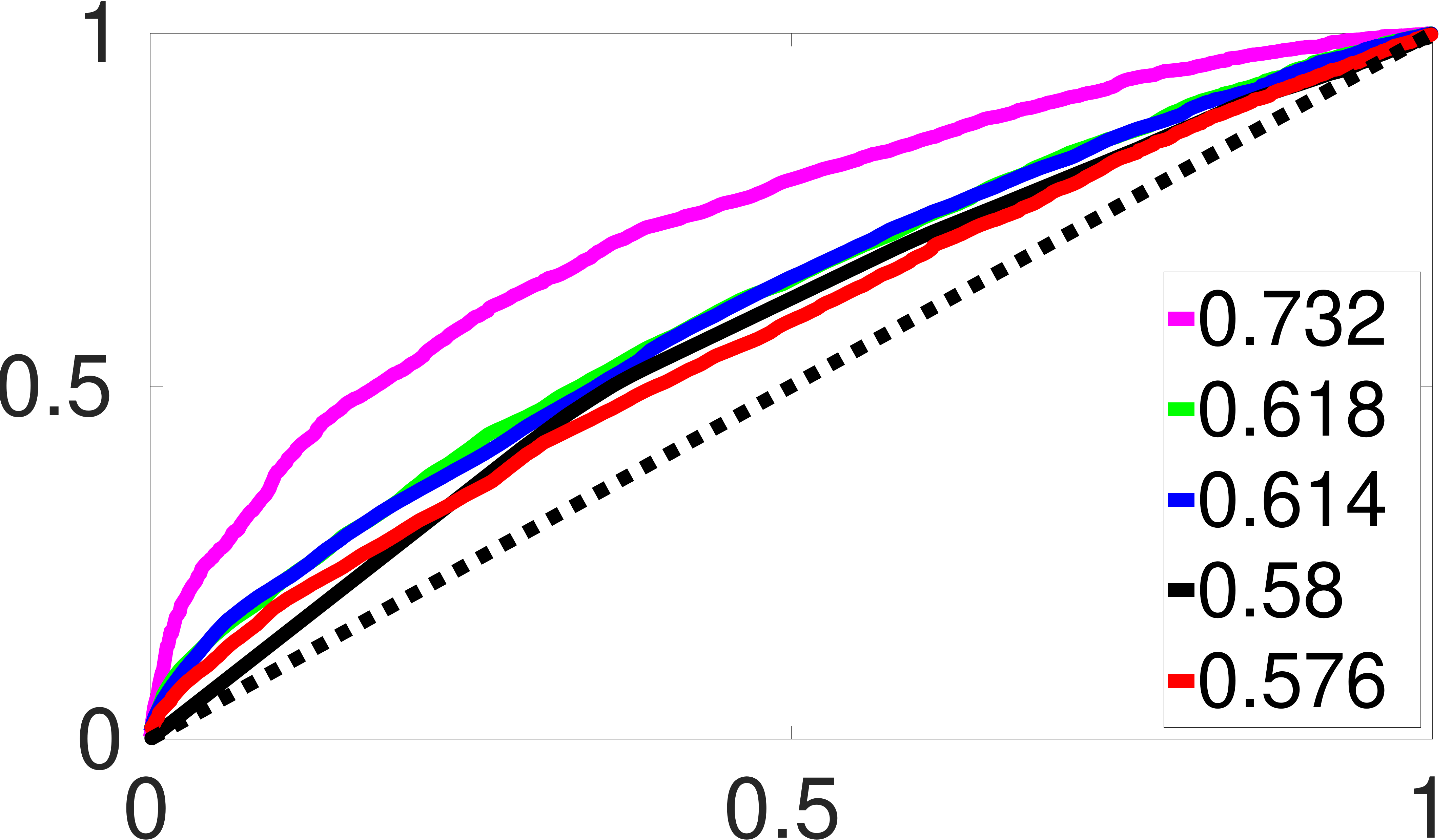}
&
\includegraphics[width=\linewidth]{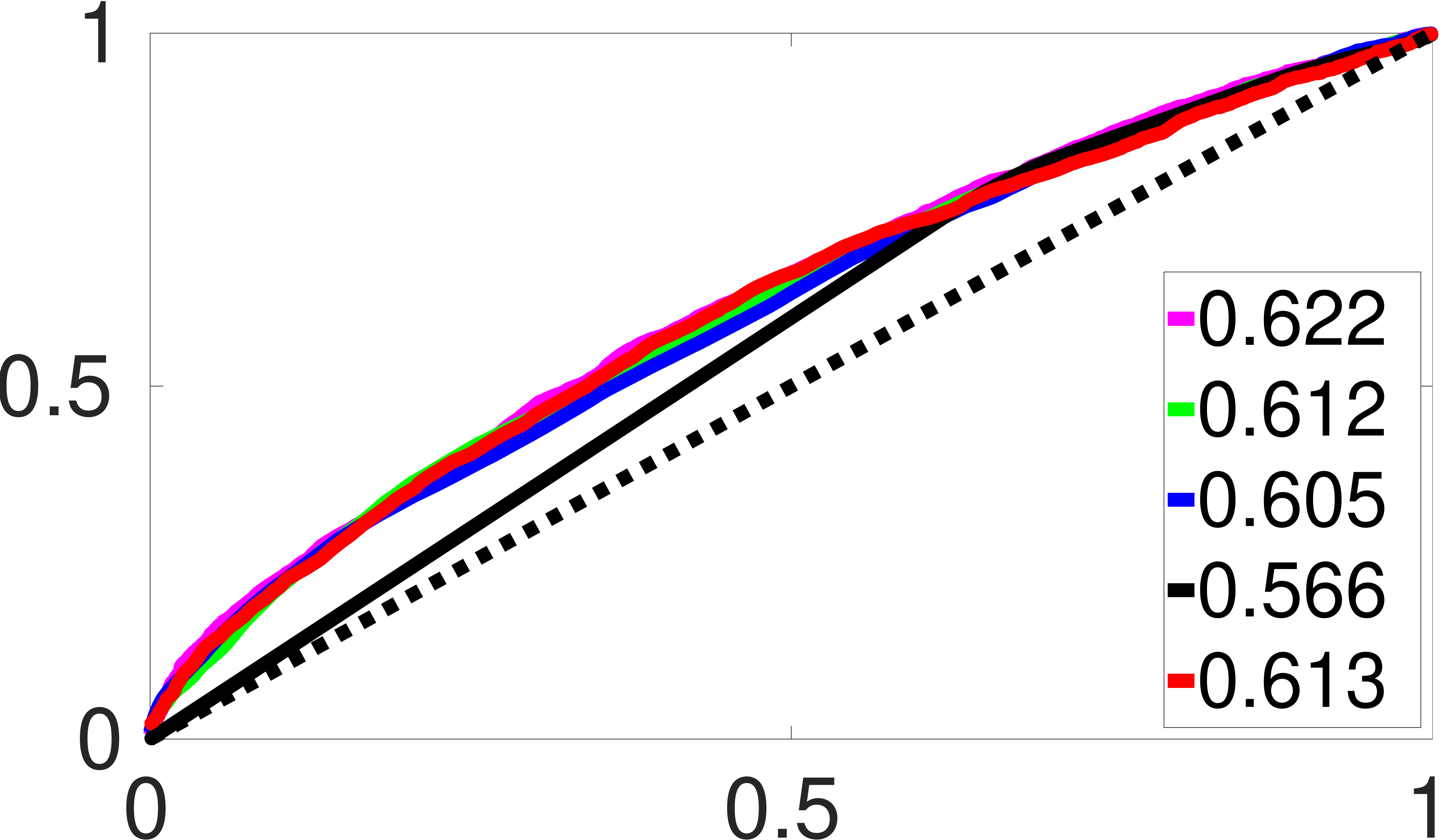}
\\
\includegraphics[width=\linewidth]{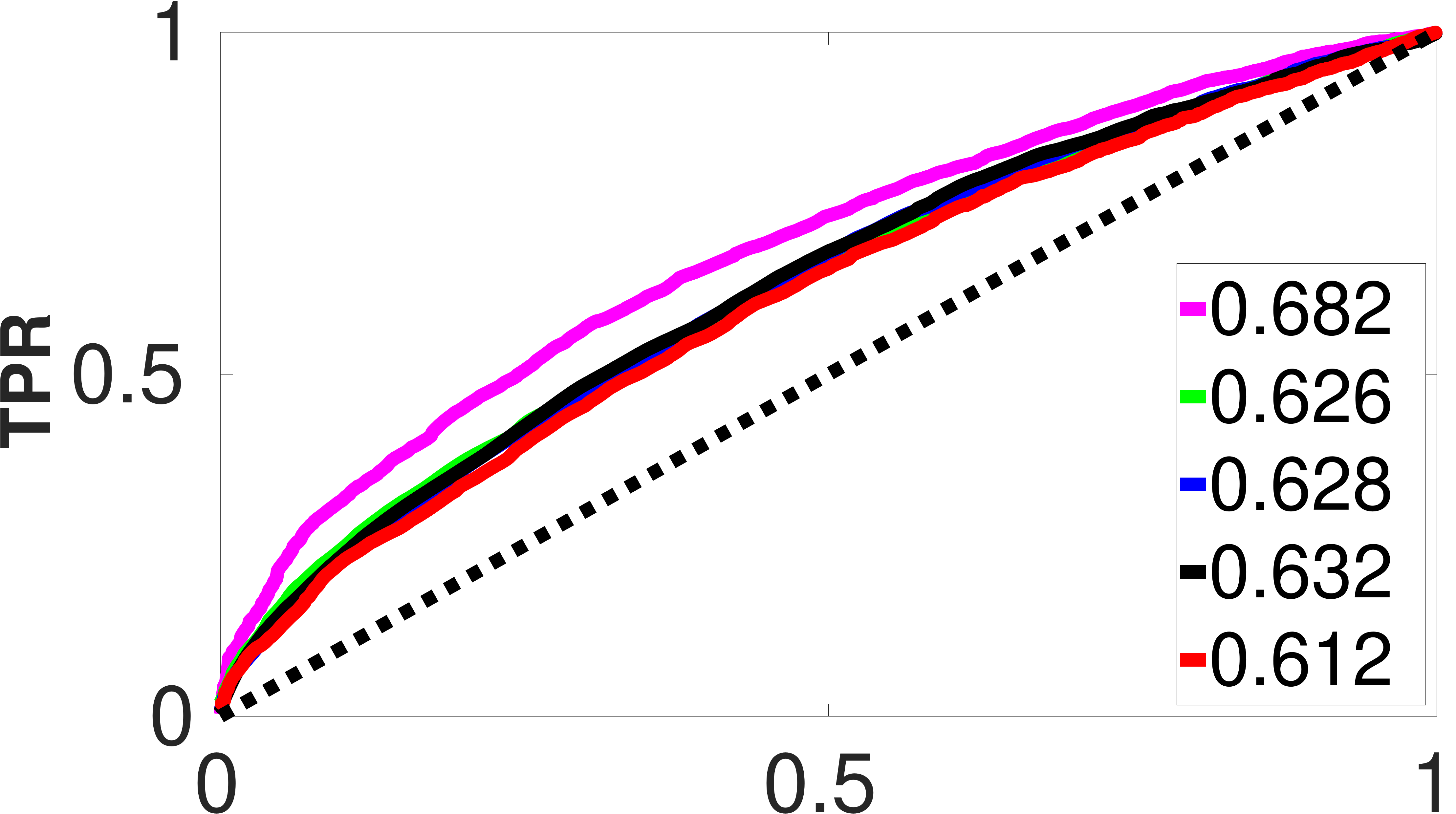}
&
 \includegraphics[width=\linewidth]{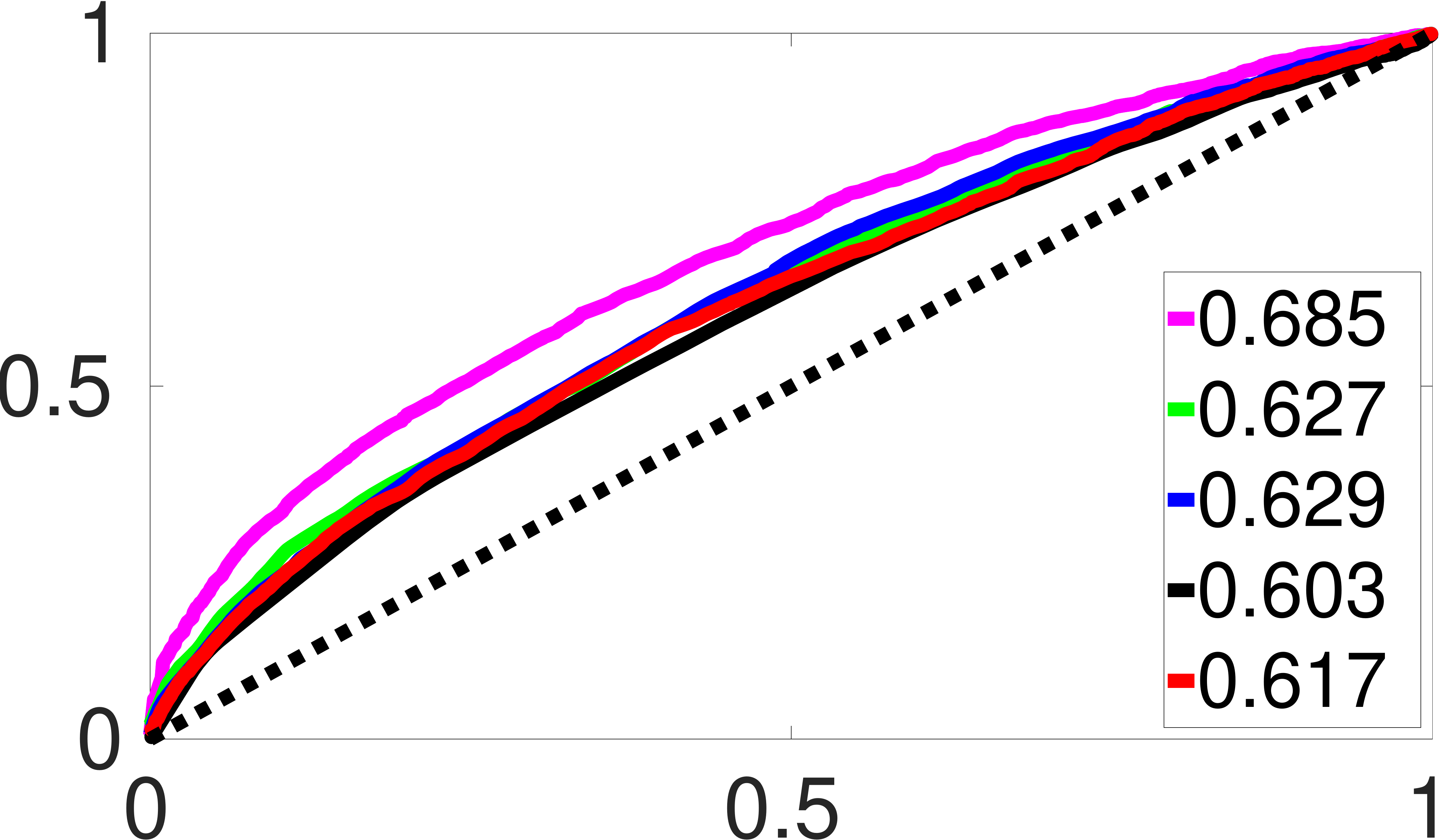}
&
\includegraphics[width=\linewidth]{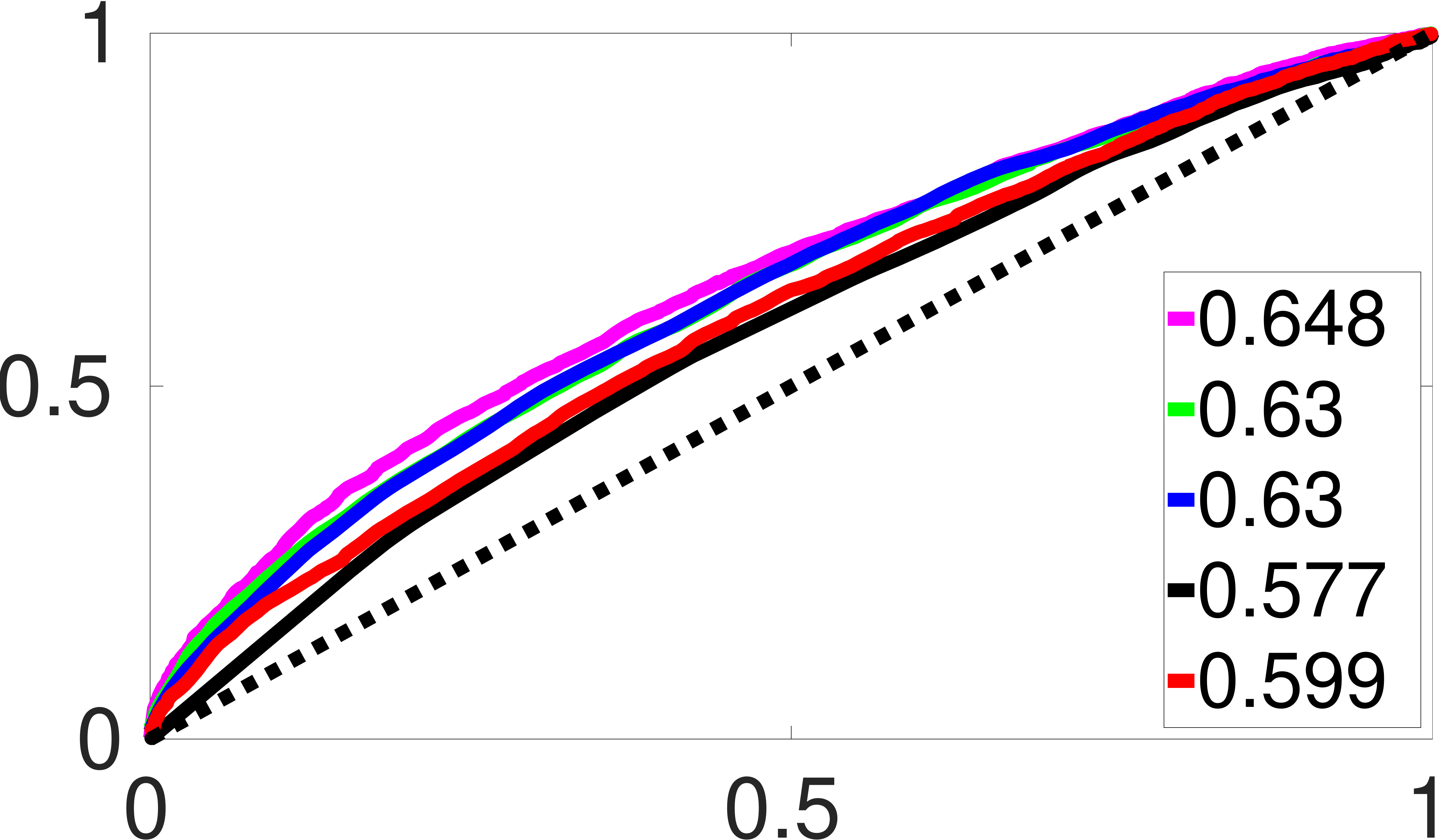}
&
\includegraphics[width=\linewidth]{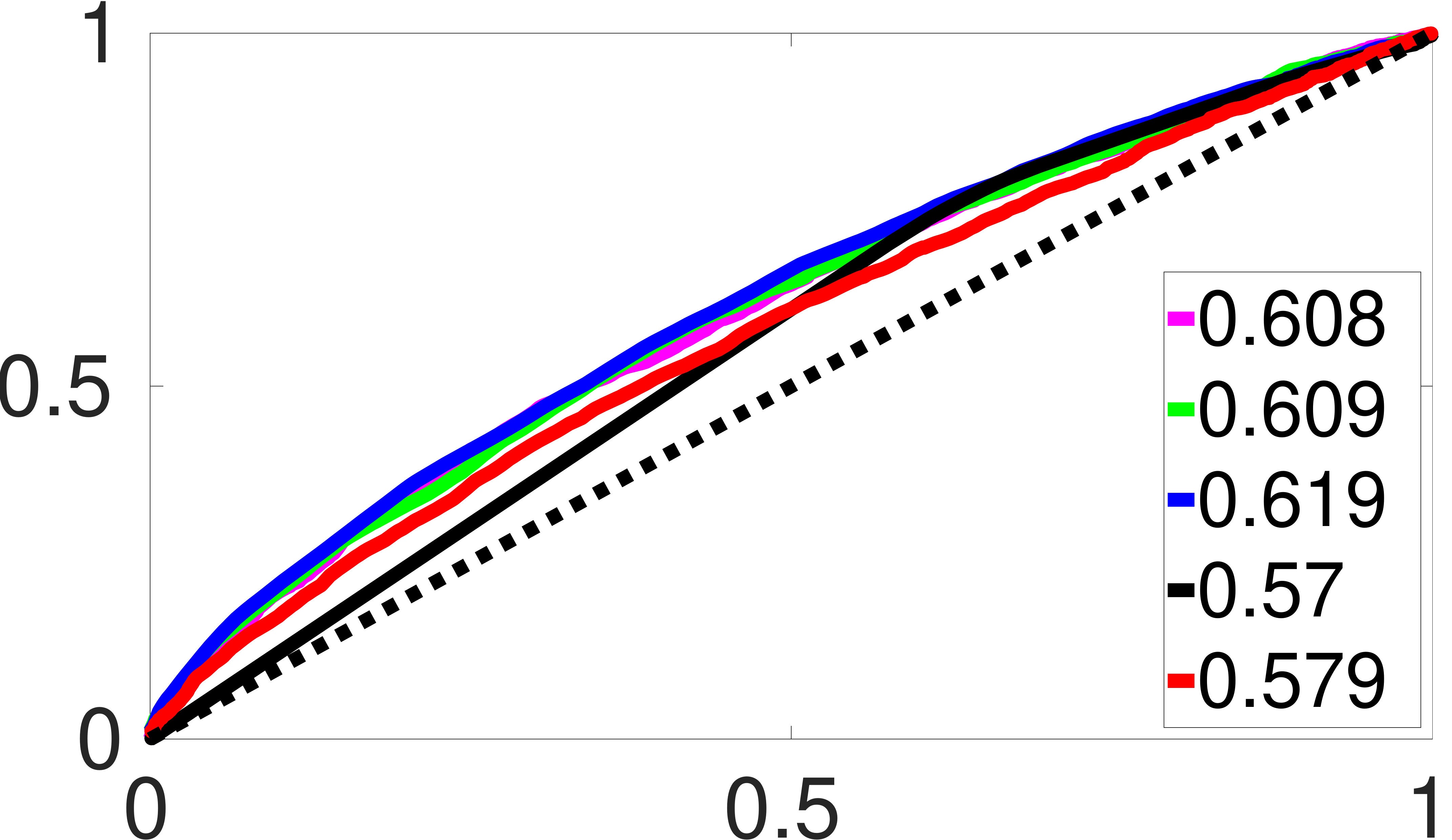}
\\
\includegraphics[width=\linewidth]{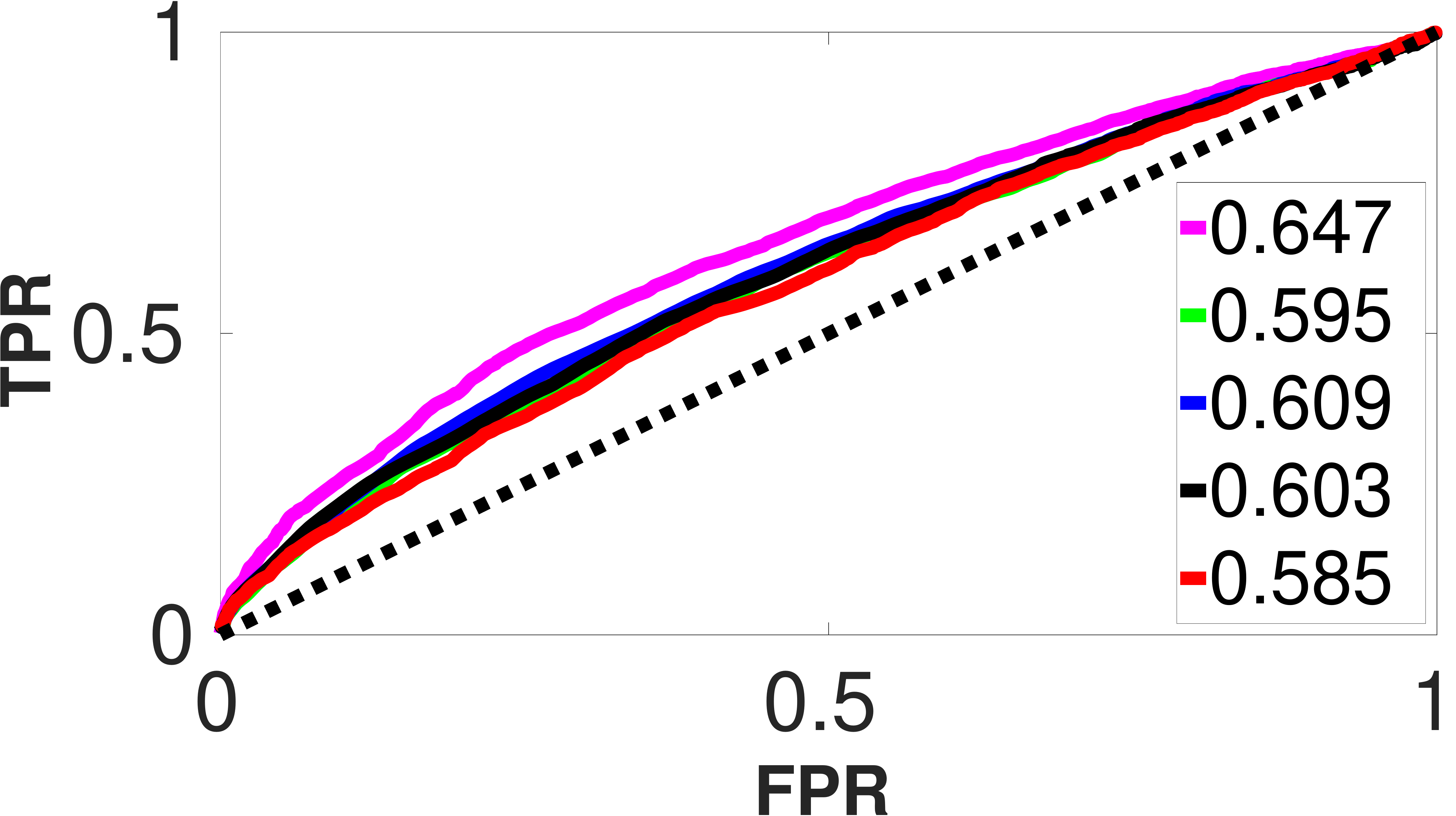}
&
\includegraphics[width=\linewidth]{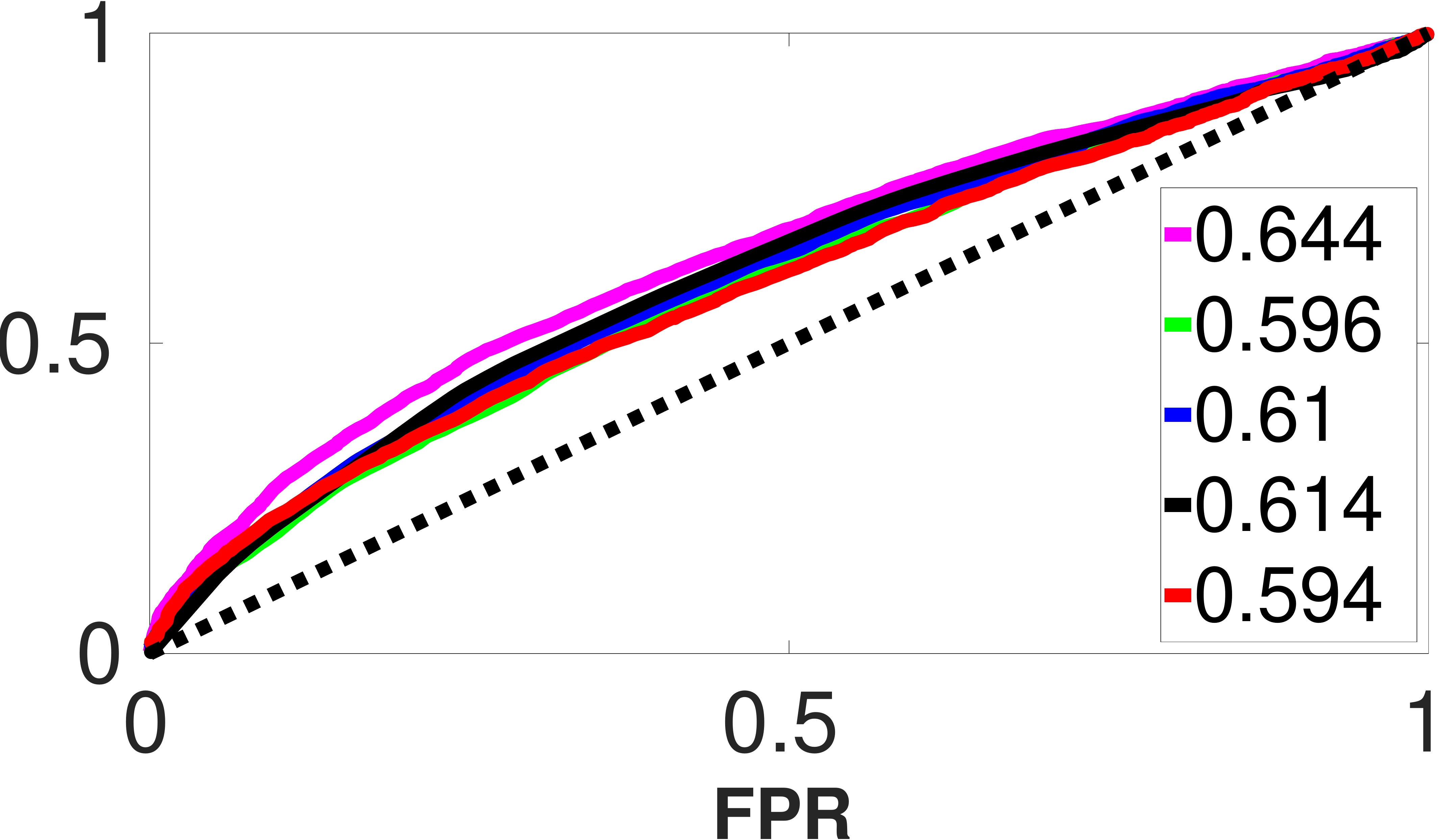}
&
\includegraphics[width=\linewidth]{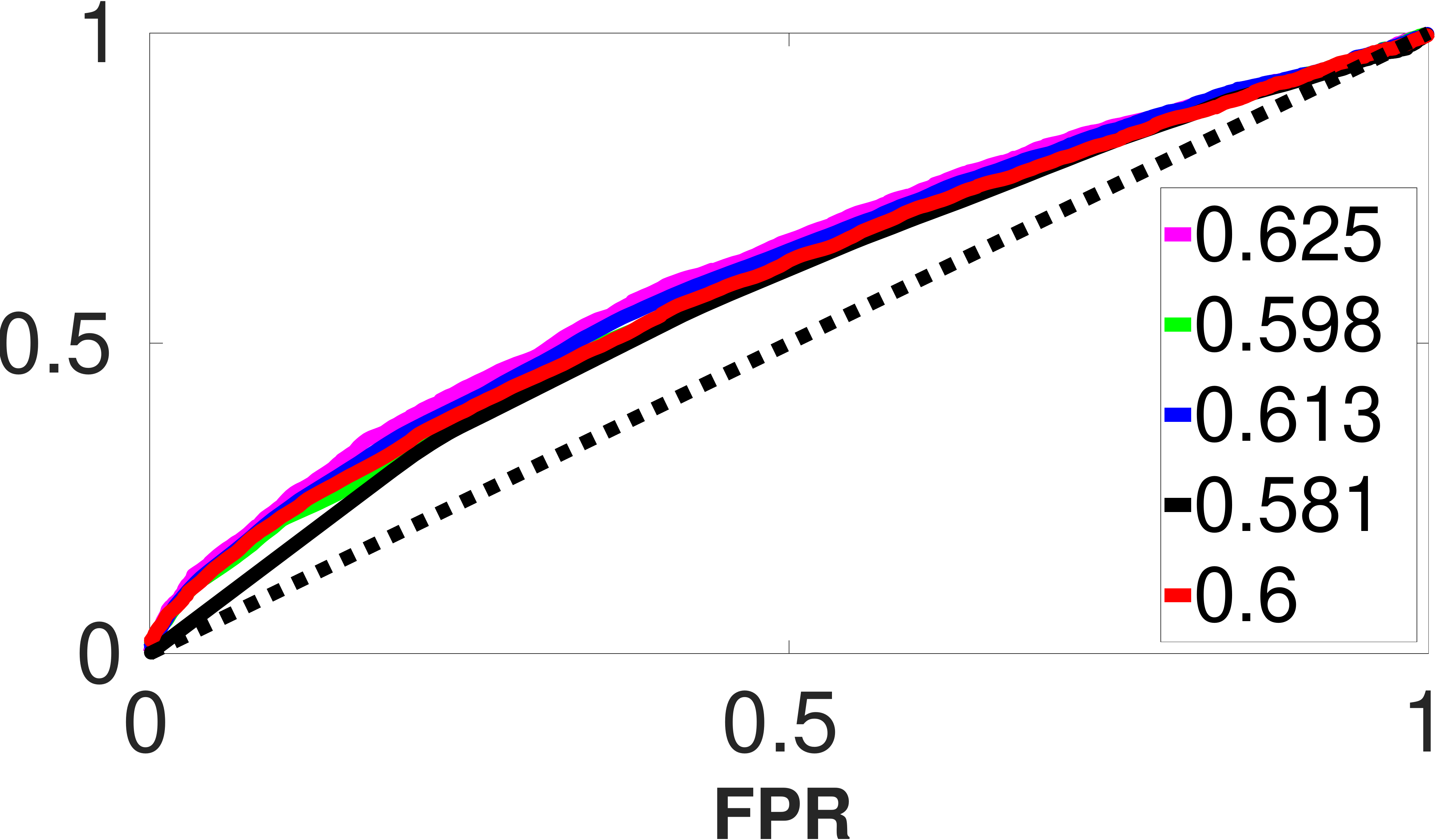}
&
\includegraphics[width=\linewidth]{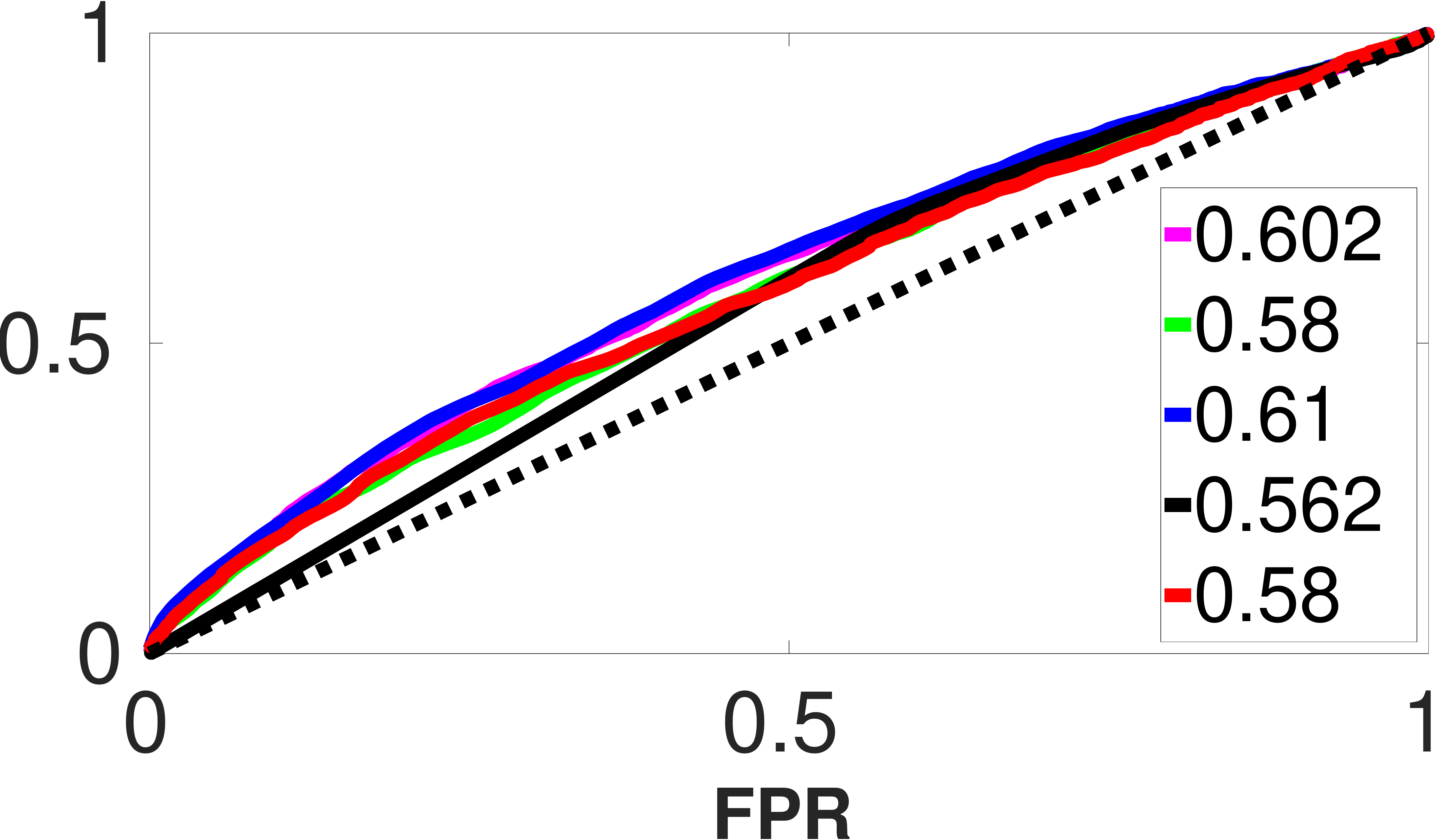}
\\
\end{tabular}
}
\caption{Receiver Operating Curves (ROCs) for the accurate AHGMM parrot-T attack at threshold $\rho_j^o=0.5$ px/cm. Each ROC is the mean of 10-curves generated by the 10-folds used for cross validation. Legend: \textcolor{magenta}{\Huge---} Unprotected, \textcolor{green}{\Huge---} \ac{AGB}, \textcolor{blue}{\Huge---} \ac{SVGB}, \textcolor{black}{\Huge---} \ac{FGB}, \textcolor{red}{\Huge ---} AHGMM. In each column, the image resolution remains constant, i.e.~first column: $96 \times 96$, second column: $48 \times 48$, third column: $24 \times 24$ and fourth column: $12 \times 12$ pixels,  while the pitch angle varies~i.e. first row: $0^\circ$, second row: $10^\circ$, third row: $20^\circ$, fourth row: $30^\circ$, fifth row: $40^\circ$, sixth row: $50^\circ$, seventh row: $60^\circ$ and eighth row: $70^\circ$. The legend values represent the Area Under Curve (AUC).}
\label{fig:ROCs_accurate_ahgmm_parrot-t}
\end{figure*}

From the na\"{i}ve-T attack, we are interested in finding the optimal threshold which defines the optimal kernel for AGB \citep{sarwar2016} (see Section \ref{sec:proposed_approach} and Eq. \ref{eq:sigma_i}). It is clear from Fig. \ref{fig:naive_attack} that the accuracy of the na\"{i}ve-T attack decreases while decreasing the threshold. When the threshold reaches $0.5$ px/cm, the difference between the accuracy achieved by AGB \citep{sarwar2016} and a random classifier ($\eta=0.5$) becomes very small except, unexpectedly, at high pitch angles. This difference further decreases at $0.4$ px/cm and $0.3$ px/cm. Thus, the optimal threshold defining the optimal kernel can be $0.5$ px/cm, $0.4$ px/cm and $0.3$ px/cm. The later two thresholds decreases the accuracy negligibly but distort the images severely. Therefore, we decide to perform a trade-off analysis of the accuracy (under na\"{i}ve, parrot attack and reconstruction attacks) and the distortion at these three thresholds. 

At these three thresholds under the na\"{i}ve-T attack, the accuracy of the AHGMM is higher as compared to the AGB \citep{sarwar2016}. The main reason for this slightly higher accuracy is due to the under blurred sub-regions of the AHGMM filtered face as it hops its kernel below and above the optimal Gaussian kernel. In contrast, the accuracy of the Space Variant Gaussian
Blur \citep{Saini2012} is always lower than AGB and AHGMM. This is because SVGB uses an isotropic Gaussian kernel which deteriorates a face more severely as compared to the anisotropic kernel of the AGB and AHGMM filter. \ac{FGB} possess the lowest accuracy at any threshold due to over blurring of all images except $96\times96$ pixels images at $0^\circ$ pitch angle.
\subsection{Parrot-T Attack}
\label{subsec:Parrot_attack}

In the parrot-T attack, we filter both gallery and probe images and then evaluate the achieved accuracy. We study the parrot-T attack on AHGMM under three sub-attacks: optimal kernel parrot-T sub-attack, pseudo \ac{AHGMM} parrot-T sub-attack and accurate \ac{AHGMM} parrot-T sub-attack. The accuracy results of these sub-attacks are given in Fig. \ref{fig:naive_attack} at different thresholds $\rho_j^o$, while \acp{ROC} for the accurate \ac{AHGMM} parrot-T sub-attack at $\rho_j^o=0.5$ px/cm are presented in Fig. \ref{fig:ROCs_accurate_ahgmm_parrot-t}. 

The parrot-T attack on state-of-the-art privacy filters increases the accuracy as compared to the na\"{i}ve-T attack. Under the optimal kernel parrot sub-attack, our AHGMM shows the least accuracy improvement at any of the three thresholds. This is because the optimal kernel Gaussian blur is a spatially invariant blur that is not helpful in recognising spatially varying Gaussian blurred images, e.g.~the AHGMM filtered images. Thus, our AHGMM provides the lowest accuracy against the parrot-T attack using the optimal kernel. 

The pseudo AHGMM parrot-T sub-attack slightly improves the accuracy further as compared to the optimal kernel parrot-T sub-attack. The main reason is that both the gallery and the probe images are now filtered using spatially varying Gaussian blur. However, under the pseudo AHGMM sub-attack, the accuracy of AHGMM remains below the other three state-of-the-art privacy filters. Thus, our AHGMM provides the highest privacy protection even against the pseudo AHGMM parrot sub-attack. 

Finally, the accurate AHGMM sub-attack improves the accuracy as compared to the optimal kernel and almost eqivalent to the pseudo AHGMM sub-attacks. Comparatively, even under the accurate AHGMM sub-attack, AHGMM performs better than FGB, AGB \citep{sarwar2016} and SVGB \citep{Saini2012} at these three thresholds with the least improvement at $\rho_j^o=0.3$ px/cm. 

From the accurate AHGMM sub-attack, it is apparent that our AHGMM permanently removes the sensitive information from the face and an attacker can not recognise it with a high accuracy even when he/she has access to the secret key. This is in contrast to the reversible filters, e.g.~encryption/scrambling based filters, which can reconstruct the original face after having the secret key. Thus, our AHGMM is robust against the brute-force attack. 
\subsection{Inverse Filter Attack}
\label{subsec:inverse_filter_attack}
\begin{figure}[tb]
\centering
\resizebox{1\columnwidth}{!}{%
\begin{tabular}{|c|cc|cc|cccc|}
\hline
Threshold& \multicolumn{2}{c|}{AGB \citep{sarwar2016}} & \multicolumn{2}{c|}{SVGB \citep{Saini2012}} &\multicolumn{4}{c|}{AHGMM}
\\
\cline{2-9}
($\rho_h^o=\rho_v^o$)&filtered & reconstructed & filtered & reconstructed & filtered & \multicolumn{3}{c|}{reconstructed}\\
\cline{7-9}
&&&&& & optimal & pseudo & accurate \\ \hline
$2.0$ px/cm & \includegraphics[width=0.2\linewidth]{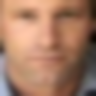}&\includegraphics[width=0.2\linewidth]{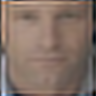}&\includegraphics[width=0.2\linewidth]{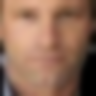}&\includegraphics[width=0.2\linewidth]{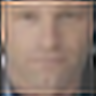}&\includegraphics[width=0.2\linewidth]{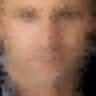}&\includegraphics[width=0.2\linewidth]{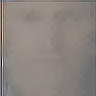}&\includegraphics[width=0.2\linewidth]{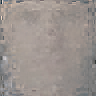}&\includegraphics[width=0.2\linewidth]{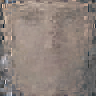}
\\
\hline
$1.2$ px/cm & \includegraphics[width=0.2\linewidth]{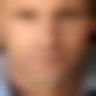}&\includegraphics[width=0.2\linewidth]{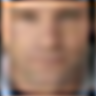}&\includegraphics[width=0.2\linewidth]{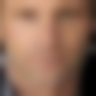}&\includegraphics[width=0.2\linewidth]{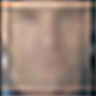}&\includegraphics[width=0.2\linewidth]{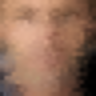}&\includegraphics[width=0.2\linewidth]{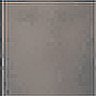}&\includegraphics[width=0.2\linewidth]{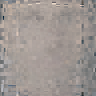}&\includegraphics[width=0.2\linewidth]{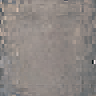}
\\
\hline
$0.4$ px/cm & \includegraphics[width=0.2\linewidth]{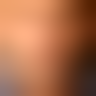}&\includegraphics[width=0.2\linewidth]{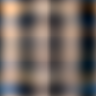}&\includegraphics[width=0.2\linewidth]{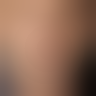}&\includegraphics[width=0.2\linewidth]{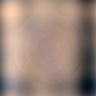}&\includegraphics[width=0.2\linewidth]{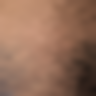}&\includegraphics[width=0.2\linewidth]{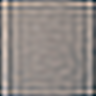}&\includegraphics[width=0.2\linewidth]{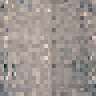}&\includegraphics[width=0.2\linewidth]{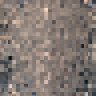}
\\
\hline
\end{tabular}
}
\caption{Inverse filtering of protected faces at different thresholds $\rho_j^o$. AGB and SVGB protected faces can be reconstructed by inverse filtering to some extent. Inverse filtering of AHGMM protected faces is hardly possible even if the hopping kernel parameters are known.} 
\label{fig:inverse_attack}
\end{figure}

\begin{figure*}[t]
\centering
\resizebox{0.97\textwidth}{!}{%
\begin{tabular}{cccc}
Na\"{i}ve-IF & Na\"{i}ve-IF & Na\"{i}ve-IF & Parrot-IF \\
(optimal kernel) & (pseudo AHGMM) & (accurate AHGMM) & (accurate AHGMM)\\
\includegraphics[width=0.25\linewidth]{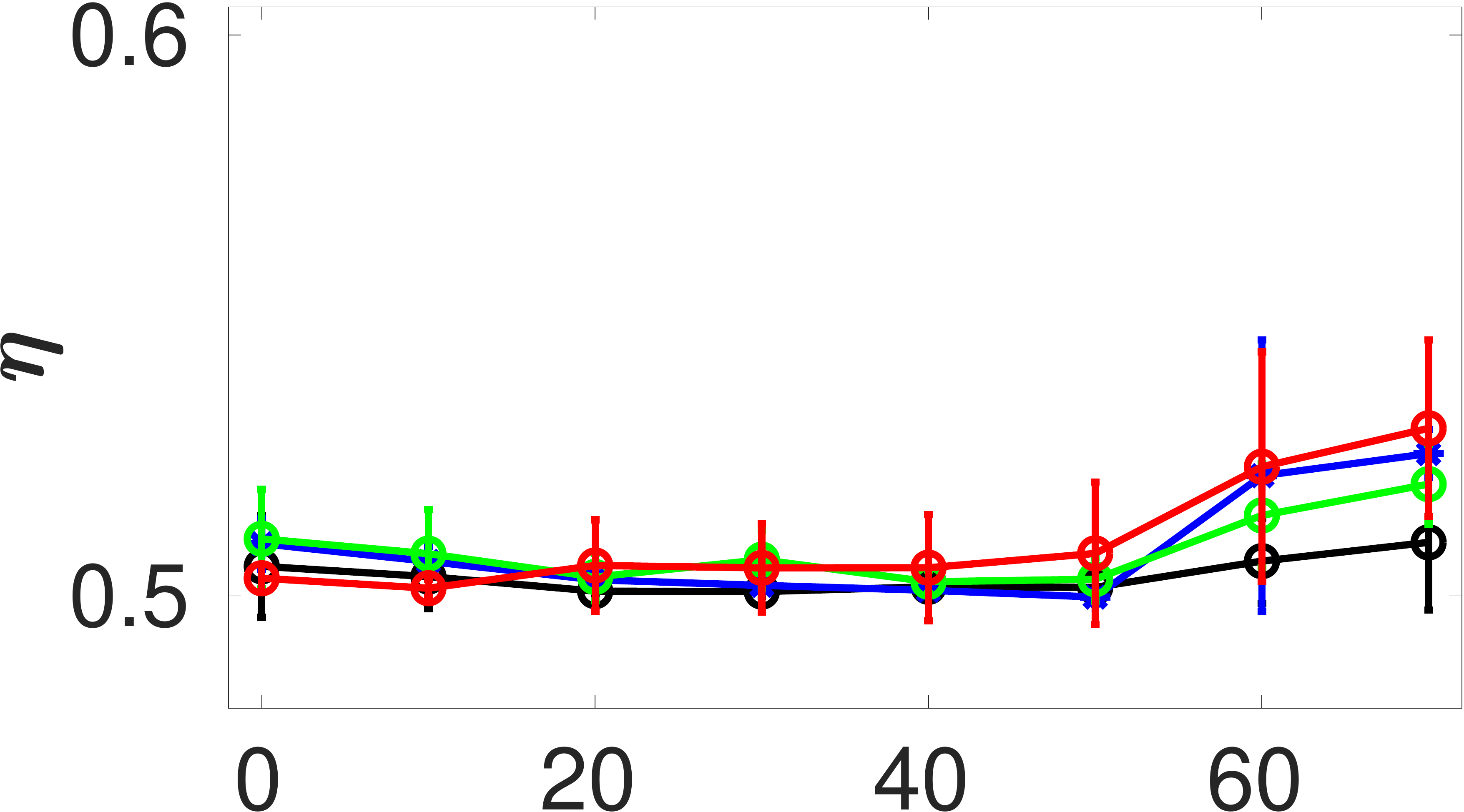}
&
\includegraphics[width=0.25\linewidth]{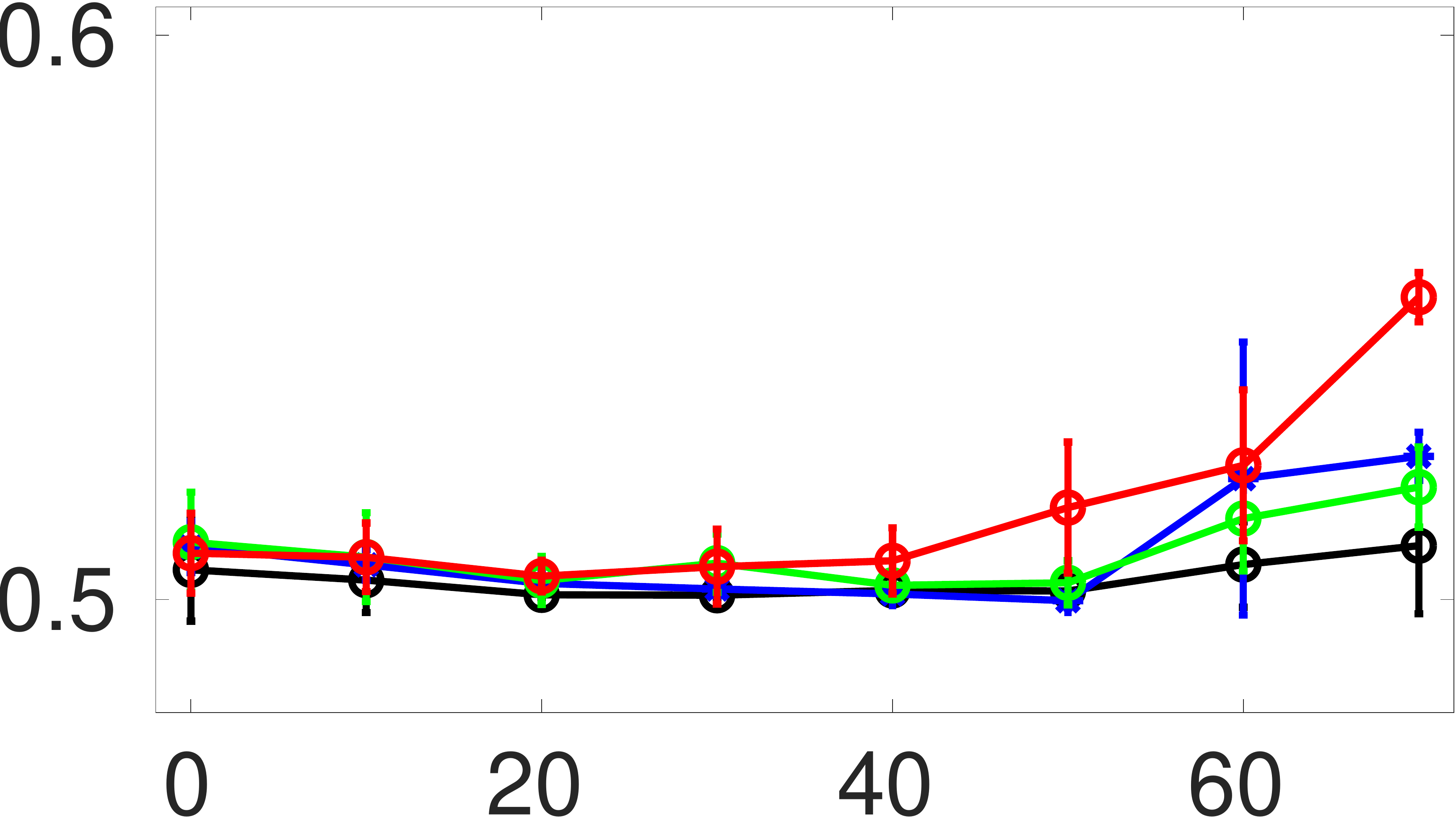}
&
\includegraphics[width=0.25\linewidth]{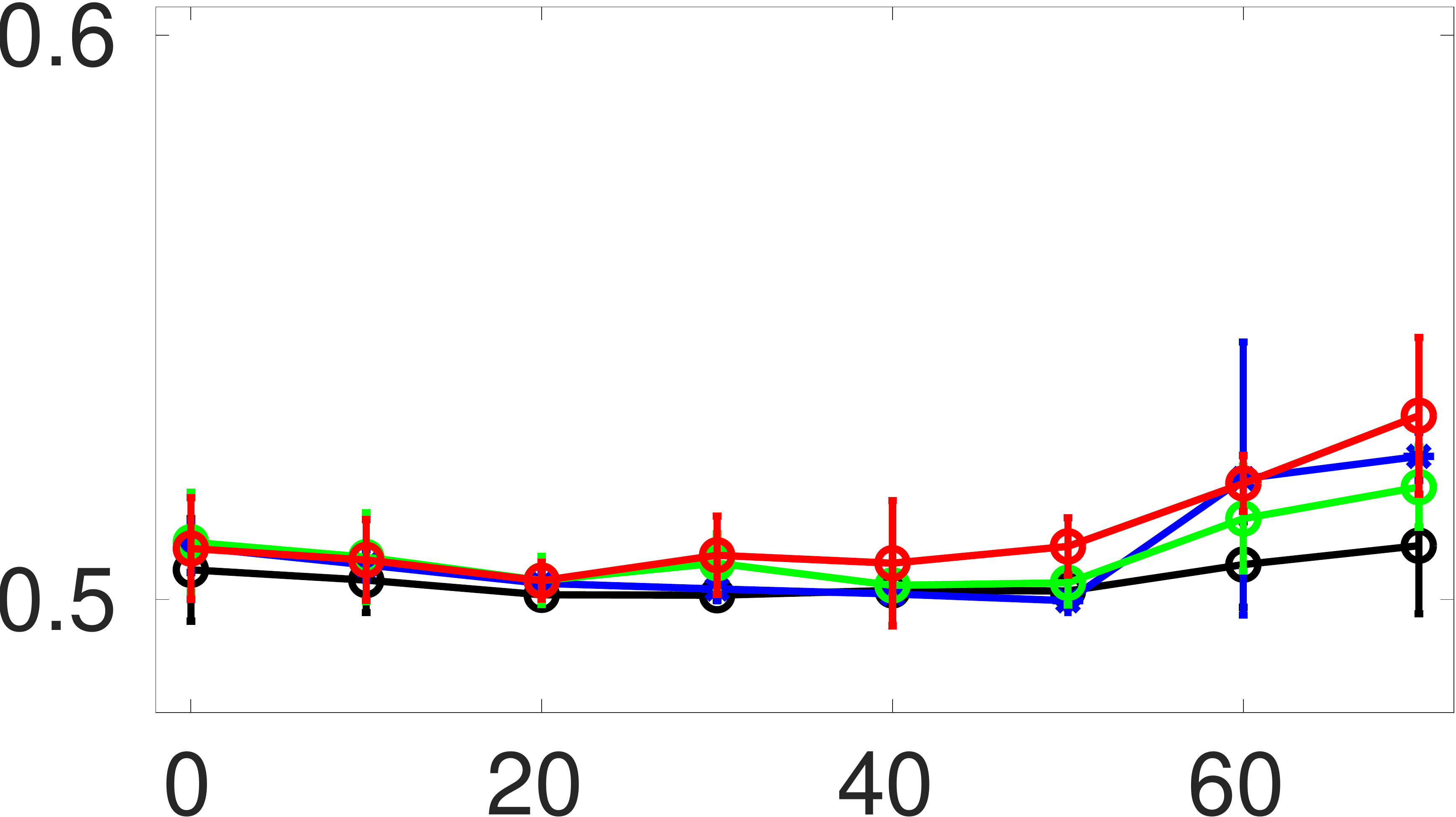}
&
\includegraphics[width=0.25\linewidth]{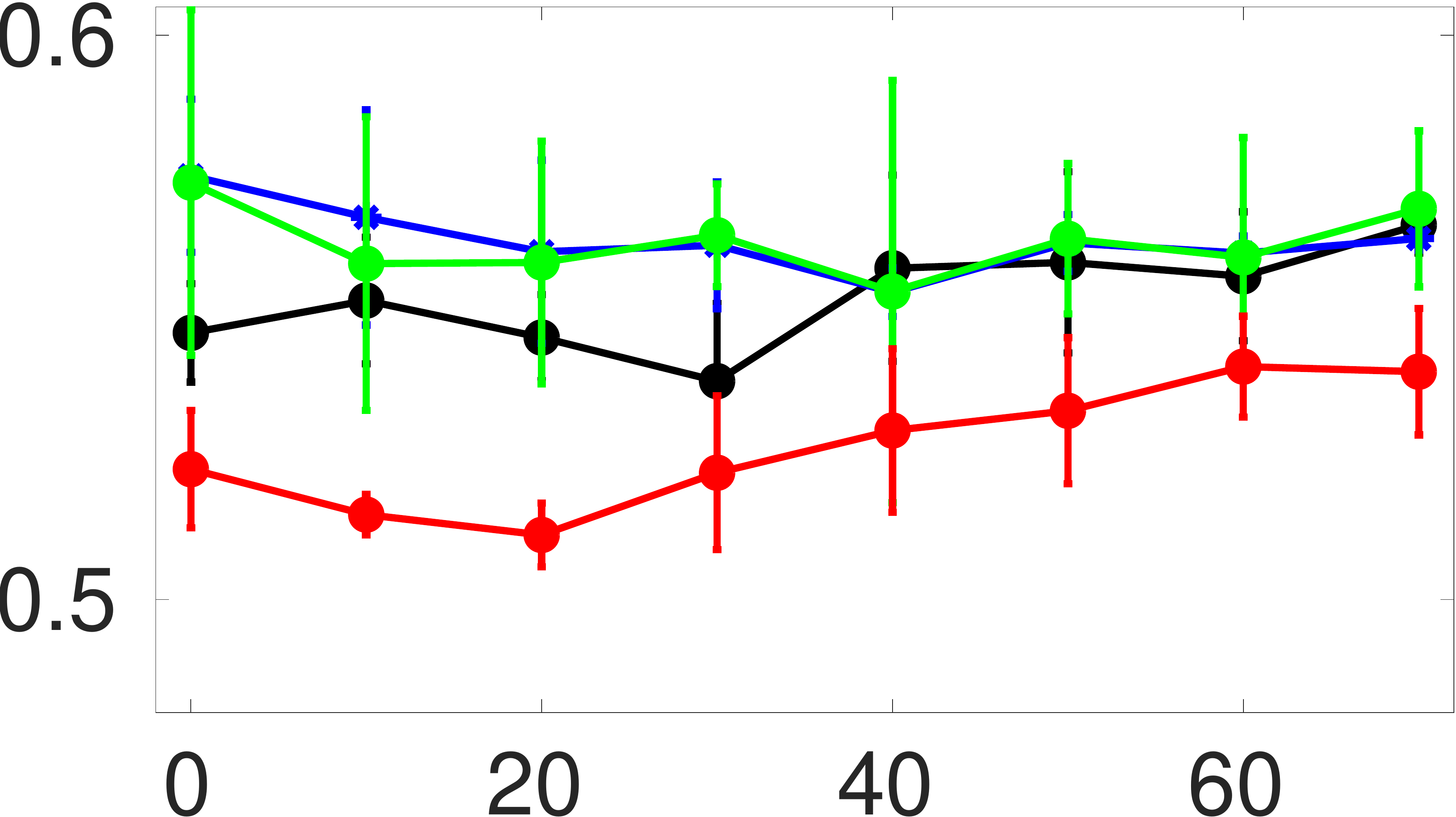}
\\
\includegraphics[width=0.25\linewidth]{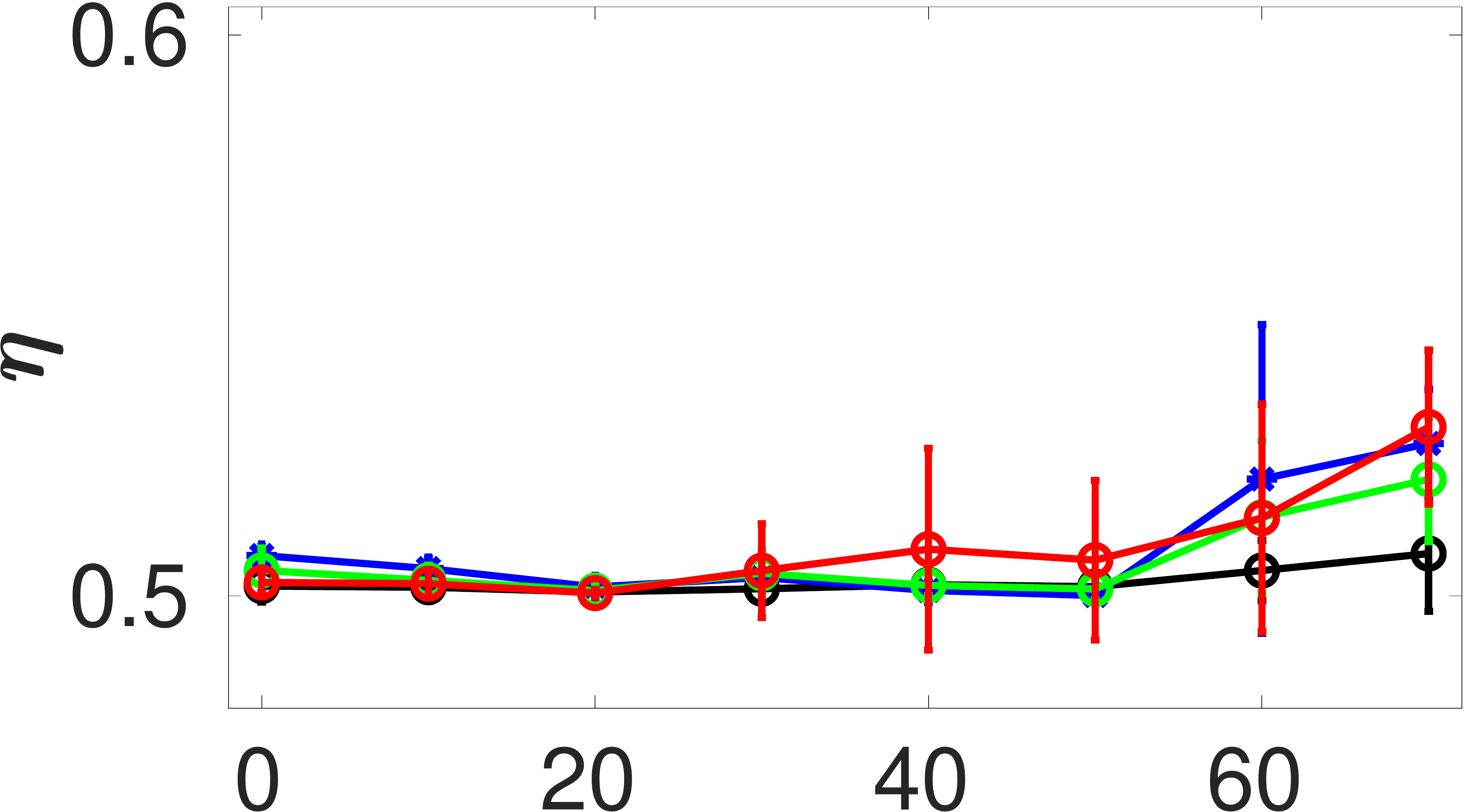}
&
\includegraphics[width=0.25\linewidth]{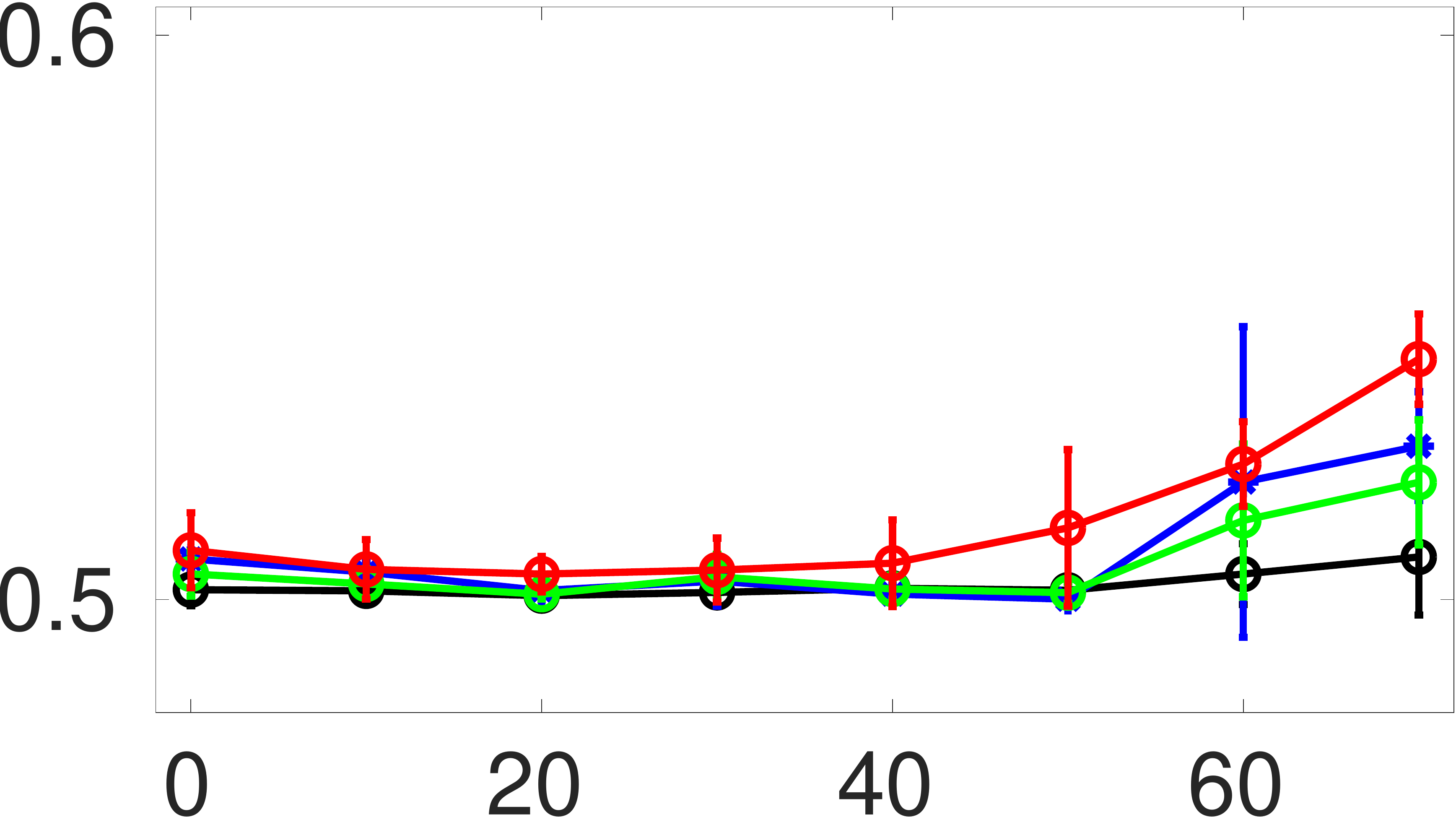}
&
\includegraphics[width=0.25\linewidth]{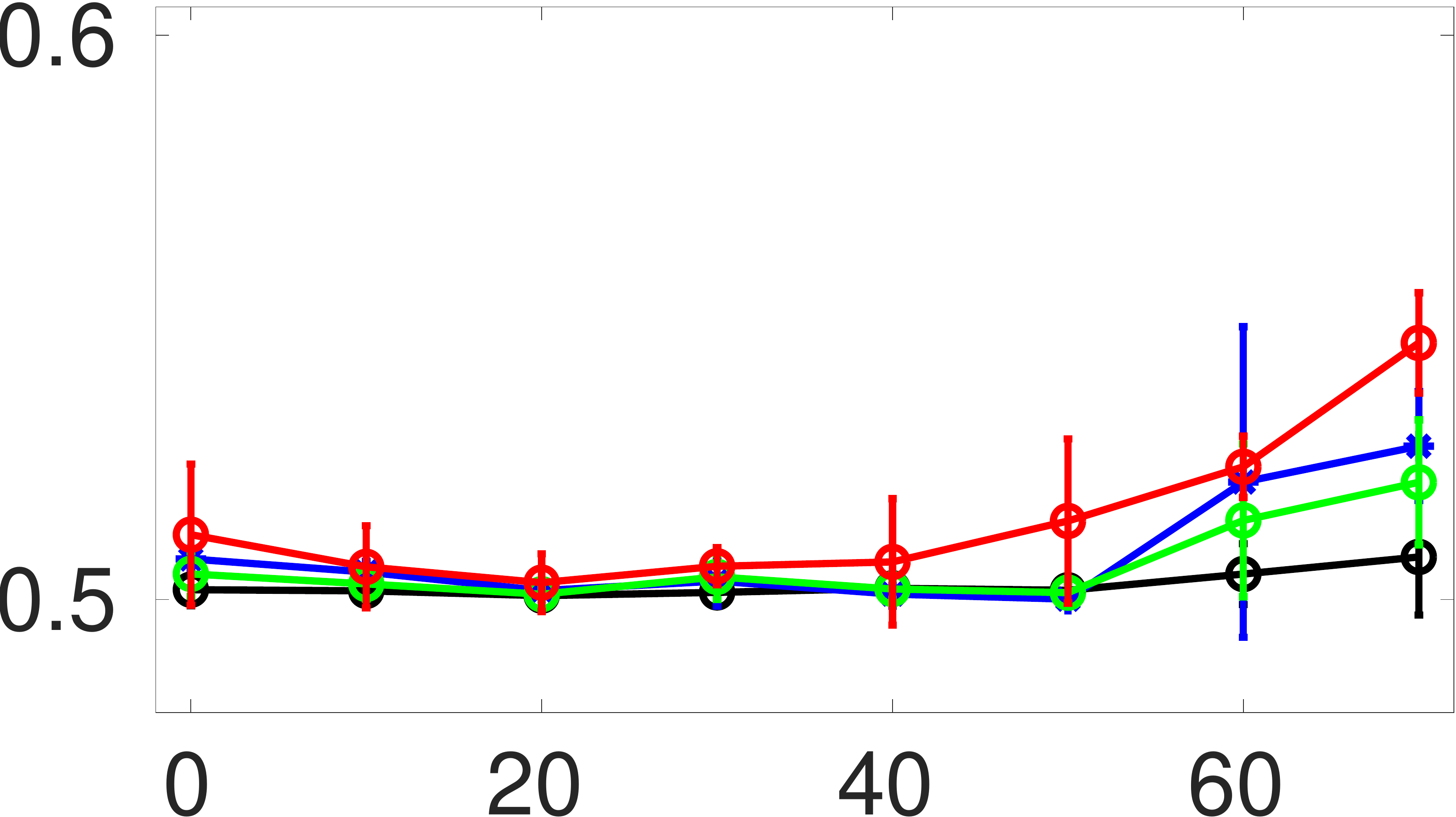}
&
\includegraphics[width=0.25\linewidth]{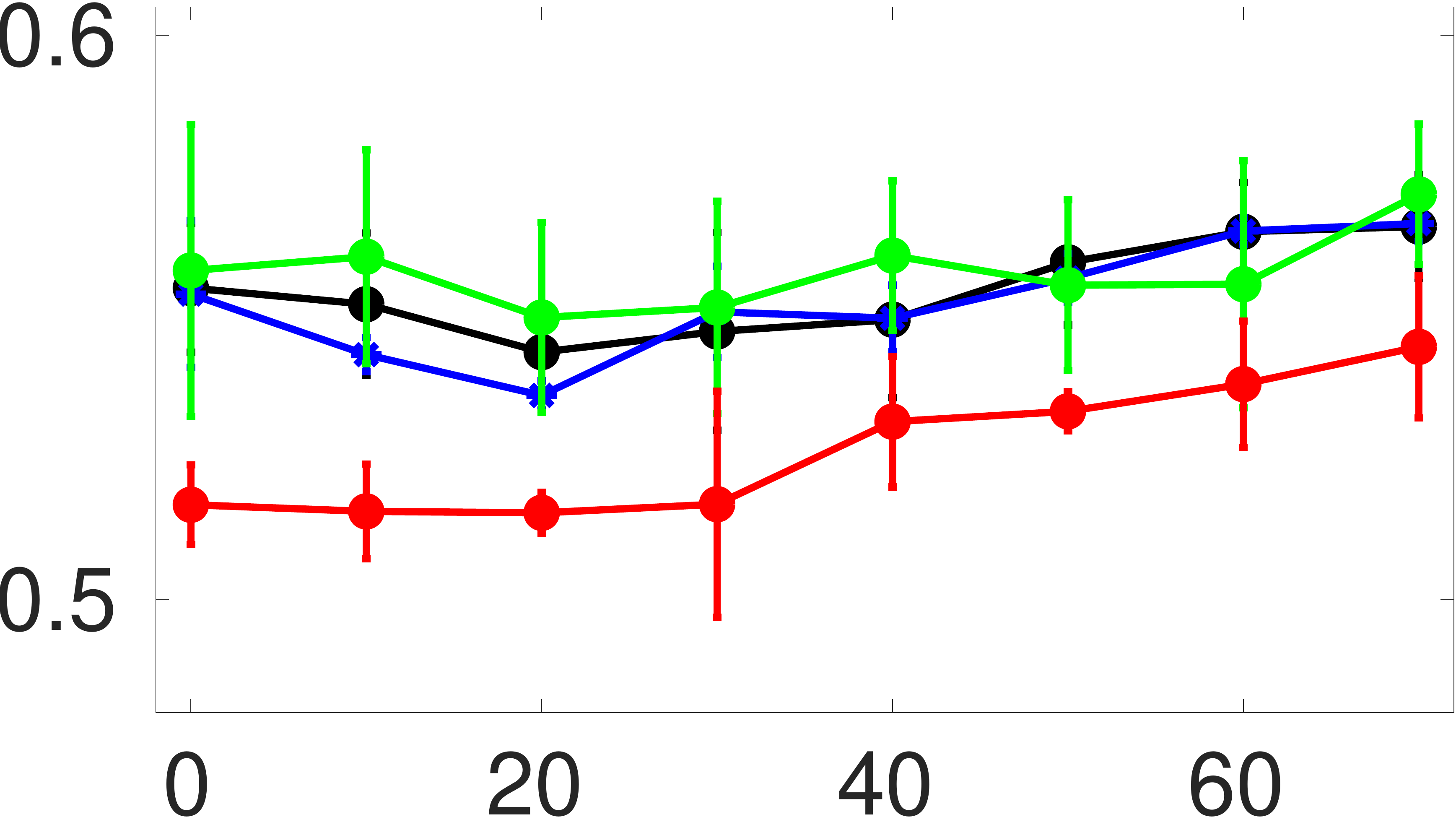}
\\
\includegraphics[width=0.25\linewidth]{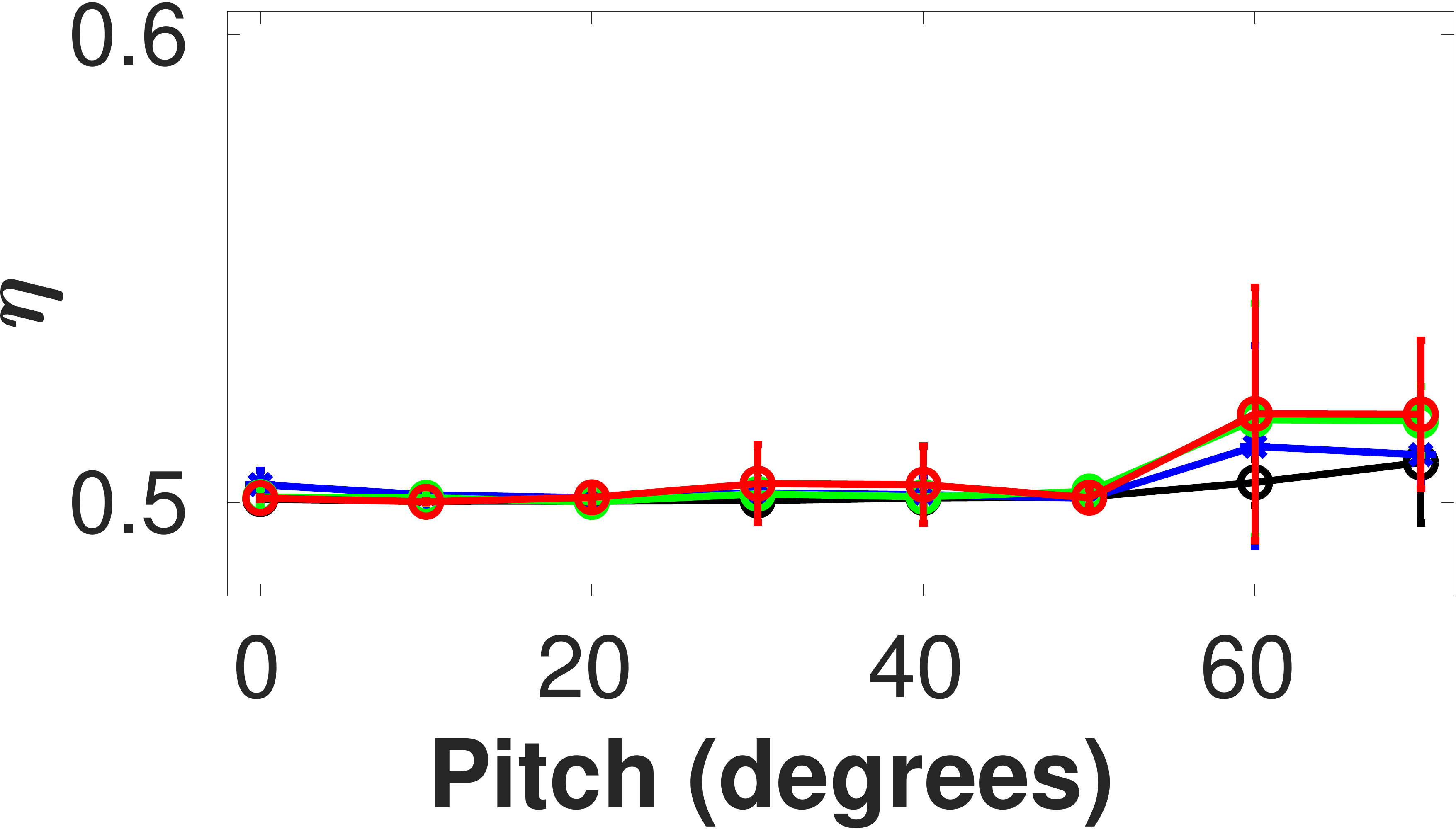}
&
\includegraphics[width=0.25\linewidth]{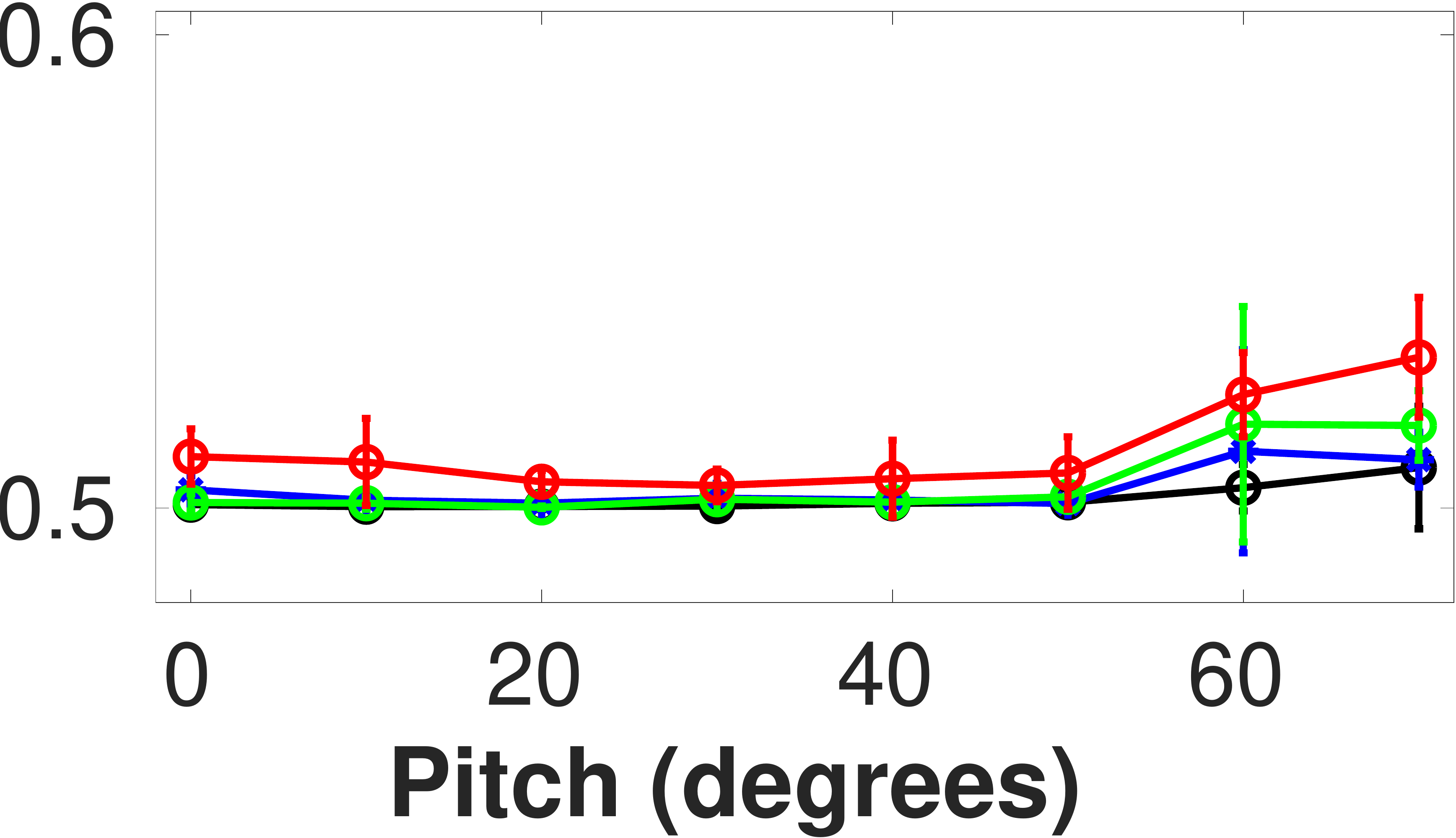}
&
\includegraphics[width=0.25\linewidth]{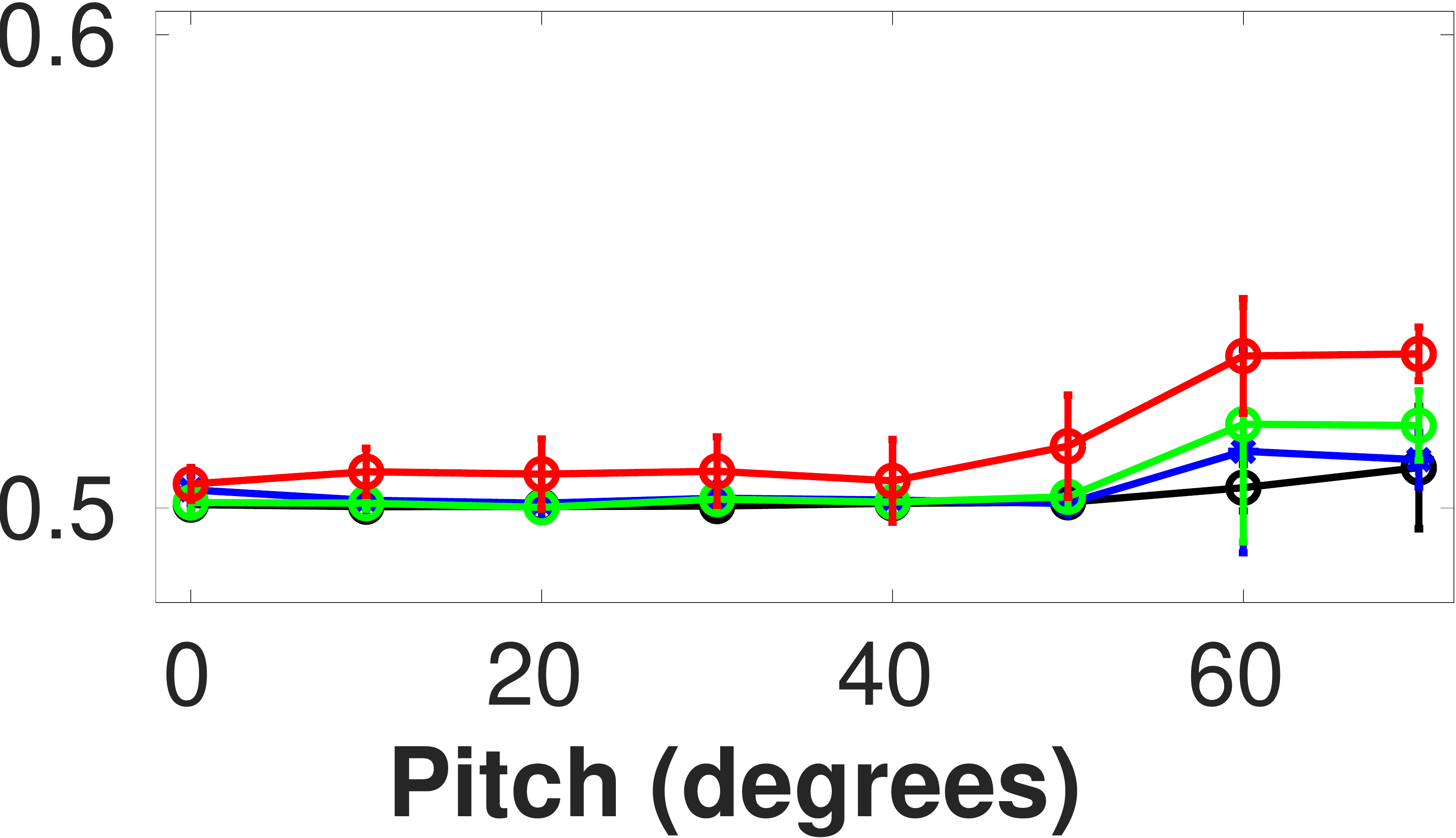}
&
\includegraphics[width=0.25\linewidth]{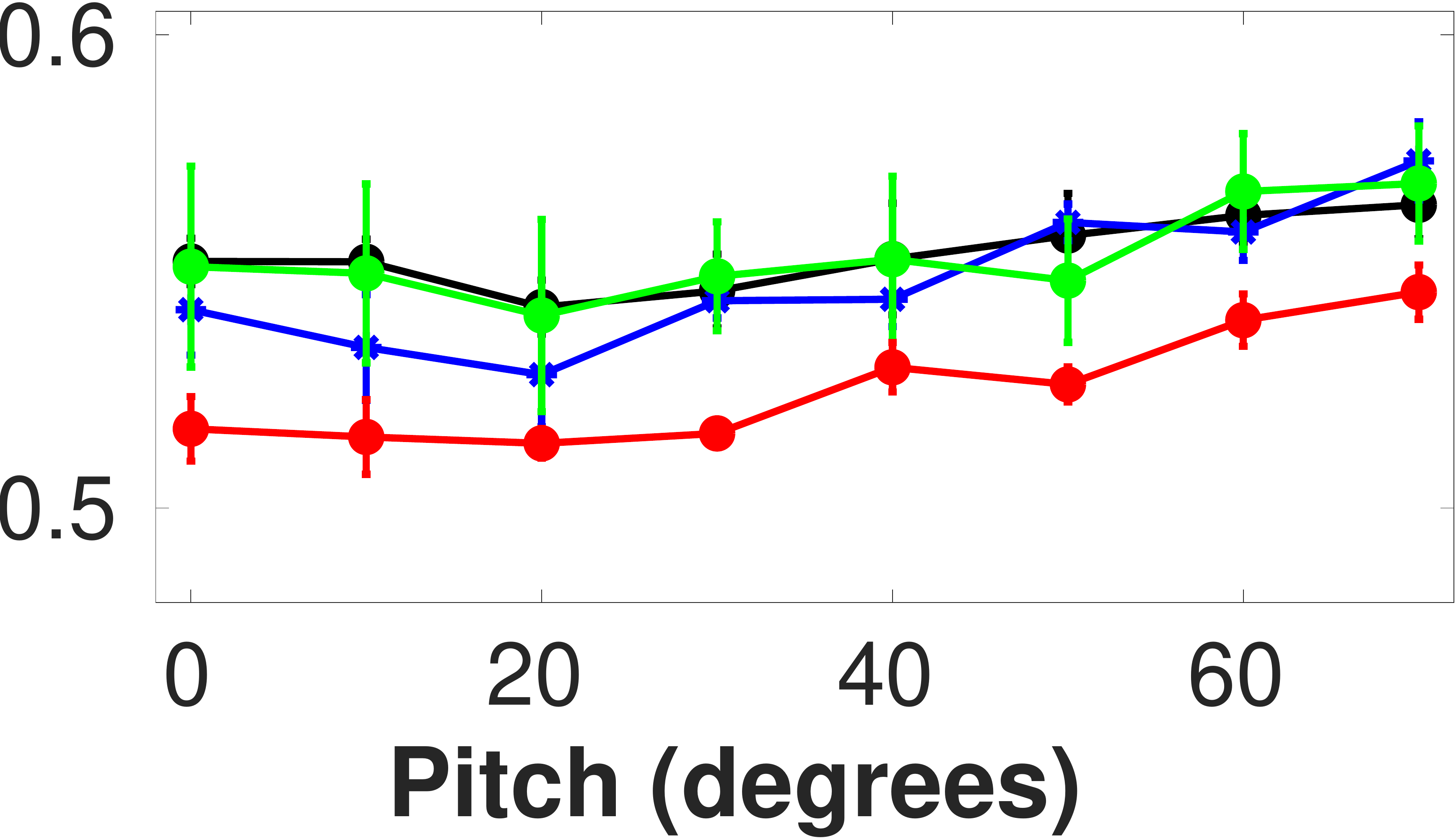}
\\
\end{tabular}
}
\caption{Face verification accuracy $\eta$ achieved by an inverse filter (IF) attack on images protected by four different privacy protection filters at different thresholds $\rho_j^o$: first row: $\rho_j^o=0.7$ px/cm, second row: $\rho_j^o=0.6$ px/cm, third row: $\rho_j^o=0.5$ px/cm. The filled marker shows the mean and the vertical bar indicates the standard deviation of $\eta$ for the multi-resolution images ($96\times96$, $48\times48$, $24\times24$, $12\times12$). Legend: \textcolor{red}{\Huge ---} AHGMM, \textcolor{green}{\Huge{---}} \ac{AGB}, \textcolor{blue}{\Huge{---}} \ac{SVGB}, \textcolor{black}{\Huge{---}} \ac{FGB}. The IF attack is investigated under four sub-attacks: optimal kernel na\"{i}ve-IF, pseudo AHGMM na\"{i}ve-IF, accurate AHGMM na\"{i}ve-IF and accurate AHGMM parrot-IF attack. The AHGMM achieves a slightly higher $\eta$ under the na\"{i}ve-IF attacks than the state-of-the-art filters, independently of the used threshold $\rho_j^o$. In contrast, AHGMM achieves the lowest $\eta$ under the parrot-IF attack. As $\eta$ is close to $0.5$ under the na\"{i}ve-IF attack for $ 0.5 \leq \rho_j^o \leq 0.7$ px/cm, we therefore do not perform experiments for $\rho_j^o < 0.5$ px/cm.}
\label{fig:inverse_attack_graphs}
\end{figure*}

\begin{figure*}[!htb]
\centering
\resizebox{0.97\textwidth}{!}{%
\begin{tabular}{cccc}
\includegraphics[width=\linewidth]{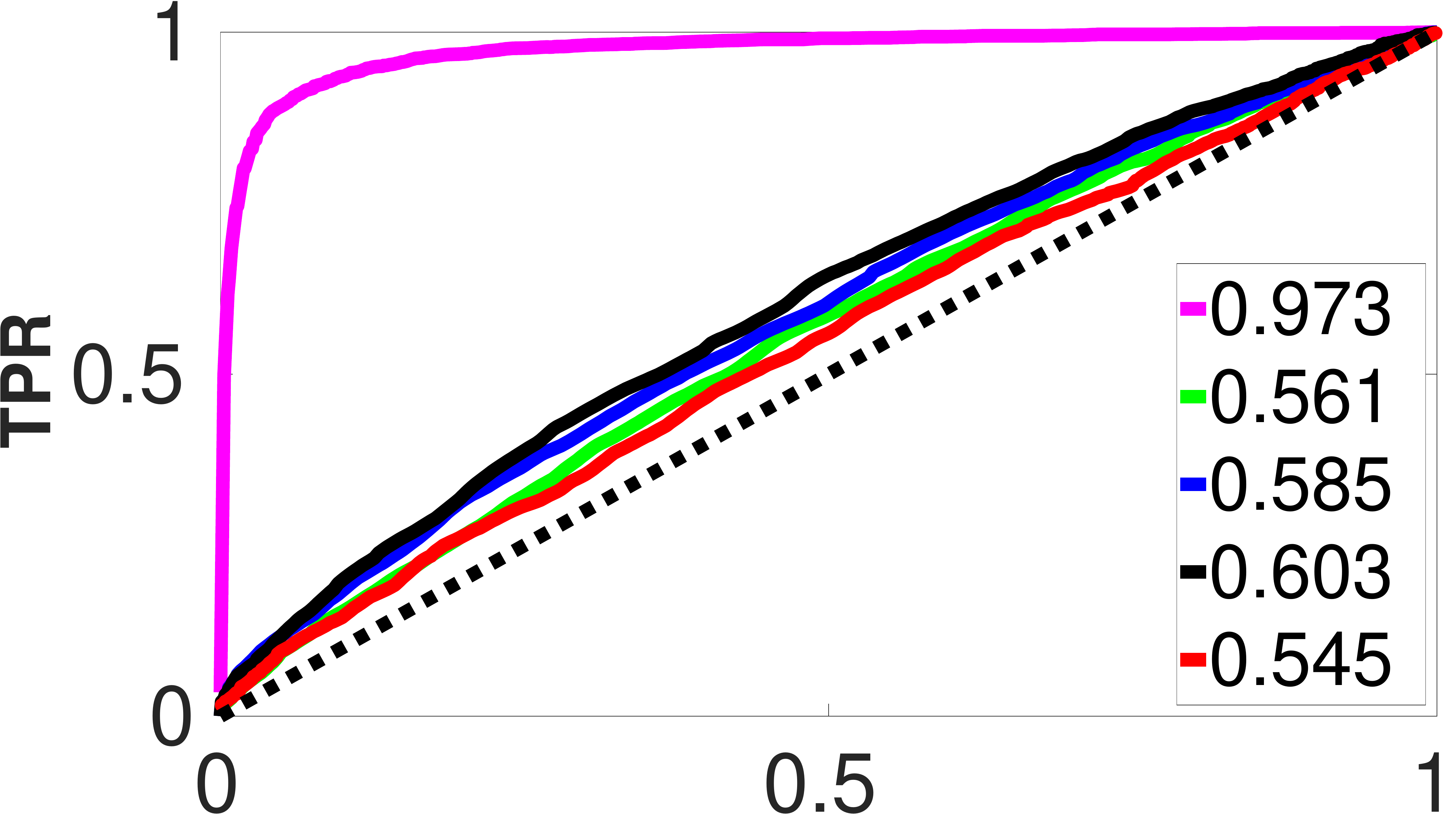}
&
\includegraphics[width=\linewidth]{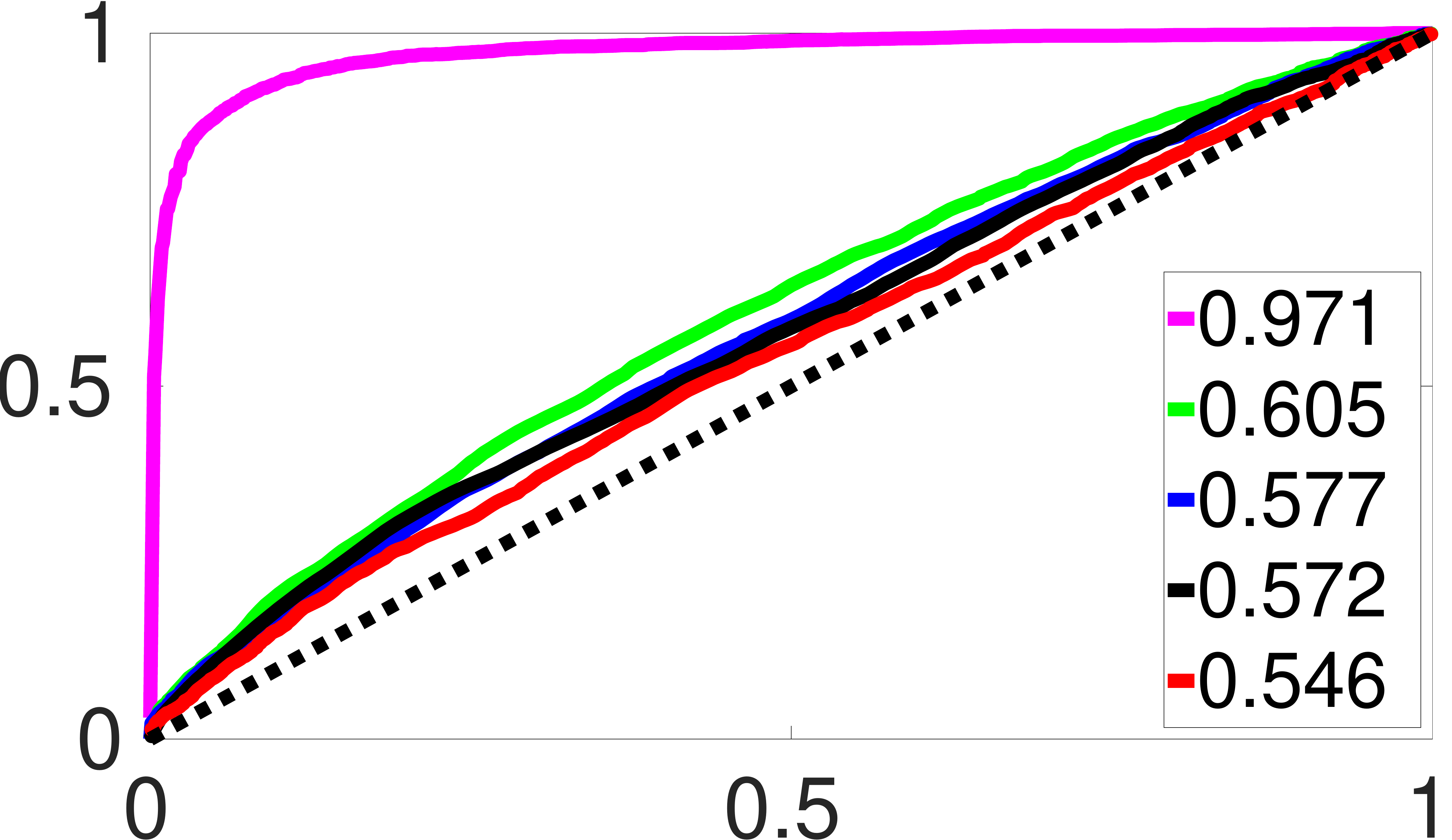}
&
\includegraphics[width=\linewidth]{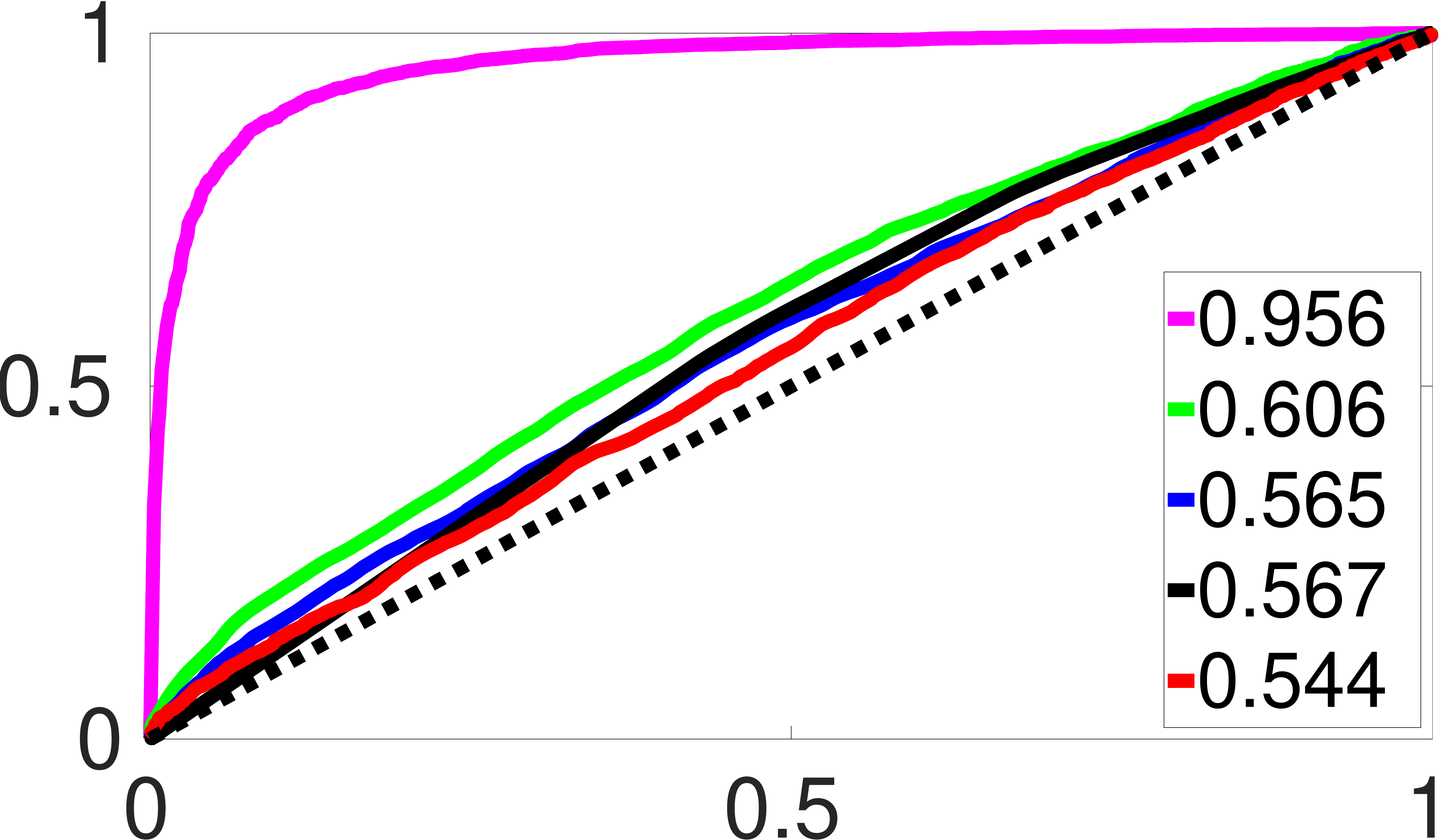}
&
\includegraphics[width=\linewidth]{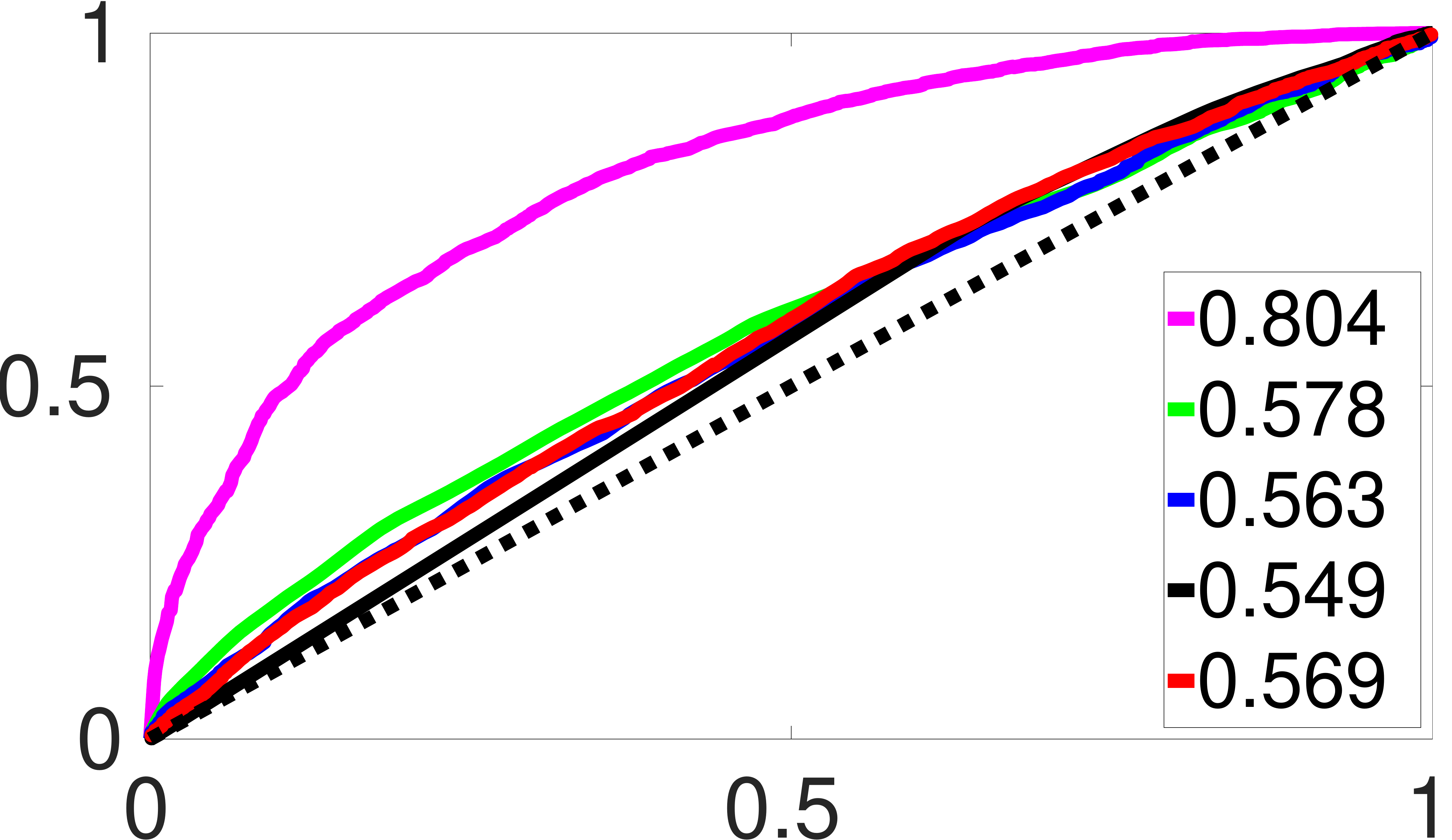}
\\
\includegraphics[width=\linewidth]{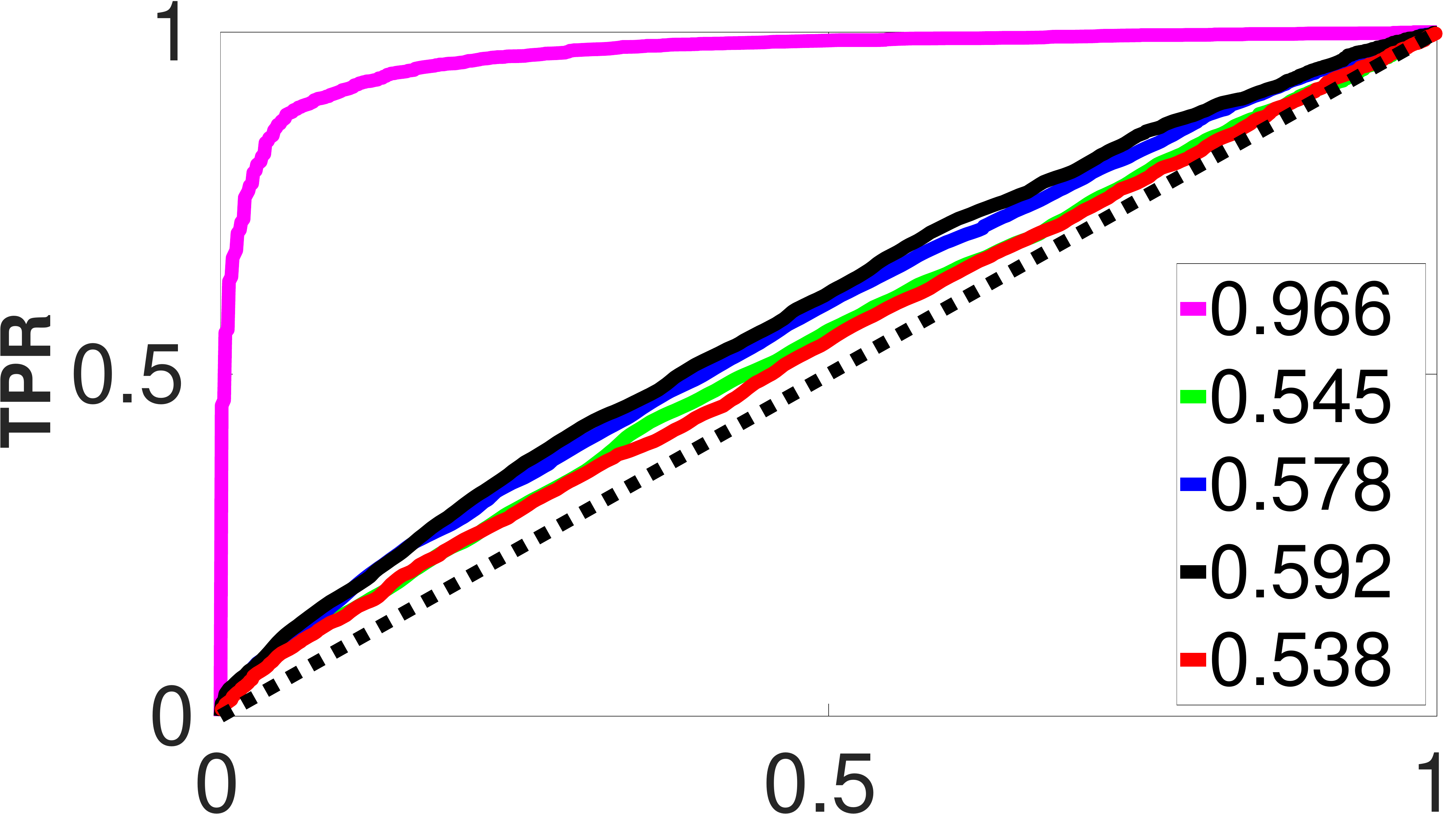}
&
\includegraphics[width=\linewidth]{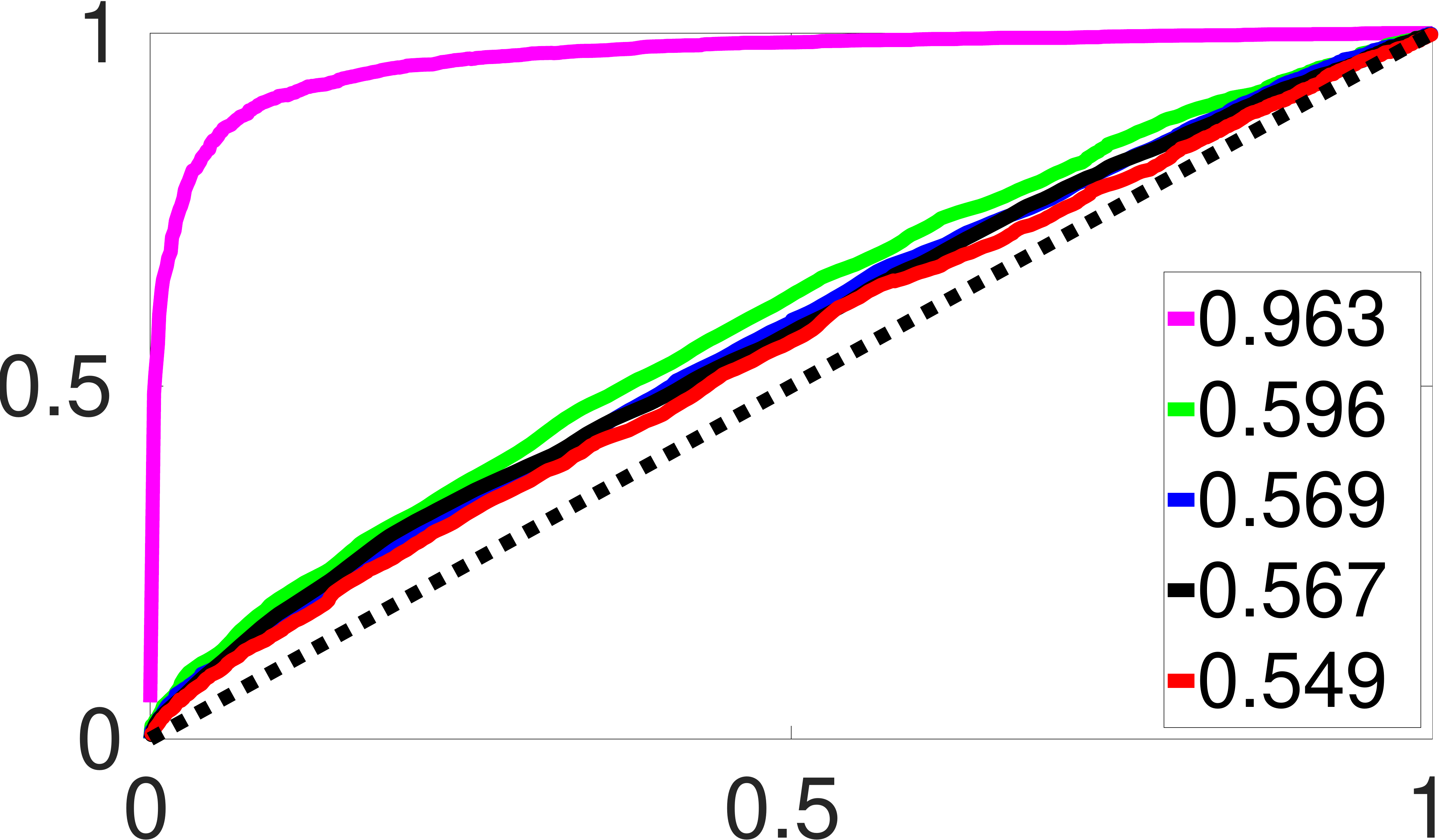}
&
\includegraphics[width=\linewidth]{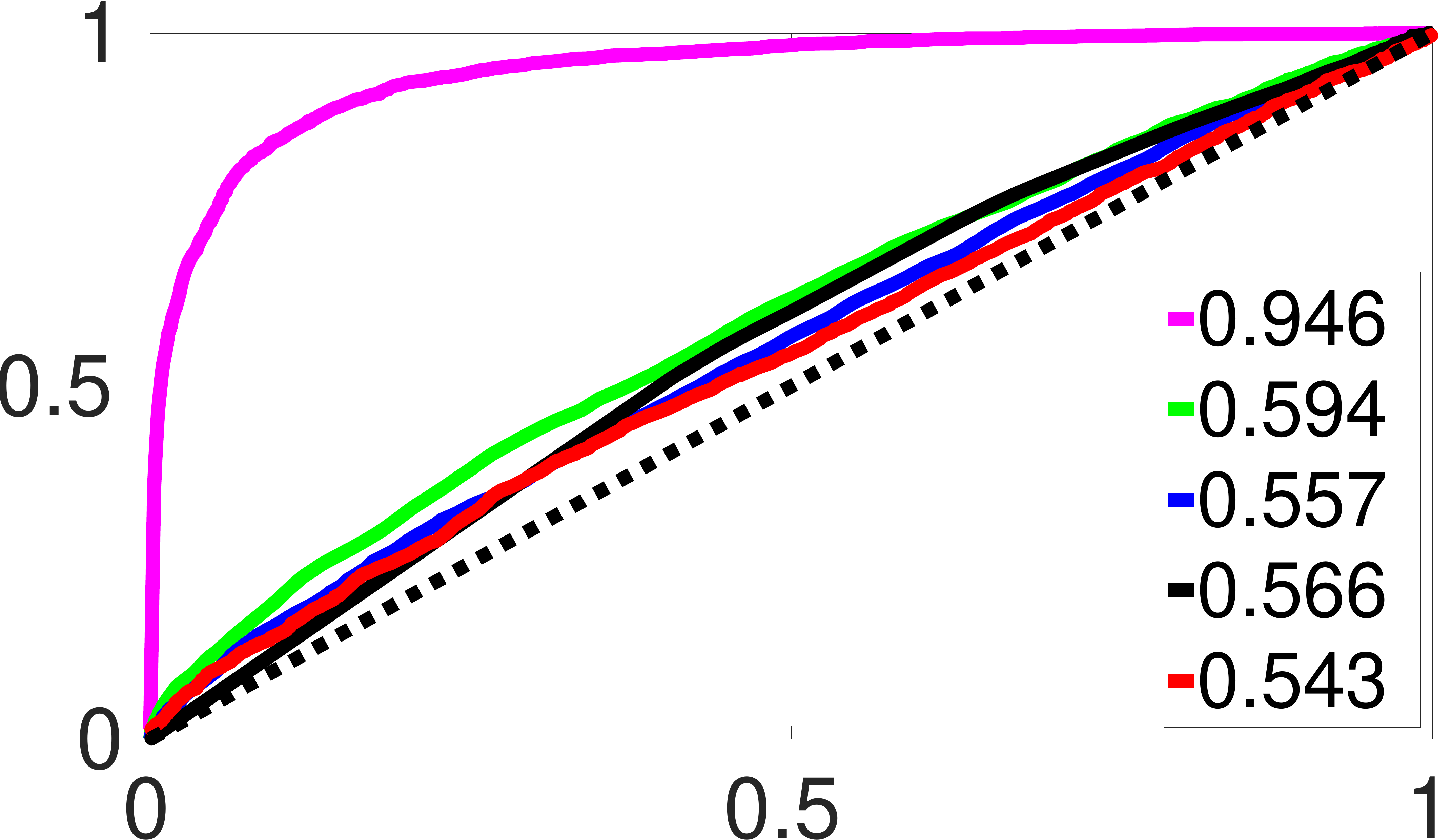}
&
\includegraphics[width=\linewidth]{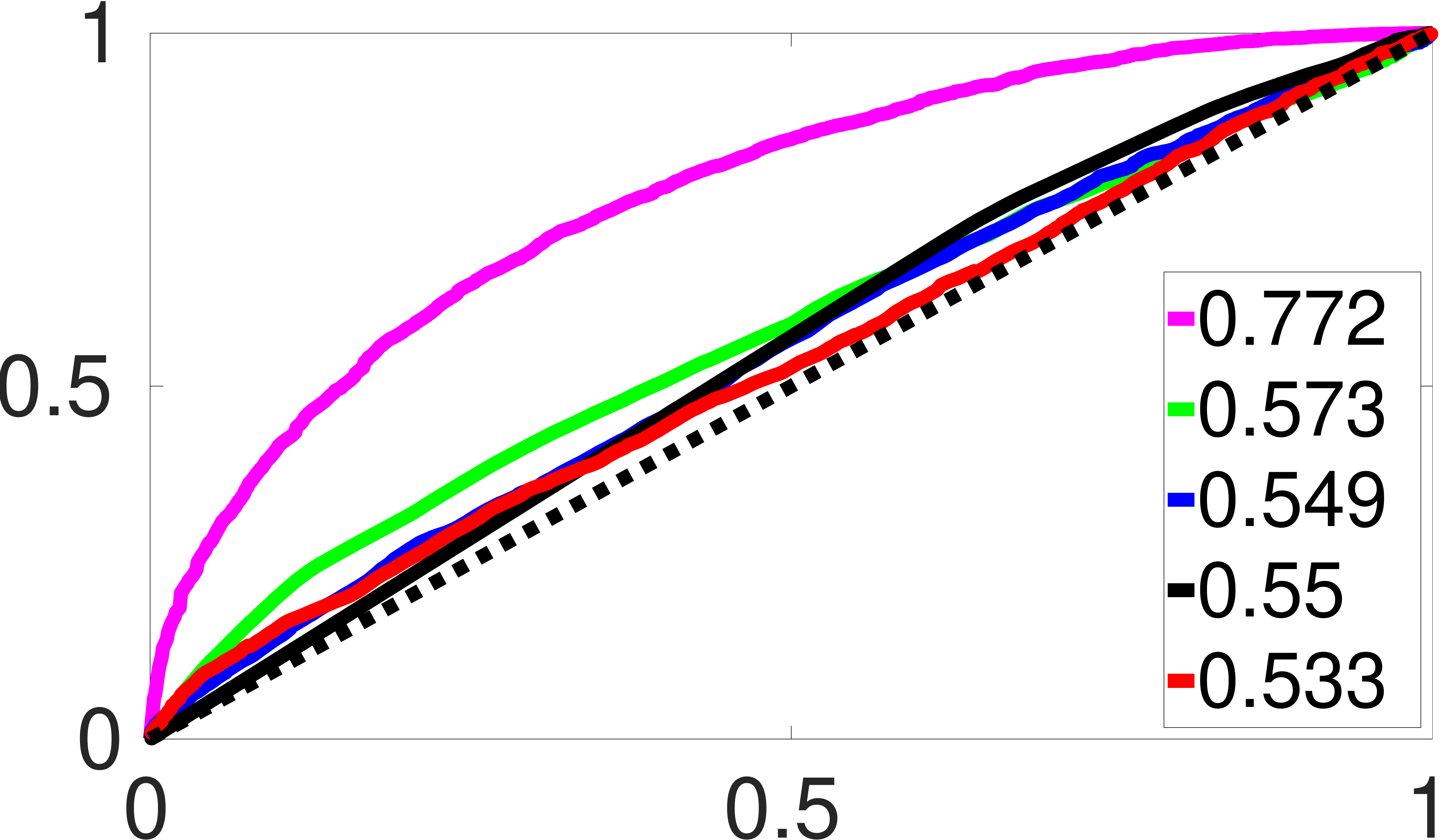}
\\
\includegraphics[width=\linewidth]{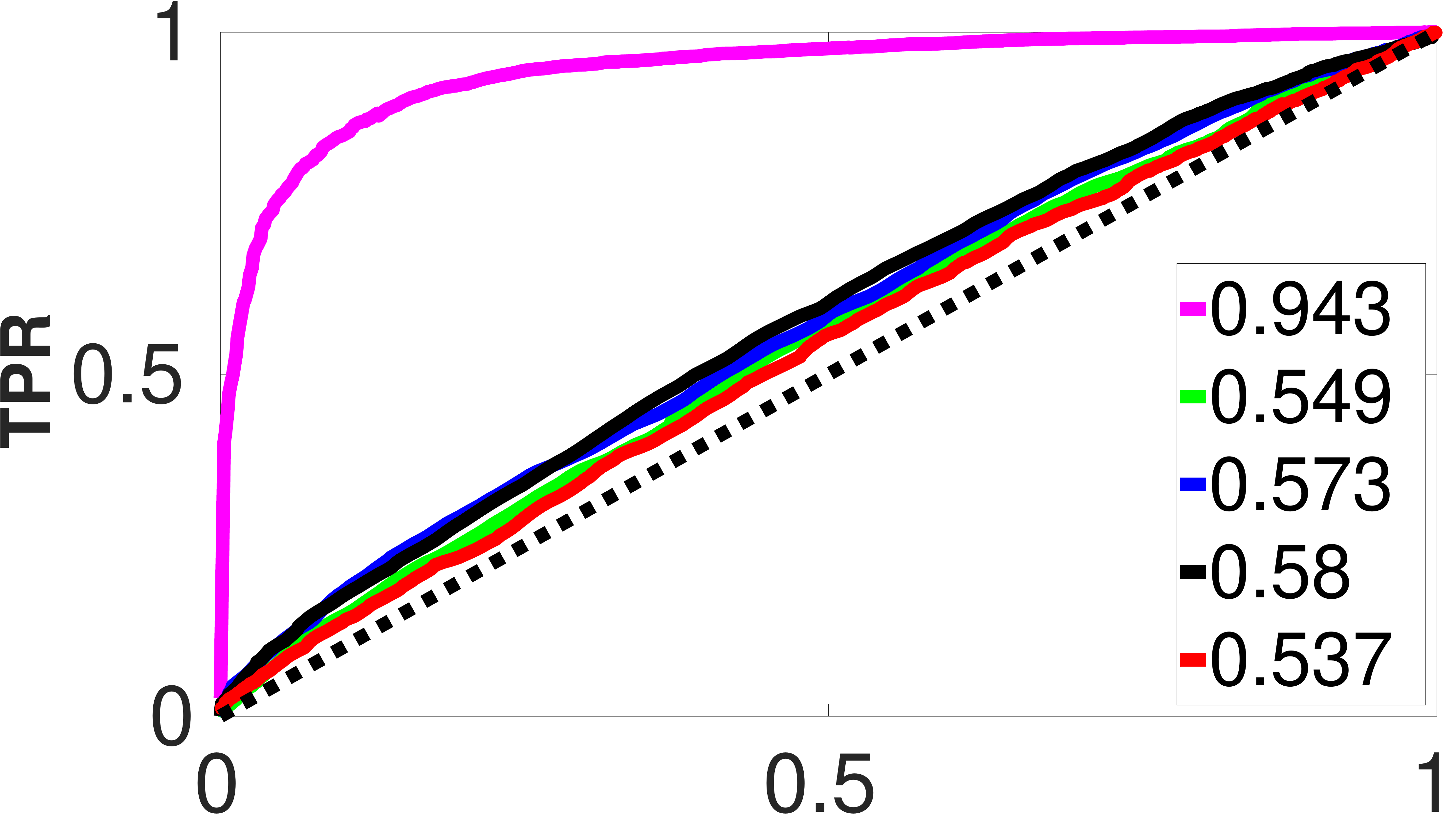}
&
\includegraphics[width=\linewidth]{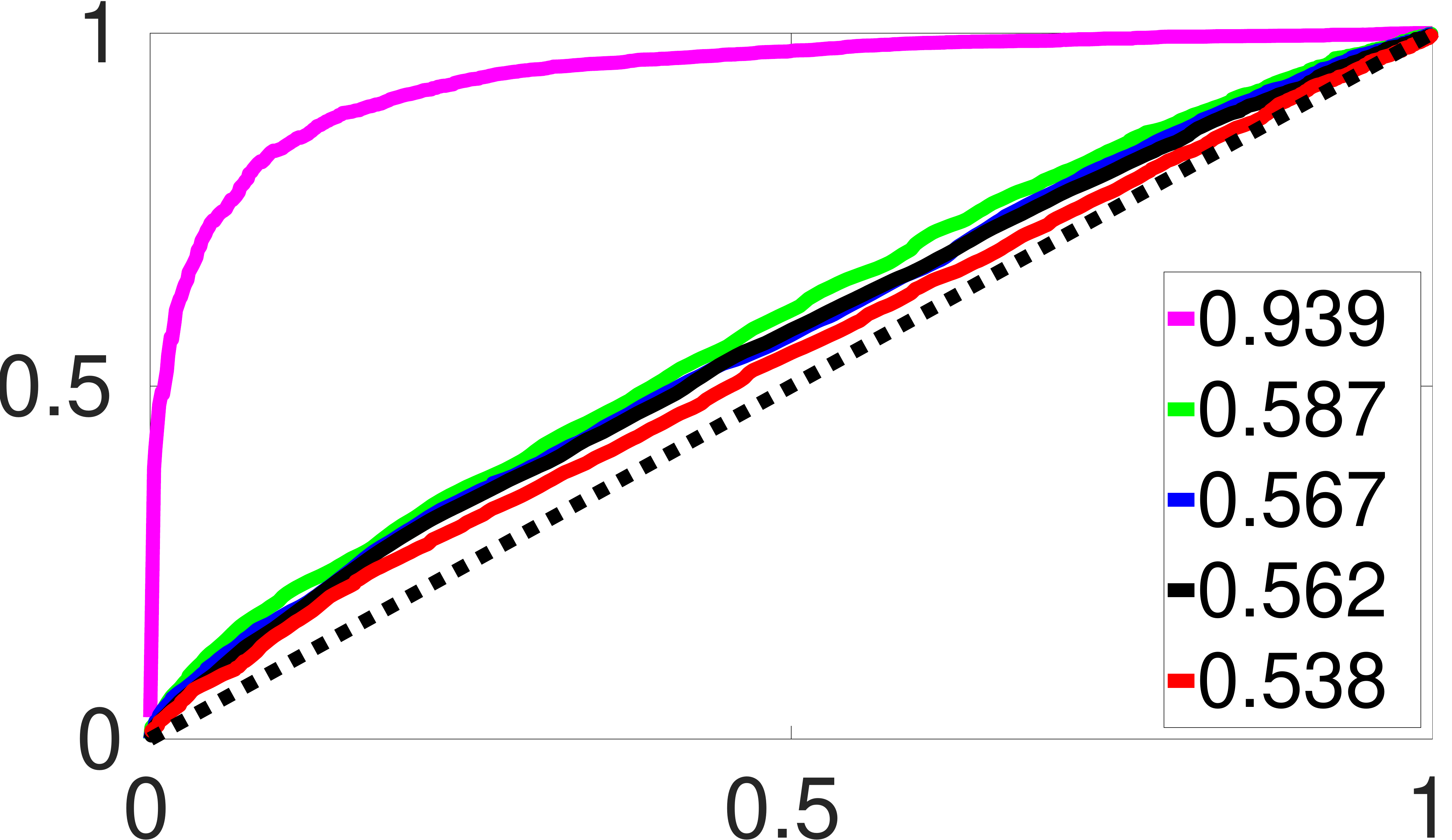}
&
\includegraphics[width=\linewidth]{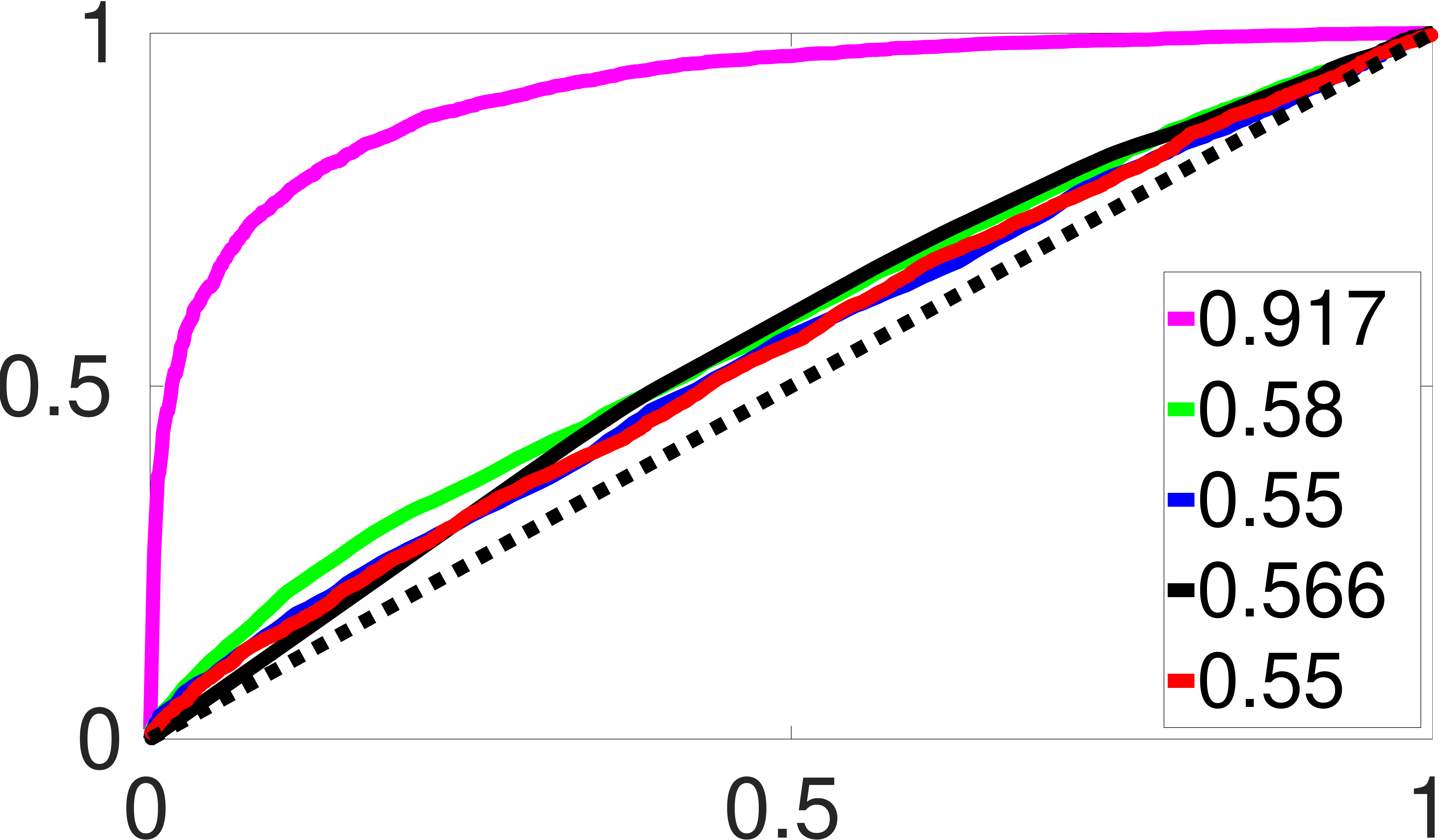}
&
\includegraphics[width=\linewidth]{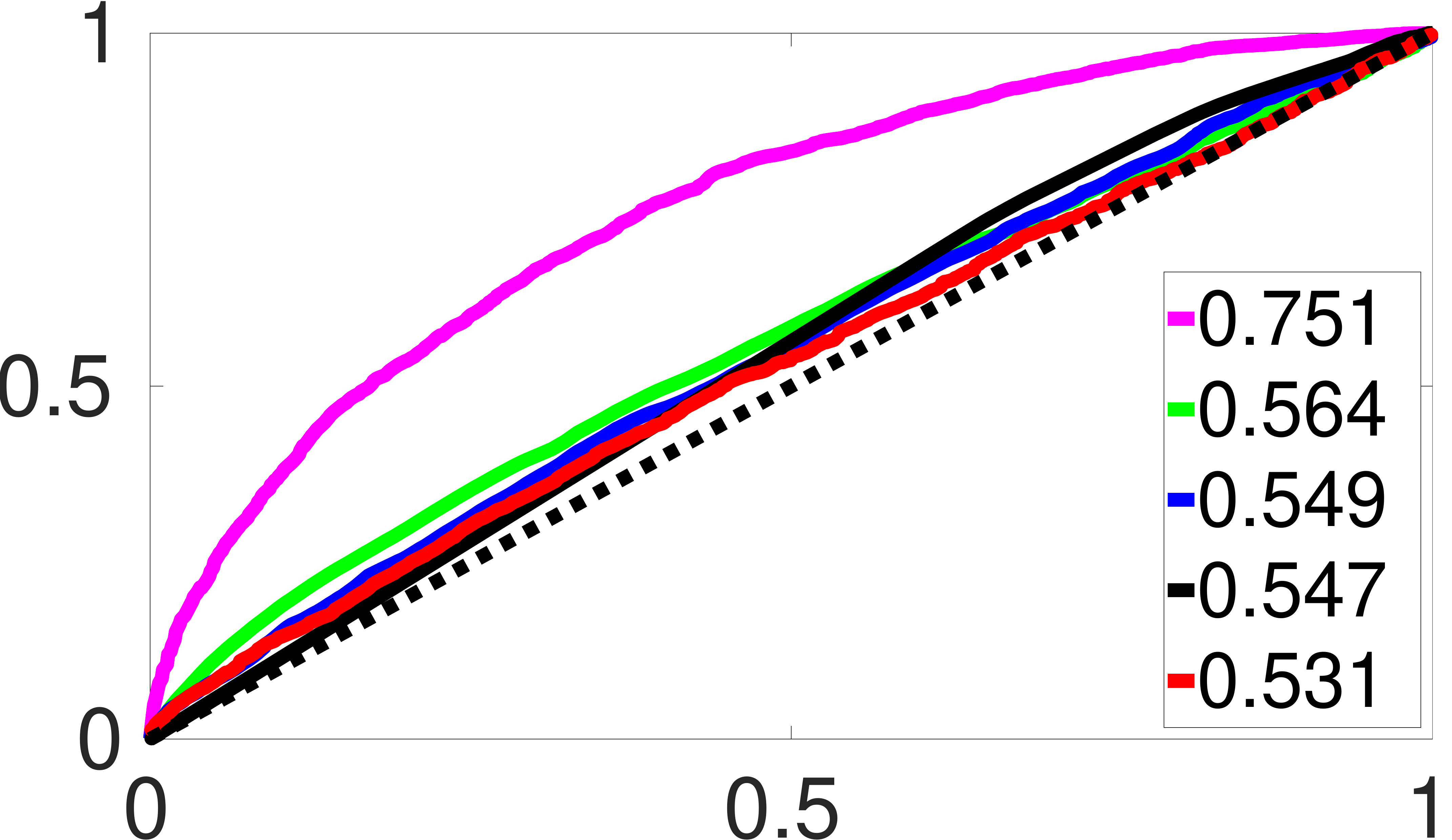}
\\
\includegraphics[width=\linewidth]{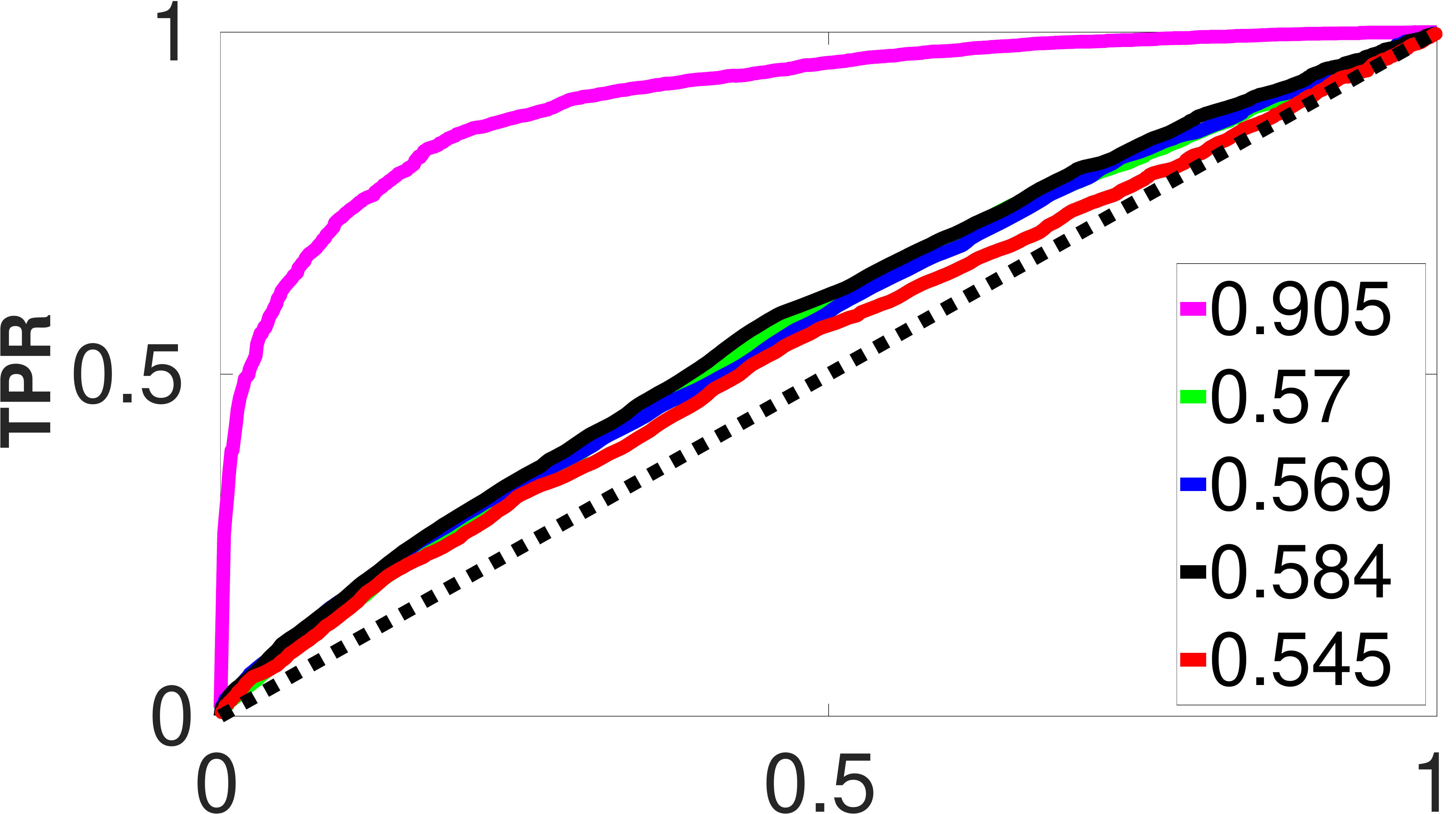}
&
\includegraphics[width=\linewidth]{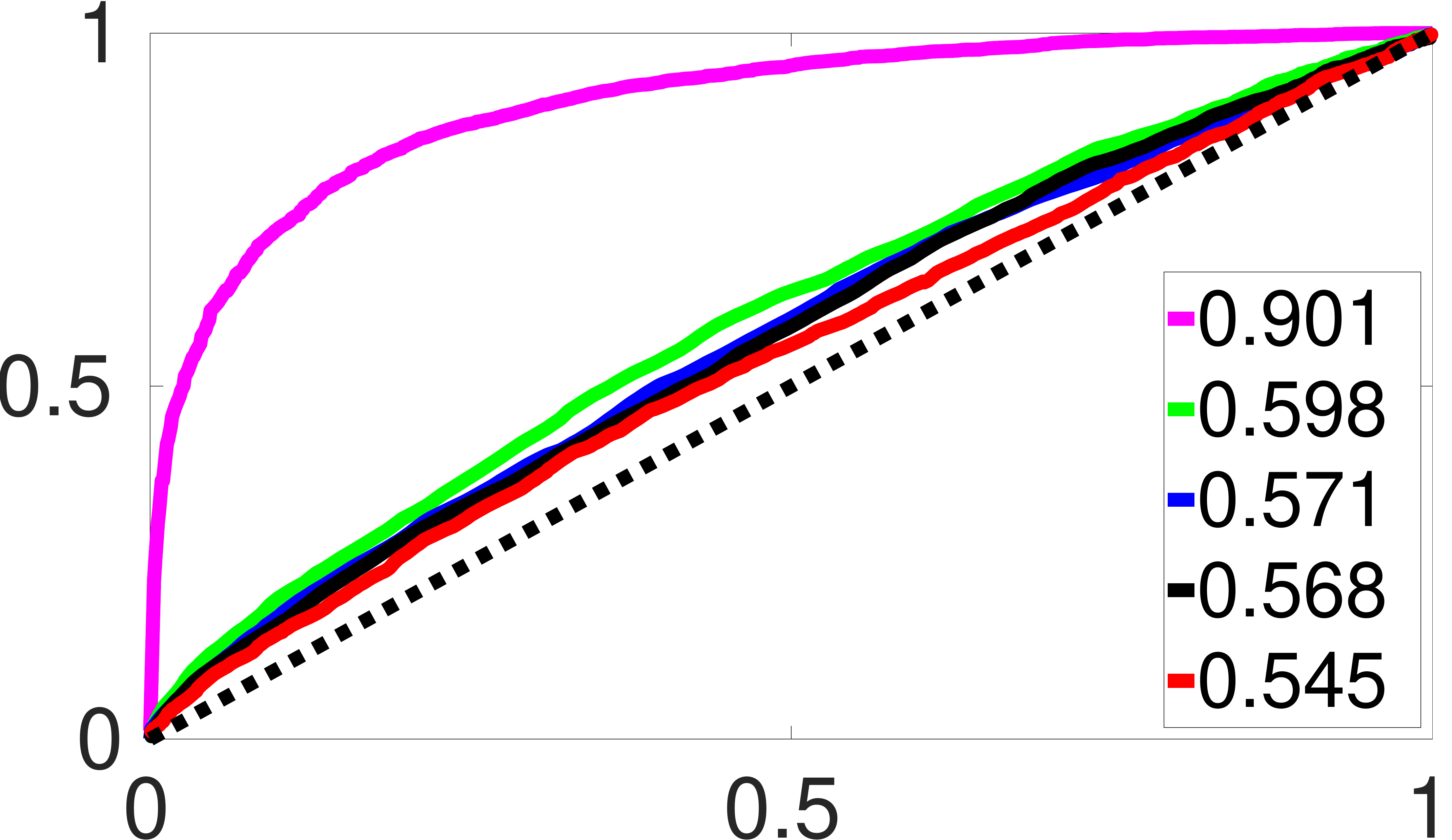}
&
\includegraphics[width=\linewidth]{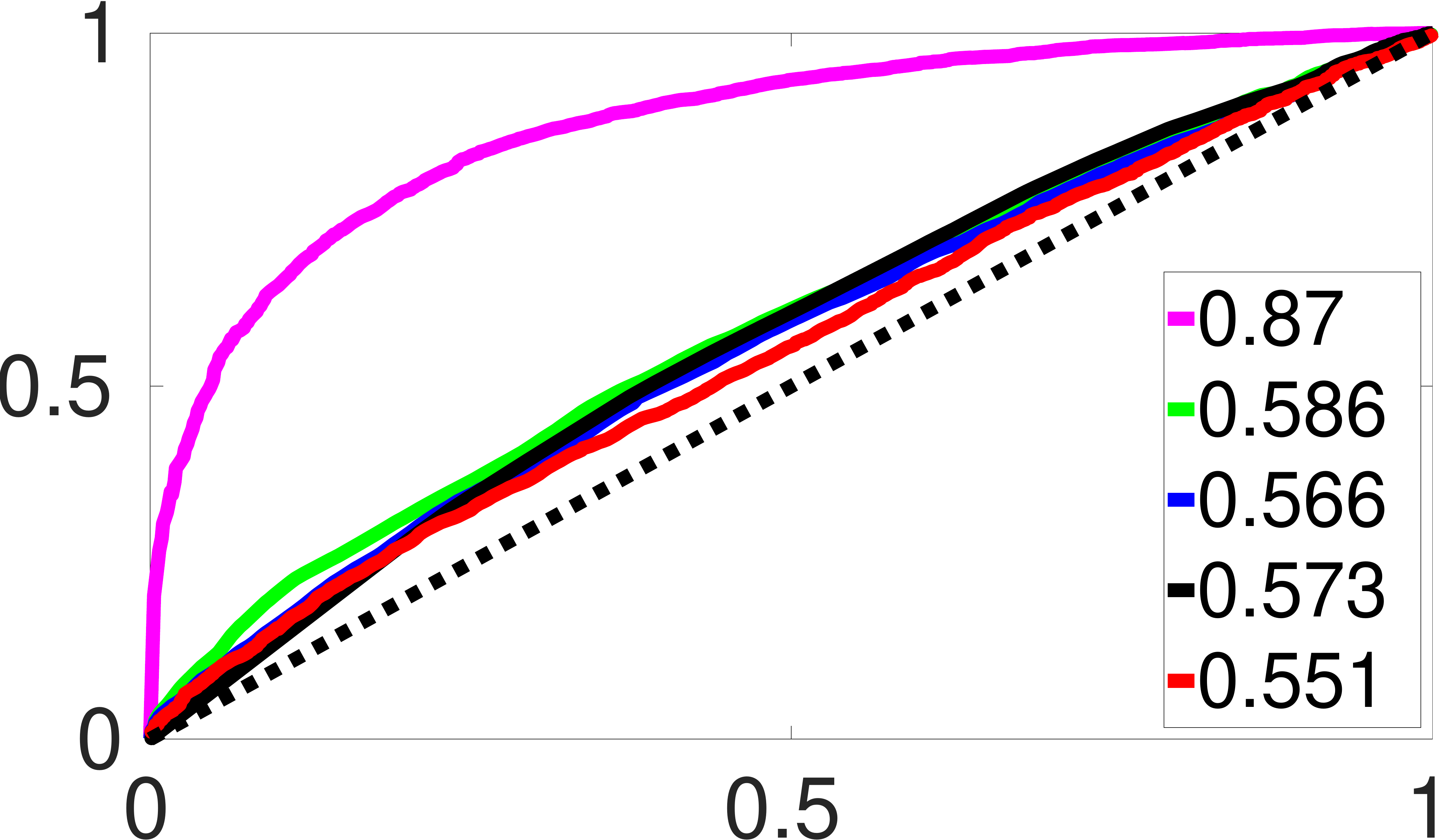}
&
\includegraphics[width=\linewidth]{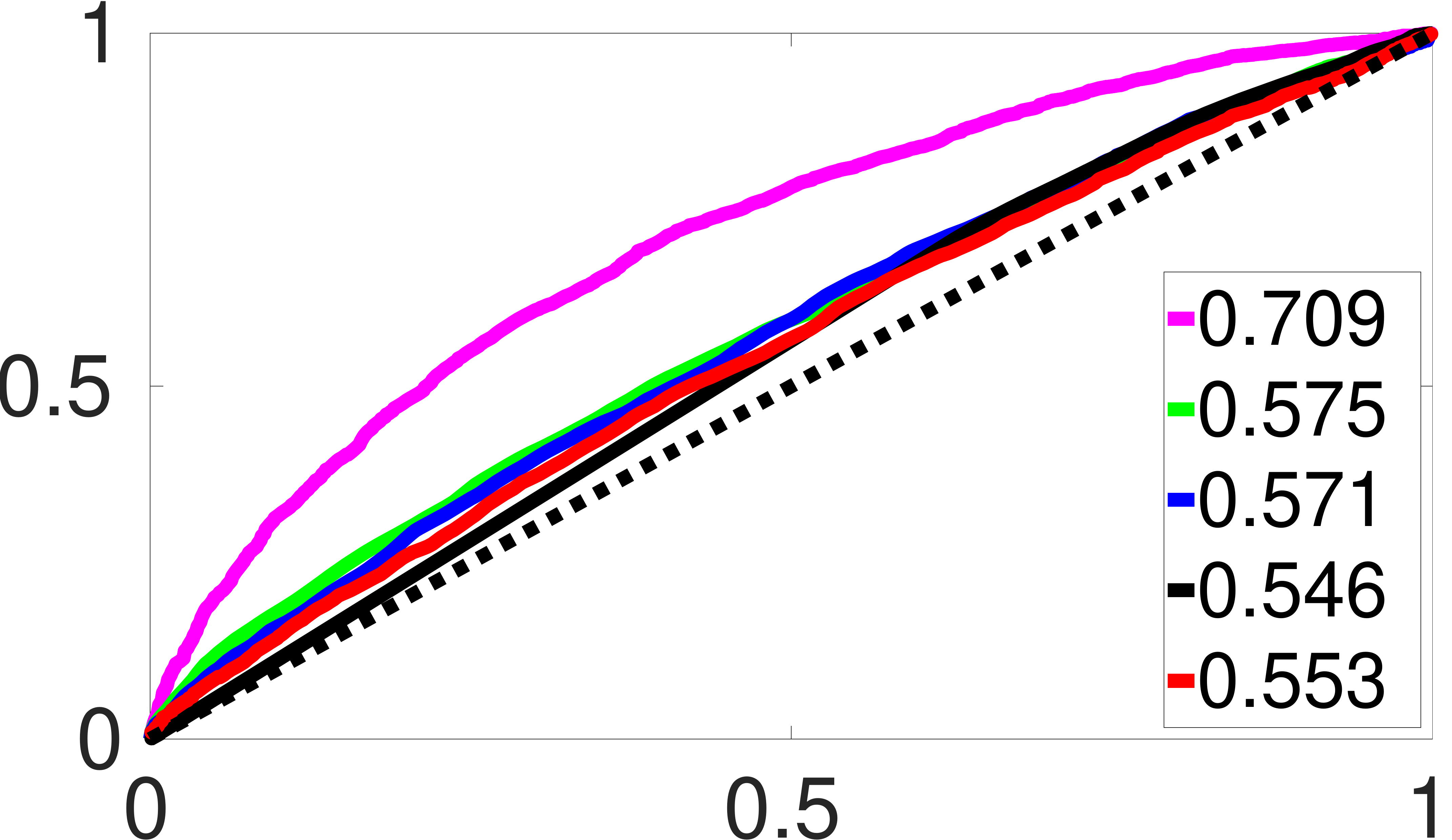}
\\
\includegraphics[width=\linewidth]{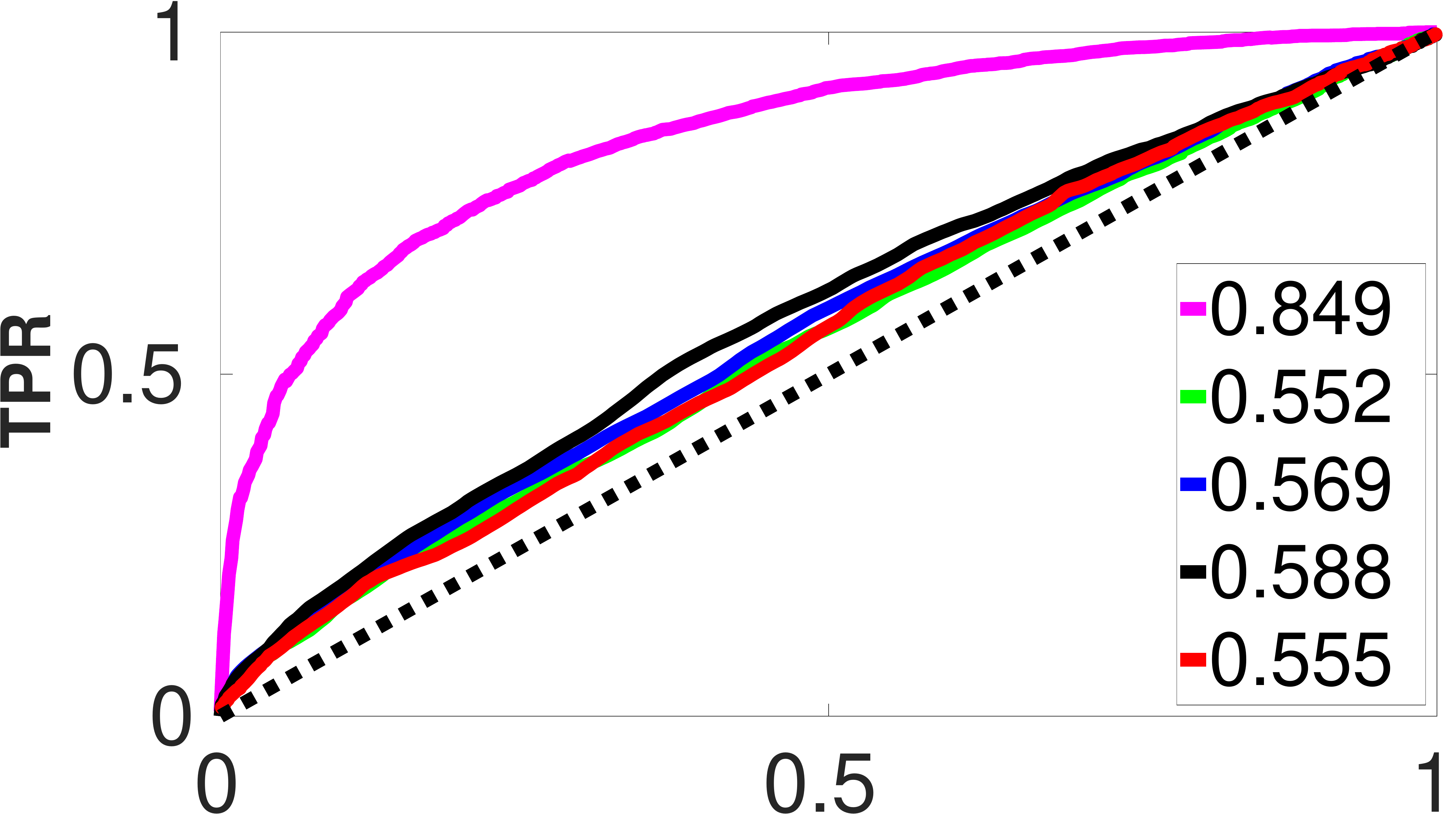}
&
\includegraphics[width=\linewidth]{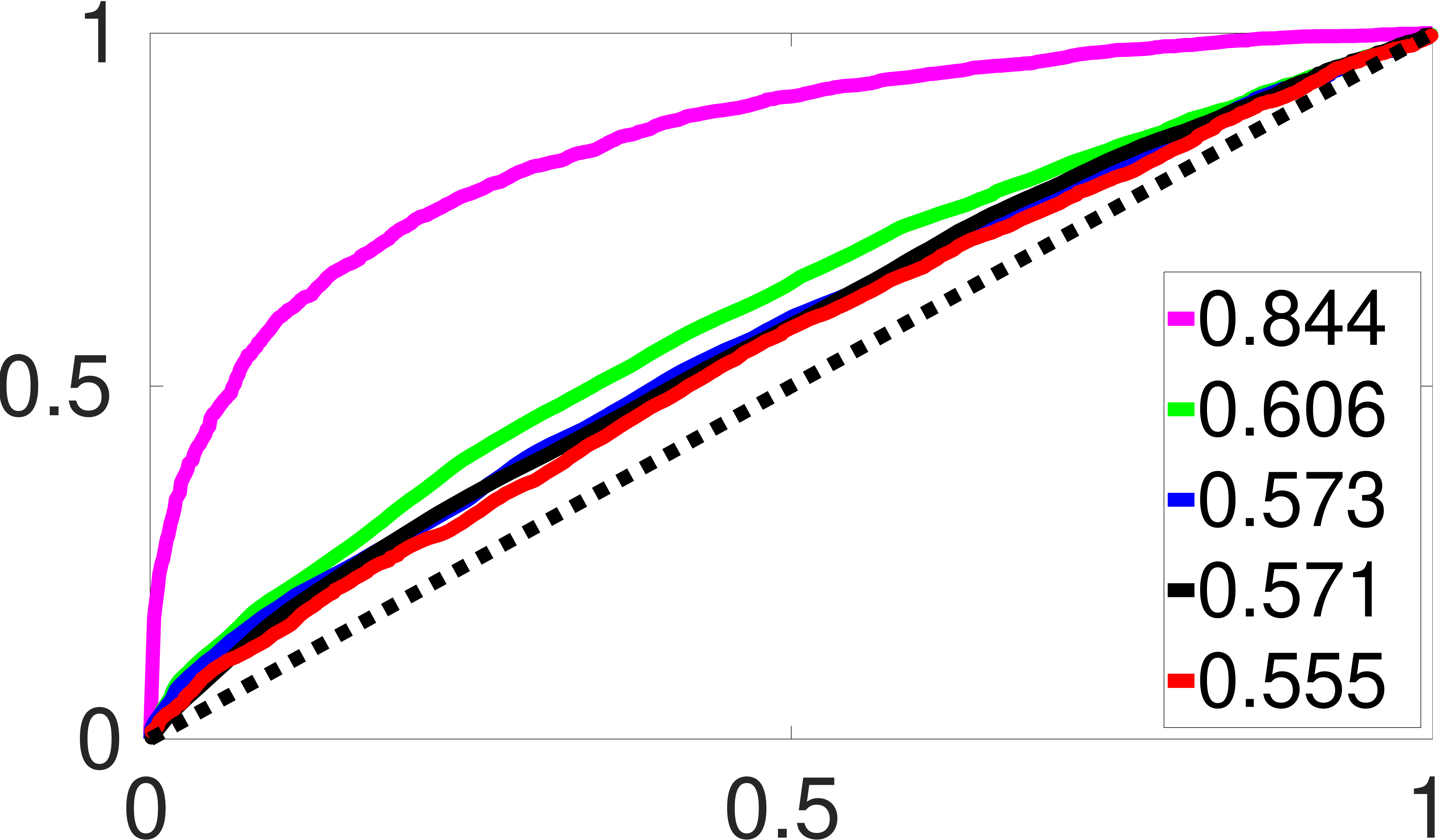}
&
\includegraphics[width=\linewidth]{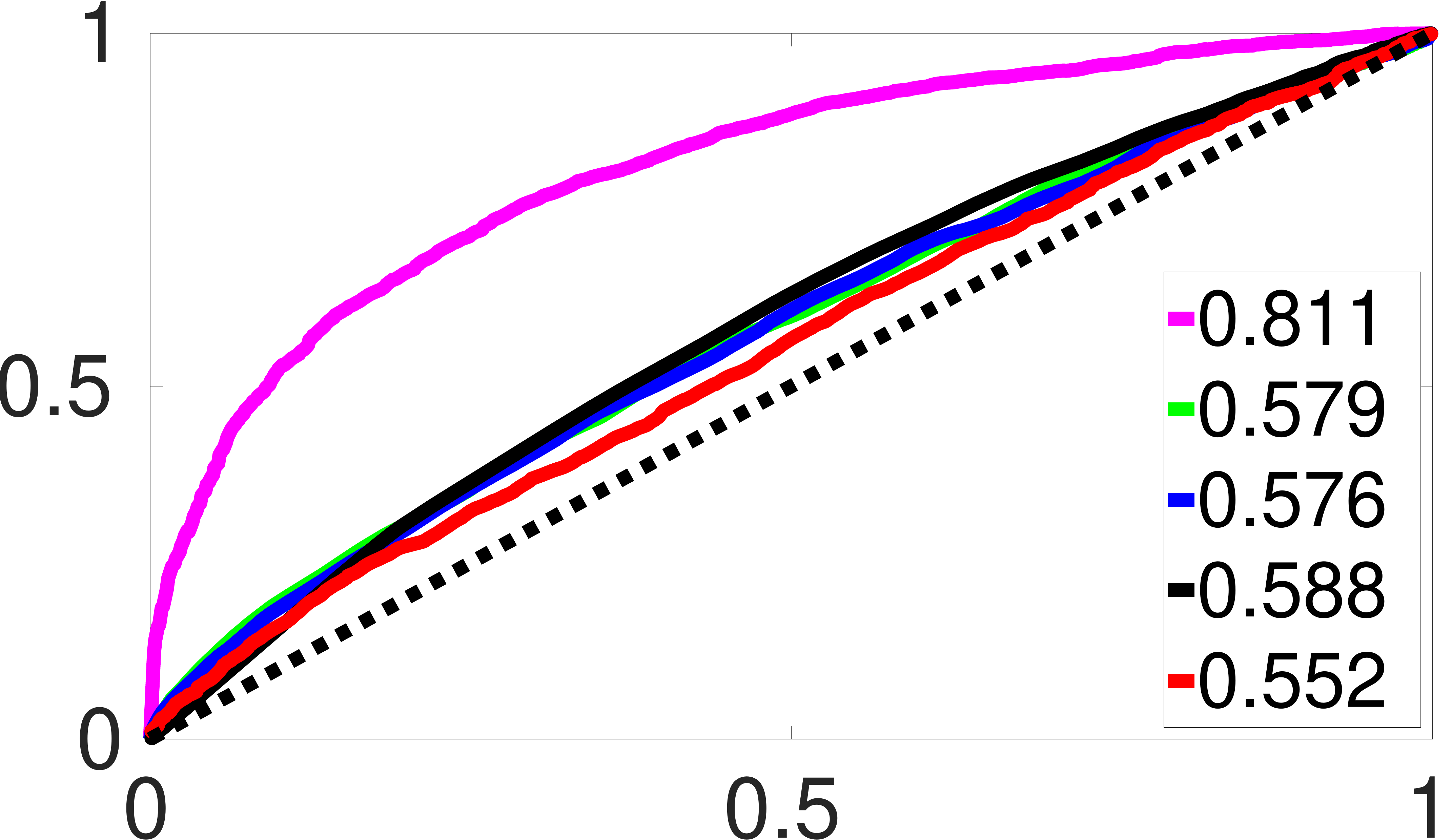}
&
\includegraphics[width=\linewidth]{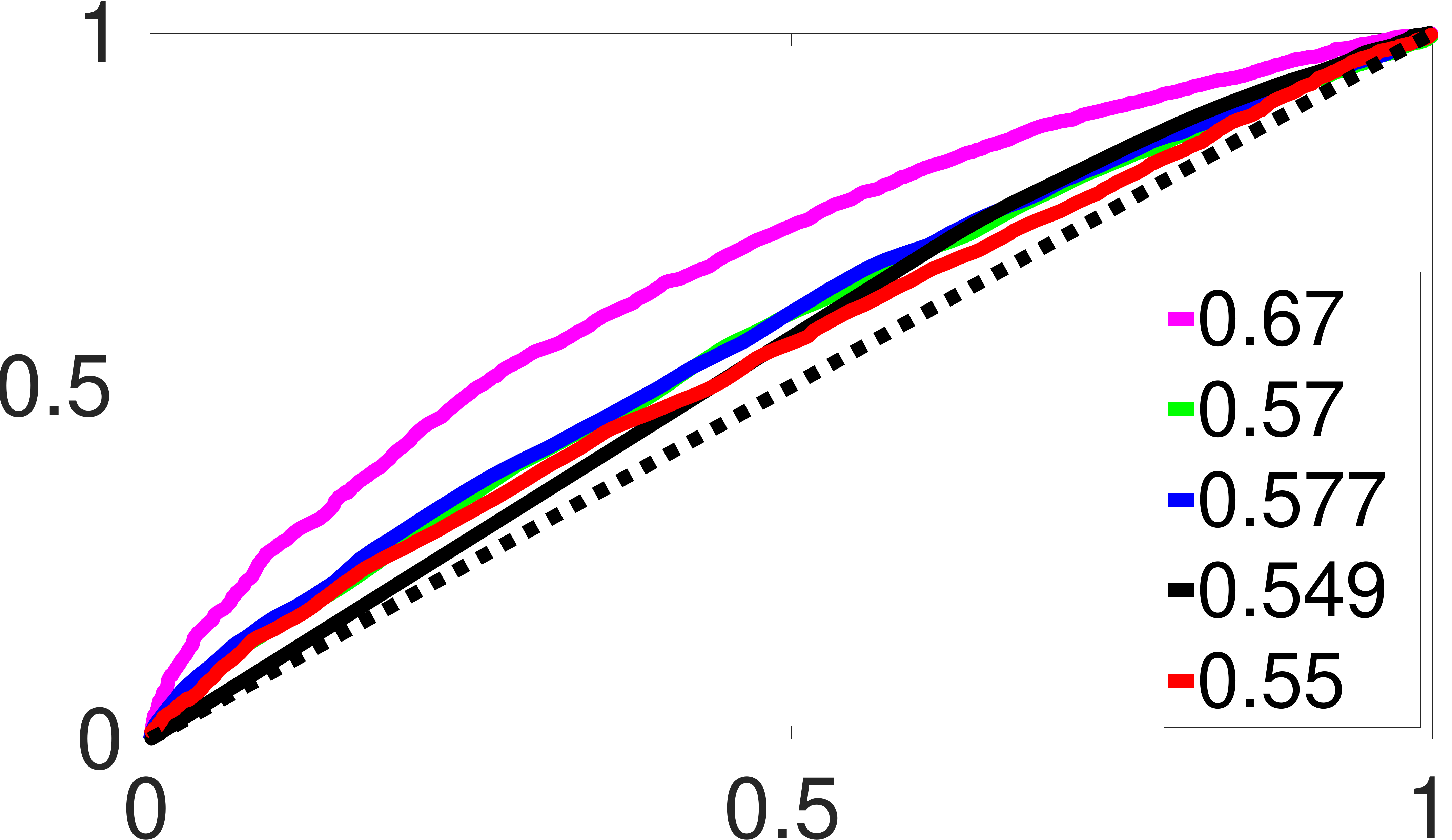}
\\
\includegraphics[width=\linewidth]{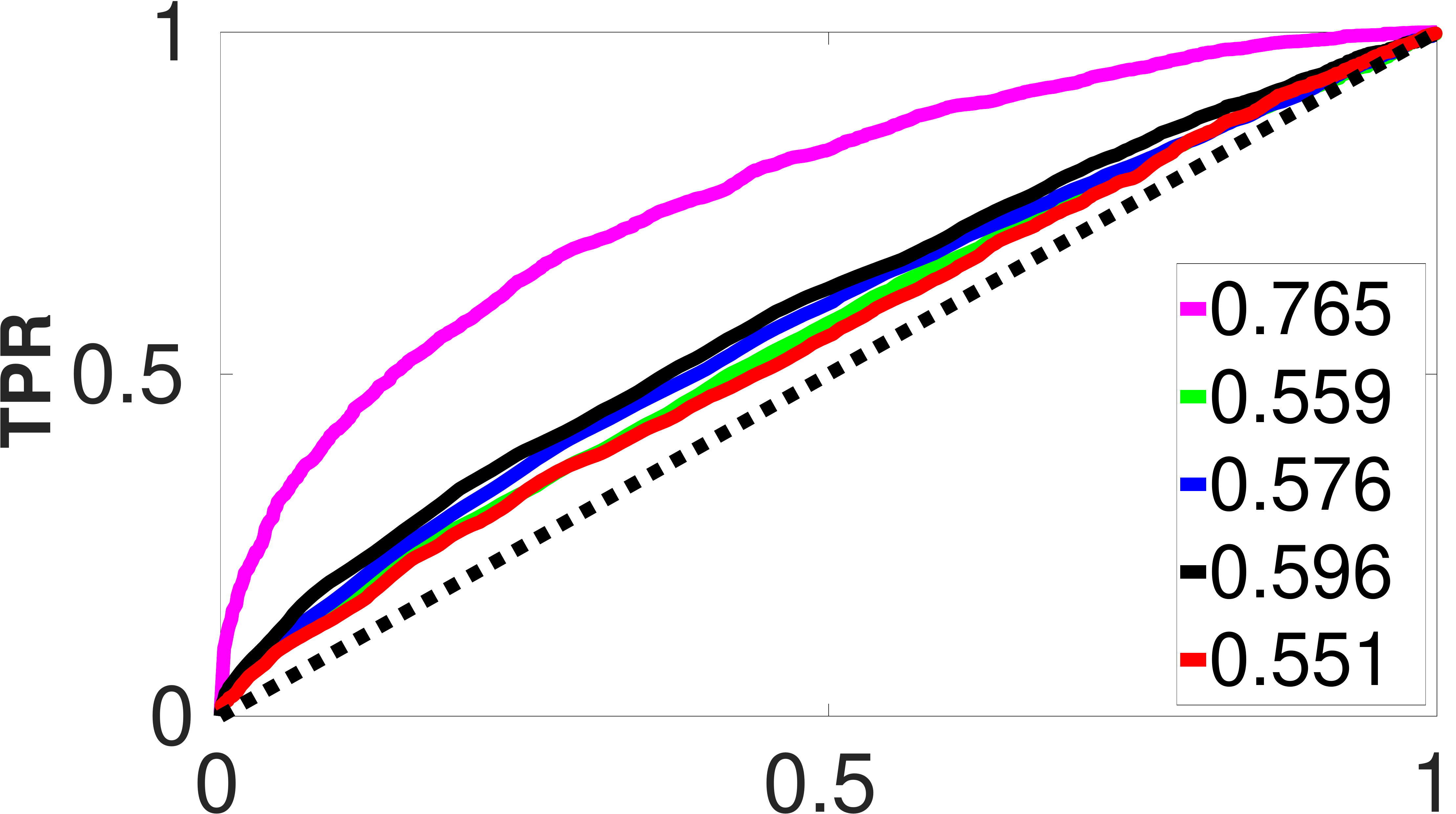}
&
\includegraphics[width=\linewidth]{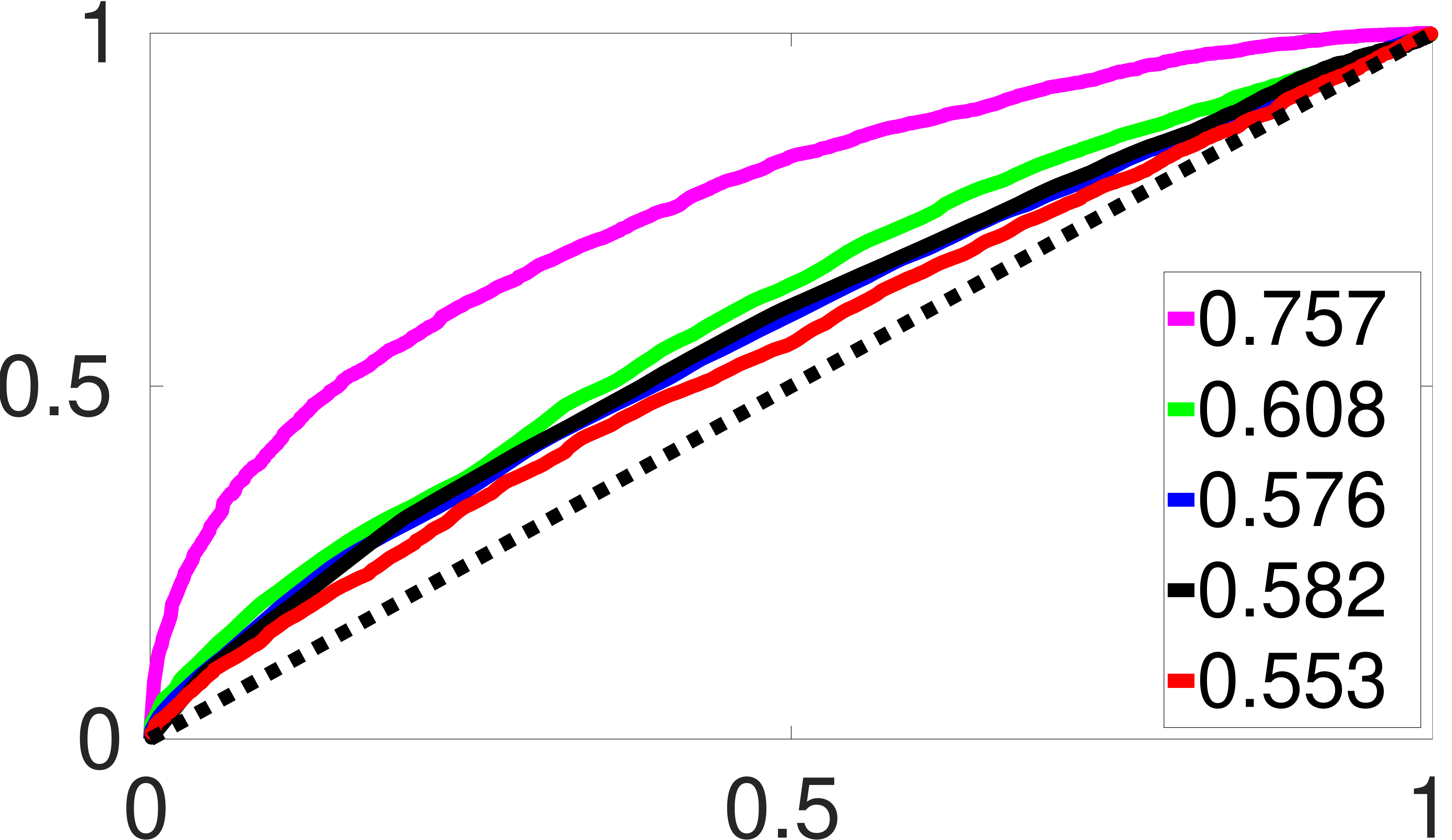}
&
\includegraphics[width=\linewidth]{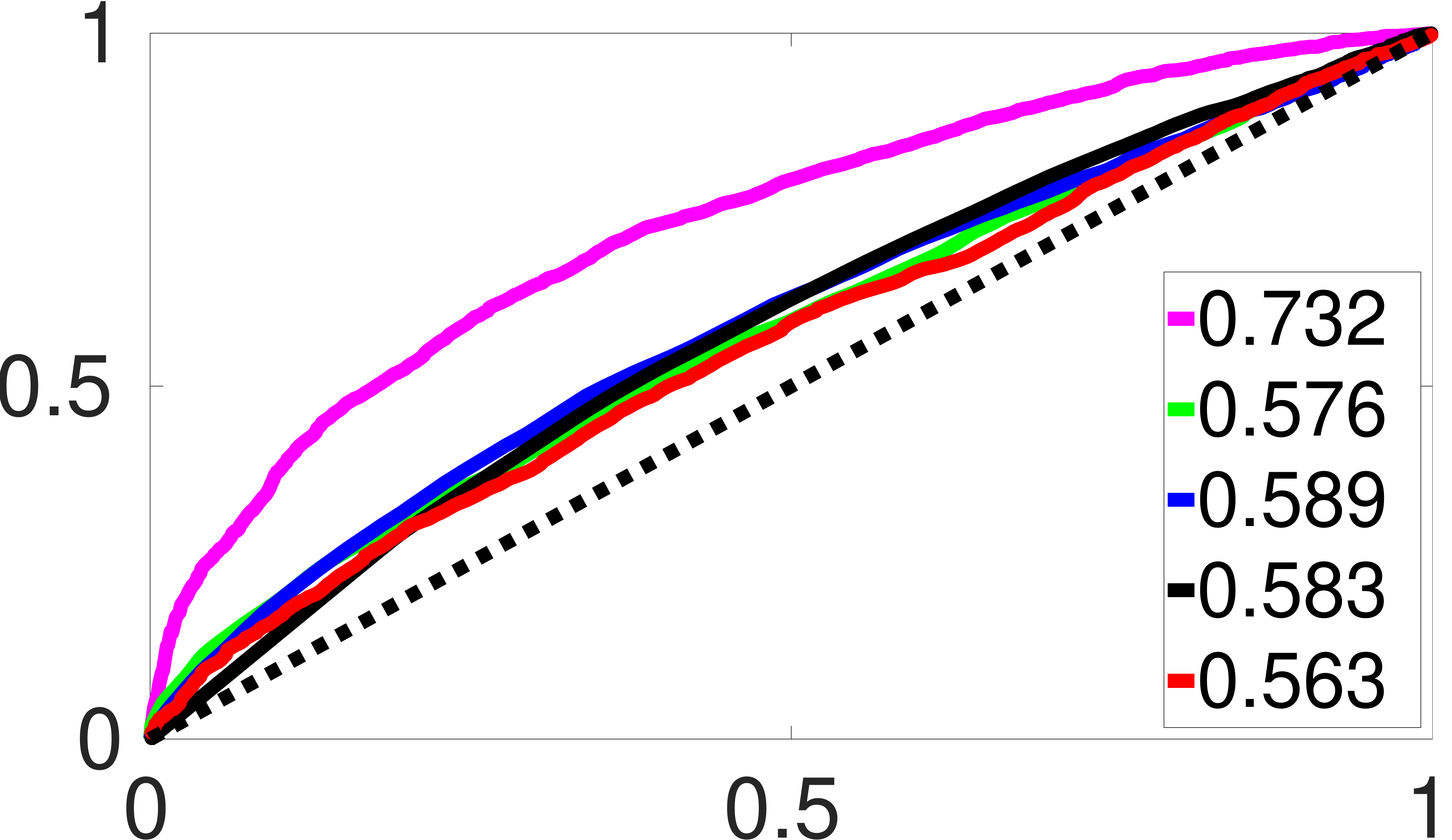}
&
\includegraphics[width=\linewidth]{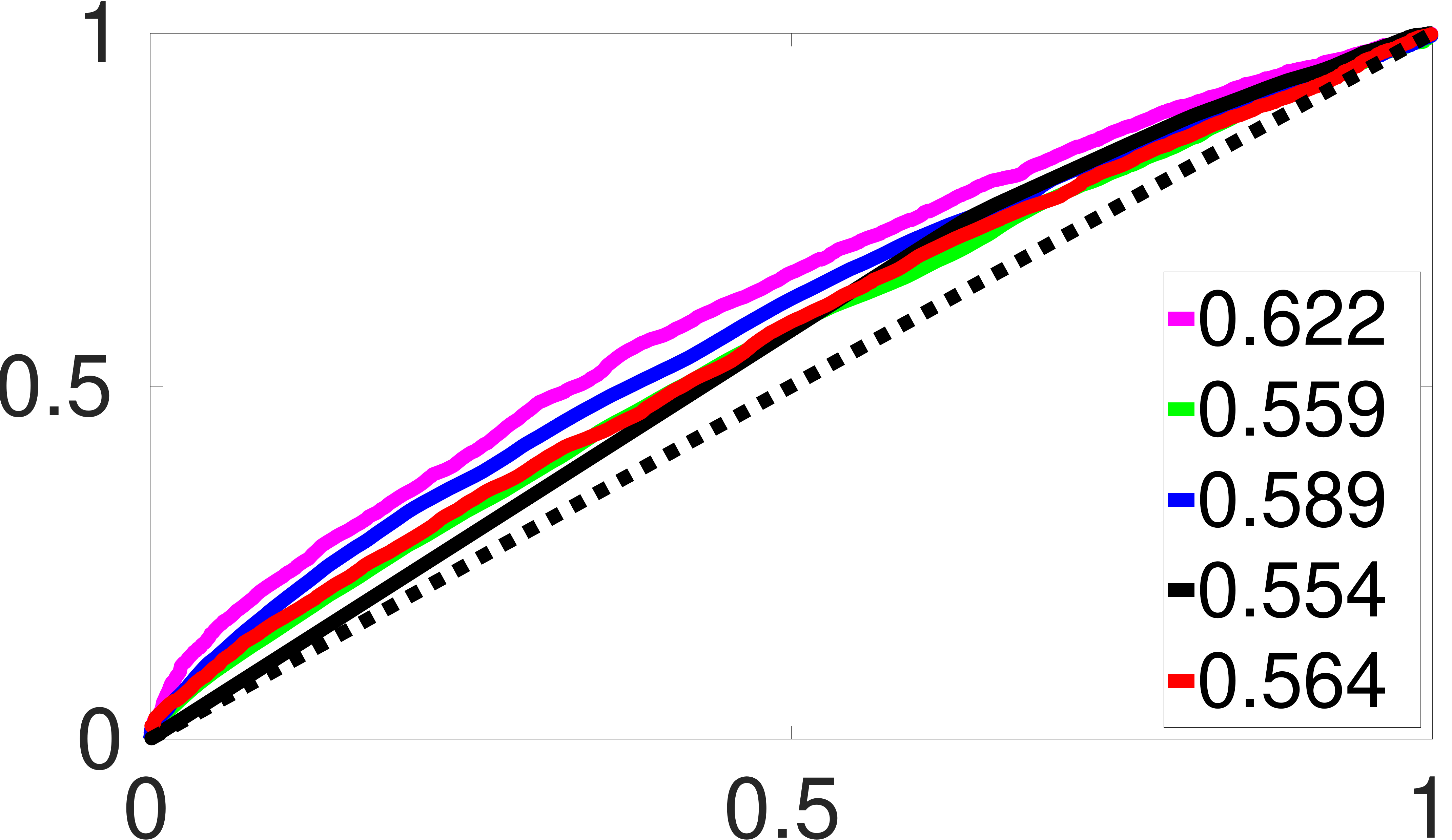}
\\
\includegraphics[width=\linewidth]{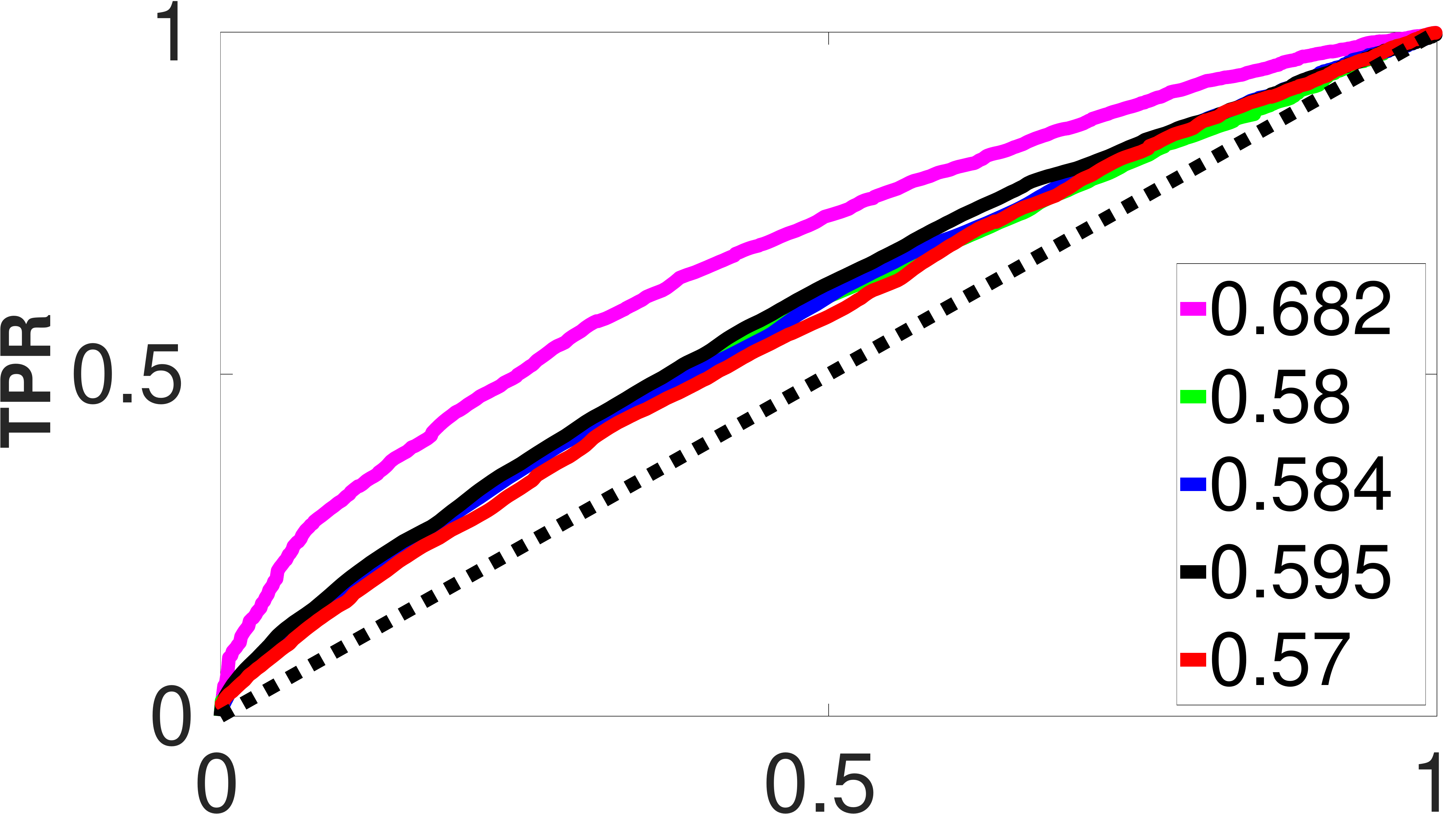}
&
\includegraphics[width=\linewidth]{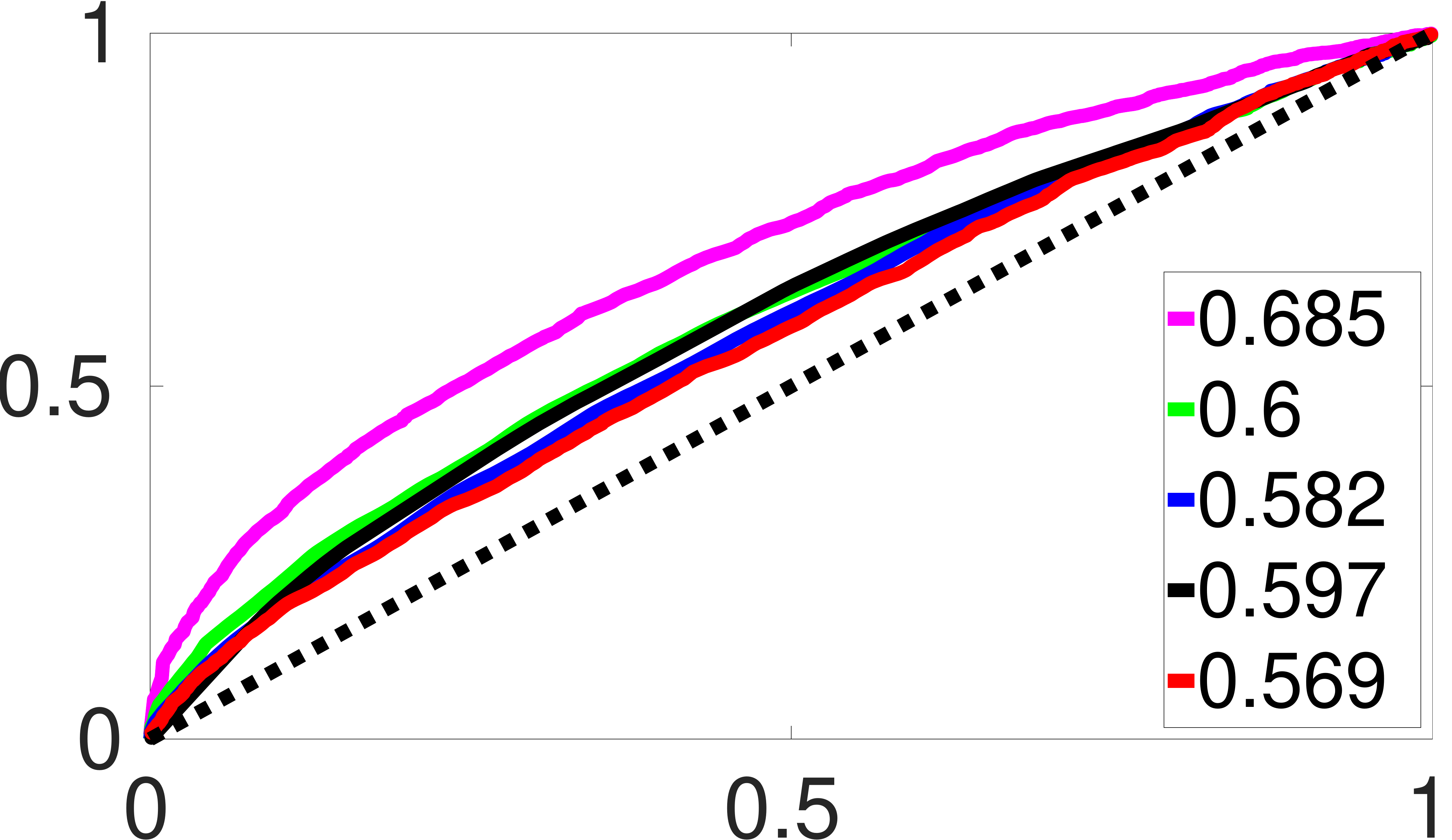}
&
\includegraphics[width=\linewidth]{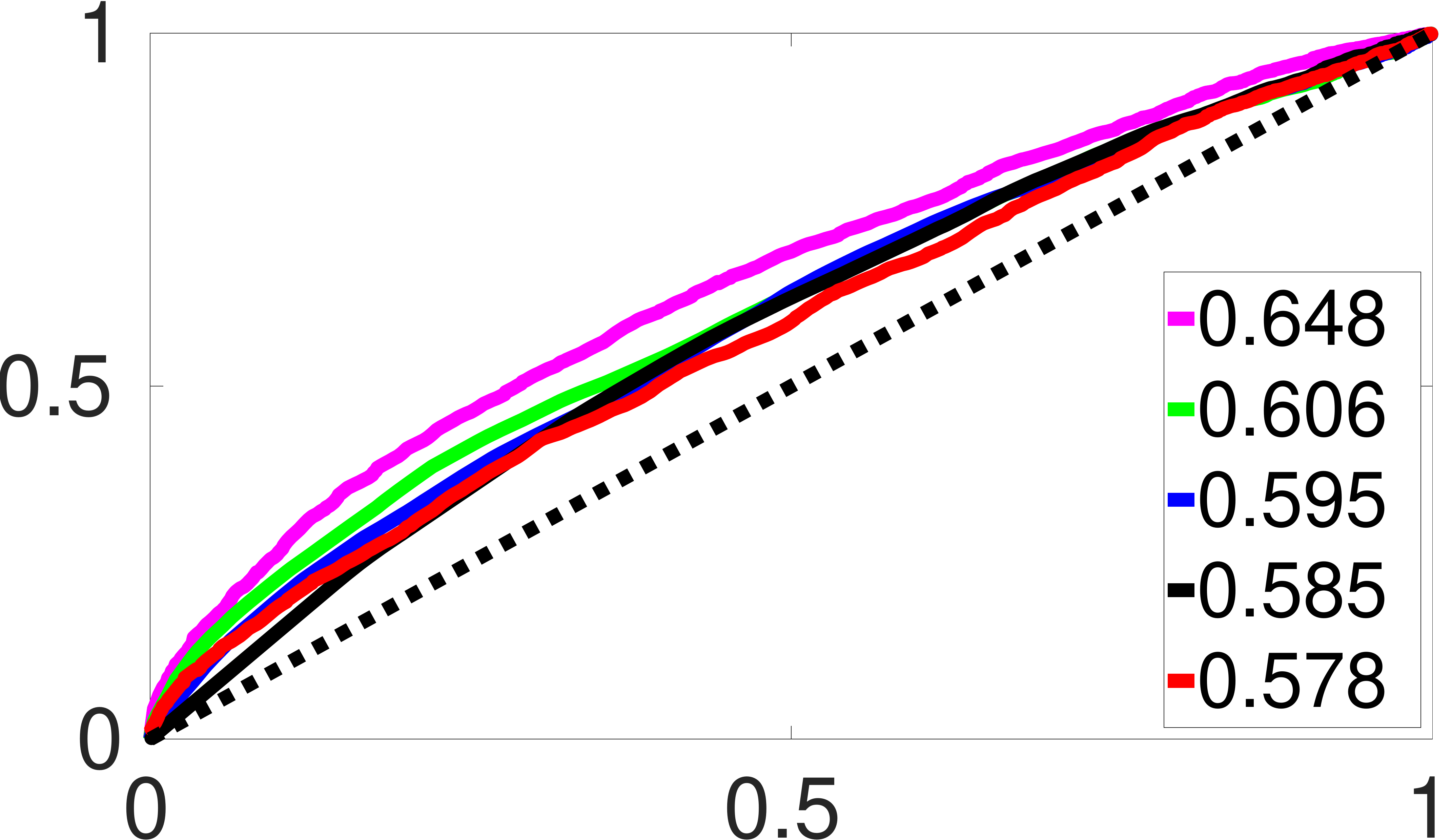}
&
\includegraphics[width=\linewidth]{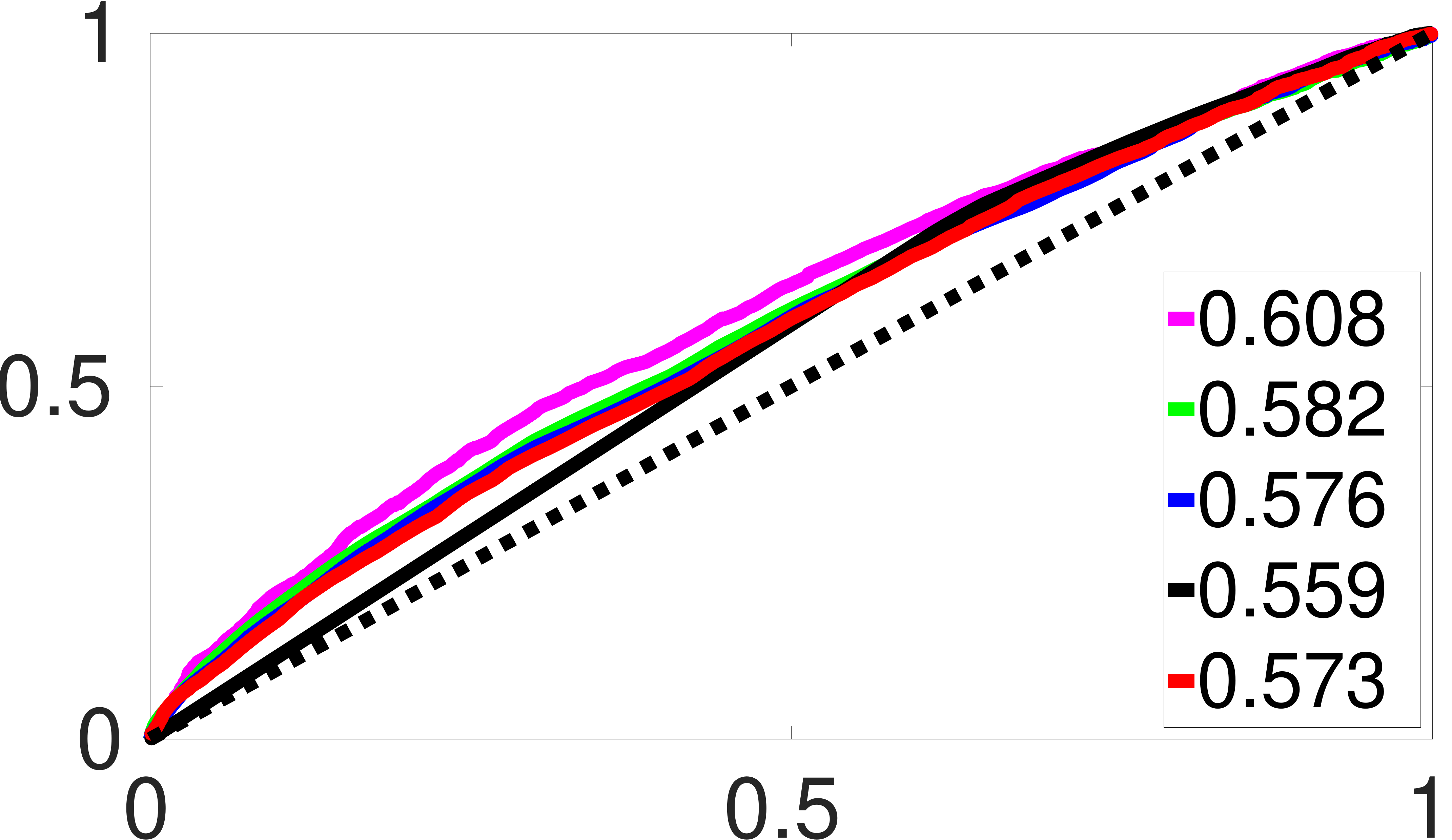}
\\
\includegraphics[width=\linewidth]{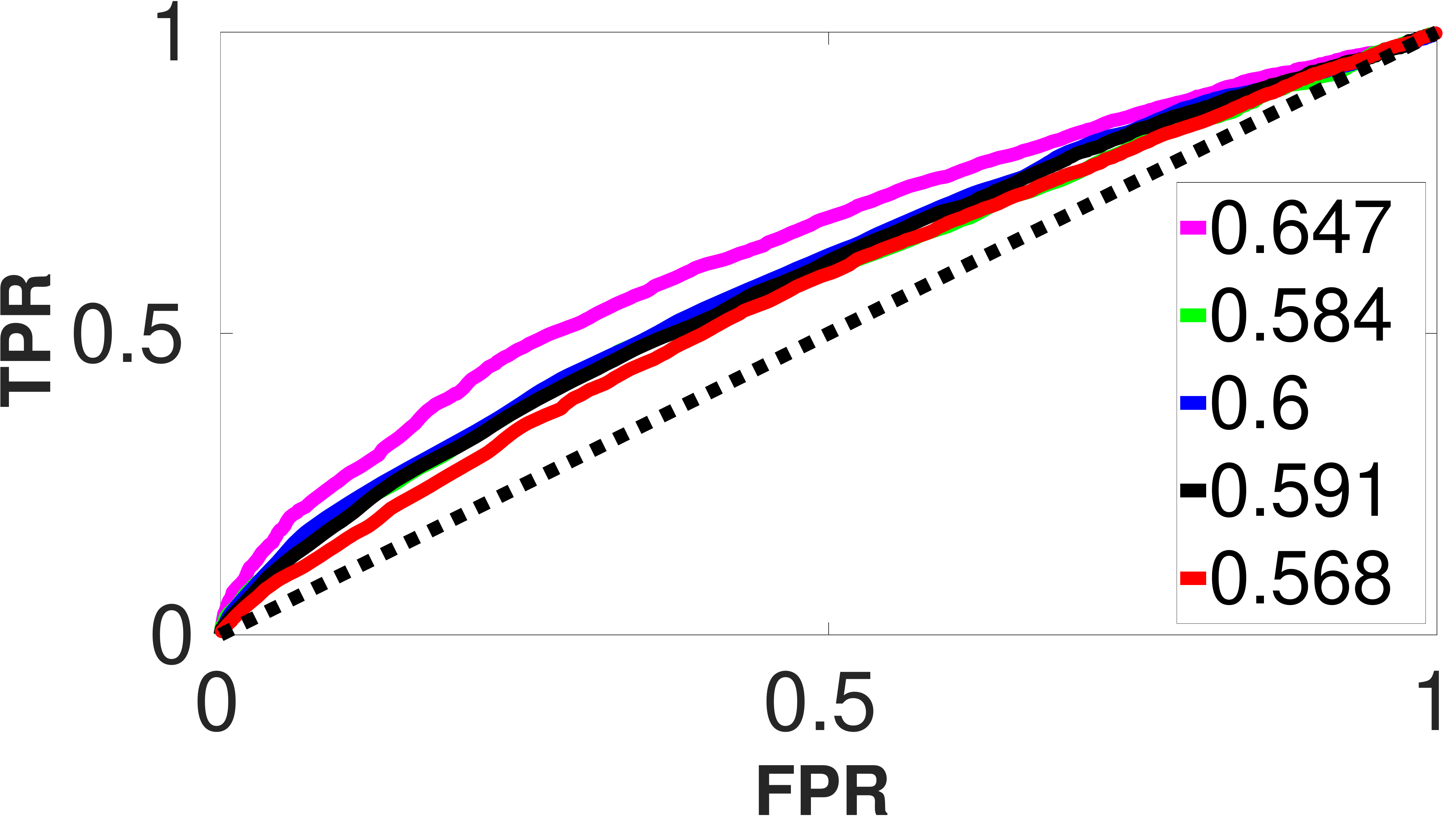}
&
\includegraphics[width=\linewidth]{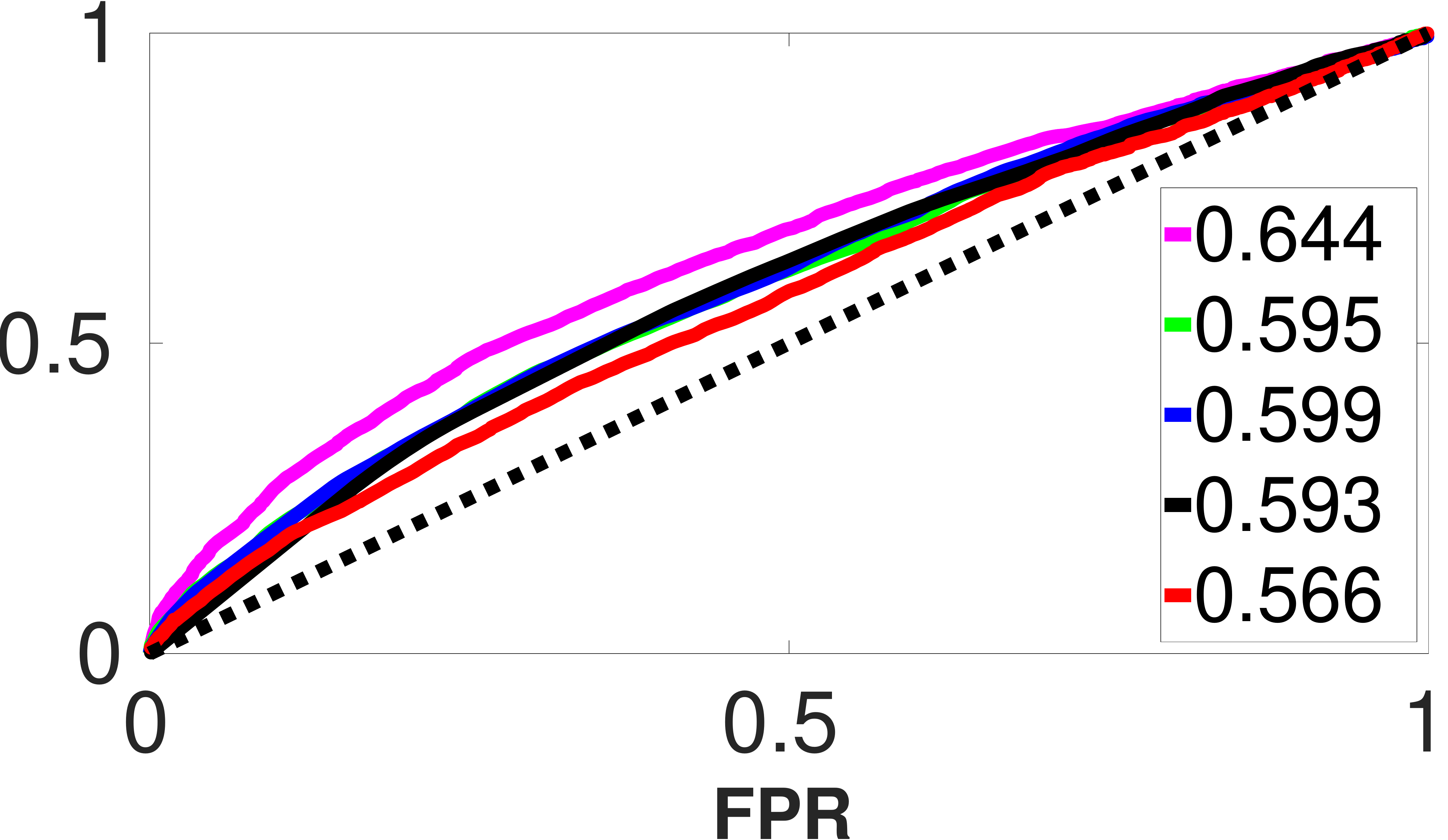}
&
\includegraphics[width=\linewidth]{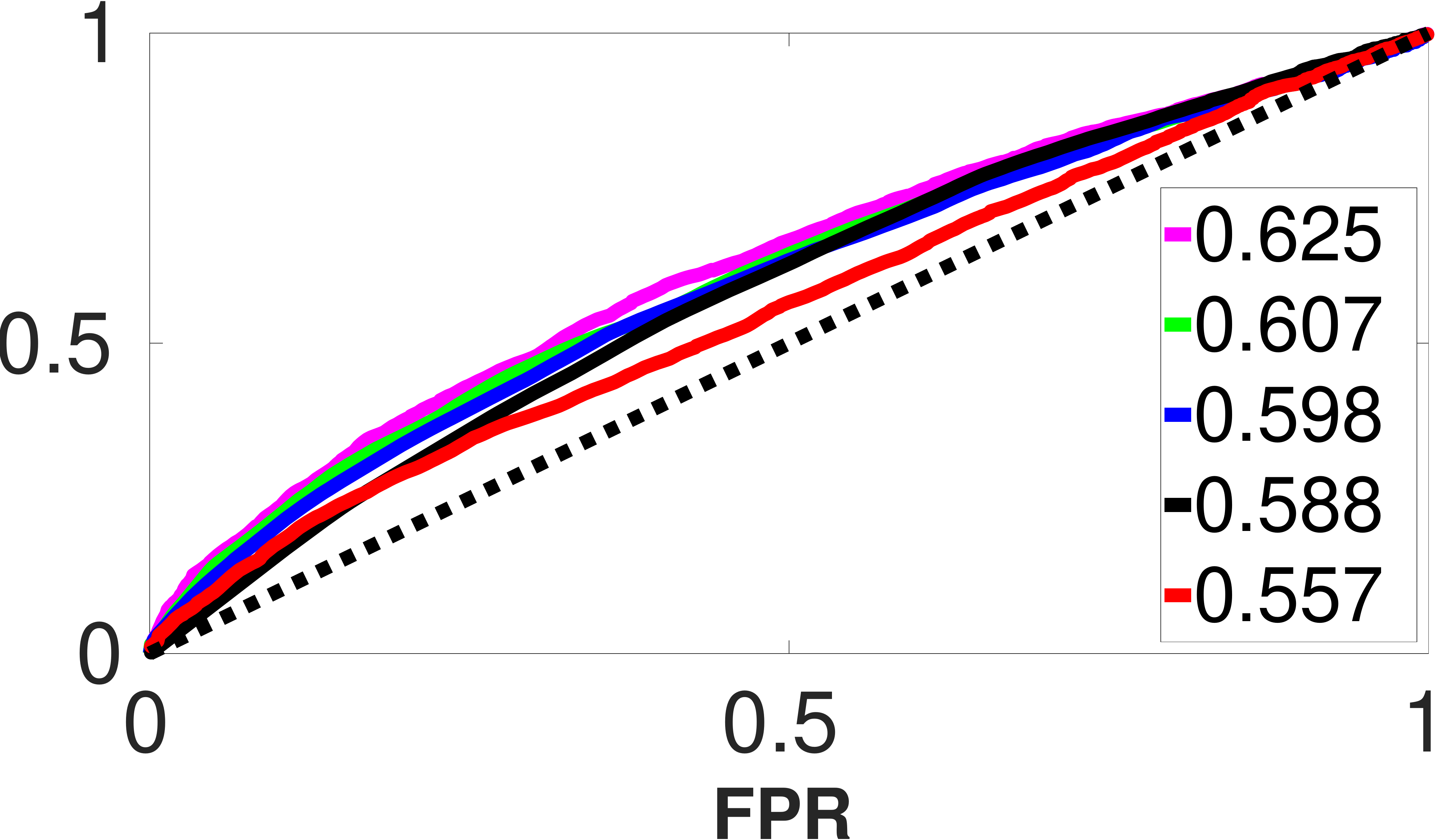}
&
\includegraphics[width=\linewidth]{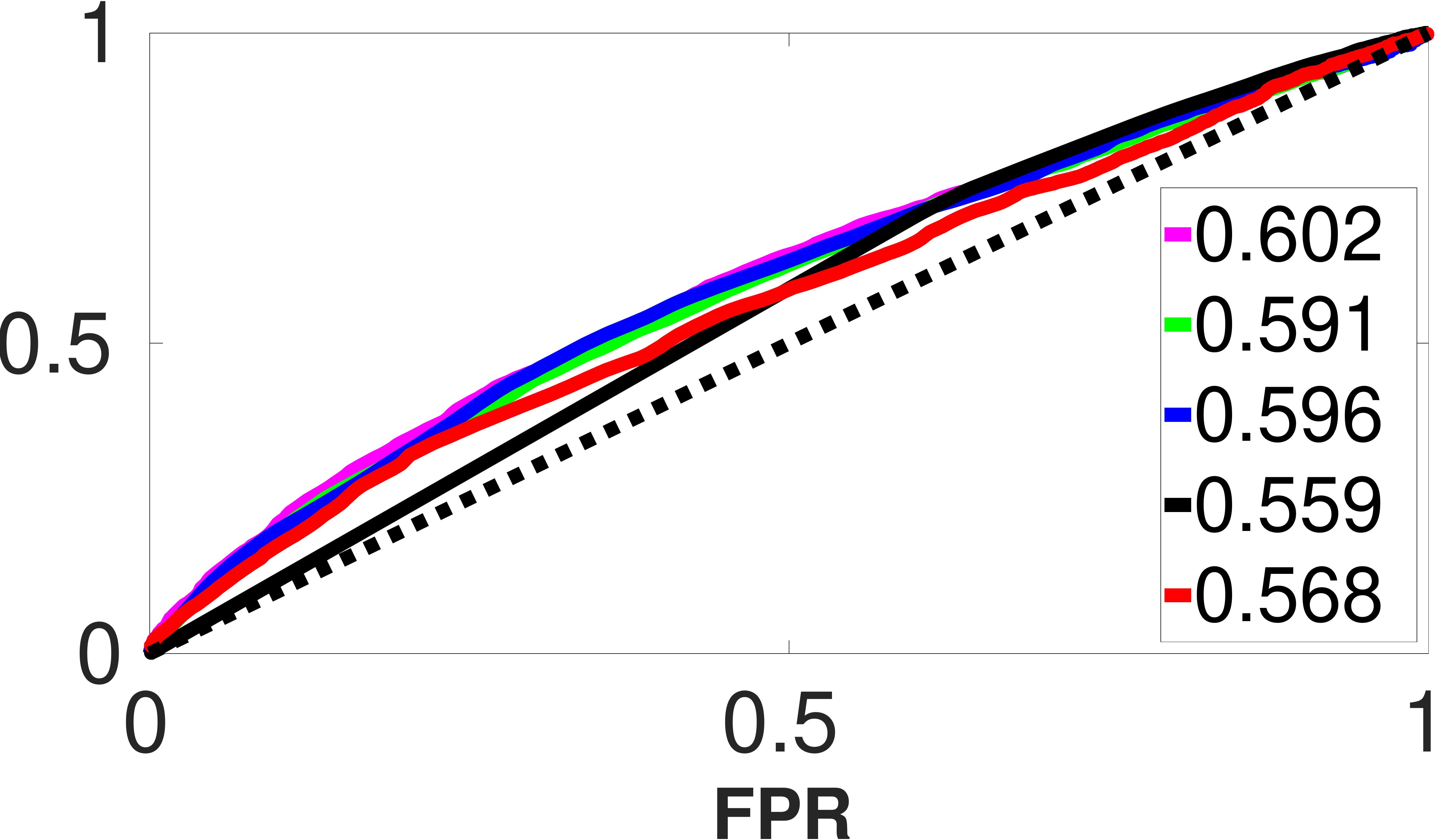} 
\\
\end{tabular}
}
\caption{Receiver Operating Curves (ROCs) for the accurate AHGMM parrot-IF attack at threshold $\rho_j^o=0.5$ px/cm. Each ROC is the mean of 10-curves generated by the 10-folds used for cross validation. Legend: \textcolor{magenta}{\Huge ---} Unprotected, \textcolor{green}{\Huge---} \ac{AGB}, \textcolor{blue}{\Huge---} \ac{SVGB}, \textcolor{black}{\Huge---} \ac{FGB}, \textcolor{red}{\Huge ---} AHGMM. In each column, the image resolution remains constant, i.e.~first column: $96 \times 96$, second column: $48 \times 48$, third column: $24 \times 24$ and fourth column: $12 \times 12$ pixels,  while the pitch angle varies~i.e. first row: $0^\circ$, second row: $10^\circ$, third row: $20^\circ$, fourth row: $30^\circ$, fifth row: $40^\circ$, sixth row: $50^\circ$, seventh row: $60^\circ$ and eighth row: $70^\circ$. The legend values represent the Area Under Curve (AUC).}
\label{fig:ROCs_accurate_ahgmm_parrot-if}
\end{figure*}

In the inverse-filter (IF) attack, we reconstruct the probe images by deconvolving the protected face with an accurate or estimated kernel. We evaluate the IF attack under four sub-attacks: optimal kernel na\"{i}ve-IF sub-attack, pseudo AHGMM na\"{i}ve-IF sub-attack, accurate AHGMM na\"{i}ve-IF sub-attack and accurate AHGMM parrot-IF sub-attack. Fig. \ref{fig:inverse_attack} depicts the effect of inverse filtering on selected sample images protected with AGB, SVGB and AHGMM. Fig. \ref{fig:inverse_attack_graphs} shows the achieved accuracies under the different sub-attacks at different values of $\rho_j^o$, while Fig. \ref{fig:ROCs_accurate_ahgmm_parrot-if} presents \acp{ROC} for the accurate \ac{AHGMM} parrot-IF sub-attack at $\rho_j^o=0.5$ px/cm. 

As can be seen in Fig. \ref{fig:inverse_attack}, the face reconstruction quality decreases when the threshold increases (increasing the filter kernel) even if the filter parameters are known. This is true for both space invariant Gaussian blur (AGB) and linear space variant Gaussian blur (SVGB). The main reason is that the boundaries of the face start propagating towards the center of the face as the threshold is decreased. Thus, it becomes difficult to distinguish between reconstructed faces at the lower thresholds (see Fig. \ref{fig:inverse_attack_graphs}).

In case of non-linear space variant blur (AHGMM), the reconstruction becomes more challenging even when the same hopping kernels are used as for the protection. The main reason, in addition to the boundary propagation, is that while deconvolving a sub-region, the IF incorrectly treats the adjacent subregions as if they were filtered with the same kernel, thus not enabling it to reconstruct the original face (see Fig \ref{fig:inverse_attack}). Consequently, it becomes difficult to accurately predict the label of the reconstructed face. 


In contrast to na\"{i}ve-IF attacks, parrot-IF attack is more severe and increases significantly the accuracy, especially for AGB, FGB and SVGB. AHGMM also shows the accuracy improvement but less than AGB, FGB and SVGB; and is more robust to an inverse filter attack even when using an accurate secret key.
\subsection{Super-resolution Attack}
\label{subsec:super_resolution_attack}

In this attack, we reconstruct the filtered probe images with SRCNN \citep{Dong2016}. SRCNN first learns a mapping between the high-resolution images and their corresponding low-resolution version, and then applies this mapping to enhance the details of a low-resolution image. We learn the SRCNN mapping for $1,000,000$ iterations between the protected images (i.e.~the low resolution) and their corresponding unprotected images (i.e.~the high resolution) using the same data sets (91-images and Set5) as used in \citep{Dong2016}. As learning of the mapping is a time consuming process, we investigate the super-resolution attack for a single point of our synthetic data set: 12000 images each with $96 \times 96$ pixels and $0^\circ$ pitch angle.

\begin{figure}[!t]
\centering
\includegraphics[width=0.6\columnwidth]{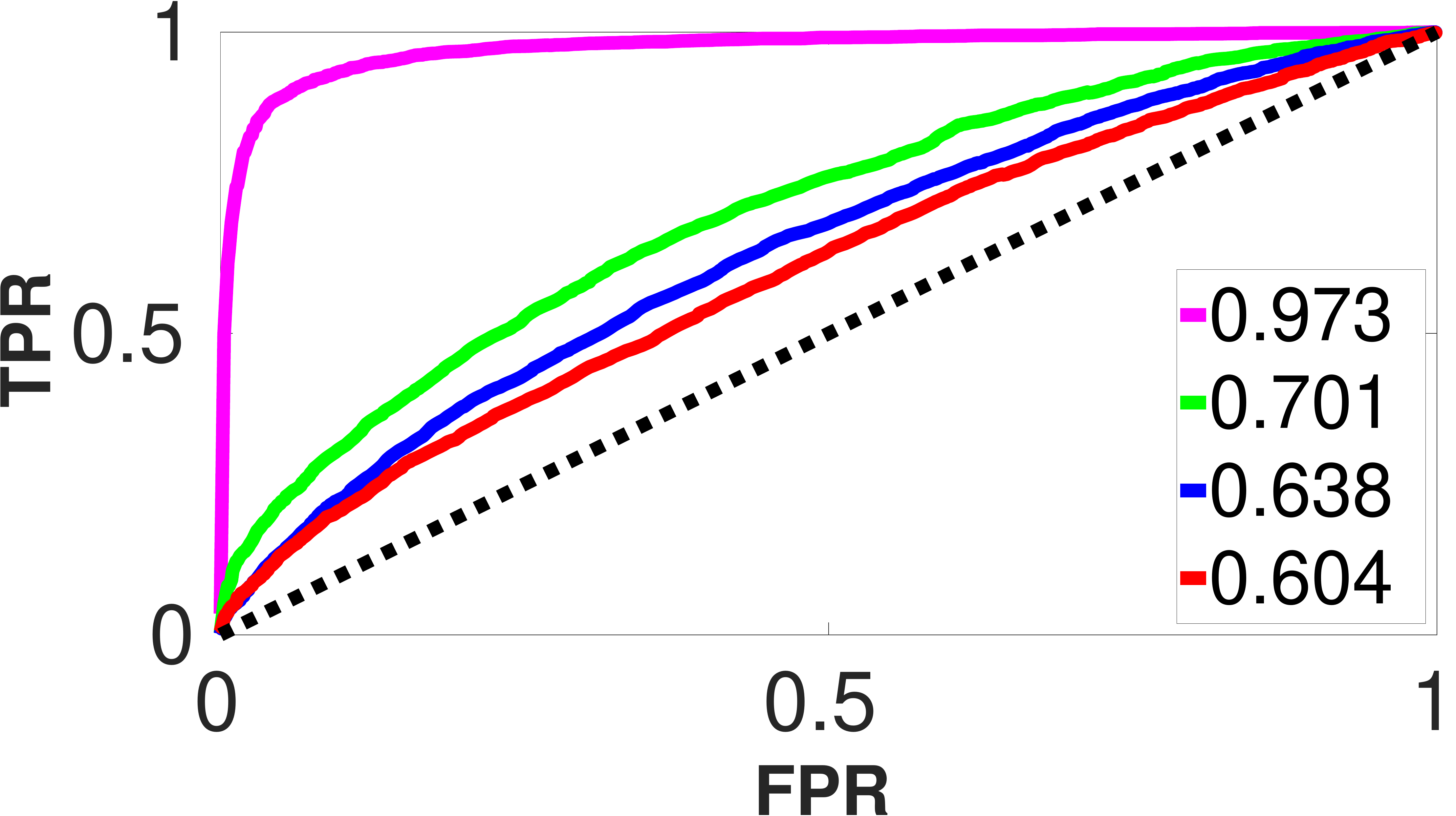}\\
\caption{Receiver Operating Curve (ROC) for the accurate AHGMM parrot-SR attack at threshold $\rho_j^o=0.5$ px/cm. Each ROC is the mean of 10-curves generated by the 10-folds used for cross validation. Legend: \textcolor{magenta}{\Huge ---} Unprotected, \textcolor{green}{\Huge---} \ac{AGB}, \textcolor{blue}{\Huge---} \ac{SVGB}, \textcolor{red}{\Huge ---} AHGMM. This test is performed only for a single resolution ($96 \times 96$ pixels) and pitch anfle ($0^\circ$). The legend values represent the Area Under Curve (AUC).}
\label{fig:ROCs_accurate_ahgmm_parrot-sr}
\end{figure}

\begin{figure}[t]
\centering
\Large
\resizebox{1\columnwidth}{!}{%
\begin{tabular}{|c|cc|cc|cccc|}
\hline
\multirow{3}{*}{Original} & \multicolumn{2}{c|}{AGB} & \multicolumn{2}{c|}{SVGB} &\multicolumn{4}{c|}{AHGMM}
\\
\cline{2-9}
&filtered & restored & filtered & restored & filtered & \multicolumn{3}{c|}{restored}\\
\cline{7-9}
&&&&& & optimal & pseudo & accurate \\ \hline

\includegraphics[width=0.2\linewidth]{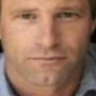}
&
\includegraphics[width=0.2\linewidth]{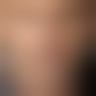}
&
\includegraphics[width=0.2\linewidth]{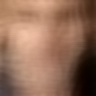}
&
\includegraphics[width=0.2\linewidth]{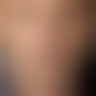}
&
\includegraphics[width=0.2\linewidth]{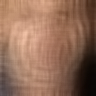}
&
\includegraphics[width=0.2\linewidth]{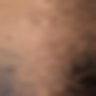}
&
\includegraphics[width=0.2\linewidth]{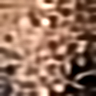}
&
\includegraphics[width=0.2\linewidth]{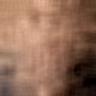}
&
\includegraphics[width=0.2\linewidth]{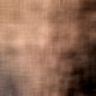}
\\
\includegraphics[width=0.2\linewidth]{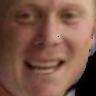}
&
\includegraphics[width=0.2\linewidth]{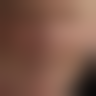}
&
\includegraphics[width=0.2\linewidth]{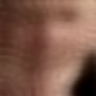}
&
\includegraphics[width=0.2\linewidth]{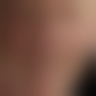}
&
\includegraphics[width=0.2\linewidth]{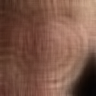}
&
\includegraphics[width=0.2\linewidth]{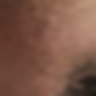}
&
\includegraphics[width=0.2\linewidth]{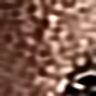}
&
\includegraphics[width=0.2\linewidth]{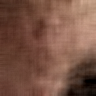}
&
\includegraphics[width=0.2\linewidth]{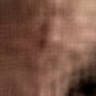}
\\
\includegraphics[width=0.2\linewidth]{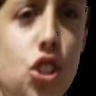}
&
\includegraphics[width=0.2\linewidth]{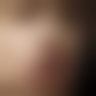}
&
\includegraphics[width=0.2\linewidth]{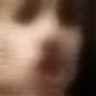}
&
\includegraphics[width=0.2\linewidth]{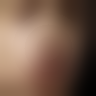}
&
\includegraphics[width=0.2\linewidth]{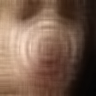}
&
\includegraphics[width=0.2\linewidth]{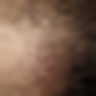}
&
\includegraphics[width=0.2\linewidth]{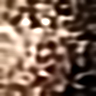}
&
\includegraphics[width=0.2\linewidth]{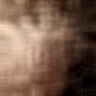}
&
\includegraphics[width=0.2\linewidth]{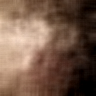}
\\
\hline
\end{tabular}
}
\caption{Visual comparison of reconstructed faces with super-resolution algorithm SRCNN \citep{Dong2016} for threshold $\rho_j^o = 0.5$ px/cm. Reconstruction performance deteriorates from AGB \citep{sarwar2016} over SVGB \citep{Saini2012} to AHGMM protected faces.} 
\label{fig:super-resolution_attack}
\end{figure}

We evaluate the super-resolution (SR) attack under four sub-attacks: optimal kernel na\"{i}ve-SR sub-attack, pseudo AHGMM na\"{i}ve-SR sub-attack, accurate AHGMM na\"{i}ve-SR sub-attack and accurate AHGMM parrot-SR sub-attack. Tab. \ref{tab:super-resolution_results} summarises the achieved accuracies under the different sub-attacks, while Fig. \ref{fig:ROCs_accurate_ahgmm_parrot-sr} presents the \ac{ROC} for the accurate AHGMM parrot-SR sub-attack. Fig. \ref{fig:super-resolution_attack} depicts a visual comparison of the super-resolution reconstruction for three sample faces protected by AGB, SVGB and AHGMM filters.

\begin{table}[t]
\centering
\caption{Face verification accuracy $\eta$ after a super-resolution attack on faces protected by adaptive Gaussian blur (AGB), space variant Gaussian blur (SVGB) and AHGMM at threshold $\rho_j^o=0.5$ px/cm. The values of $\eta$ are given as $\tilde{\mu}(\tilde{\sigma})$, where $\tilde{\mu}$ indicates the mean and $\tilde{\sigma}$ the standard deviation for the 10-fold cross validations. In the na\"{i}ve-SR attack, the reconstructed probe faces are compared against the unprotected gallery images, while both the probe and the gallery images are super-resolved in the parrot-SR attack. }
\label{tab:super-resolution_results}
\resizebox{\columnwidth}{!}{%
\begin{tabular}{|c|c|c|c|}
\hline
Attack type  & AGB & SVGB & AHGMM \\
\hline
optimal na\"{i}ve-SR & 0.592 (0.012) & 0.566 (0.016) & \textbf{0.515 (0.014)}\\
pseudo AHGMM na\"{i}ve-SR & -- & -- & \textbf{0.520 (0.006)}\\
accurate AHGMM na\"{i}ve-SR & --& -- & \textbf{0.532 (0.018)}\\
\hline
accurate AHGMM parrot-SR & 0.634 (0.015) & 0.583(0.034) & \textbf{0.546 (0.018)}\\
\hline
\end{tabular}
}
\end{table}  
%

For the space invariant Gaussian blur (AGB), it is apparent from Fig. \ref{fig:super-resolution_attack} that the SR attack can reconstruct the faces more effectively, even when the kernel size is quite high (i.e. $\rho_j^o=0.5$ px/cm). Therefore, the faces protected by AGB achieves a higher accuracy (see Tab. \ref{tab:super-resolution_results}). In contrast, faces protected by linear space variant Gaussian blur (SVGB) are difficult to reconstruct. The main reason is that the SR mapping becomes erroneous especially for patches which contain parts processed by different kernels. However, SR can effectively reconstruct patches where the Gaussian blur is locally invariant (e.g.~compare the areas around eyes of the SVGB restored faces in Fig. \ref{fig:super-resolution_attack}). The overall reconstruction is worse than for AGB and thus the achieved accuracy is lower.

Reconstruction by super-resolution is even more challenging for AHGMM protected faces. The main reason is that a single patch for learning the mapping contains several sub-regions each filtered with pseudo-randomly correlated Gaussian mixture models. Thus, the error in the learned SR mapping increases resulting in the lowest accuracy as compared to AGB and SVGB.   

Similarly to parrot-IF attack, the accuracy improves for the parrot-SR attack where SR-reconstruction is also performed for the gallery images. Especially for AGB and SVGB, the similarity between (protected and reconstructed) gallery images and the (reconstructed) probe images increases. Thus, the accuracy increases. As for the other attacks, AHGMM is more robust to parrot attacks than AGB and SVGB, and achieves the lowest accuracy.  
\begin{figure}
\centering

\begin{subfigure}{.49\linewidth}
  \centering
  \includegraphics[width=\linewidth]{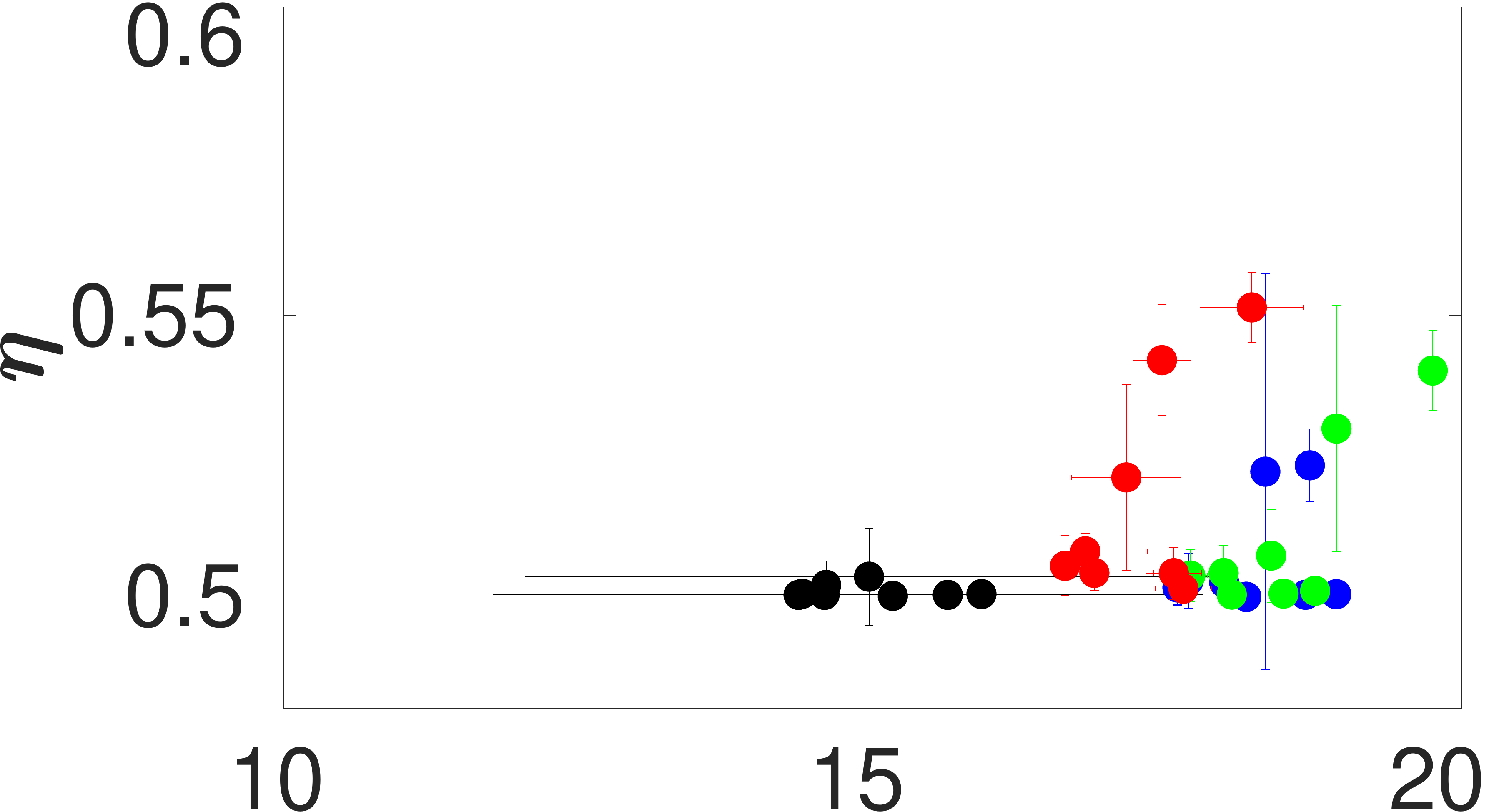}
   \caption{}
   \label{fig:privacy_fidelity_adaptive_psnr_th_0_5}
\end{subfigure} 
\begin{subfigure}{.49\linewidth}
  \centering
  \includegraphics[width=\linewidth]{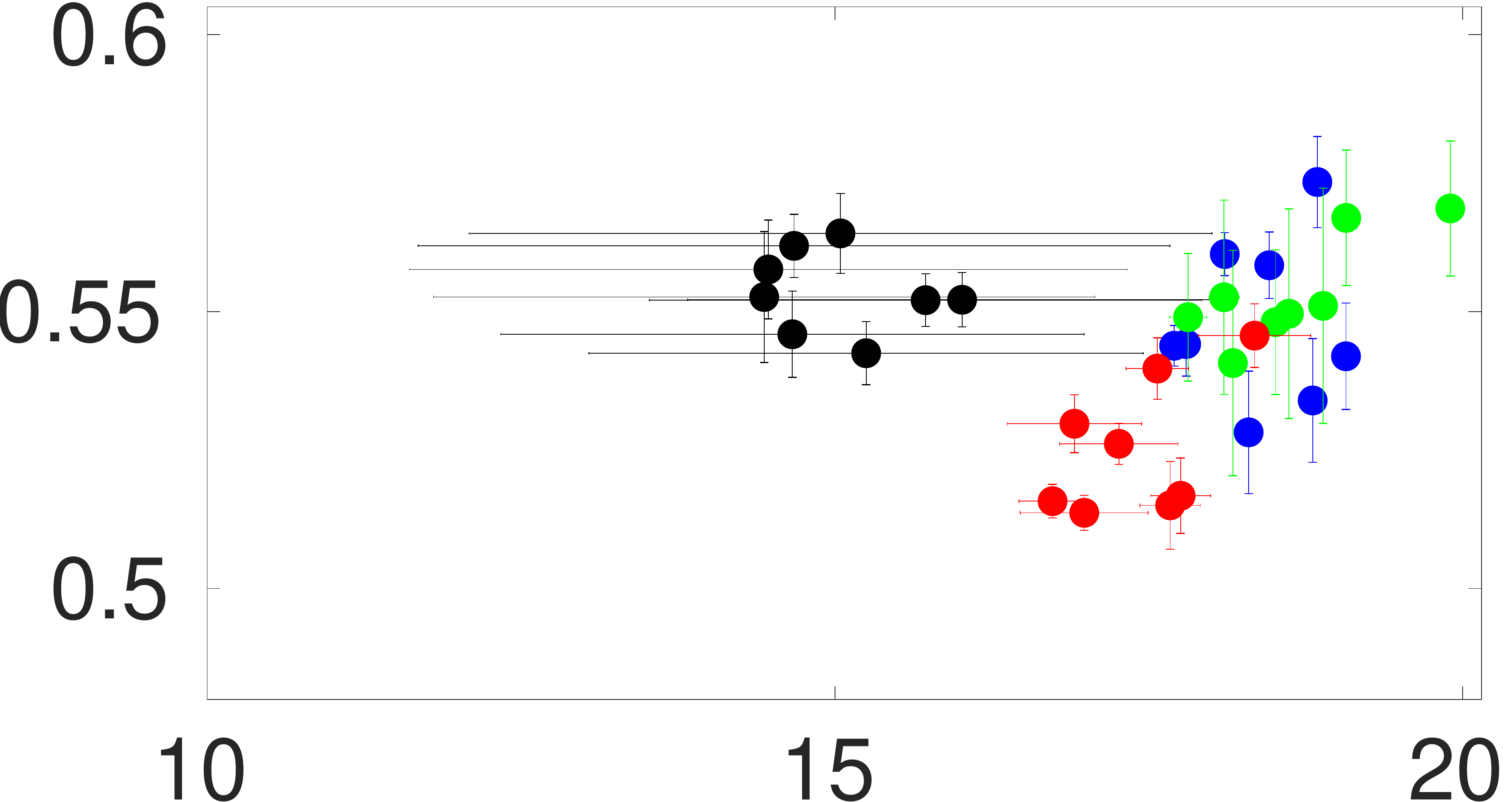}
   \caption{}
   \label{fig:privacy_fidelity_adaptive_psnr_th_0_5}
\end{subfigure} 
\\
%
\begin{subfigure}{.49\linewidth}
  \centering
  \includegraphics[width=\linewidth]{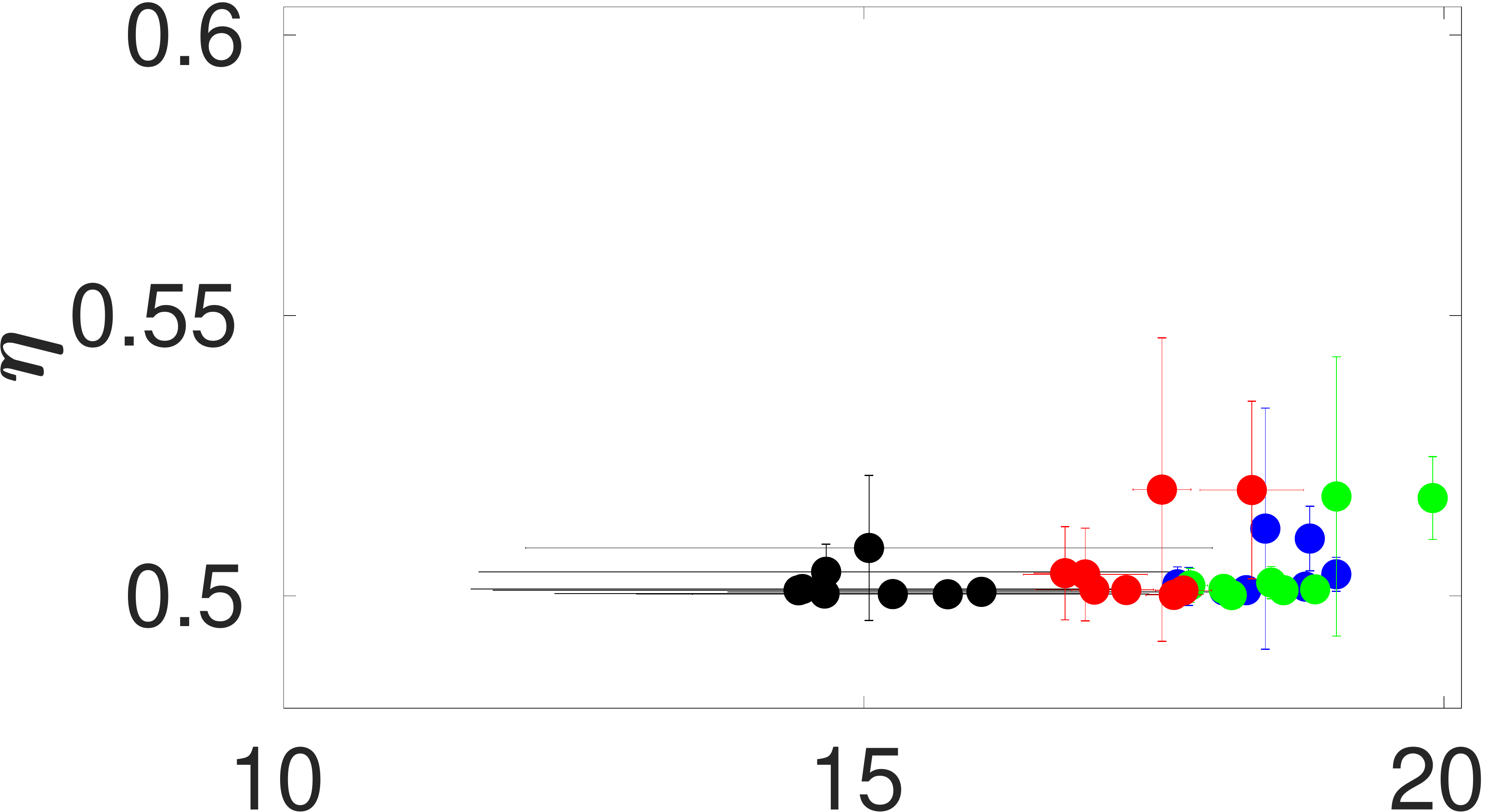}
   \caption{}
   \label{fig:privacy_fidelity_adaptive_psnr_th_0_5}
\end{subfigure} 
\begin{subfigure}{.49\linewidth}
  \centering
  \includegraphics[width=\linewidth]{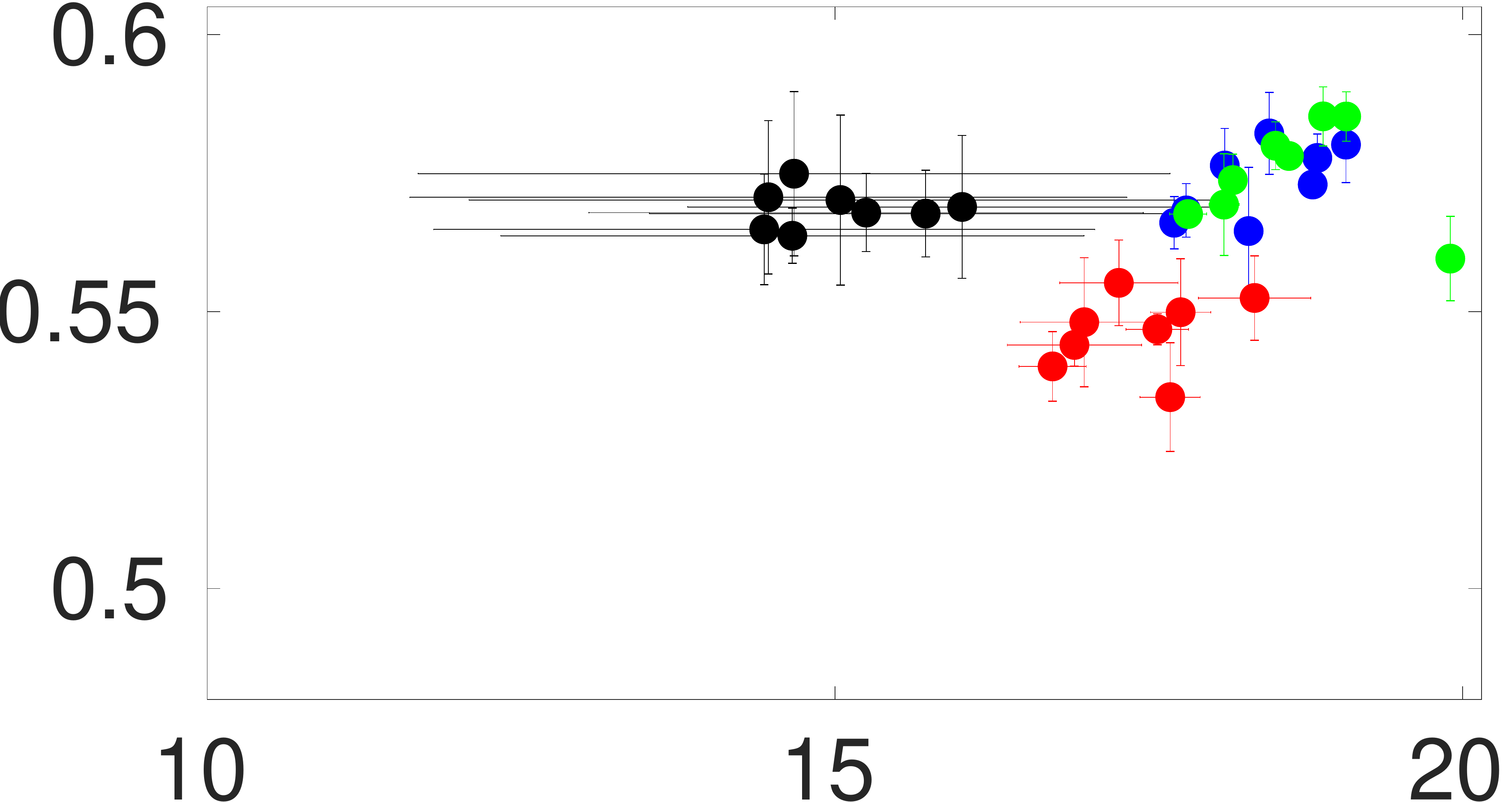}
   \caption{}
   \label{fig:privacy_fidelity_adaptive_psnr_th_0_5}
\end{subfigure}
\\
%
\begin{subfigure}{.49\linewidth}
  \centering
  \includegraphics[width=\linewidth]{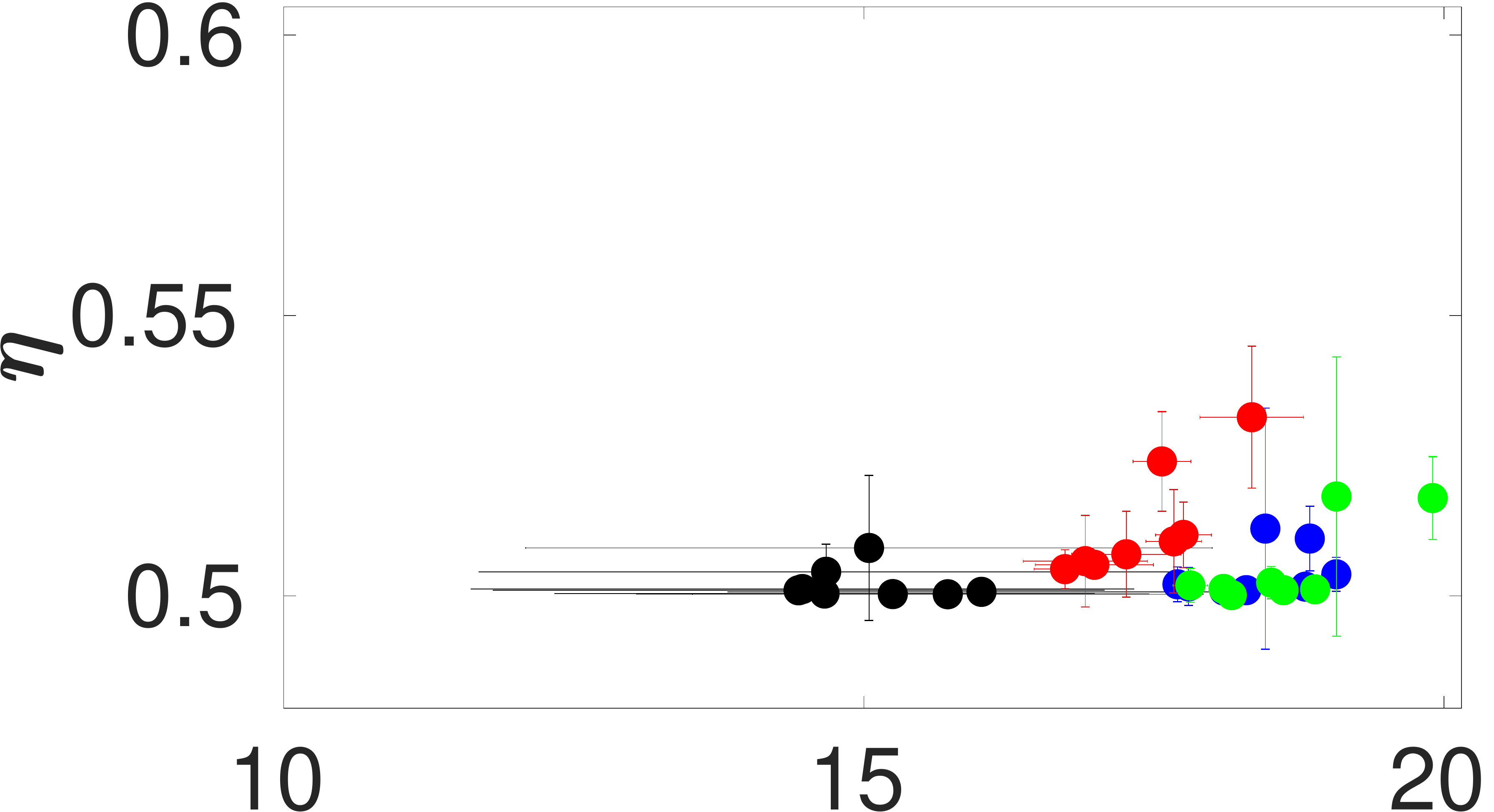}
   \caption{}
   \label{fig:privacy_fidelity_pseudoAhgmm_psnr_th_0_5}
\end{subfigure}
\begin{subfigure}{.49\linewidth}
  \centering
  \includegraphics[width=\linewidth]{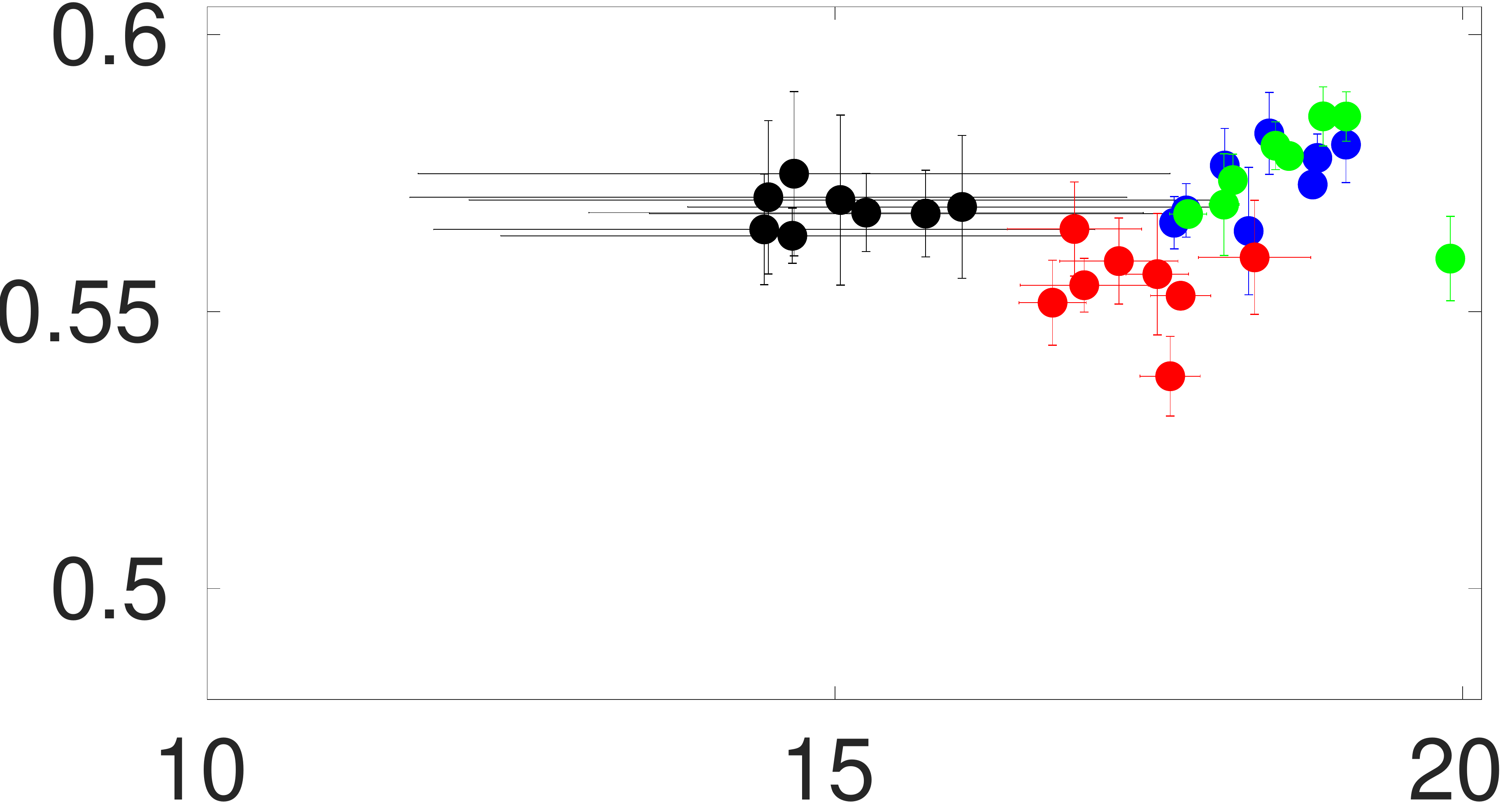}
  
   \caption{}
   \label{fig:privacy_fidelity_pseudoAhgmm_psnr_th_0_5}
\end{subfigure}
\\
%
\begin{subfigure}{.49\linewidth}
  \centering
  \includegraphics[width=\linewidth]{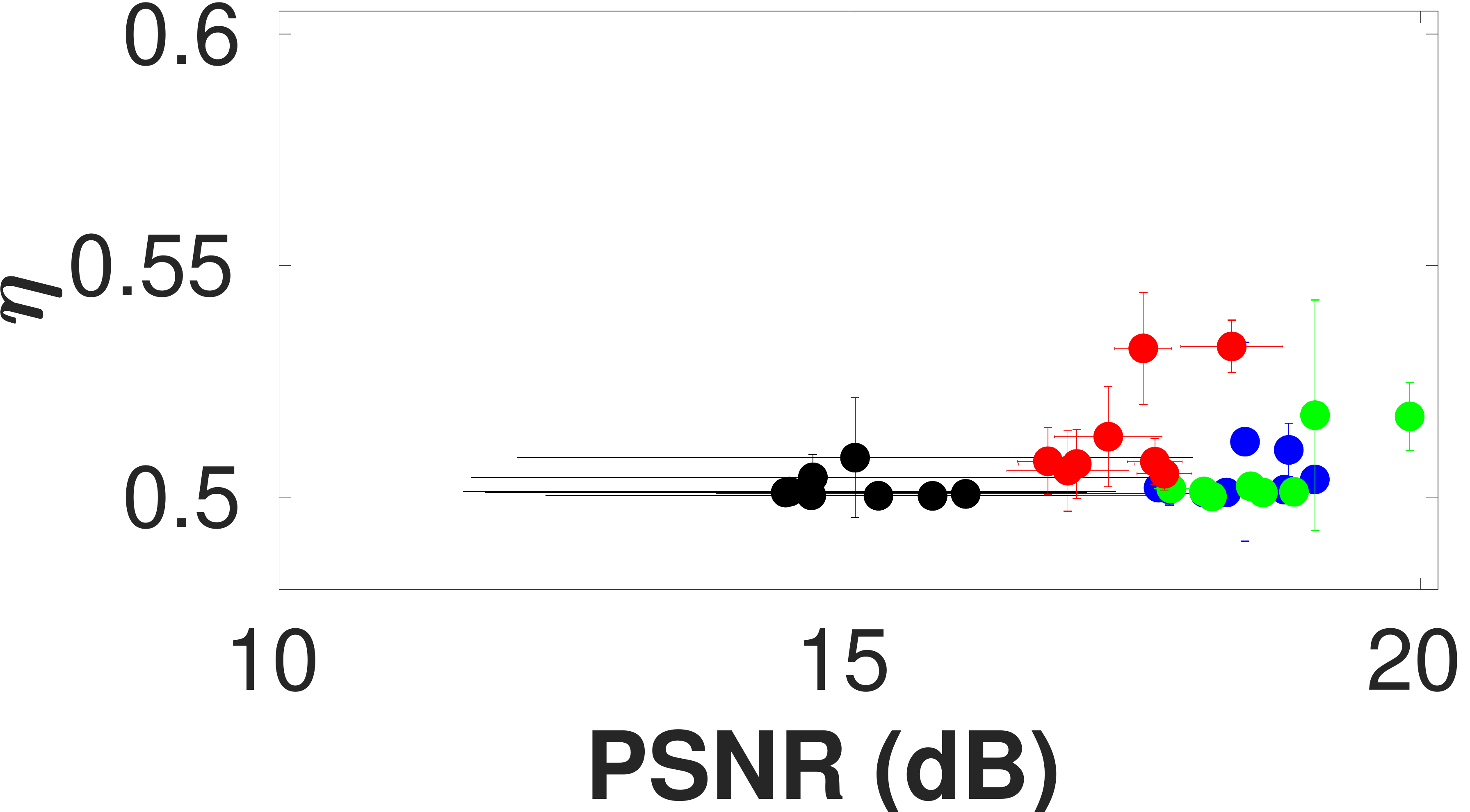}
   \caption{}
   \label{fig:privacy_fidelity_inverse_AccurateAhgmm_psnr}
\end{subfigure}
\begin{subfigure}{.49\linewidth}
  \centering
  \includegraphics[width=\linewidth]{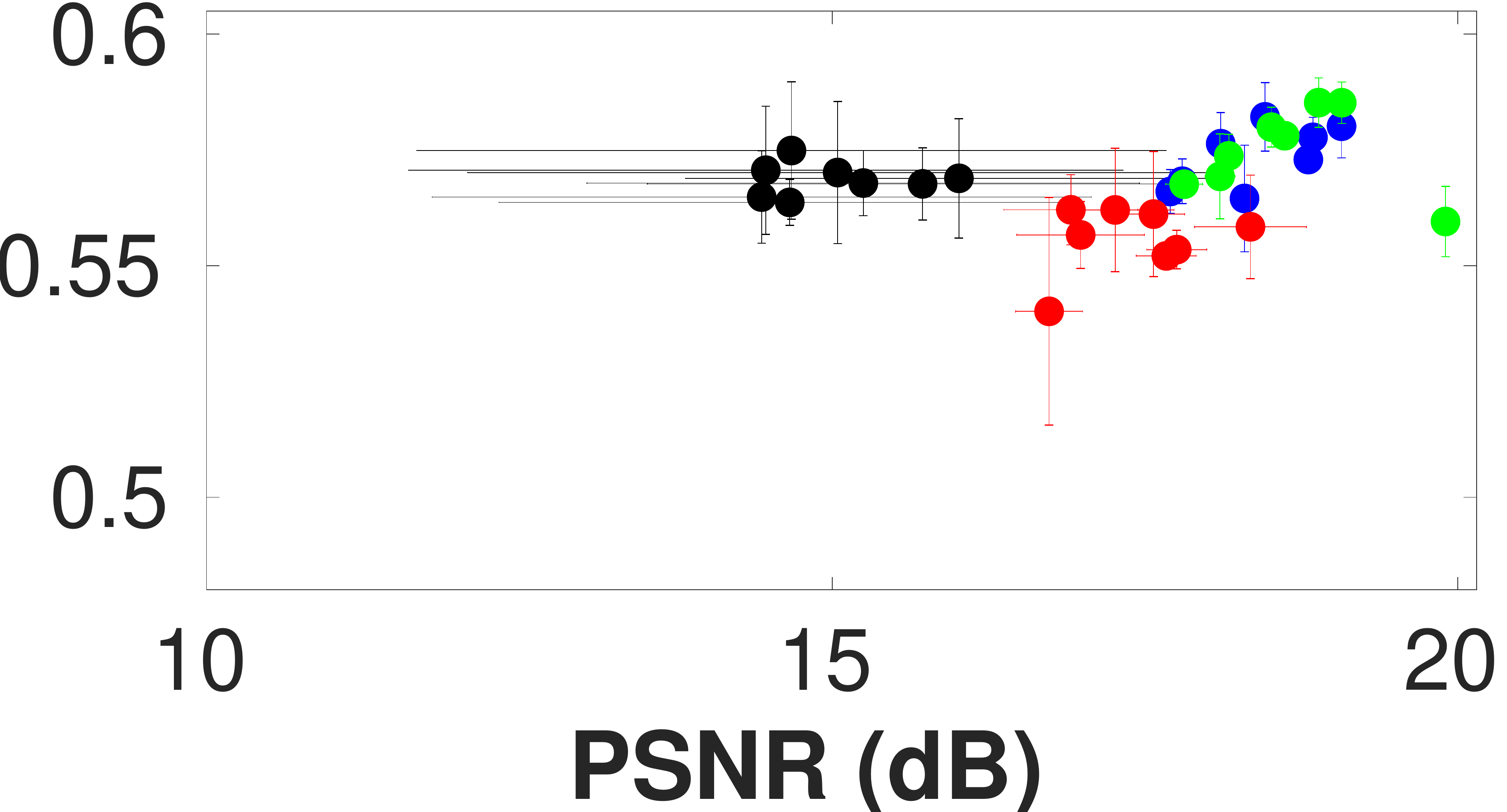}
   \caption{}
   \label{fig:privacy_fidelity_AccurateAhgmm_psnr_th_0_5}
\end{subfigure}

\caption{Trade-off analysis between the Face verification accuracy $\eta$ and the distortion provided by the different privacy filters under the na\"{i}ve-T, parrot-T and inverse filter (IF) attacks at threshold $\rho_j^o=0.5$ px/cm. The distortion is measured by the Peak Signal to Noise Ratio (PSNR). Legend: \textcolor{red}{\Huge ---} AHGMM, \textcolor{green}{\Huge{---}} \ac{AGB} \citep{sarwar2016}, \textcolor{blue}{\Huge{---}} \ac{SVGB} \citep{Saini2012}, \textcolor{black}{\Huge{---}} \ac{FGB}. Under the na\"{i}ve-T attack, our proposed AHGMM possesses $\eta$ almost equivalent to the state-of-the-art filter, but lowest under the parrot-T attacks. However, AHGMM has slightly lower PSNR as compared to AGB and SVGB, but much higher than FGB. (a) na\"{i}ve-T attack, (b) accurate AHGMM parrot-IF attack, (c) optimal kernel na\"{i}ve-IF attack, (d) optimal kernel parrot-T attack, (e) pseudo AHGMM na\"{i}ve-IF attack, (f) pseudo AHGMM parrot-T attack, (g) accurate AHGMM na\"{i}ve-IF attack and (h) accurate AHGMM parrot-T attack. For the last three na\"{i}ve-IF and parrot-T attacks, the results of AGB, SVGB and FGB are the same and have been superimposed for the comparison. Please see Section \ref{subsec:naive_attack}, Section \ref{subsec:Parrot_attack} and Section \ref{subsec:inverse_filter_attack} for the details of the attacks.}
\label{fig:ssim_psnr_th_0_5}
\end{figure}
\subsection{Distortion Analysis}
\label{subsec:fidelity_analysis} 
We measure the distortion of the FGB, SVGB \citep{Saini2012},  AGB \citep{sarwar2016} and AHGMM using \ac{PSNR}. For a trade-off analysis between distortion and privacy, we plot the face verification accuracy against PSNR. The results of this trade-off analysis are presented in Fig. \ref{fig:ssim_psnr_th_0_5}.

AGB \citep{sarwar2016} has the highest average PSNR values followed by SVGB \citep{Saini2012}, AHGMM and FGB. The main reason is that AGB uses a single anisotropic kernel instead of spatially linearly varying kernel used by SVGB \citep{Saini2012}. Although AHGMM also uses an anisotropic kernel like AGB, the spatial hopping phenomena of the Gaussian mixture model of the AHGMM results in high distortion (PSNR values) as compared to AGB and SVGB (see Fig. \ref{fig:visual_comparison}). FGB has the highest distortion as it does not change its parameters depending upon the resolution of the face.  

\section{Conclusion}
\label{sec:conclusion}

We presented an irreversible visual privacy protection filter which is robust against a parrot, an inverse-filter and a super-resolution attack that are faced by an adhoc blurring of sensitive regions. The proposed filter is based on an adaptive hopping Gaussian mixture model. Depending upon the captured resolution of a sensitive region, the filter globally adapts the parameters of the Gaussian mixture model to minimise the distortion, while locally hop them pseudo-randomly so that an attacker is unable to estimate these parameters. We evaluated the validity of the \ac{AHGMM} using a state-of-the-art face recognition algorithm and a synthetic face data set with faces at different pitch angles and resolutions emulating faces as captured from an \ac{MAV}. The proposed algorithm provides the highest privacy level under a parrot, an inverse-filter and a super-resolution attack and an almost equivalent level of privacy to state-of-the-art privacy filters under a na\"{i}ve attack. 

Unlike face-de-identification approaches (\citep{Newton2005, Gross2006, Du2014, Lin2012,  Letournel2015, Chriskos2015}), we do not depend on an auxiliary visual detector (i.e. pose, facial expression, age, gender, race) to counter a parrot, an inverse-filter or a super-resolution attack. Moreover, unlike the encryption/scrambling filters (\citep{Dufaux2006, Dufaux2008, Baaziz2007, Sohn2011, Korshunov2013, Korshunov2013b, Boult2005, Chattopadhyay2007, Rahman2010, Winkler2011}), AHGMM prevents the recovery of the original face even with access to the seed of the PRNG.

We will make available to the research community the  face dataset of 4281 subjects we generated to emulate faces captured from an \ac{MAV} under varying poses and illumination conditions.

%
\appendix
\label{ap:notations}
All the symbols used in the paper along with their meanings are summarised in Table \ref{tab:list_of_notations}.
\begin{table}[!t]
\caption{List of notations.}
\label{tab:list_of_notations}
\resizebox{1\columnwidth}{!}{%
\begin{tabular}{l|l}
\hline
Notation & Meaning \\
\hline
$R, \bar{R}, \hat{R}$ & \emph{unprotected, protected and reconstructed face region} \\
$C_R$ & \emph{centre of $R$} \\
$W, H$ & \emph{width and height of $R$} \\
$\mathcal{D}$ & \emph{A data set including both gallery and probe data sets} \\
$\mathcal{R_G}, \mathcal{\bar{R}_G}, \mathcal{\hat{R}_G}$ & \emph{unprotected, protected and reconstructed  gallery data set} \\
$\mathcal{R_P}, \mathcal{\bar{R}_P}, \mathcal{\hat{R}_P}$ & \emph{unprotected, protected and reconstructed  probe data set} \\
$K, \tilde{K}$ & \emph{original and predicted identity labels} \\
$F_{\Omega_j}$ & \emph{a privacy filter of parameter $\Omega_j$} \\
$G$ & \emph{a function that an attacker exploits} \\
$D$ & \emph{distortion introduced by $F_{\Omega_j}$} \\
$P$ & \emph{probability of predicting the label of a face}\\
$\eta$ & \emph{face verification accuracy} \\
$\epsilon$ & \emph{verification accuracy of a random classifier} \\
$f$ & \emph{focal length of the camera} \\
$p_j$ & \emph{physical dimension of a pixel in $j$ direction} \\
$h_1, h_2$ & \emph{height of a camera and face from ground level} \\
$\vect{N, \vect{P}}$ & \emph{vectors representing Nadir and principal axis of a camera} \\
$\theta_R, \theta_P$ & \emph{angle between $\vect{N}$, $R$ and $\vect{N}$, $\vect{P}$} \\
$N$ & \emph{Number of sub-regions of $R$} \\
$M$ & \emph{Number of supplementary Gaussian functions}\\
$\rho_j$ & \emph{pixel density (px/cm)}, where $j\in\{h,v\}$\\
$\rho_j^o$ & \emph{threshold pixel density for privacy filtering}\\
$\mu_j, \sigma_j$ & \emph{mean and standard deviation of a Gaussian PSF}\\
$\mu_j^o, \sigma_j^o$ & \emph{mean and standard deviation of an optimal Gaussian PSF}\\
$\mu_{jm}, \sigma_{jm}$ & \emph{randomly modified $\mu_j^o$ and $\sigma_j^o$ for $m^{th}$ Gaussian PSF}\\
$\alpha_{jm}, \beta_{jm}$ & \emph{randomly generated numbers for $\mu_{jm}$ and $\sigma_{jm}$}\\
$\Omega_j, \Omega_j^o, \bar{\Omega}_j^o$ & \emph{a tuple ($\mu_j$, $\sigma_j$), ($\mu_j^o$, $\sigma_j^o$)} and ($\mu_{jm}, \sigma_{jm}$)\\
$f_s, f_s^o$ & \emph{Nyquist frequency of $\rho_j$ and $\rho_j^o$}\\
$\acute{\sigma_j^o}$ & \emph{frequency domain standard deviation corresponding to $\sigma_j^o$}\\
$\gamma_{jm}$ & \emph{scaling factor for $\sigma_j^o$} \\
$\mathcal{X}$ & \emph{set of tuple containing parameters of Gaussian functions}\\
$\mathcal{G}$ & \emph{a set of Gaussian functions}\\
$G_{nm}$ & \emph{an element of $\mathcal{G}$ }\\
$\mathcal{\phi}$ & \emph{a set of weights for Gaussian mixture model}\\
$\phi_{nm}$ & \emph{an element of $\mathcal{\phi}$}\\
$\mathcal{M}$ & \emph{Gaussian mixture model}\\
$M_{n}$ & \emph{an element of $\mathcal{M}$}\\
$Q_j$ & \emph{sub-region size in pixels} \\
$\bar{\sigma}_j$ & \emph{standard deviation of global smoothing filter} \\
\hline
\end{tabular}
}
\end{table}

\section*{Acknowledgment}
O. Sarwar was supported in part by Erasmus Mundus Joint Doctorate in Interactive and Cognitive Environment, which is funded by the Education, Audio-visual \& Culture Executive Agency under the FPA no 2010-0015.



%
%
\bibliographystyle{spbasic}
\bibliography{ref}

\end{document}